%% file: _PaperDeeb.tex
\documentclass{article}
\usepackage[
	backend=biber,
	style=alphabetic,
	natbib=true,
	url=false,
	doi=true,
	eprint=true,
	giveninits=true,
	isbn=false,
	maxbibnames=12,
]{biblatex}
\usepackage[a4paper, margin=3cm]{geometry}
\usepackage{hyperref}
\usepackage{authblk}
\usepackage[T1]{fontenc}
\input{packages}
\input{macros_general}
\input{macros_specific}
\newbox{\myorcidthanksbox}
\sbox{\myorcidthanksbox}{\large\includegraphics[height=1.8ex]{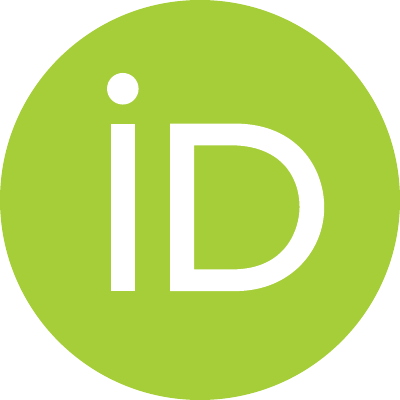}}
\newcommand{\orcidthanks}[1]{%
    \href{https://orcid.org/#1}{\raisebox{-0.5ex}{\usebox{\myorcidthanksbox}}\,#1}}
\title{Machine Learning for Predicting Chaotic Systems}
\date{}
\author[1,2]{Christof Schötz\thanks{math@christof-schoetz.de, \orcidthanks{0000-0003-3528-4544}}}
\author[1,2]{Alistair White\thanks{alistair.white@tum.de, \orcidthanks{0000-0003-3377-6852}}}
\author[1,2]{Maximilian Gelbrecht\thanks{maximilian.gelbrecht@tum.de, \orcidthanks{0000-0002-0729-6671}}}
\author[1,2,3]{Niklas Boers\thanks{n.boers@tum.de, \orcidthanks{0000-0002-1239-9034}}}
\affil[1]{Potsdam Institute for Climate Impact Research}
\affil[2]{Technical University of Munich}
\affil[3]{University of Exeter}
\begin{document}
\maketitle
\begin{abstract}
	\input{abstract.tex}
\end{abstract}
\tableofcontents
\input{sec_intro.tex}
\input{sec_methodology.tex}
\input{sec_results.tex}
\input{sec_conclusion.tex}
\newpage
\begin{appendix}
\input{sec_app_data.tex}\input{sec_app_methods.tex}\input{sec_app_smape.tex}
\input{sec_app_results.tex}
\end{appendix}
\printbibliography

\newpage

\input{app_tables_figures.tex}
\end{document}

%% file: packages.tex
\usepackage{amsthm,amsmath,amsfonts,amssymb}
\usepackage{mathtools}
\usepackage{stmaryrd}
\usepackage[capitalize,nameinlink,noabbrev]{cleveref} 
\usepackage{enumitem} 
\usepackage{graphicx} 
\usepackage{dsfont} 
\usepackage{tikz} 
\usetikzlibrary{positioning}
\usepackage{booktabs, caption, longtable, colortbl, array, anyfontsize, multirow}
\usepackage{helvet}
\usepackage{graphicx,calc}
\usepackage{caption}
\usepackage{subcaption}

%% file: macros_general.tex
\newcommand*{\mc}[1]{\mathcal{#1}}

\newcommand*{\ms}[1]{\mathsf{#1}}
\newcommand*{\mo}[1]{\mathbf{#1}}

\newcommand{\N}{\mathbb{N}}

\newcommand{\R}{\mathbb{R}}
\newcommand{\Rp}{\R_{\geq0}}
\newcommand{\Rpp}{\R_{>0}}

\newcommand{\nnset}[1]{{\left\llbracket #1\right\rrbracket}}

\newcommand{\lcb}{\left\lbrace} 
\newcommand{\rcb}{\right\rbrace} 
\newcommand{\cb}[1]{\lcb #1 \rcb} 
\newcommand{\lb}{\left(} 
\newcommand{\rb}{\right)} 
\newcommand{\br}[1]{\lb #1 \rb} 
\newcommand{\brOf}[1]{\!\br{#1}} 
\newcommand{\abs}[1]{\left| #1 \right|} 
\newcommand{\euclof}[1]{\Vert#1\Vert_2}
\newcommand{\euclOf}[1]{\left\Vert#1\right\Vert_2}

\newcommand{\sizedMid}[2]{#1 \, \kern-\nulldelimiterspace\mathopen{}\left| \vphantom{#1}\,#2\right.\mathclose{}\kern-\nulldelimiterspace}
\newcommand{\setByEle}[2]{\cb{\sizedMid{#1}{#2}}}

\newcommand{\tr}{^{\!\top}\!} 
\newcommand{\pr}{^\prime}

\newcommand{\eqcm}{\,,} 
\newcommand{\eqfs}{\,.}
\newcommand{\dl}{\mathrm{d}} 
\DeclareMathOperator*{\argmin}{arg\,min}

%% file: macros_specific.tex
\newcommand{\PlotDirPath}{img/}
\newcommand{\method}[1]{\normalfont\texttt{#1}}
\newcommand{\methodh}[2]{\texorpdfstring{\protect\hyperlink{method#1}{\normalfont\texttt{#2}}}{}}
\newcommand{\methodNewh}[1]{\hypertarget{method#1}{}}
\newcommand{\model}[1]{{\normalfont\textsc{#1}}}

\newcommand{\tvalid}{t_{\ms{valid}}}
\newcommand{\smape}{\ms{sMAPE}}
\newcommand{\cme}{\ms{CME}}
\newcommand{\sd}{\ms{sd}}
\newcommand{\stepsize}{\Delta\!t}
\newcommand{\DeebLorenz}{{\normalfont\textit{DeebLorenz}}}
\newcommand{\Dysts}{{\normalfont\textit{Dysts}}}
\newlength\myheight
\newlength\mydepth
\newcommand*\inlinegraphics[1]{%
    \settototalheight\myheight{SXygp}%
    \settodepth\mydepth{SXygp}%
    \raisebox{-\mydepth}{\includegraphics[height=\myheight]{#1}}%
}
\newcommand{\CmeScale}{%
    \raisebox{-1.4ex}{%
    \begin{tikzpicture}
        \def\scalelen{2.2}
        \draw (0,0) -- (\scalelen,0);
        \draw (0,0.0) -- (0, -0.1);
        \draw (0.5*\scalelen,0.0) -- (0.5*\scalelen, -0.1);
        \draw (\scalelen,0.0) -- (\scalelen, -0.1);
        \node[above,yshift=-0.5mm] at (0.0*\scalelen,0.0) {$\mathsf{0}$};
        \node[above,yshift=-0.5mm] at (0.5*\scalelen,0.0) {$\mathsf{0.5}$};
        \node[above,yshift=+3.5mm,outer sep=0pt,inner sep=0pt] at (0.5*\scalelen,0.0) {$\cme$};
        \node[above,yshift=-0.5mm] at (1.0*\scalelen,0.0) {$\mathsf{1}$};
    \end{tikzpicture}}%
}
\definecolor{colorLorenz63random}{HTML}{F8766D}
\definecolor{colorLorenz63std}{HTML}{00BA38}
\definecolor{colorLorenz63nonpar}{HTML}{619CFF}
\newcommand{\SymbolLorenzRandom}{\tikz[baseline]{\node[inner sep=0.5mm,outer sep=0, font=\sffamily,fill=colorLorenz63random,anchor=base] {R}}}
\newcommand{\SymbolLorenzStandard}{\tikz[baseline]{\node[inner sep=0.5mm,outer sep=0,font=\sffamily,fill=colorLorenz63std,anchor=base] {S}}}
\newcommand{\SymbolLorenzNonparam}{\tikz[baseline]{\node[inner sep=0.5mm,outer sep=0,font=\sffamily,fill=colorLorenz63nonpar,anchor=base] {N}}}
\definecolor{colorClassDirect}{HTML}{FFA0FF}
\definecolor{colorClassGrDesc}{HTML}{F0C0C0}
\definecolor{colorClassFitPro}{HTML}{B0B0FF}
\definecolor{colorClassFitSol}{HTML}{00D8D8}
\newcommand{\ClassDirect}{\tikz[baseline]{\node[inner sep=0.5mm,outer sep=0,font=\sffamily,fill=colorClassDirect,anchor=base] {Direct\vphantom{pg}}}}
\newcommand{\ClassGrDesc}{\tikz[baseline]{\node[inner sep=0.5mm,outer sep=0,font=\sffamily,fill=colorClassGrDesc,anchor=base] {Gradient Descent\vphantom{pg}}}}
\newcommand{\ClassFitPro}{\tikz[baseline]{\node[inner sep=0.5mm,outer sep=0,font=\sffamily,fill=colorClassFitPro,anchor=base] {Fit Propagator\vphantom{pg}}}}
\newcommand{\ClassFitSol}{\tikz[baseline]{\node[inner sep=0.5mm,outer sep=0,font=\sffamily,fill=colorClassFitSol,anchor=base] {Fit Solution\vphantom{pg}}}}

%% file: abstract.tex
Predicting chaotic dynamical systems is critical in many scientific fields, such as weather forecasting, but challenging due to the characteristic sensitive dependence on initial conditions. Traditional modeling approaches require extensive domain knowledge, often leading to a shift towards data-driven methods using machine learning. However, existing research provides inconclusive results on which machine learning methods are best suited for predicting chaotic systems. In this paper, we compare different lightweight and heavyweight machine learning architectures using extensive existing benchmark databases, as well as a newly introduced database that allows for uncertainty quantification in the benchmark results. In addition to state-of-the-art methods from the literature, we also present new advantageous variants of established methods. Hyperparameter tuning is adjusted based on computational cost, with more tuning allocated to less costly methods. Furthermore, we introduce the cumulative maximum error, a novel metric that combines desirable properties of traditional metrics and is tailored for chaotic systems. Our results show that well-tuned simple methods, as well as untuned baseline methods, often outperform state-of-the-art deep learning models, but their performance can vary significantly with different experimental setups. These findings highlight the importance of aligning prediction methods with data characteristics and caution against the indiscriminate use of overly complex models.

%% file: sec_intro.tex
\section{Introduction}\label{sec:intro}
Chaos is a ubiquitous phenomenon in nature and technology, arising from nonlinearities, complex interactions, or feedbacks within the system \cite{Ott1993, Strogatz2024}. Predicting the behavior of such dynamical systems is a critical goal in many scientific disciplines. Chaotic systems are characterized by their sensitive dependence on initial conditions, making them notoriously difficult to predict and posing a significant challenge to researchers studying them.

A traditional approach to addressing this problem involves using domain knowledge to model the dynamical system. However, this approach requires detailed understanding of the specific system that is often not be available. In recent years, system-agnostic, data-driven approaches have emerged as a viable alternative. These methods, which leverage techniques from statistics and machine learning, are trained on observational data to predict the future behavior of the system.

These algorithms range from relatively simple and computationally inexpensive methods, such as fitting polynomials, to highly sophisticated models that utilize complex neural network architectures, demanding significant computational power for training. To evaluate these approaches, various comparison studies have been conducted.

For instance, \cite{Han2021} focus on complex machine learning models with mostly classical neural network architectures. They do not include the sophisticated transformer model \cite{Vaswani2017}, which has gained popularity due to its effectiveness for large language models. Such models are surveyed in \cite{wen2022transformers}, where different variants of transformer models for time series forecasting are reviewed. Remarkably, \cite{zeng2023transformers} demonstrate how simple linear methods can outperform these complex models. Additionally, \cite{vlachas2020backpropagation} compare reservoir computers (RCs) and recurrent neural networks (RNNs), favoring RNNs, whereas \cite{Shahi2022} conduct a similar comparison, adding further machine learning methods and concluding that RCs are preferable.

Previous studies typically compare methods on a limited number of systems. However, \cite{Gilpin21, Gilpin23} emphasize the need to use extensive databases and benchmarks to evaluate the performance of different methods. They introduce a database of more than 130 different chaotic dynamical systems and, in their comparison study, find that the sophisticated NBEATS method \cite{oreshkin2020nbeatsneuralbasisexpansion} performs best. Hyperparameter tuning in this study is minimal, despite its importance for algorithm performance.

All in all, there is no consensus in the literature on which method should be preferred. Even a distinction in the quality of results between lightweight and heavyweight methods is not clear.
Apart from using only a few systems (and often different data in each study), the lack of robustness of existing comparisons may stem from an inherent randomness in the evaluations.  The results of the error metrics used may vary when different initial conditions are applied. If experiments are repeated only a few times with randomly drawn initial conditions, the results may still not be robust. If there is only one trial, then uncertainty cannot even be judged.

With this work, we aim to address several questions left open in previous studies. We compare various lightweight and heavyweight algorithms, including simple statistical methods that have shown promising performance but have received limited attention \cite{Ramsay_2017}. We also present modifications of algorithms based on propagator map estimates, such as RCs, where we target the increment instead of the next state, and show the advantages of these for certain prediction tasks. We tune hyperparameters guided by the computational cost of each algorithm, allowing a broader exploration of hyperparameter sets for lightweight methods to facilitate a fairer comparison with heavyweight models.

As one source of synthetic data, we use the \Dysts{} database of \cite{Gilpin21}, allowing tests on a wide range of chaotic systems and direct comparison with the results in \cite{Gilpin23}. Additionally, we introduce a new database, \DeebLorenz{}, which encompasses three variations of the well-known Lorenz63 system \cite{Lorenz1963}, allowing for an easy comparison with other studies on the Lorenz63 system. Our database includes $100$ repetitions of simulated time series for each system to increase the robustness of results and to allow uncertainty quantification. The database also features a nonparametric version of Lorenz63 to assess the performance of polynomial-based fitting algorithms on non-polynomial targets. All the systems studied have state-space dimensions ranging from $3$ to $10$, categorizing them as low-dimensional. Different observation schemes are used for each system to examine the impact of noise and time step intervals on performance. Furthermore, we propose a new error metric for forecasting tasks on dynamical systems, the cumulative maximum error ($\cme$).
It combines the benefits of integrated error metrics (as used in, e.g., \cite{Gilpin23, godahewa2021monashtimeseriesforecasting}) and measures of forecast horizon, such as the valid time \cite{Ren2009, pathak2018hybrid}.

Our study demonstrates significantly better performance of various tuned and untuned simple methods compared to complex machine learning models on forecasting tasks for low-dimensional chaotic systems. Additionally, we find that the relative performance of methods is highly dependent on the experimental setup, though some trends, such as the strong performance of simple methods, remain consistent. Notably, introducing random time steps between consecutive observations dramatically alters performance. In this case, a Gaussian process-based method \cite{Rasmussen2006Gaussian,heinonen18} that ranks mid-tier under constant time steps, outperforms all other methods.

The remaining parts of the paper are structured as follows: \cref{sec:methodology} describes the databases, the forecasting task, the methods compared, the $\cme$ error metric, and the hyperparameter tuning. The results and key insights of the simulation study are described in \cref{sec:results}, followed by a short conclusion in \cref{sec:conclusion}. \cref{sec:app:data} provides additional details on the databases, followed by a discussion of the estimation methods in \cref{sec:app:methods}. Finally, additional tables and plots detailing the simulation study results are available in \cref{sec:app:tabresults}.

%% file: sec_methodology.tex
\section{Methodology}\label{sec:methodology}
\subsection{Data}\label{ssec:data}
We use two different databases: \DeebLorenz\footnote{Differential Equation Estimation Benchmark -- Lorenz, available at \url{https://doi.org/10.5281/zenodo.12999941}}, a database created specifically for this study, and the \Dysts{} database\footnote{\url{https://github.com/williamgilpin/dysts}, \url{https://github.com/williamgilpin/dysts_data}} from \cite{Gilpin21}. An overview of their structure is given in \cref{fig:data:overview}, and further details are described below.

Both databases consist of solutions $u \colon \R\to \R^d$ of autonomous, first-order ordinary differential equations (ODEs), i.e., $\dot u(t) = f(u(t))$, where $\dot u$ is the derivative of $u$ and $f\colon\R^d\to\R^d$ is the vector field of the ODE describing the dynamics of the system. All considered systems are chaotic, i.e., they exhibit a sensitive dependence on initial conditions, an aperiodic long-term behavior, and a fractal structure in the state space \cite{Ott1993, Strogatz2024}.

The two databases are complementary in nature. \Dysts{} consists of many different chaotic systems with only one time series per system and dataset, and only a rather small amount of samples per time series. \DeebLorenz{} focuses on different versions of the Lorenz63 system, using multiple different observation schemes and an order of magnitude more samples per time series than \Dysts{}.
This allows for a more detailed case study of forecasting chaotic systems, specifically using the Lorenz63 system, and enables the investigation of the effects of different settings under directly comparable conditions.
As 100 time series per system and observation scheme are available with randomly drawn initial conditions, system parameters, and noise values, an evaluation of methods can be made robust in a statistical sense. To obtain any kind of robustness in the evaluation using \Dysts{}, we have to aggregate over all of its different systems, which makes the results depend strongly and somewhat obscurely on the systems chosen for \Dysts{}.
\begin{figure}
    \begin{center}
        \begin{tikzpicture}
            \tikzset{every node/.style={font=\sffamily\small}}

            \def\yup{1}
            \def\ydown{0}
            \def\ymid{0.5}
            \node at (-2, \ymid) {Database};
            \draw (0,\yup) rectangle node[align=center] {\Dysts{}\\\cite{Gilpin21}}  (4, \ydown);
            \draw (5,\yup) rectangle node[align=center] {\DeebLorenz{}\\(new)}  (12, \ydown);

            \def\yup{-1}
            \draw (1,\ydown) -- (0.25,\yup);
            \draw (1.5,\ydown) -- (1.25,\yup);
            \draw (3,\ydown) -- (3.75,\yup);
            \draw (7,\ydown) -- (6,\yup);
            \draw (8.5,\ydown) -- (8.5,\yup);
            \draw (10,\ydown) -- (11,\yup);

            \def\ydown{-2.2}
            \def\ymid{-1.6}
            \node at (-2, \ymid) {System};
            \draw (0,\yup) rectangle node[] {\rotatebox{90}{\model{Aizawa}}} (0.5, \ydown);
            \draw (1,\yup) rectangle node[] {\rotatebox{90}{\model{Anish.}}} (1.5, \ydown);
            \node[anchor=south] at (2.5, \ymid) {133 systems};
            \fill[black] (2, \ymid) circle (2pt);
            \fill[black] (2.5, \ymid) circle (2pt);
            \fill[black] (3, \ymid) circle (2pt);
            \draw (3.5,\yup) rectangle node[] {\rotatebox{90}{\model{ZhouC.}}} (4, \ydown);
            \draw (5,\yup) rectangle node[align=center] {\model{Lorenz63-}\\\model{std}}  (7, \ydown);
            \draw (7.5,\yup) rectangle node[align=center] {\model{Lorenz63-}\\\model{random}}  (9.5, \ydown);
            \draw (10,\yup) rectangle node[align=center] {\model{Lorenz63-}\\\model{nonpar}}  (12, \ydown);

            \def\yup{-3.2}
            \draw (0.2,\ydown) -- (0.5,\yup);
            \draw (1.2,\ydown) -- (0.75,\yup);
            \draw (3.7,\ydown) -- (1.25,\yup);
            \draw (0.3,\ydown) -- (2.75,\yup);
            \draw (1.3,\ydown) -- (3.0,\yup);
            \draw (3.8,\ydown) -- (3.5,\yup);
            \foreach \i/\xa in {1/5, 2/7.5, 3/10}{
                \foreach \j/\xb in {1/5, 2/6.5+0.333333, 3/8+0.666666, 4/10.5}{
                    \draw (\xa+0.5-0.333333+0.333333*\j,\ydown) -- (\xb+0.25+0.25*\i,\yup);
                }
            }

            \def\ydown{-4.5}
            \def\ymid{-3.75}
            \node[align=center] at (-2, \ymid) {Observation\\Scheme};
            \draw (0,\yup) rectangle node[align=center] {noisefree\\\\$\stepsize$ const} (1.75, \ydown);
            \draw (2.25,\yup) rectangle node[align=center] {noisy\\(system)\\$\stepsize$ const} (4, \ydown);
            \draw (5,\yup) rectangle node[align=center] {noisefree\\\\$\stepsize$ const}  (6.5, \ydown);
            \draw (6.5+0.3333333,\yup) rectangle node[align=center] {noisy\\(measure)\\$\stepsize$ const}  (6.5+1.5+0.3333333, \ydown);
            \draw (6.5+1.5+0.666666,\yup) rectangle node[align=center] {noisefree\\\\$\stepsize$ rand}  (6.5+3+0.666666, \ydown);
            \draw (10.5,\yup) rectangle node[align=center] {noisy\\(measure)\\$\stepsize$ rand}  (12, \ydown);

            \def\yup{-5.5}
            \foreach \i/\xa in {0/0, 1/2.25}{
                \foreach \j/\xb in {0/0, 1/2.25}{
                    \draw (\xa+0.5+0.75*\j,\ydown) -- (\xb+0.5+0.75*\i,\yup);
                }
            }
            \foreach \i/\xa in {0/5, 1/6.5+0.333333, 2/8+0.666666, 3/10.5}{
                \foreach \j/\xb in {0/5, 1/8.75}{
                    \draw (\xa+0.5+0.5*\j,\ydown) -- (\xb+0.5+0.75*\i,\yup);
                }
            }

            \def\ydown{-6.5}
            \def\ymid{-6}
            \node[align=center] at (-2, \ymid) {Datasets};
            \draw (0,\yup) rectangle node[align=center] {validation\\$1$ repetition} (1.75, \ydown);
            \draw (2.25,\yup) rectangle node[align=center] {testing\\$1$ repetition} (4, \ydown);
            \draw (5,\yup) rectangle node[align=center] {validation\\$10$ repetitions}  (8.25, \ydown);
            \draw (8.75,\yup) rectangle node[align=center] {testing\\$100$ repetitions}  (12, \ydown);

            \def\ymid{-8.25}
            \node[align=center] at (-2, \ymid) {Time Series};

            \def\yup{-7.5}
            \foreach \x in {0.5*1.75, 2.25+0.5*1.75, 5+0.5*3.25, 8.75+0.5*3.25}{
                \draw (\x,\ydown) -- (\x,\yup);
            }
            \def\ydown{-8}
            \draw (0,\yup) rectangle node[align=center] {states from solution of ODE}  (4, \ydown);
            \draw (5,\yup) rectangle node[align=center] {states from solution of ODE}  (12, \ydown);

            \def\yup{-8}
            \def\ydown{-9}
            \draw (0,\yup) rectangle node[align=center] {training\\$n=10^3$}  (1.5, \ydown);
            \draw (1.5,\yup) rectangle node[align=center] {validation/testing\\$m=200$}  (4, \ydown);
            \draw (5,\yup) rectangle node[align=center] {training\\$n=10^4$}  (8.5, \ydown);
            \draw (8.5,\yup) rectangle node[align=center] {validation/testing\\$m=10^3$}   (12, \ydown);

            \def\ymid{-10}
            \node at (2,\ymid)[align=center] {$133 \cdot 2 \cdot (1 + 1)$ time series\\each of length $10^3 + 200$};
            \node at (8.5,\ymid)[align=center] {$3 \cdot 4 \cdot (10 + 100)$ time series\\each of length $10^4 + 10^3$};
        \end{tikzpicture}
    \end{center}
    \caption{Visual overview over the databases used in this study. We consider two different databases, \Dysts{} and \DeebLorenz{}. Each database consists of different (dynamical) systems (133 for \Dysts{}, 3 for \DeebLorenz{}). Systems may come with parameters, including the initial conditions. After random initialization of the parameters, ground truth data is generated by solving the differential equations numerically. From ground truth data, we generate synthetic 'observation' time series using different observations schemes (with or without noise; constant or random timestep $\stepsize$). In \Dysts{}, noise is treated as system noise, i.e., it influences the evolution of the state of the system; in \DeebLorenz{}, the noise is treated as measurement noise, i.e., the noisy data is observed but the system is evolved based on the ground truth. For each system and observation scheme, we create two different datasets: a validation dataset for hyperparameter tuning and a testing dataset for evaluation of performance. Each dataset consists of one (\Dysts{}) or several (\DeebLorenz{}) time series. Each time series is divided into two parts: the first part is designated as training data, while the second part serves different purposes depending on the dataset. For time series in the validation dataset, the second part is used as validation data, and for time series in the testing dataset, it is used as testing data. It is important to note that we use the terms 'testing' and 'validation' both to differentiate between datasets and to distinguish between data used for training and ground truth data to which the predictions are compared. In this setup, the training data from the validation dataset, the validation data from the validation dataset, the training data from the testing dataset, and the testing data from the testing dataset are all mutually exclusive.}
    \label{fig:data:overview}
\end{figure}
\subsubsection{DeebLorenz}
\begin{figure}
    \begin{center}%
        \begin{tabular}{ccc}
            \SymbolLorenzStandard{} -- \model{Lorenz63std\vphantom{p}}&
            \SymbolLorenzRandom{} -- \model{Lorenz63random\vphantom{p}}&
            \SymbolLorenzNonparam{} -- \model{Lorenz63nonpar}
            \\
            \includegraphics[width=0.3\textwidth]{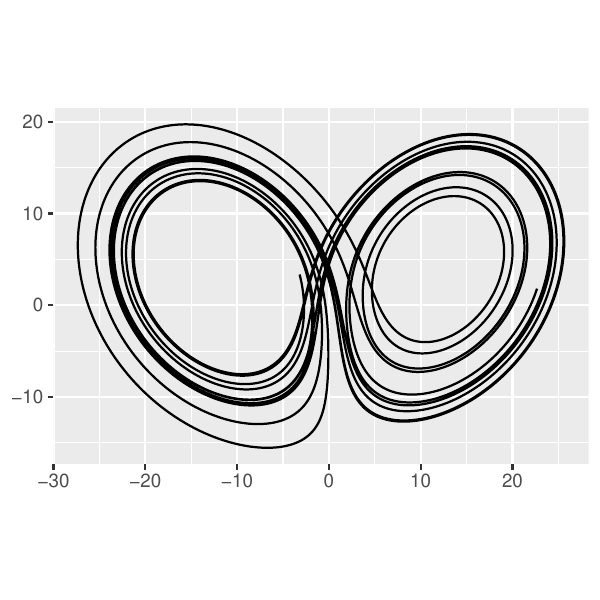}&
            \includegraphics[width=0.3\textwidth]{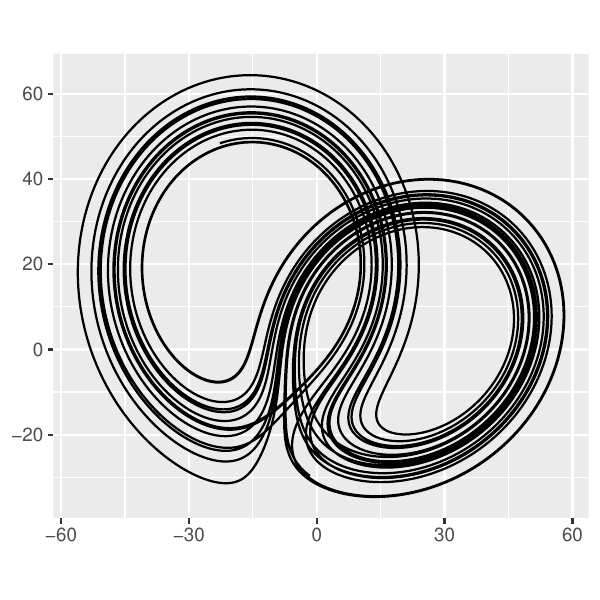}&
            \includegraphics[width=0.3\textwidth]{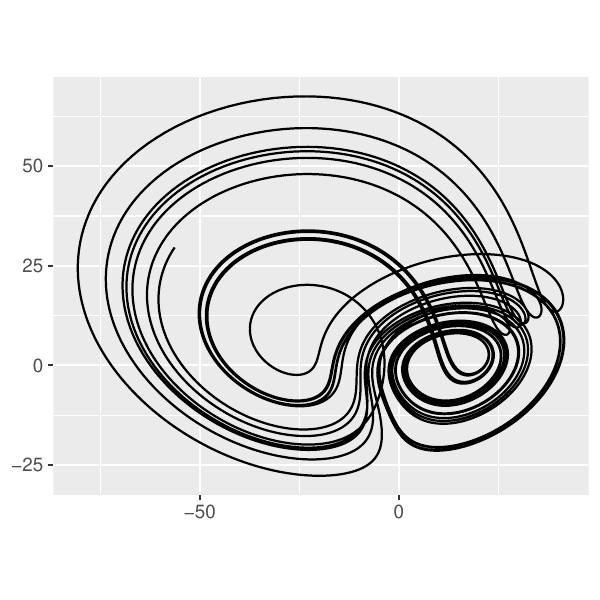}
            \\
            \includegraphics[width=0.3\textwidth]{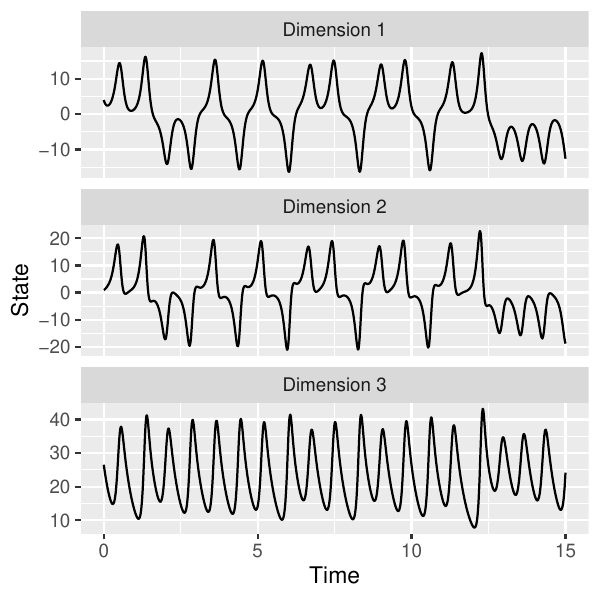}&
            \includegraphics[width=0.3\textwidth]{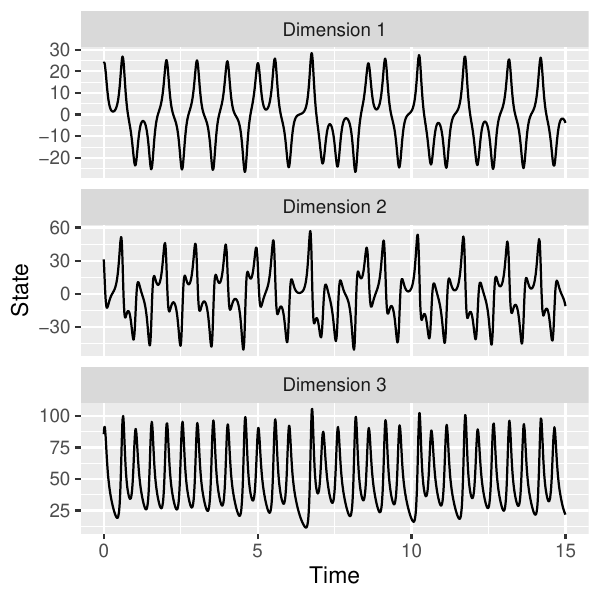}&
            \includegraphics[width=0.3\textwidth]{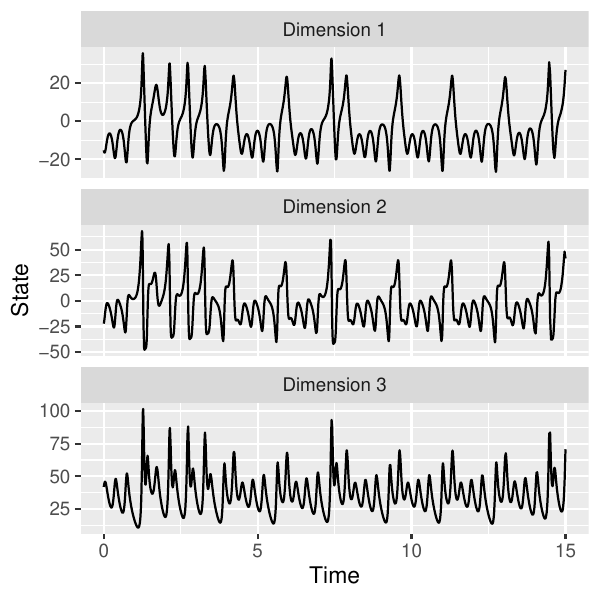}
        \end{tabular}
    \end{center}
    \caption{The three systems of the database \DeebLorenz{}. The plots show one example of a ground truth for each of the systems for an interval of 15 time units. The top row illustrates the attractor for each system in a 2D-projection of the state space. The bottom row shows individual time series plots for the three dimensions of the state space. All three systems are related to the Lorenz63 system \cite{Lorenz1963}, which can be described with a sparse polynomial of degree two as vector field $f$. The vector field $f$ has three parameters, which are set to the default values found in the literature for \model{Lorenz63std}. In \model{Lorenz63random}, for each time series, we sample those three parameters from a probability distribution centered around the default values. In \model{Lorenz63nonpar}, the parameter values have a functional dependence on the state that, for each time series, is drawn randomly from a Gaussian process.}\label{fig:lorenz:example}
\end{figure}

All three systems considered in \DeebLorenz{} are related to the Lorenz63 system \cite{Lorenz1963}, in which $f\colon\R^3\to\R^3$ is a sparse polynomial of degree 2 with 3 parameters. In \model{Lorenz63std}, the parameters are fixed to their default values used in the literature. In \model{Lorenz63random} the parameters are drawn randomly. In \model{Lorenz63nonpar} the parameters change with the state using a random nonparametric function drawn from a Gaussian process. See \cref{fig:lorenz:example} for examples of ground truth data from these systems. Notably, \model{Lorenz63nonpar} introduces a novel approach to randomly sampling nonpolynomial dynamical systems, all of which appear to exhibit chaotic behavior.

The data are generated under different observation schemes: In all cases, initial conditions $u(0)$ are drawn randomly from the attractor of the system. For times $t_1, \dots, t_n \in [0, T]$, $T\in\Rpp$, the state $u(t_i)$ is recorded either directly as $Y_i = u(t_i)$ or with random measurement error $\varepsilon_i$, i.e., $Y_i = u(t_i) + \varepsilon_i$. The observation times $t_i$ either have a constant timestep so that $t_i = iT/n$ or timesteps are drawn randomly from an exponential distribution $t_{i+1} - t_i \sim \ms{Exp}(\lambda)$, $\lambda = n/T$. All combinations of these timestep and measurement options make up the four different observation schemes for \DeebLorenz{}.

For each combination of the three systems and four observation schemes, the generation of the training data $(t_i, Y_i)_{i=1,\dots,n}$, $t_i\in[0, T]$, with $n = 10^4$ (in expectation if time steps are random), and of the (noisefree) testing data $(t_j, u(t_j))_{j= n+1,\dots,n+m}$, $t_j\in[T, T+S]$, $S\in\Rpp$, with $m = 10^3$, is repeated with different random seeds to create a validation dataset (10 repetitions) and a testing dataset (100 repetitions). Further details can be found in \cref{sec:app:data:deenlorenz}.
\subsubsection{Dysts}
The \Dysts{} \cite{Gilpin21} database consists of 133 chaotic systems with dimensions between 3 and 10, observed with constant timesteps, i.e., $t_i = iT/n$, and two different noise settings: one without noise $Y_i = u(t_{i})$, $\dot u(t) = f(u(t))$; the other with system noise, i.e., $Y_i = u(t_{i})$ with
\begin{equation}
    \dl u(t) = f(u(t))\dl t + \sigma\dl B(t)
    \eqcm
\end{equation}
where $B(t)$ is a standard Brownian motion and $\sigma\in\Rpp$ is the scale of the noise. For both observation schemes, data are available in a validation dataset and a testing dataset, which each consist of one repetition of generated training data $(t_i, Y_i)_{i=1,\dots,n}$ with $n=10^3$ and testing data $(t_j, u(t_j))_{j= n+1,\dots,n+m}$ with $m=200$ for each of the 133 systems.
\subsection{Prediction Task and Evaluation}
For all systems, observation schemes, datasets, and repetitions, we impose the following task: Given the observations of the past, $(t_i, Y_i)_{i=1,\dots,n}$ in the time interval $t_i\in[0, T]$ (training time), and the true current state $u(T)$, predict the future states $u(t_j)$ at given times $t_j\in[T, T+S]$ (prediction time) for $j = 1,\dots, m$.

Allowing the forecasting methods to know the true current state $u(T)$ --- even in the case of otherwise noisy observations --- eliminates the dependence of our simulation results on the (random) specific noise value of the final observation. This is crucial to increase the reliability of our comparison study, since prediction errors in chaotic systems typically grow exponentially with time. Additionally, it allows us to concentrate solely on the forecasting task (predicting future states) rather than on the data assimilation task (estimating the true value of a nosily observed state).

We apply and compare different statistical and machine learning methods on this task. The methods are described below in \cref{ssec:methods}.
For a given method, we first tune its hyperparameters on the validation dataset for each system individually: For each time series in the validation dataset, we train a given method and hyperparameter combination on the training data and calculate the error metrics described below on the validation data. We select the hyperparameter combination with the best (lowest) mean cumulative maximum error ($\cme$, see \cref{ssec:cummaxerr}), which creates the tuned method. Then, for each time series of the testing dataset, we judge performance the same way as in the validation dataset: We train the (tuned) method on the training data and create the predictions $\hat u(t_j)$ for the prediction time. The predictions are compared with the ground truth $u(t_j)$ using different metrics.

We calculate the following three metrics for comparing an estimate $\hat u(t)$ with the ground truth $u(t)$ in the prediction time $t\in[T, T+S]$:
\begin{itemize}
    \item
    The \textit{valid time} $\tvalid$, e.g., \cite{Ren2009, pathak2018hybrid}, for a given threshold $\kappa\in\Rp$ is defined as the duration such that the normalized error stays below $\kappa$, i.e.,
    \begin{equation}\label{eq:validtime}
        \tvalid(\hat u, u) := \inf\setByEle{
            t \in[T,T+S]
        }{
            \frac{
                \euclOf{\hat u(t) - u(t)}
            }{
                \sd(u)
            }
            > \kappa
        } - T
    \end{equation}
    where $\euclof{\cdot}$ denotes the Euclidean norm and $\sd(u)$ is the standard deviation of $u$ in the interval $[T, T+S]$, i.e.,
    \begin{equation}\label{eq:sd}
        \sd(u) := \sqrt{\frac{1}{S}\int_{T}^{T+S}\euclOf{u(t) - \mu(u)}^2 \dl t}\eqcm
        \qquad
        \mu(u) := \frac{1}{S}\int_{T}^{T+S}u(t) \dl t
        \eqfs
    \end{equation}
    If the threshold is not reached, we set $\tvalid = S$. In our evaluation, we set the threshold as $\kappa = 0.4$. Note that the normalization, i.e., division by $\sd(u)$, is sometimes done differently in the literature: \cite{Ren2009} do not integrate but normalize by $\euclof{u(t)}$, and \cite{pathak2018hybrid} do not remove the mean $\mu(u)$ in \eqref{eq:sd}. However, only the version described in \eqref{eq:validtime} and \eqref{eq:sd} is translation invariant, which is an advantage as described in \cref{ssec:cummaxerr} below.
    \item
    The \textit{symmetric mean absolute percent error} $\smape$, e.g., \cite{Gilpin23, godahewa2021monashtimeseriesforecasting}, is defined as
    \begin{equation}
        \smape(\hat u, u) := \frac{2\cdot 100}{S} \int_{T}^{T+S}\frac{\euclOf{\hat u(t) - u(t)}}{\euclOf{\hat u(t)} + \euclOf{{u(t)}}} \dl t
        \eqfs
    \end{equation}
    It is the standard mean absolute error, symmetrically normalized with respect to both the forecast and ground truth, and expressed as a percentage. A $\smape$ of 0 represents a perfect forecast and a forecast that diverges to infinity approaches a $\smape$ value of $200$, see \cref{sec:app:smape}.
    \item
    For this study, we design a new error metric, the cumulative maximum error $\cme$. It is defined as
    \begin{equation}\label{eq:cme}
        \cme(\hat u, u) := \frac1{S} \int_{T}^{T+S} \max_{s \in[T, t]} \min\brOf{1,\, \frac{\euclOf{\hat u(s) - u(s)}}{\sd(u)}} \dl t
    \end{equation}
    with $\sd$ as in \eqref{eq:sd}.
    It combines the advantages of $\tvalid$ and $\smape$, see \cref{ssec:cummaxerr} for details.
\end{itemize}
In practice, the canonical discrete-time versions of these metrics are applied, i.e., integrals are replaced by suitable sums.
\subsection{Estimation Methods}\label{ssec:methods}
We apply a wide range of different methods. Additionally, for the noisefree version of the \Dysts{} database, we also include the methods evaluated in \cite{Gilpin23} in our benchmark (pre-calculated predictions are not available for the noisy version). Names of these reference methods start with an underscore, e.g., \method{\_NBEAT}. In this section, we provide a high-level description of the methods specifically applied for this study. A more detailed description can be found in \cref{sec:app:methods}.

The methods can be separated into \textit{propagators},  \textit{recurrent architectures}, \textit{solution smoothers}, and other methods that do not belong to any of the first three groups.

We start with the last category, which also contains our baselines:
\begin{itemize}
    \item \methodNewh{constm}\method{ConstM} (\textit{constant mean}, also called \textit{climatology}): The prediction $\hat u(t)$ is the constant mean of the observed states, i.e., $\frac{1}n \sum_{i=1}^{n} Y_i$.
    \item \methodNewh{constl}\method{ConstL} (\textit{constant last}, also called \textit{persistence}): In our prediction task, the noisefree true state at time $T$ is always available to the methods, even in an observation scheme with noise. For this method, we set $\hat u(t) := u(T)$ for all $t$.
    \item \methodNewh{analog}\method{Analog}: For the method of the analog, we find the observed state $Y_{i_0}$ with the closest Euclidean distance to $u(T)$ and predict $\hat u(T + k\stepsize) = Y_{i_0 + k}$, or a linear interpolation thereof if timesteps are not constant.
    \item \methodNewh{node}\method{Node1}, \method{Node32}: For the neural ODE \cite{chen18}, a neural network $f_\theta$ is fitted to the data such that solving the ODE $\dot{\hat u}(t) = f_\theta(\hat u(t))$ minimizes the error $\euclof{\hat u(t_i)-Y_i}$.
    \item \methodNewh{trafo}\method{Trafo}, \method{TrafoT}: A transformer network \cite{Vaswani2017} is applied that uses attention layers to compute the next state from a sequence of previous states. The suffix \texttt{T} in \method{TrafoT} indicates that the timestep $t_{i+1} - t_{i}$ is appended to the observation vector $Y_i$, which is relevant if timesteps are non-constant.
\end{itemize}
The methods \method{ConstM}, \method{ConstL}, and \method{Analog} can be treated as trivial baselines as no fitting or learning is required.

Next, we consider the group of \textit{propagators}: For these methods, the basic idea is to find an estimate for the so-called propagator map
\begin{equation}\label{eq:propgator}
    \mc P_{\stepsize}\colon u(t) \mapsto u(t + \stepsize) \qquad\text{or equivalently}\qquad \mc P_{\stepsize}\colon \R^d\to\R^d,\, x \mapsto u_x(\stepsize)
    \eqcm
\end{equation}
where $u_x$ solves the initial value problem $\dot u_x(t) = f(u_x(t))$, $u_x(0) = x$.
This task can be formulated as a regression problem on the data $(Y_i, Y_{i+1})_{i=1,\dots,n-1}$ assuming constant timesteps $t_{i+1} - t_i = \stepsize$. After the propagator is estimated, recursive application yields the predictions
\begin{equation}\label{eq:propa:predict}
	\hat u(T + k\stepsize) = \underbrace{\mc P_{\stepsize}(\dots\mc P_{\stepsize}(}_{k \text{ times}}u(T)) \dots )
	\eqfs
\end{equation}
All methods listed next are variations on this basic idea. Several of them come in different flavors marked by the suffix \method{S}, \method{D}, and/or \method{T} in their name: \method{S} indicates the next \textbf{s}tate as a target, see \eqref{eq:propgator}, whereas \method{D} means that the \textbf{d}ifference quotient
\begin{equation}
    x \mapsto \frac{u_x(\stepsize) - x}{\stepsize}
\end{equation}
is estimated and the prediction formula is adapted suitably. The latter approach is analogous to residual learning commonly used in the deep learning literature. If \method{T} is part of the suffix, the timestep $t_{i+1} - t_{i}$ is available for the method as an input for the prediction of $u(t_{i+1})$. This is not useful for constant timesteps, but crucial for random timesteps. The symbol \method{*} below indicates that a method is available in different variants denoted with the suffixes \method{T}, \method{S}, \method{D}, or a combination thereof.
\begin{itemize}
    \item \methodNewh{pg}\method{PgGp*}, \method{PgNet*}, \method{PgLl*}: Fit $\mc P_{\stepsize}$ with a Gaussian process \cite{Rasmussen2006Gaussian}, a feed-forward neural network \cite{rumelhart1986learning}, or a local linear estimator \cite{Ruppert1994}, respectively.
    \item \methodNewh{lin}\method{Lin*}: The propagator $\mc P_{\stepsize}$ is estimated via ridge regression \citep[chapter 3.4.1]{hastie2009elements} using polynomial features. For a prediction of $u(t_{i+1})$, state values at times  $t_{i-sk}$, $k=0, \dots, K$ for a fixed $s\in\N$ are available as inputs to the polynomial features. This method is also called a \textit{nonlinear vector autoregressive model} \cite{Gauthier2021}.
    \item \methodNewh{linpo}\method{LinPo4}, \method{LinPo6}, \method{LinPo}{LinPo4T}, \method{LinPo6T}: Tuning-free versions of \method{LinD} and \method{LinDT} with linear regression (ridge penalty parameter equal to $0$), without past states as predictors ($K=0$), and with the polynomial degree fixed to $4$ and $6$, respectively.
    \item \methodNewh{rafe}\method{RaFe*}: For \textit{Random Features} \cite{Rahimi2008}, features are created from the predictors (the current state variables) by applying a once randomly created and untrained neural network. Then ridge regression \citep[chapter 3.4.1]{hastie2009elements} is applied to estimate the mapping from features to targets (the next state).
\end{itemize}

The \textit{recurrent architectures} are similar to the propagators, as they essentially also estimate the propagator map $P_{\stepsize}$. But instead of viewing the task purely as a standard regression problem, where the ordering of the data $(Y_i, Y_{i+1})_{i=1,\dots,n-1}$ has no influence of the result, these methods keep an internal state that is updated while processing the data sequentially, ordered by time. The internal state influences the prediction of the next system state.
\begin{itemize}
    \item \methodNewh{esn}\method{Esn*}: The Echo State Network (ESN) \cite{jaeger2001echo} is a specific type of \textit{reservoir computer}. It is similar to \textit{Random Features}, but features of the previous timestep are also inputs for the feature calculation of the current timestep. This structure creates an internal state, allowing the ESN to retain memory of the information from previous inputs.
    \item \methodNewh{rnn}\method{Rnn}, \method{RnnT}: The Recurrent Neural Network (RNN) \cite{rnn} has cyclic connections of neurons, allowing information from previous inputs to persist for future predictions.
    \item \methodNewh{lstm}\method{Lstm}, \method{LstmT}: The Long Short--Term Memory network (LSTM) \cite{lstm} is a variation of the standard RNN designed to capture long-term dependencies in sequential data by using gating mechanisms to regulate the flow of information and prevent issues like vanishing gradients.
    \item \methodNewh{gru}\method{Gru}, \method{GruT}: A Gated Recurrent Unit (GRU) \cite{gru} is a simplified variant of LSTM that uses gating mechanisms to control the flow of information, but with fewer parameters, making it more efficient while still addressing the vanishing gradient problem in sequential data.
\end{itemize}

The final category of methods are the \textit{solution smoothers} \cite[chapter 8]{Ramsay_2017}. In certain settings (that are slightly different from our experiments), these methods are shown to be optimal, see \cite[section 4]{schoetz2024upper} and \cite[section 4]{schoetz2024lower}. Solution smoothers work in three steps: First, estimate $u(t)$ in the observation time $[0, T]$ with a regression estimator $\tilde u(t)$. Then, view the data $(\tilde u(t), \dot{\tilde u}(t))_{t\in[0,T]}$ as observations of $(x, f(x))$ and use a second regression estimator to obtain an estimate $\hat f$. Lastly, solve the ODE $\dot{\hat u}(t) = \hat f(\hat u(t))$ on the time interval $[T, T+S]$ with initial conditions $\hat u(T) = u(T)$.
\begin{itemize}
    \item \methodNewh{pwnn}\method{PwNn}: The observed solution is interpolated with a piece-wise linear function $\tilde u$, from which $\hat f$ is generated by a nearest neighbor interpolation.
    \item \methodNewh{spnn}\method{SpNn}: As \method{PwNn}, but $\tilde u$ is cubic spline interpolation \cite{forsythe1977computer}.
    \item \methodNewh{llnn}\method{LlNn}: As \method{PwNn}, but $\tilde u$ is local linear regression \cite{Fan1993}.
    \item \methodNewh{sppo}\method{SpPo}:  As \method{SpNn}, but $\hat f$ is ridge regression \citep[chapter 3.4.1]{hastie2009elements} with polynomial features.
    \item \methodNewh{sppon}\method{SpPo2}, \method{SpPo4}: Tuning-free version of \method{SpPo} with linear regression (ridge penalty parameter equal to $0$) and with fixed polynomial degree 2 and 4, respectively.
    \item \methodNewh{spgp}\method{SpGp}: As \method{SpNn}, but $\hat f$ is Gaussian process regression \cite{Rasmussen2006Gaussian}.
    \item \methodNewh{gpgp}\method{GpGp}: As \method{SpGp}, but $\tilde u$ is also Gaussian process regression. The method is similar to the one proposed in \cite{heinonen18}.
    \item \methodNewh{sindy}\method{SINDy}, \method{SINDyN}: As \method{SpPo} without $L_2$-penalty, but sparsity is enforced for the polynomial $\hat f$ via thresholding \cite{Brunton2016}. As for all other methods, data normalization is applied before the actual estimation method for \method{SINDyN}. Only for \method{SINDy} it is turned off to not destroy potential sparsity in the original data. The name \method{SINDy} is short for \textit{Sparse Identification of Nonlinear Dynamics}.
\end{itemize}

In the following, we will additionally classify the methods according to their central computational concept:
\begin{itemize}
	\item \ClassDirect: Predictions for \method{ConstL}, \method{ConstM}, \method{Analog} are computed directly from the observations.
	\item \ClassGrDesc: \method{Node*}, \method{Trafo*}, \method{PgNet*}, \method{Rnn*}, \method{Lstm*}, \method{Gru*} are trained using a variant of gradient descent.
	\item \ClassFitPro: \method{Lin*}, \method{LinPo2}, \method{LinPo4}, \method{RaFe*}, \method{Esn*}, \method{PgGp*}, \method{PgLl*} fit the propagator map by solving a system of linear equations.
	\item \ClassFitSol: \method{PwNn}, \method{SpNn}, \method{LlNn}, \method{SpPo}, \method{SpPo2}, \method{SpPo4}, \method{SpGp}, \method{GpGp}, \method{SINDy}, \method{SINDyN} fit the solution and/or the vector field $f$ by solving a system of linear equations.
\end{itemize}
\subsection{Cumulative Maximum Error -- CME}\label{ssec:cummaxerr}
Although different error metrics are calculated to evaluate the results of this simulation study, we focus on the $\cme$ \eqref{eq:cme}, which we newly introduce for this study. If time is discrete, \eqref{eq:cme} becomes
\begin{equation}\label{eq:cme:disc}
    \cme_m(\hat u, u)
    :=
    \frac1{m} \sum_{j=1}^{m} \max_{k =1,\dots, j}\min\brOf{1,\, \frac{\euclOf{\hat u(t_k) - u(t_k)}}{\sd_m(u)}}
    \eqfs
\end{equation}
where
\begin{equation}
    \sd_m(u) := \sqrt{\frac1m\sum_{j=1}^{m}\euclOf{u(t_j) - \mu_m(u)}^2}\eqcm
    \qquad
    \mu_m(u) := \frac1m\sum_{j=1}^{m} u(t_j)
    \eqfs
\end{equation}
Similar to the $\smape$, errors are integrated over time, and similar to $\tvalid$, the time and error of the worst prediction so far are essential for the value of the overall error metric.
The $\cme$ has many desirable properties:
\begin{enumerate}[label=(\roman*)]
    \item
        It is translation and scale invariant, in contrast to $\smape$ and the version of the valid time used in \cite{Ren2009, pathak2018hybrid}, but similar to our definition of $\tvalid$.
        This is desirable as every ODE system can be modified --- without adding complexity --- to create any translation and scaling of its solutions. Thus, the evaluation of an estimator $\hat u$ should not depend on the location and scale of the system.
    \item
        We focus on a task that involves evaluating the precise state within a near-future time interval, as in weather forecasting. In contrast, one could also consider a task where the long-term behavior of the prediction should resemble the ground truth, without emphasizing the error at any single point in time, as in climate modeling. For the former task, predictive power deteriorates quickly over time due to the chaotic nature of the systems considered. Therefore, if the prediction interval $[T, T+S]$ is long, we are primarily interested in the early times when accurate prediction is possible. Accurate predictions at the beginning of the interval should not be discounted by divergence at the end, where accurate predictions are not possible anyway. This requirement is captured by $\cme$ and, to some extent, by $\tvalid$, but not by $\smape$.
    \item
        The value of the $\cme$ is always between 0 and 1, where 0 is only achieved by perfect prediction. In contrast, the scale of $\tvalid$ depends strongly on the system's time scale and the best value $\tvalid = S$ can be achieved for imperfect predictions.
    \item
        Dealing with missing values when calculating the $\cme$ is canonical: If $\hat u(t)$ is not available for some $t$, we set the value of the minimum in \eqref{eq:cme} to $1$. Consistently, if all of the prediction is missing, we set the $\cme$ to $1$.
    \item
        The $\cme$ is parameter-free in contrast to $\tvalid$, where the threshold $\kappa$ has to be specified.
    \item
        If time is discrete, $\tvalid$ can only attain a finite number of values. In contrast, for all values in $[0,1]$ there is a prediction $\hat u$ such that $\cme_m(\hat u, u)$ attains this value. Furthermore, $\hat u_1$ and $\hat u_2$ may have the same $\tvalid$, but $\euclof{\hat u_1(t) - u(t)} < \euclof{\hat u_2(t) - u(t)}$ for all $t\in [T, T + \tvalid)$. If time is discrete, we can even have $\euclof{\hat u_1(t) - u(t)} < \euclof{\hat u_2(t) - u(t)}$ for all $t \in [T, T+S]$ while $\tvalid(\hat u_1, u) = \tvalid(\hat u_2, u)$. This cannot happen with $\cme$: If $\euclof{\hat u_1(t) - u(t)} < \euclof{\hat u_2(t) - u(t)}$ for all $t\in [T, t^*]$ for some $t^*>T$ and $\euclof{\hat u_1(t) - u(t)} \leq \euclof{\hat u_2(t) - u(t)}$ for all $t\in [t^*, T+S]$, we have  $\cme(\hat u_1, u) < \cme(\hat u_2, u)$.
\end{enumerate}
Let us also describe potential drawbacks of the $\cme$.
\begin{enumerate}[label=(\roman*)]
    \item
        The $\cme$ is not symmetric, $\cme(\hat u, u) \neq \cme(u, \hat u)$ in general, in contrast to $\smape$. But in our use case, the relationship between its two arguments $\hat u$ and $u$ is also not symmetric. Thus, there does not seem to be a benefit to symmetry.
    \item
        If $u$ is constant, then $\sd(u) = 0$ and \eqref{eq:cme} is not valid. In this case, a reasonable definition is to set $\cme$ to $1$ for all $\hat u$ except the perfect prediction $\hat u = u$ in which case $0$ is the appropriate value. Note that $\tvalid$ (and to some extent $\smape$) also suffers from this \textit{division by zero}-problem.
    \item
        If a prediction gets far away from $u(t)$ but recovers at a later point in time, the recovery is not accredited in $\cme$. As future states in our dynamical systems are independent of the past given the present (Markov property), a time point with a good forecast following a previous time point with a poor forecast should be considered spurious and attributed to chance rather than skill. This is especially true in chaotic systems, where the dependence of the future on the present is particularly strong. But note that the same distance $\delta = \euclof{\hat u(t_0) - u(t_0)}$ can translate to different future predictability at times $t > t_0$ for different states  $u(t_0) = x_1$ or $u(t_0) = x_2$, $x_1 \neq x_2$. I.e., we can have
        \begin{equation}
            \euclOf{u_{x_1}(t) - u_{x_1\pr}(t)} \neq \euclOf{u_{x_2}(t) - u_{x_2\pr}(t)}
        \end{equation}
        for $t > t_0$ even if $\euclof{x_1 - x_1\pr} = \euclof{x_2 - x_2\pr}$.
    \item
    	On one hand, the value $1$ in the $\min$-term of \eqref{eq:cme} could be replaced by a threshold parameter $\kappa$ as in $\tvalid$, removing the advantage of being parameter-free for the $\cme$. On the other hand, $\kappa = 1$ is a natural choice for the $\cme$ (in contrast to $\tvalid$), as the trivial baseline \method{ConstM}, $\hat u(t) = \frac1n\sum_{i=1}^n Y_i$, achieves $\euclOf{\hat u(s) - u(s)} / \sd(u) = 1$ on average. Furthermore, for every constant prediction there is a finite time point $s_0$ such that $\euclOf{\hat u(s) - u(s)} / \sd(u) \geq \kappa$ for $\kappa = 1$, and $\kappa = 1$ is the largest value with this property.
    \item
        The value of the $\cme$ depends on the testing duration $S$ (similar to $\smape$, but in contrast to $\tvalid$). This could be mitigated by setting $S = \infty$ and multiplying the integrand in \eqref{eq:cme} by a weighting function $w(t-T)$, e.g., $w(\tau) = \exp(-\alpha \tau)$, but this introduces a new parameter choice $\alpha\in\Rpp$. To make the weighting choice-free and time-scale adaptive, $\alpha$ needs to be set depending on $u$, e.g., to the inverse correlation time.
\end{enumerate}
\subsection{Hyperparameter Tuning}
Each pair of a dynamical system and an observation scheme comes as a \textit{validation} dataset and a \textit{testing} dataset. For hyperparameter tuning only the \textit{validation} dataset is used. Each such dataset consists of $1$ (\Dysts) or $10$ (\DeebLorenz) repetitions/time series. Each repetition consists of training and validation data.

For a given method $\hat u_{\mo a}$ with hyperparameters $\mo a$, we train it, predict, and calculate the $\cme$ for different $\mo a$. Let us denote the average $\cme$ of all repetitions by $\cme(\mo a)$ and the best hyperparameters among the tested ones as $\mo a^* = \argmin_{\mo a} \cme(\mo a)$. Then $\hat u_{\mo a^*}$ is applied to the testing dataset for the final results presented in \cref{sec:results}. See \cref{sec:app:methods} for a description of the tuned hyperparameters for each method.

To decide which hyperparameters $\mo a$ to evaluate, we follow a local grid search procedure: For a given estimation method, decide on which parameters should be tuned  $\mo a = (a_1, \dots, a_\ell)$. Each parameter $a_j$ has a domain of possible values $\mc A_j$. The domain can be \textit{categorical} or \textit{scalar}.
For categorical parameters, decide whether they are \textit{persistent} (always evaluate all options) or \textit{yielding} (only evaluate best).
For scalar parameters, decide on a \textit{linear} or \textit{exponential scale} and a \textit{stepsize} $s_j \in \Rpp$. Decide on finite sets of initial hyperparameter values $A_{0,j}$. Now, for step $k\in\N_0$ in the optimization procedure, evaluate all elements of the grid of hyperparameters $(a_1, \dots, a_\ell) \in A_{k,1} \times \dots \times A_{k,\ell}$ that have not been evaluated before. If there are none, stop the search. Denote the best hyperparameter combination evaluated so far as $(a_{k,1}^*, \dots, a_{k,\ell}^*)$.  Generate $A_{k+1,j}$ from $a_{k,j}^*$ as follows:
If the $j$-th variable is categorical, set $A_{k+1,j} = A_{0,j}$ for persistent parameters and $A_{k+1,j} = \{a_{k,j}^*\}$ for yielding ones.
If the $j$-th variable is scalar, set $A_{k+1,j} = \{a_{k,j}^* - s_j, a_{k,j}^*, a_{k,j}^* + s_j\} \cap \mc A_j$ if it has a linear scale and $A_{k+1,j} = \{a_{k,j}^* / s_j, a_{k,j}^*, s_j a_{k,j}^*\} \cap \mc A_j$ if the scale is exponential.

The local grid search finds locally optimal hyperparameter combinations with a reasonable amount of evaluations assuming the number of tuned hyperparameters $\ell$ is small enough. In this study, the algorithm is applied to $\ell \leq 4$ hyperparameters to limit the computational costs. We tune more parameters in case of methods with low computational demand and limit to one categorical variable (no adaptive search) for the most expensive methods. See the main results \cref{fig:results:dysts,fig:results:lorenz:const,fig:results:lorenz:rand} in the next section (column \textsf{Tune}) for the computational cost of hyperparameter tuning. By adjusting the number of evaluated hyperparameters, we try to make the comparison fairer regarding total computational costs. The resulting total compute time can still be rather different between methods. This is partially due to the adaptive nature of the local grid search algorithm (i.e., the total number of predictions cannot be known beforehand) and partially due to the lack of meaningful tunable hyperparameters for some methods (in particular for very simple methods that typically have low computational costs). Furthermore, perfect fairness does not seem achievable, as the choice of hyperparameters to be tuned, the choice of initial hyperparameters, and the design of the scales all influence the result.

In contrast to our approach, \cite{Gilpin23} tune one hyperparameter for each method, independent of computational cost. The parameter selected is always related to the number of past time series elements that are inputs for the prediction of the next step. Note that in theory only the current state is required to predict a future state (Markov property) if there is no measurement noise, and as long as one has access to all system variables.

%% file: sec_results.tex
\section{Results and 10 Key Insights}\label{sec:results}
\begin{figure}
    \begin{center}
    	\vspace{-1cm}
    	\begin{tikzpicture}[
    			every node/.style={inner sep=0,outer sep=0},
    			zoomLine/.style={darkgray, thick}
    		]
    		\node (S0) at (0,0) {\SymbolLorenzStandard{} -- \textsc{Lorenz63std\vphantom{p}}};
    		\node[below=1mm of S0] (S1) {\includegraphics[width=\textwidth]{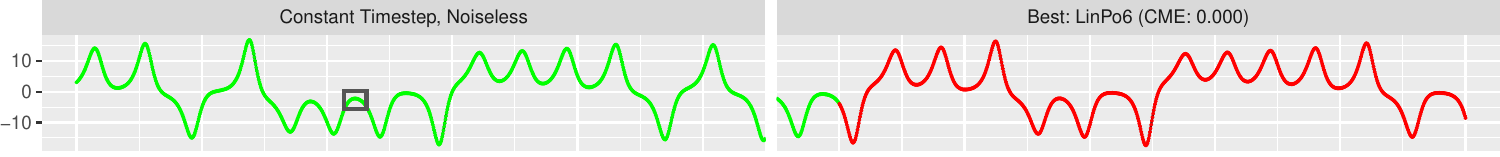}};
    		\node[below=0mm of S1] (S2) {\includegraphics[width=\textwidth]{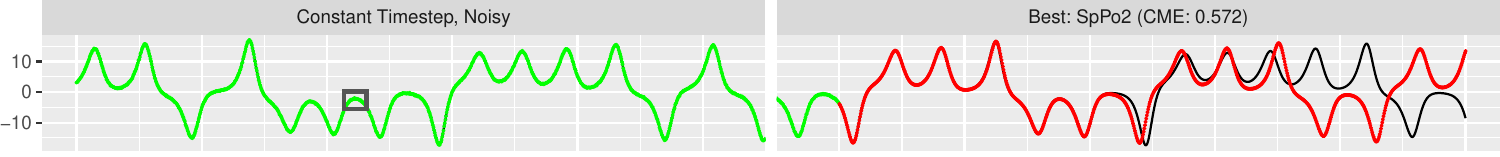}};
    		\node[below=0mm of S2] (S3) {\includegraphics[width=\textwidth]{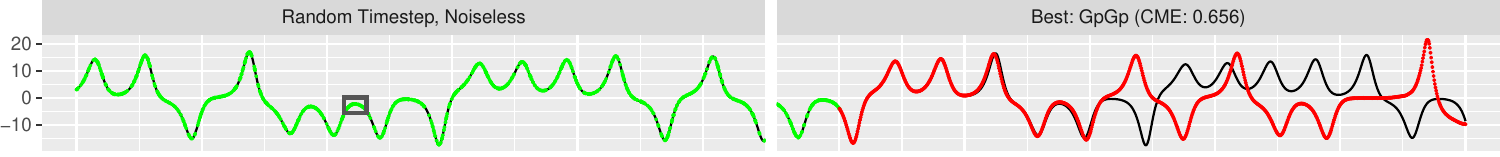}};
    		\node[below=0mm of S3] (S4) {\includegraphics[width=\textwidth]{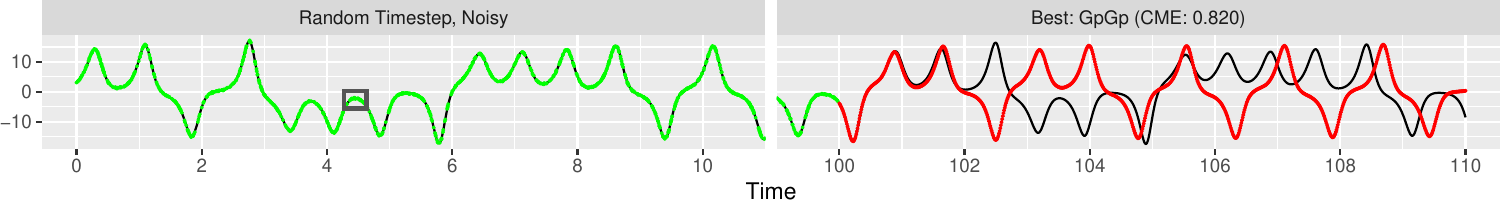}};
    		\node[below=3mm of S4] (R0) {\SymbolLorenzRandom{} -- \textsc{Lorenz63random\vphantom{p}}};
    		\node[below=1mm of R0] (R1) {\includegraphics[width=\textwidth]{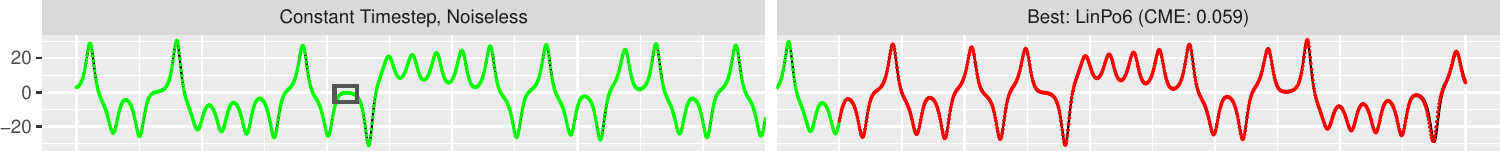}};
    		\node[below=0mm of R1] (R2) {\includegraphics[width=\textwidth]{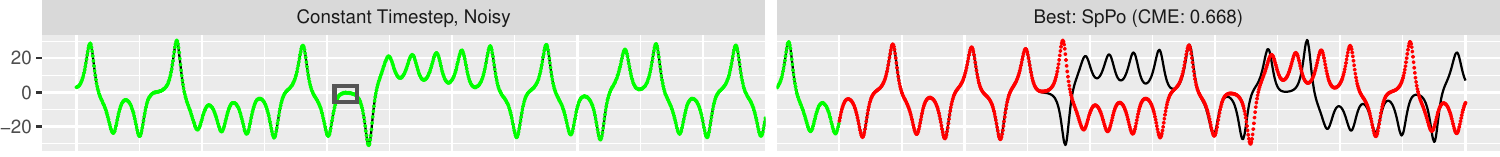}};
    		\node[below=0mm of R2] (R3) {\includegraphics[width=\textwidth]{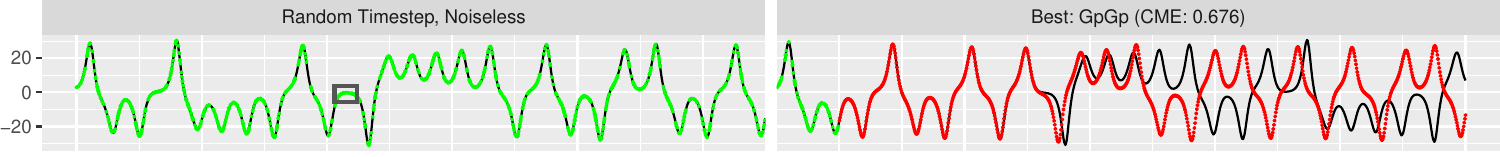}};
    		\node[below=0mm of R3] (R4) {\includegraphics[width=\textwidth]{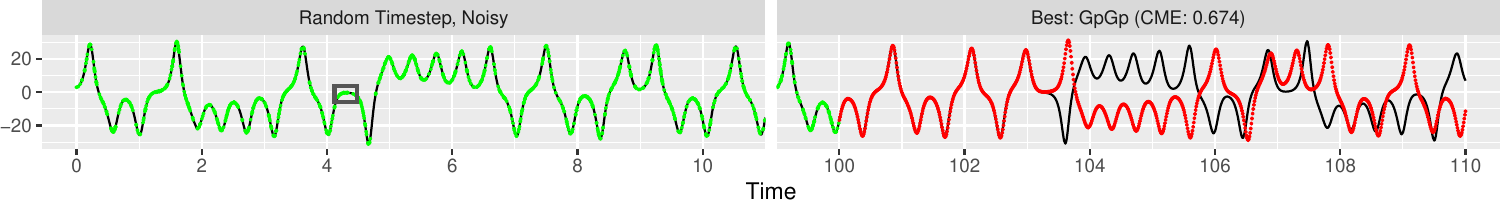}};
    		\node[below=3mm of R4] (N0) {\SymbolLorenzNonparam{} -- \textsc{Lorenz63nonpar}};
    		\node[below=1mm of N0] (N1) {\includegraphics[width=\textwidth]{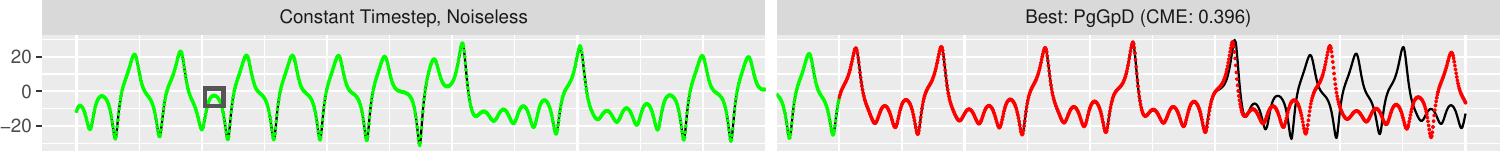}};
    		\node[below=0mm of N1] (N2) {\includegraphics[width=\textwidth]{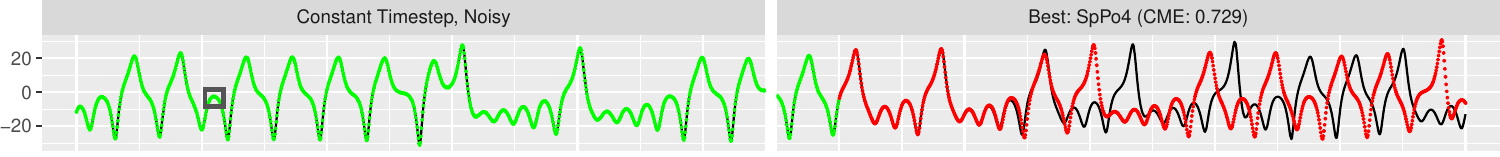}};
    		\node[below=0mm of N2] (N3) {\includegraphics[width=\textwidth]{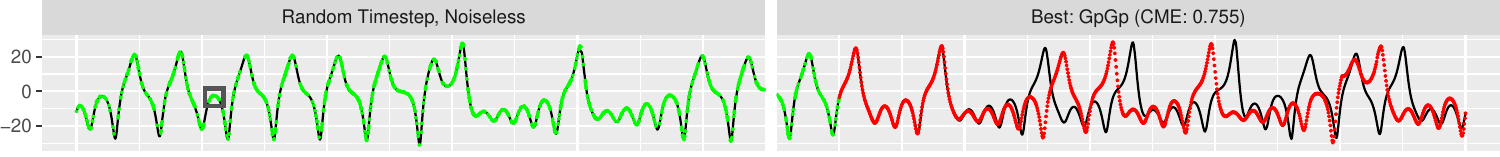}};
    		\node[below=0mm of N3] (N4) {\includegraphics[width=\textwidth]{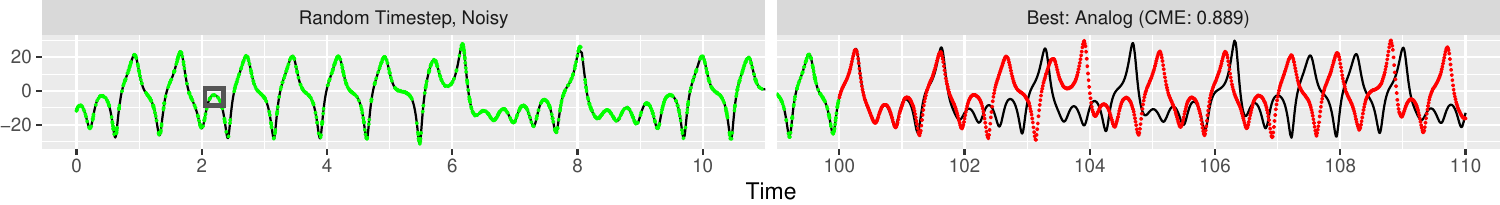}};
    		\node[anchor=north east] (S1z) at ([xshift=7.65cm,yshift=-0.33cm]S1.north west) {\includegraphics[width=2.4cm]{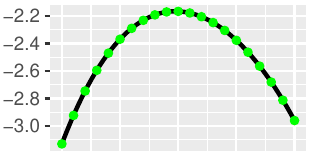}};
    		\draw[zoomLine,dashed] (S1z.south west) rectangle (S1z.north east);
    		\draw[zoomLine] (S1z.north west) -- ([xshift=34.4mm, yshift=-9.1mm]S1.north west);
    		\draw[zoomLine] (S1z.south west) -- ([xshift=34.4mm, yshift=-10.9mm]S1.north west);
    		\node[anchor=north east] (S2z) at ([xshift=7.65cm,yshift=-0.33cm]S2.north west) {\includegraphics[width=2.4cm]{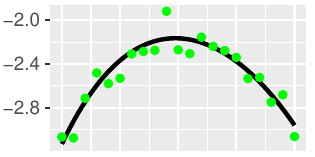}};
    		\draw[zoomLine,dashed] (S2z.south west) rectangle (S2z.north east);
    		\draw[zoomLine] (S2z.north west) -- ([xshift=34.4mm, yshift=-9.1mm]S2.north west);
    		\draw[zoomLine] (S2z.south west) -- ([xshift=34.4mm, yshift=-10.9mm]S2.north west);
    		\node[anchor=north east] (S3z) at([xshift=7.65cm,yshift=-0.33cm]S3.north west) {\includegraphics[width=2.4cm]{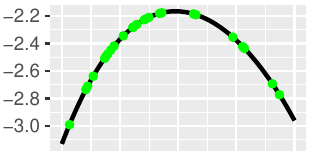}};
    		\draw[zoomLine,dashed] (S3z.south west) rectangle (S3z.north east);
    		\draw[zoomLine] (S3z.north west) -- ([xshift=34.4mm, yshift=-9.6mm]S3.north west);
    		\draw[zoomLine] (S3z.south west) -- ([xshift=34.4mm, yshift=-11.4mm]S3.north west);
    		\node[anchor=north east] (S4z) at ([xshift=7.65cm,yshift=-0.33cm]S4.north west) {\includegraphics[width=2.4cm]{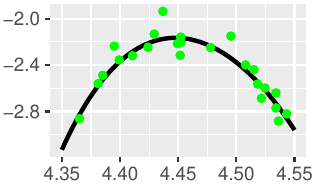}};
    		\draw[zoomLine,dashed] (S4z.south west) rectangle (S4z.north east);
    		\draw[zoomLine] (S4z.north west) -- ([xshift=34.4mm, yshift=-9.1mm]S4.north west);
    		\draw[zoomLine] (S4z.south west) -- ([xshift=34.4mm, yshift=-10.9mm]S4.north west);
    		\node[anchor=north east] (R1z) at ([xshift=7.65cm,yshift=-0.33cm]R1.north west) {\includegraphics[width=2.4cm]{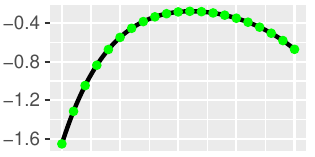}};
    		\draw[zoomLine,dashed] (R1z.south west) rectangle (R1z.north east);
    		\draw[zoomLine] (R1z.north west) -- ([xshift=33.4mm, yshift=-8.6mm]R1.north west);
    		\draw[zoomLine] (R1z.south west) -- ([xshift=33.4mm, yshift=-10.3mm]R1.north west);
    		\node[anchor=north east] (R2z) at ([xshift=7.65cm,yshift=-0.33cm]R2.north west) {\includegraphics[width=2.4cm]{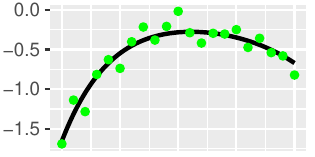}};
    		\draw[zoomLine,dashed] (R2z.south west) rectangle (R2z.north east);
    		\draw[zoomLine] (R2z.north west) -- ([xshift=33.4mm, yshift=-8.6mm]R2.north west);
    		\draw[zoomLine] (R2z.south west) -- ([xshift=33.4mm, yshift=-10.3mm]R2.north west);
    		\node[anchor=north east] (R3z) at ([xshift=7.65cm,yshift=-0.33cm]R3.north west) {\includegraphics[width=2.4cm]{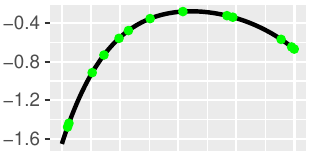}};
    		\draw[zoomLine,dashed] (R3z.south west) rectangle (R3z.north east);
    		\draw[zoomLine] (R3z.north west) -- ([xshift=33.4mm, yshift=-8.6mm]R3.north west);
    		\draw[zoomLine] (R3z.south west) -- ([xshift=33.4mm, yshift=-10.2mm]R3.north west);
    		\node[anchor=north east] (R4z) at ([xshift=7.65cm,yshift=-0.33cm]R4.north west) {\includegraphics[width=2.4cm]{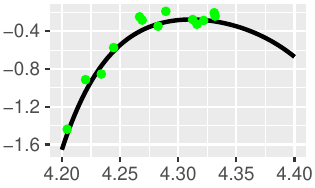}};
    		\draw[zoomLine,dashed] (R4z.south west) rectangle (R4z.north east);
    		\draw[zoomLine] (R4z.north west) -- ([xshift=33.4mm, yshift=-8.6mm]R4.north west);
    		\draw[zoomLine] (R4z.south west) -- ([xshift=33.4mm, yshift=-10.3mm]R4.north west);
    		\node[anchor=north east] (N1z) at ([xshift=7.65cm,yshift=-0.33cm]N1.north west) {\includegraphics[width=2.4cm]{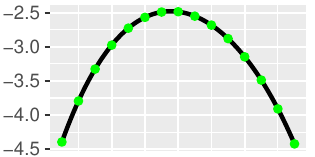}};
    		\draw[zoomLine,dashed] (N1z.south west) rectangle (N1z.north east);
    		\draw[zoomLine] (N1z.north west) -- ([xshift=20.4mm, yshift=-8.9mm]N1.north west);
    		\draw[zoomLine] (N1z.south west) -- ([xshift=20.4mm, yshift=-10.6mm]N1.north west);
    		\node[anchor=north east] (N2z) at ([xshift=7.65cm,yshift=-0.33cm]N2.north west) {\includegraphics[width=2.4cm]{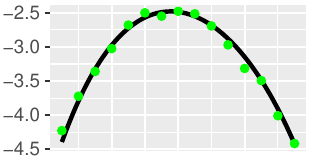}};
    		\draw[zoomLine,dashed] (N2z.south west) rectangle (N2z.north east);
    		\draw[zoomLine] (N2z.north west) -- ([xshift=20.4mm, yshift=-8.9mm]N2.north west);
    		\draw[zoomLine] (N2z.south west) -- ([xshift=20.4mm, yshift=-10.6mm]N2.north west);
    		\node[anchor=north east] (N3z) at ([xshift=7.65cm,yshift=-0.33cm]N3.north west) {\includegraphics[width=2.4cm]{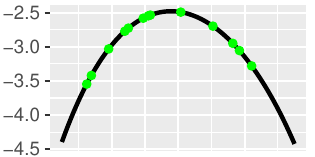}};
    		\draw[zoomLine,dashed] (N3z.south west) rectangle (N3z.north east);
    		\draw[zoomLine] (N3z.north west) -- ([xshift=20.4mm, yshift=-8.9mm]N3.north west);
    		\draw[zoomLine] (N3z.south west) -- ([xshift=20.4mm, yshift=-10.6mm]N3.north west);
    		\node[anchor=north east] (N4z) at ([xshift=7.65cm,yshift=-0.33cm]N4.north west) {\includegraphics[width=2.4cm]{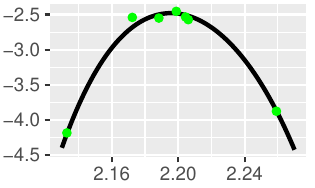}};
    		\draw[zoomLine,dashed] (N4z.south west) rectangle (N4z.north east);
    		\draw[zoomLine] (N4z.north west) -- ([xshift=20.4mm, yshift=-8.9mm]N4.north west);
    		\draw[zoomLine] (N4z.south west) -- ([xshift=20.4mm, yshift=-10.6mm]N4.north west);
    	\end{tikzpicture}
    \end{center}
    \caption{The plots show one example (out of 100) of train and test data from the testing dataset of \DeebLorenz{}. Only the first state dimension (out of three) is shown. Observations are drawn in green, predictions of the best method in red, and the ground truth in black. As the noise is relatively small, a part of the observation time series is zoomed in to make it visible.}\label{fig:lorenz:best}
\end{figure}

\begin{figure}
	\begin{center}
		{Cumulative Maximum Error for Test Data of \model{Lorenz63random} with Constant Timestep}

		\vspace{0.5cm}

		\includegraphics[width=\textwidth]{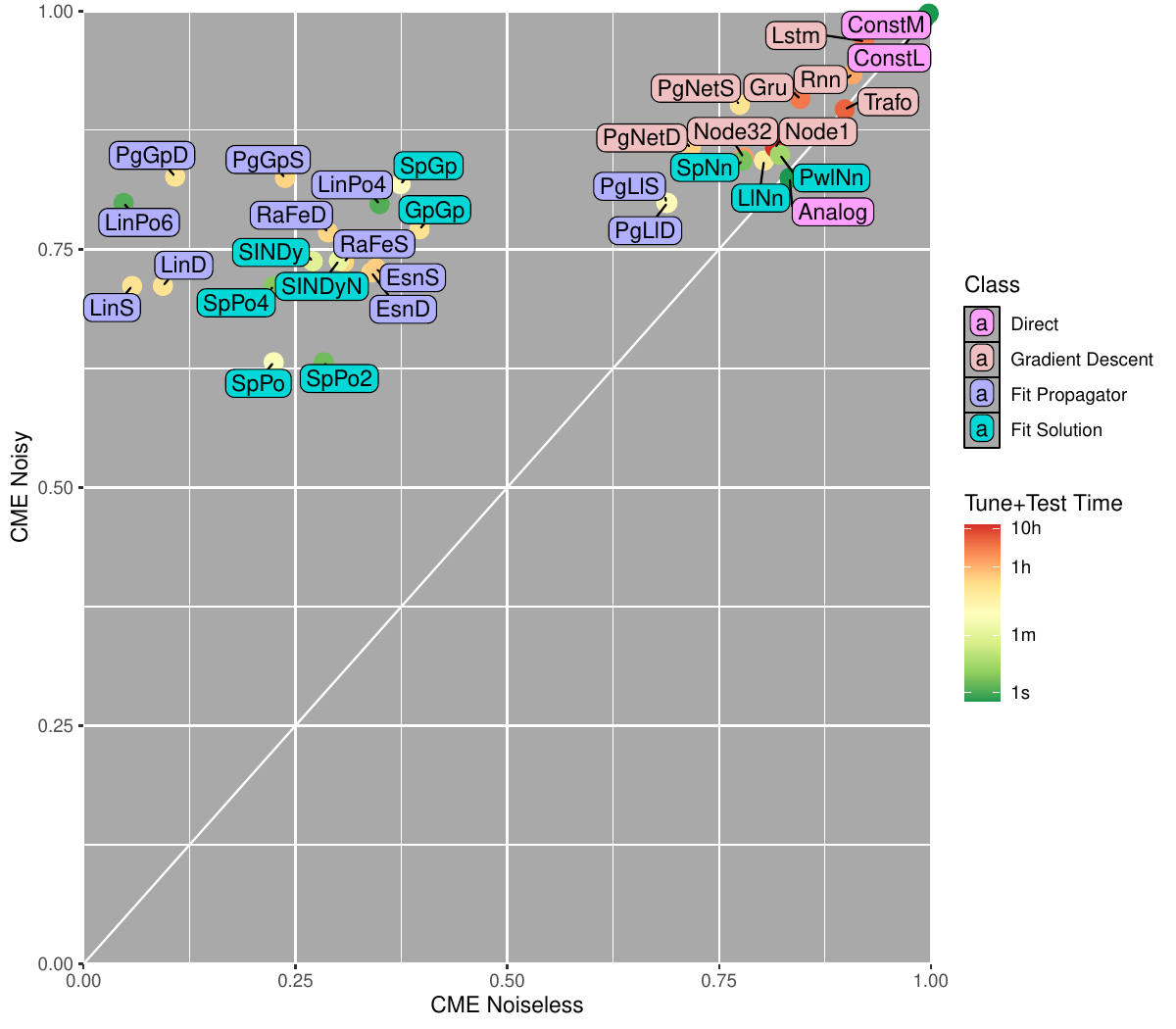}
	\end{center}
	\caption{The Cumulative Maximum Error ($\cme$) for noiseless and noisy test data with constant timestep $\stepsize$ of system \model{Lorenz63random} from database \DeebLorenz{}. The dots are colored according to the sum of average tune and test time of the respective method. Distance to the diagonal indicates sensitivity to noise. The plot shows clustering into a low error and a high error cluster, where all gradient descent based methods are in the high error group. The respective plots for the remaining system and timestep setting combinations are in the appendix, see \cref{fig:plane:const:lorenzStd,fig:plane:const:lorenzNonparam,fig:plane:rand:lorenzStd,fig:plane:rand:lorenzRandom,fig:plane:rand:lorenzNonparam}.
	Only some of them show a similarly clear clustering.}\label{fig:plane:const:lorenzRandom}
\end{figure}

Examples of training data and the prediction of the best methods are shown in \cref{fig:lorenz:best} for \DeebLorenz{}. The results of the simulation study in terms of the error metric $\cme$ are depicted in \cref{fig:results:lorenz:const} for \DeebLorenz{} with constant timestep, in \cref{fig:results:lorenz:rand} for \DeebLorenz{} with random timestep, and in \cref{fig:results:dysts} for the $\Dysts$ database. Additional details including score values and ranks for the error metrics $\cme$, $\smape$, and $\tvalid$ can be found in the appendix \cref{tbl:values:lorenz:cme,tbl:values:lorenz:smape,tbl:values:lorenz:tvalid,tbl:ranks:lorenz:cme,tbl:ranks:lorenz:smape,tbl:ranks:lorenz:tvalid,tbl:Dysts:medi,tbl:ranks:allscores}. Finally, \cref{fig:plane:const:lorenzRandom} shows the results for noiseless and noisy test data with constant timestep $\stepsize$ of system \model{Lorenz63random} from database \DeebLorenz{}. Below, we describe 10 key insights these results provide. Further details on the results can be found in \cref{sec:app:tabresults}.
\begin{table}
    \begin{center}
    	\input{tbl/tableLorenzFxdResults.tex}

    \end{center}
    \caption{Results on the \DeebLorenz{} database with constant timestep. In the first four columns, the methods are described by abbreviated names, their \textsf{Class} (Direct, Gradient Descent, Fit Propagator, Fit Solution), \textsf{Model} and \textsf{Variant} (see \cref{ssec:methods}). The average \textsf{Compute} time of the tuned method for training and testing on one time series (\textsf{Test}) and of hyperparameter tuning per time series (\textsf{Tune}) are shown. Missing tune times indicate methods that are not tuned. The last columns describe the results for each tuned method applied to the testing dataset based on the Cumulative Maximum Error ($\cme$) with forecasting horizon $S = 10$ system time units: S\# (green), R\# (red), N\# (blue) indicate the rank of the mean $\cme$ over 100 repetitions for the systems \model{Lorenz63std}, \model{Lorenz63random}, and \model{Lorenz63nonpar}, respectively. The column right after the ranks shows plots of the mean $\cme$ of 100 repetitions, with $95\%$ confidence intervals (color coded as in the rank column titles).}
    \label{fig:results:lorenz:const}
\end{table}
\begin{table}
    \begin{center}
        \input{tbl/tableLorenzRndResults.tex}

    \end{center}
    \caption{Same as \cref{fig:results:lorenz:const} but with the observation schemes that have random timesteps. Where applicable, we show the variants of the methods that incorporate the timestep as an input (marked with \textsf{T}). For the Neural ODE, we include only the variant with a batch size of $1$ (\method{Node1}) and exclude the variant with a batch size of $32$ (\method{Node32}). This exclusion is due to the requirement of equal computational steps across all elements within a batch, which is not feasible when varying timesteps are used.}
    \label{fig:results:lorenz:rand}
\end{table}
\begin{table}
    \vspace*{-1.0cm}
    \begin{center}
        \input{tbl/tableDystsResults.tex}

    \end{center}
    \vspace*{-0.2cm}
    \caption{Results on the \Dysts{} database. In the first four columns, the methods are described by abbreviated names, their \textsf{Class} (Direct, Gradient Descent, Fit Propagator, Fit Solution), \textsf{Model} and \textsf{Variant} (see \cref{ssec:methods}). The average \textsf{Compute} time of the tuned method for training and testing (\textsf{Test}) and of hyperparameter tuning (\textsf{Tune}) are shown. For results from \cite{Gilpin23} (class \textsf{Dysts}) this information is not available. For all other rows, missing tune times indicate methods that are not tuned. The last columns describe the results for each tuned method applied to the testing dataset, based on the Cumulative Maximum Error ($\cme$) with forecast horizon of $m=200$ timesteps: \# indicates the rank of the median (\textsf{medi} and black vertical line) over all systems. Each colored point corresponds to one system of \Dysts{}. The colors are chosen according to the median $\cme$ over all applied methods.}
    \label{fig:results:dysts}
\end{table}

\subsection{Lightweight Methods Outperform Heavyweight Methods}
In all settings, almost all of the best methods (top 5) are comparably simple (e.g., \methodh{sppo}{SpPo}, \methodh{lin}{LinS}, \methodh{esn}{EsnS}) and have low computational costs, as they do not rely on gradient descent learning. In contrast, computationally more complex learning methods (e.g., \methodh{node}{Node*}, \methodh{trafo}{Trafo}, \methodh{rnn}{Rnn}) perform rather poorly. The only exceptions are \methodh{gru}{Gru} and \methodh{trafo}{Trafo} for the noisy \Dysts{} data and \methodh{node}{Node1} for \model{Lorenz63random} and \model{Lorenz63nonpar} in the noisy setting with random time steps. Nonetheless, in these select cases, simple methods with similar or better performance are also available. See also \cref{fig:plane:const:lorenzRandom} for an illustration of this distinction on \model{Lorenz63random} with constant timestep. Furthermore, \cref{fig:results:lorenz:length} shows that lightweight methods also outperform heavyweight methods under different sample sizes (at least on the system \model{Lorenz63std} with constant timestep).

Even some of the tuning-free methods (\methodh{sppon}{SpPo2}, \methodh{sppon}{SpPo4}, \methodh{pwnn}{PwNn}, \methodh{spnn}{SpNn}, \methodh{linpo}{LinPo4}, \methodh{linpo}{LinPo6}) have mostly lower errors than heavyweight algorithms. We recommend using at least some of these methods as baselines for tasks similar to those shown in this study, as these methods are conceptually simple, easy to implement, have negligible computational costs, and good performance.

Altogether, performance, computational demand, and implementation complexity clearly favor lightweight methods. This calls into question the use of complex machine learning algorithms for the prediction of low-dimensional chaotic dynamical systems.
\subsection{Polynomial Fits to Noisefree Data Are Accurate Emulators}
\methodh{linpo}{LinPo6} applied to the noisefree data of \model{Lorenz63std} delivers near-perfect forecasts, achieving a $\cme$ of $6.6 \cdot 10^{-6}$ (see \cref{tbl:values:lorenz:cme}). This method employs a degree-6 polynomial to approximate the propagator map $u(t) \mapsto u(t + \stepsize)$. Our results are in line with the experiments on the Lorenz63 system shown in \cite{Gauthier2021} from which \methodh{lin}{LinS} (fitting polynomial propagator with time-delay) is adapted, even though \methodh{linpo}{LinPo6} does not use time delay. \methodh{lin}{LinS} achieves a $\cme$ of $1.9\cdot 10^{-4}$ in our setting.

While the vector field $f$ of the ODE in \model{Lorenz63std} is a degree-2 polynomial, the propagator map itself is not necessarily polynomial. However, since the propagator is derived from solving an ODE with smooth vector field, it is a smooth function. Taylor's theorem ensures that polynomials can provide increasingly accurate approximations to such smooth functions. Note that we obtain our ground truth trajectory $u(t)$ using a 4th order Runge-Kutta solver with constant solver timestep $\stepsize/10$, where $\stepsize$ is the observation timestep, to ensure the higher order terms are present in $u(t)$. If we were to use the Euler method with solver timestep equal to $\stepsize$ instead, the propagator would indeed be a degree-2 polynomial.

\begin{figure}
	\vspace*{-1cm}
	\begin{subfigure}{\textwidth}
		\centering
		\includegraphics[width=0.8\textwidth]{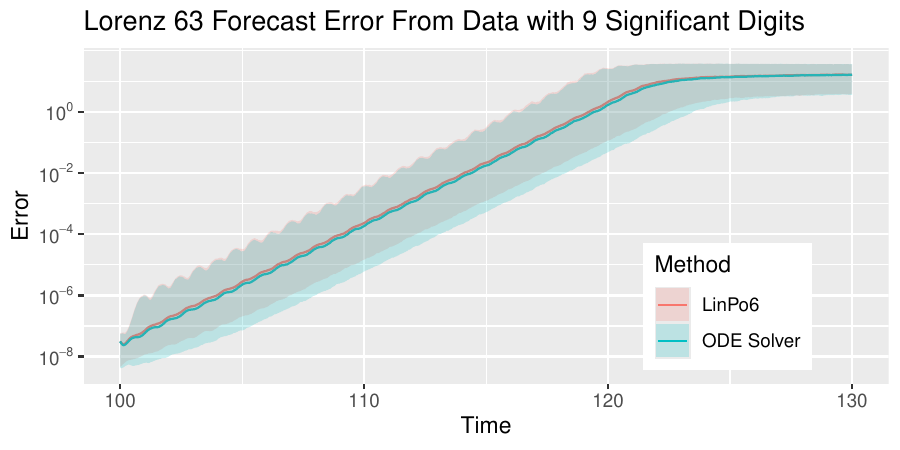}
		\vspace*{-0.5cm}
		\caption{%
			Median and central 90\% range of the error $\euclof{\hat u(t) - u(t)}$, $t \geq 100$ over $10^4$ repetitions.
			The true trajectory $u(t)$ is generated by a RK4 ODE solver (with constant timestep $10^{-1}\stepsize$) with full double-precision accuracy (64 bit), starting from a random point on the attractor of \model{Lorenz63std} and using the exact \model{Lorenz63std} vector field.
			For the red curve, $\hat u(t)$ is computed using a degree-six polynomial propagator (\method{LinPo6}) trained on $\bar u(t_i)$ with $i \in \{1, \dots, 10^4\}$, $t_i = i\stepsize$, $\stepsize = 10^{-2}$. Here, $\bar u(t_i)$ is obtained by rounding $u(t_i)$ to 9 significant digits. Forecasting begins at $t_n = 100$ with initial conditions $\bar u(t_n)$.
			For the blue curve, $\hat u(t)$ is obtained by applying a RK4 ODE solver (double-precision) to the true \model{Lorenz63std} vector field, starting from the same initial conditions $\bar u(t_n)$.
			The nearly identical errors of the polynomial propagator and the ODE solver demonstrate that at 9-digit accuracy, \method{LinPo6} effectively emulates the ODE solver on the \model{Lorenz63std} system.
		}\label{fig:poly:a}
	\end{subfigure}
	
	\vspace{0.5cm}
	
	\begin{subfigure}{\textwidth}
		\centering
		\includegraphics[width=0.8\textwidth]{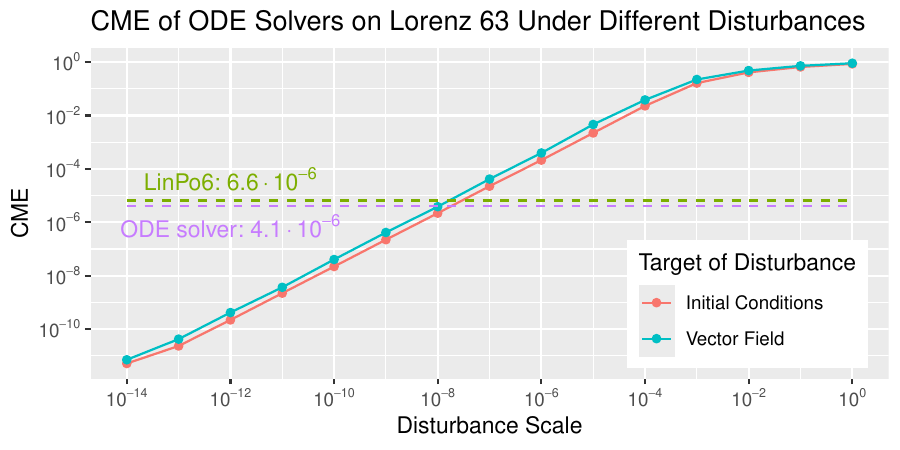}
		\caption{%
			The Cumulative Maximum Error $\cme(\hat u, u)$ on a time range $[0, S]$ with $S=10$ of \model{Lorenz63std} for ODE solvers (median over $10^3$ repetitions).
			The true trajectory $u(t)$ is generated by a RK4 ODE solver (with constant timestep $10^{-1}\stepsize$) with full double-precision accuracy (64 bit), starting from a random point on the attractor of \model{Lorenz63std} and using the exact vector field.
			For the red curve, $\hat u(t)$ is obtained by applying a RK4 ODE solver (double-precision) to the true \model{Lorenz63std} vector field, starting from $u(0) + \delta$. Here, $\delta$ is drawn uniformly at random from the sphere in $\R^3$ centered at the origin with a radius specified on the $x$-axis.
			For the blue curve, $\hat u(t)$ is obtained by applying a RK4 ODE solver (double-precision) to the true initial conditions $u(0)$ with a perturbed version of the \model{Lorenz63std} vector field. The perturbed dynamics are created by modifying the three parameters of \model{Lorenz63std}, $(\hat\sigma, \hat\rho, \hat\beta) = (\sigma, \rho, \beta) + \delta$, where $\sigma = 10$, $\rho = 28$, $\beta = 8/3$, see \cref{sec:app:data:deenlorenz}. Again, $\delta$ is drawn uniformly at random from the sphere in $\R^3$ with the radius specified on the $x$-axis.
			In our noisefree settings, all methods have only access to a rounded version $\bar u(t)$ of $u(t)$, as we only write $8$ digits after the decimal point in the CSV files that contain the training, validation, and testing data. The purple dashed line indicates the $\cme$ of a RK4 ODE solver applied to initial conditions from such a file format and using the true \model{Lorenz63std} vector field. For comparison, the green dashed line shows the $\cme$ of $\method{LinPo6}$, the best performing method on \model{Lorenz63std}.
		}\label{fig:poly:b}
	\end{subfigure}
	\caption{Comparison between polynomial fit and ODE solver.}
	\label{fig:poly}
\end{figure}

In \cref{fig:poly}, we demonstrate that \methodh{linpo}{LinPo6} performs essentially as well as directly solving the ODE for \model{Lorenz63std} starting from initial conditions specified with 9 significant digits. This shows that \methodh{linpo}{LinPo6} surpasses the accuracy of 32-bit single-precision machine arithmetic, which provides between 7 and 8 significant digits of accuracy. Moreover, as our simulation study stores data in CSV files with only 8 digits written after the decimal point, \methodh{linpo}{LinPo6} is close to achieving the best possible accuracy given the limitations of the data (for noisefree \model{Lorenz63std}).

\subsection{Method Selection and Hyperparameter Optimization Are Essential}
Compared to the results of \cite{Gilpin23} (the method names starting with an underscore in \cref{fig:results:dysts}), several of the methods considered here have much lower errors: For the noisefree \Dysts{} data, the median $\cme$ of \method{\_NBEAT}, the best performing method in \cite{Gilpin23}, is 0.56. The lowest error achieved in our simulations is 0.0041 (\methodh{sppo}{SpPo}, \cref{tbl:Dysts:medi}).

One reason for the improved performance is the more extensive hyperparameter tuning in our study compared to \cite{Gilpin23}, where only a single parameter was optimized for each method. For methods with low computational costs, we optimize several parameters. To illustrate the effect, we compare the median $\cme$ (in parenthesis) for selected methods on the noisefree testing dataset from \Dysts{}, as shown in \cref{fig:results:dysts} and \cref{tbl:Dysts:medi}: The Nonlinear Vector Autoregression \method{\_Nvar} (0.80) and the Echo State Network \method{\_Esn} (0.79) from \cite{Gilpin23} are essentially the same as our \methodh{lin}{LinS} (0.0054) and \methodh{esn}{EsnS} (0.030), respectively. Our simulations yield a median $\cme$ that is orders of magnitude lower than for the corresponding methods in \cite{Gilpin23}. This indicates that optimizing key hyperparameters is critical for achieving good performance in these methods.

But note that \methodh{sppon}{SpPo4} and \methodh{linpo}{LinPo4} are tuning-free methods and achieve median $\cme$'s of 0.0079 and 0.040, respectively. Furthermore, from all methods of \cite{Gilpin23}, only \method{\_NBEAT} outperforms the baseline method \methodh{analog}{Analog}. This highlights the critical role of selecting appropriate methods and baselines for a given task.
\subsection{Noise and Timestep Design Influence Absolute and Relative Performance}
In the absence of noise, forecasting becomes an interpolation task, whereas noisy observations yield a regression problem.
Generally, as expected, performance degrades across methods strongly when noise is added to the observations. Moreover, methods that are best for the noisefree interpolation task are typically not the best for the noisy regression task. This distinction is evident when comparing the noisefree and noisy columns in \cref{fig:results:lorenz:const,fig:results:lorenz:rand,fig:results:dysts}.

The Echo State Network \methodh{esn}{Esn*} has access to states of the past, whereas the otherwise similar method Random Features \methodh{rafe}{RaFe*} does not. Because of the Markovian nature of our dynamical systems, knowledge of the past is not required in the noisefree case as the next state is equal to the (true) propagator applied to the current state. This is reflected in our simulation results, where \methodh{rafe}{RaFe*} tends to perform better than \methodh{esn}{Esn*} in noisefree settings and vice versa for the settings with measurement noise. As the noisy systems of \Dysts{} are created with system noise, they are Markovian. Surprisingly, the \methodh{esn}{Esn*} seems to be at an advantage over \methodh{rafe}{RaFe*} here, too.

Under noise, the lower degree polynomial solution smoother \methodh{sppon}{SpPo2} performs relatively better than the higher degree polynomial \methodh{sppon}{SpPo4}, and vice versa in the noisefree case. Overfitting to noise, which cannot happen in the noisefree case, seems to be the reason for this observation.

There is a noticeable difference in performances of methods between fixed and random timesteps (\cref{fig:results:lorenz:const,fig:results:lorenz:rand}). Generally, random timesteps for the training data make forecasting (at fixed timesteps) more difficult. But also the ranking of methods according to their $\cme$ changes: The Gaussian process based method \methodh{gpgp}{GpGp} takes the lead for random timesteps. For fixed timesteps, this method is only mediocre.

As expected, propagator methods that do not have the timestep as an input, do not fare well if the timestep is not constant. With the additional knowledge of the timestep (\method{*ST} and \method{*DT}), the performance improves (\cref{tbl:ovsT}).

The combination of a random timestep with noisy observations seems to make the learning task extremely difficult. In the case of \model{Lorenz63random}, only \methodh{gpgp}{GpGp} (Gaussian process based solution smoother) and \methodh{lin}{LinST} (ridge regression fit of polynomial propagator) have better scores than the baseline \methodh{analog}{Analog}. For \model{Lorenz63nonpar} no method beats \methodh{analog}{Analog}.
\subsection{Repeating Experiments is Crucial for a Robust Evaluation}
As there is only one time series per system available in the \Dysts{} database, reported differences between the performance of methods might not be robust on the same system. Furthermore, confidence intervals based on the results on the 133 systems are hard to justify as one typically has to assume an independent, identical distribution of the considered values, whereas the selection of the specific systems for \Dysts{} was a deterministic non-independent process. But note that the code for generating more data for \Dysts{} is available at the respective repository.

In contrast, for \DeebLorenz{}, experiments for the same system and observation scheme are repeated 100 times with randomly drawn initial conditions, noise values, and timesteps (if applicable). This allows us to judge whether reported differences in $\cme$ values are statistically significant. In \cref{fig:pValues}, we visualize the results of paired $t$-Tests for the null hypothesis $\cme(\text{\method{Method1}}) \geq \cme(\text{\method{Method2}})$. If the null can be rejected, it shows that \method{Method1} is significantly better \method{Method2} on the given system with the given observation scheme using the $\cme$ as error metric. We see that the ranking of the algorithms as shown in \cref{tbl:ranks:lorenz:cme} is mostly robust up to permutation of two or three consecutive ranks. In comparison, \cref{fig:pValues10} shows that only 10 repetitions would yield largely indistinguishable performances.

All in all, it seems prudent to repeat experiments to obtain more reliable results and be able to judge the confidence of these results.
\subsection{Evaluation Lacks Robustness When Only Assessing a Fixed System}
Although \model{Lorenz63std} and \model{Lorenz63random} share the same functional form of the vector field $f$, differing only in their polynomial coefficients, the absolute errors and relative rankings of methods vary significantly between the two systems (\cref{fig:results:lorenz:const,fig:results:lorenz:rand}).

Tuning for \model{Lorenz63random} (and \model{Lorenz63nonpar}) is hard in that the tuned hyperparameters must perform well across varying, randomly chosen system parameters. In contrast, \model{Lorenz63std} has fixed dynamics, allowing not only the functional form of $f$ but also the specific polynomial coefficients to be (partially) inferred during tuning.

Thus, a low error value for \model{Lorenz63random} is a more robust indicator of a method's ability to generalize than a low error value for \model{Lorenz63std}.
\subsection{The Functional Form of a Parametric Method Should Fit the System}
Most systems in \Dysts{} and all systems but \model{Lorenz63nonpar} in \DeebLorenz{} have a vector field $f$ that is a polynomial of low degree. Thus, it is not surprising that methods based on this assumption, such as \methodh{sppo}{SpPo} (polynomial solution smoother) and \methodh{sindy}{SINDy} (similar to \methodh{sppo}{SpPo}, but assumes sparsity of target polynomial), perform well on these tasks. Note that a polynomial vector field $f$ does not imply that the propagator $\mc P_{\stepsize}$ for a given timestep $\stepsize$ is also polynomial. Still, there might be a reasonable polynomial approximation, which would explain the good performance of \methodh{lin}{Lin*} (ridge regression fit of polynomial propagator).

For \model{Lorenz63std} and \model{Lorenz63random}, the vector field $f$ is a quadratic polynomial. For these systems \methodh{sppon}{SpPo2}, which fits such a polynomial, works very well (when excluding the extremely difficult combination of random timesteps and noisy observations). But the method shows poor performance for the non-polynomial \model{Lorenz63nonpar}. The same is true in the noisefree, constant timestep setting for methods that can fit higher degree polynomials (\methodh{sppo}{SpPo}, \methodh{sppon}{SpPo4}, \methodh{sindy}{SINDy}, \methodh{sindy}{SINDyN}). In the noisy setting, where performance in general is worse, this effect is not visible, likely as a polynomial approximation (Taylor's theorem) is accurate enough relative to the other methods.

The method \methodh{sindy}{SINDyN} (with normalization) is usually worse than \methodh{sindy}{SINDy}. This makes sense, as many systems studied here have sparse polynomial dynamics and normalization (which in our case includes a rotation) destroys sparsity.

All in all, low degree polynomial approximation works only well if the target function can be described sufficiently well as such a polynomial. Inherently nonparametric methods, such as \methodh{gpgp}{GpGp} or \methodh{pg}{PgGp*}, which both fit Gaussian processes, seem to be more suitable if the target is non-polynomial.
\subsection{Choosing the Right Variant of Propagators Improves Performance}
As expected, the errors of propagators with and without timestep input (\method{*T}, \cref{tbl:ovsT}) are similar when the timestep is constant.
An exception is observed with \methodh{lin}{LinD}, \methodh{linpo}{LinPo4}, and \methodh{linpo}{LinPo6}, which fit a polynomial propagator using ridge regression, in the noisefree case, where performance degrades when the timestep is (unnecessarily) included as an additional input. This decline is likely due to numerical issues arising from co-linear predictors. For random timesteps, however, providing timestep information improves the results in most cases.

Between the state (\method{*S*}) and the difference quotient (\method{*D*}) versions of the propagator estimators (\cref{tbl:SvsD}), there is no clear overall winner. However, in the case of random timesteps without noise, difference-quotient based methods consistently yield lower errors.  This may be due to the reduced dependence of the target on the timestep when estimating the difference quotient. Additionally, both \methodh{pg}{PgGp*} and \methodh{pg}{PgNet*}, which fit the propagator map using a Gaussian process and a neural network, respectively, exhibit a clear preference for the difference quotient as the target (except in the noisy setting with random timesteps).

To explain this preference, note that system states change smoothly over time, so one-step-ahead predictions should be close to persistence (where future states equal the current state). When the target is the state, persistence is achieved by the identity function as the propagator. When the target is the difference quotient, persistence corresponds to the constant zero function.
The Gaussian Process \methodh{pg}{PgGp*} has a prior centered at zero, biasing it toward persistence in the difference quotient setup. In contrast, when the state is the target, its predictions are biased toward the mean of observations (climatology), as the input data is normalized. For the neural network \methodh{pg}{PgNet*}, the constant zero function is easier to learn since only the weights in the last layer need suitable values, while the identity function requires appropriate weights across all layers.
Thus, both the Gaussian process and the neural network are better suited to the difference quotient setup due to their inherent bias towards zero.
\subsection{Using The Cumulative Maximum Error Is Advantageous}
When ranking different methods based on their performance, the error metrics $\cme$, $\smape$, and $\tvalid$ generally produce similar results. However, a notable difference arises with $\smape$ for more challenging tasks, where strong initial performance can be overshadowed by a divergence of the forecast later on (instead of, e.g., defaulting to a constant), see \cref{tbl:ranks:allscores}. As seen from the explicit error values in \cref{tbl:Dysts:medi}, $\tvalid$ is less effective at distinguishing well-performing methods when the testing duration is too short or the threshold parameter is too lenient, as several methods can achieve the maximum valid time. The $\cme$ does not suffer from any of these limitations.
\subsection{Scarce Validation Data Can Lead to Over-Tuning}
Unsurprisingly, the error measures of tuned methods typically perform better on the validation dataset than on the testing datasets, see \cref{tbl:Dysts:medi}. For some methods, the difference is rather large (for the Echo State Network \methodh{esn}{EsnS}, we have a median $\cme$ of $0.0040$ on the noisefree \Dysts{} validation data and $0.030$ on the test data). In combination with the high number of parameters tested for these methods (for \methodh{esn}{EsnS} on the noisefree \Dysts{} data, $288.5$ per system on average), this suggests that the hyperparameter tuning has the potential to introduce overfitting to the validation data.
This affects \Dysts{} more than \DeebLorenz{} as there is much less tuning data for the former and no repetitions, making the validation errors less robust.

%% file: tbl/tableLorenzFxdResults.tex
\begin{center}
\caption*{
{\large Cumulative Maximum Error for Test Data of \DeebLorenz{} with Constant Timestep}
} 
\fontsize{8.0pt}{10pt}\selectfont
\fontfamily{phv}\selectfont
\renewcommand{\arraystretch}{1.05}
\setlength{\tabcolsep}{0.3em}
\rowcolors{2}{gray!20}{white}
\begin{tabular}{lcllrrrrrcrrrc}
\toprule
\multicolumn{4}{c}{Method} & \multicolumn{2}{c}{Compute} & \multicolumn{4}{c}{Noisefree} & \multicolumn{4}{c}{Noisy} \\ 
\cmidrule(lr){1-4} \cmidrule(lr){5-6} \cmidrule(lr){7-10} \cmidrule(lr){11-14}
Name & Class & Model & Variant & Test & Tune & \cellcolor[HTML]{00BA38}{S\#} & \cellcolor[HTML]{F8766D}{R\#} & \cellcolor[HTML]{619CFF}{N\#} & \CmeScale{} & \cellcolor[HTML]{00BA38}{S\#} & \cellcolor[HTML]{F8766D}{R\#} & \cellcolor[HTML]{619CFF}{N\#} & \CmeScale{} \\ 
\midrule\addlinespace[2.5pt]
\method{LinPo6} & {\cellcolor[HTML]{B0B0FF}{\textcolor[HTML]{000000}{Fit Prop.}}} & Lin & Poly6 & {\cellcolor[HTML]{168F4C}{\textcolor[HTML]{FFFFFF}{1s}}} & {\cellcolor[HTML]{808080}{\textcolor[HTML]{FFFFFF}{}}} & \textbf{\underline{1}} & \textbf{\underline{1}} & 8 & \inlinegraphics{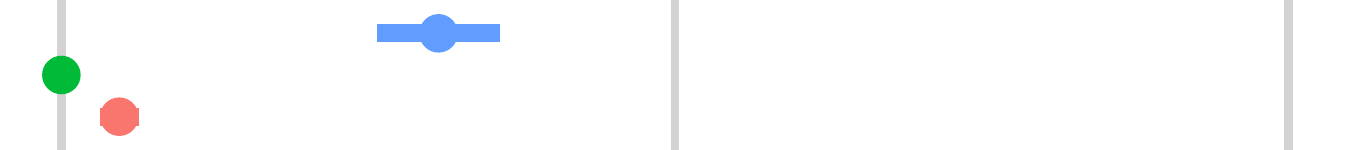} & 15 & 14 & 15 & \inlinegraphics{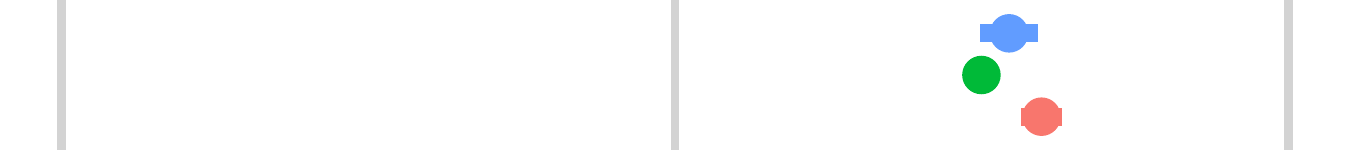} \\ 
\method{LinD} & {\cellcolor[HTML]{B0B0FF}{\textcolor[HTML]{000000}{Fit Prop.}}} & Lin & D & {\cellcolor[HTML]{77C465}{\textcolor[HTML]{000000}{7s}}} & {\cellcolor[HTML]{EEF8A8}{\textcolor[HTML]{000000}{17m}}} & \textbf{2} & \underline{3} & \textbf{2} & \inlinegraphics{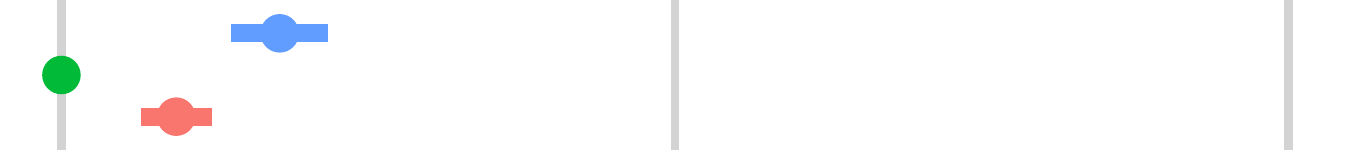} & 4 & 5 & 6 & \inlinegraphics{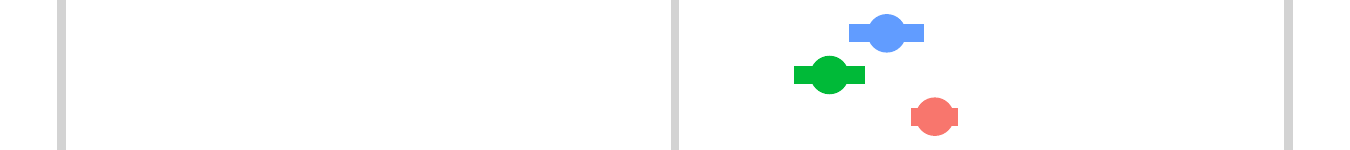} \\ 
\method{LinS} & {\cellcolor[HTML]{B0B0FF}{\textcolor[HTML]{000000}{Fit Prop.}}} & Lin & S & {\cellcolor[HTML]{5AB65F}{\textcolor[HTML]{000000}{4s}}} & {\cellcolor[HTML]{EFF8A9}{\textcolor[HTML]{000000}{17m}}} & \underline{3} & \textbf{2} & 4 & \inlinegraphics{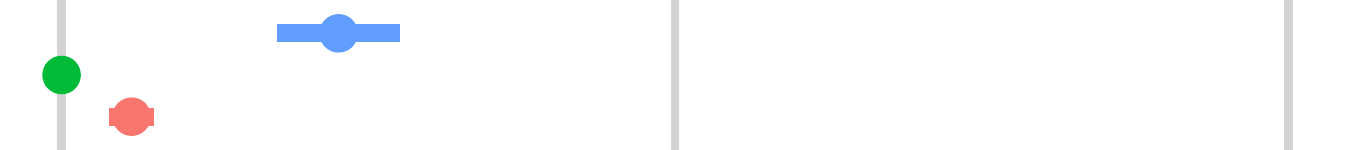} & 5 & 4 & 4 & \inlinegraphics{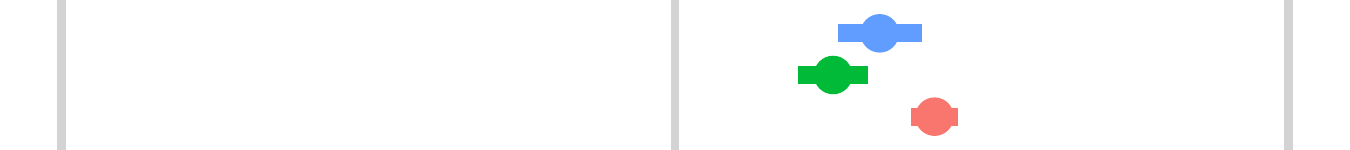} \\ 
\method{SpPo4} & {\cellcolor[HTML]{00D8D8}{\textcolor[HTML]{000000}{Fit Solu.}}} & Spline & Poly4 & {\cellcolor[HTML]{5AB65F}{\textcolor[HTML]{000000}{4s}}} & {\cellcolor[HTML]{808080}{\textcolor[HTML]{FFFFFF}{}}} & 4 & 6 & 14 & \inlinegraphics{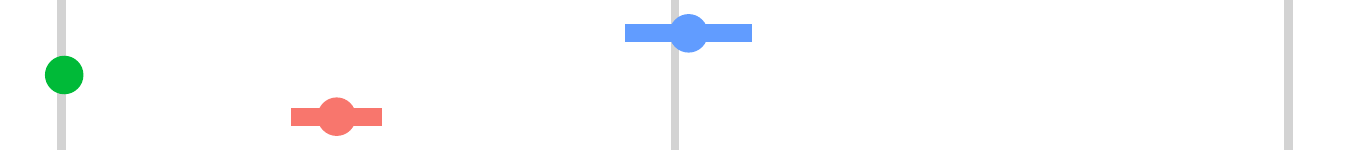} & 9 & \underline{3} & \textbf{\underline{1}} & \inlinegraphics{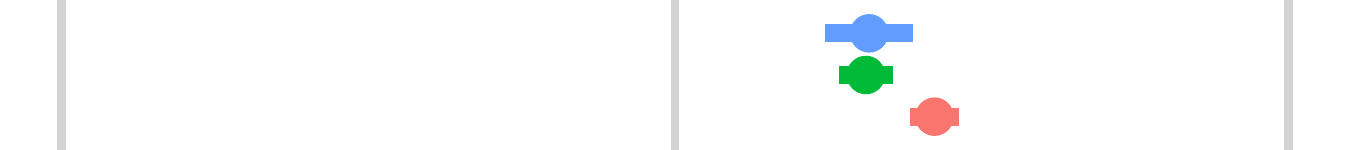} \\ 
\method{SpPo} & {\cellcolor[HTML]{00D8D8}{\textcolor[HTML]{000000}{Fit Solu.}}} & Spline & Poly & {\cellcolor[HTML]{66BD63}{\textcolor[HTML]{000000}{5s}}} & {\cellcolor[HTML]{60B961}{\textcolor[HTML]{000000}{3m}}} & 5 & 5 & 7 & \inlinegraphics{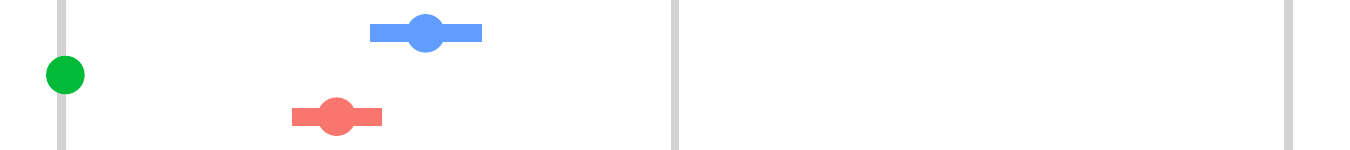} & \textbf{2} & \textbf{\underline{1}} & 5 & \inlinegraphics{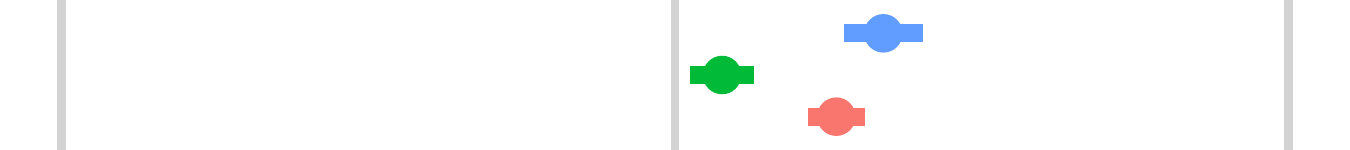} \\ 
\method{SINDy} & {\cellcolor[HTML]{00D8D8}{\textcolor[HTML]{000000}{Fit Solu.}}} & SINDy &  & {\cellcolor[HTML]{6CBF64}{\textcolor[HTML]{000000}{6s}}} & {\cellcolor[HTML]{006837}{\textcolor[HTML]{FFFFFF}{51s}}} & 6 & 8 & 11 & \inlinegraphics{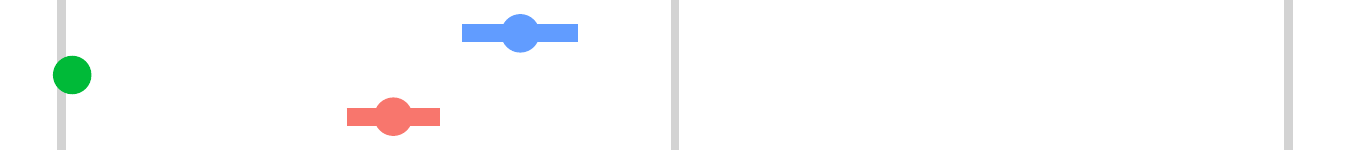} & 11 & 9 & \textbf{2} & \inlinegraphics{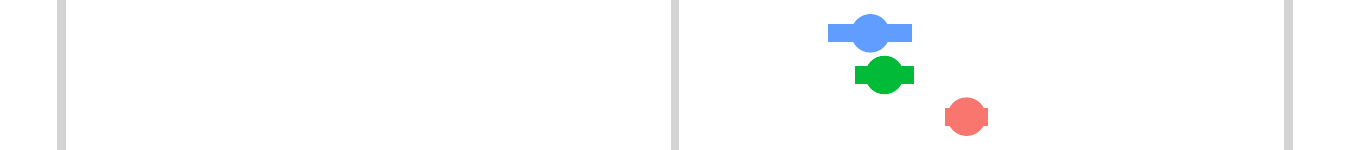} \\ 
\method{SpPo2} & {\cellcolor[HTML]{00D8D8}{\textcolor[HTML]{000000}{Fit Solu.}}} & Spline & Poly2 & {\cellcolor[HTML]{3DA657}{\textcolor[HTML]{FFFFFF}{2s}}} & {\cellcolor[HTML]{808080}{\textcolor[HTML]{FFFFFF}{}}} & 7 & 9 & 28 & \inlinegraphics{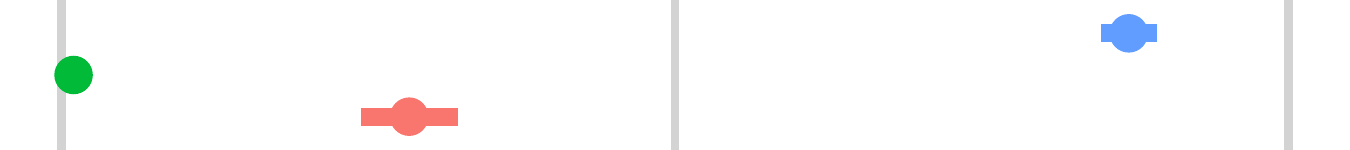} & \textbf{\underline{1}} & \textbf{\underline{1}} & 26 & \inlinegraphics{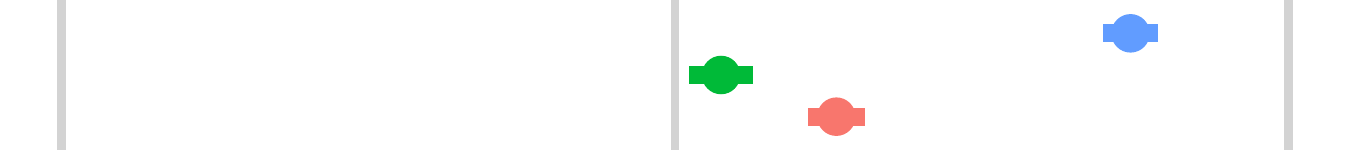} \\ 
\method{PgGpD} & {\cellcolor[HTML]{B0B0FF}{\textcolor[HTML]{000000}{Fit Prop.}}} & GP & D & {\cellcolor[HTML]{85CA66}{\textcolor[HTML]{000000}{9s}}} & {\cellcolor[HTML]{EEF8A8}{\textcolor[HTML]{000000}{17m}}} & 8 & 4 & \textbf{\underline{1}} & \inlinegraphics{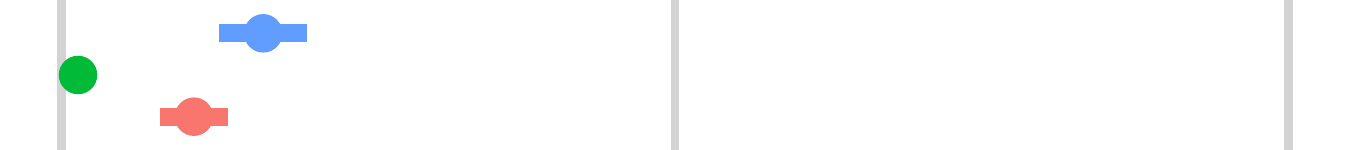} & 14 & 20 & 19 & \inlinegraphics{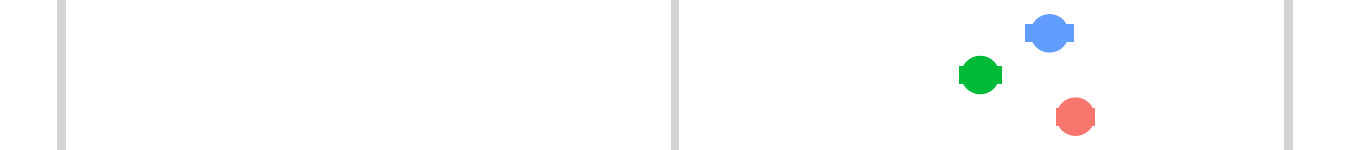} \\ 
\method{SINDyN} & {\cellcolor[HTML]{00D8D8}{\textcolor[HTML]{000000}{Fit Solu.}}} & SINDy & norm & {\cellcolor[HTML]{69BE63}{\textcolor[HTML]{000000}{5s}}} & {\cellcolor[HTML]{108446}{\textcolor[HTML]{FFFFFF}{1m}}} & 9 & 11 & 16 & \inlinegraphics{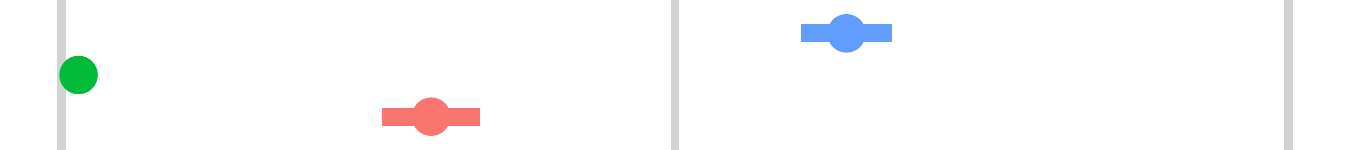} & 10 & 10 & 8 & \inlinegraphics{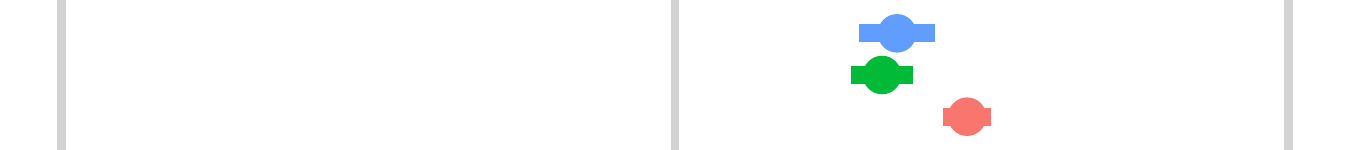} \\ 
\method{LinPo4} & {\cellcolor[HTML]{B0B0FF}{\textcolor[HTML]{000000}{Fit Prop.}}} & Lin & Poly4 & {\cellcolor[HTML]{158E4B}{\textcolor[HTML]{FFFFFF}{1s}}} & {\cellcolor[HTML]{808080}{\textcolor[HTML]{FFFFFF}{}}} & 10 & 15 & 15 & \inlinegraphics{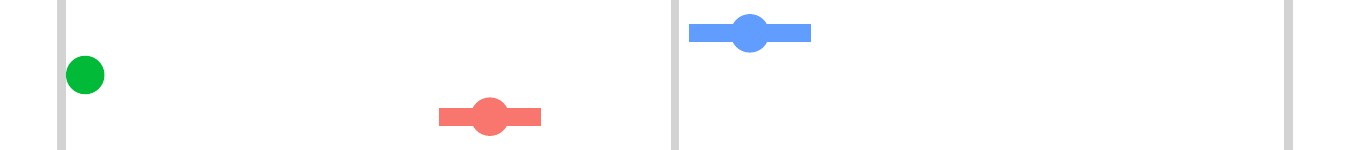} & 13 & 13 & 12 & \inlinegraphics{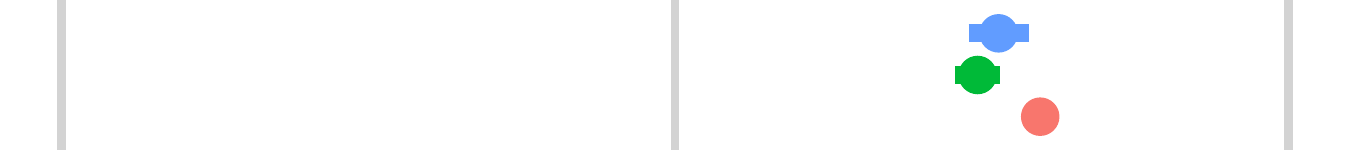} \\ 
\method{EsnD} & {\cellcolor[HTML]{B0B0FF}{\textcolor[HTML]{000000}{Fit Prop.}}} & ESN & D & {\cellcolor[HTML]{66BD63}{\textcolor[HTML]{000000}{5s}}} & {\cellcolor[HTML]{FFF7B2}{\textcolor[HTML]{000000}{27m}}} & 11 & 13 & 9 & \inlinegraphics{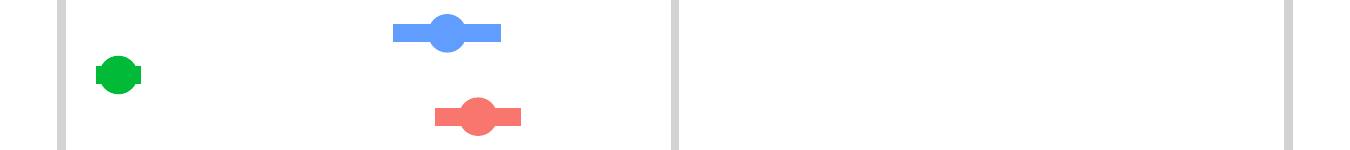} & 8 & 6 & 7 & \inlinegraphics{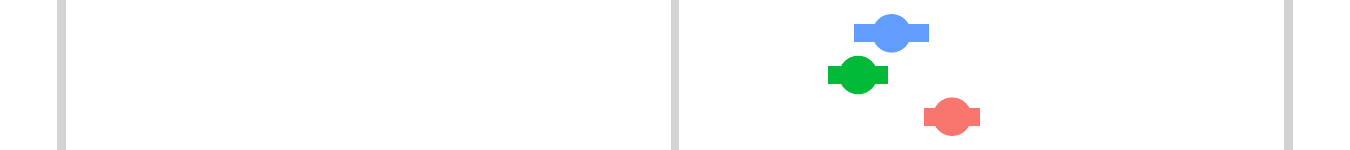} \\ 
\method{PgGpS} & {\cellcolor[HTML]{B0B0FF}{\textcolor[HTML]{000000}{Fit Prop.}}} & GP & S & {\cellcolor[HTML]{AADB6C}{\textcolor[HTML]{000000}{15s}}} & {\cellcolor[HTML]{FFF6B0}{\textcolor[HTML]{000000}{28m}}} & 12 & 7 & \underline{3} & \inlinegraphics{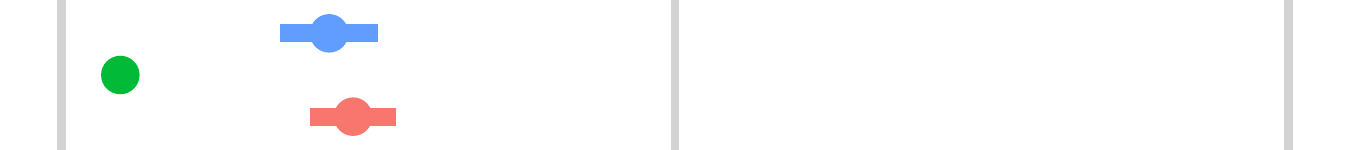} & 24 & 18 & 17 & \inlinegraphics{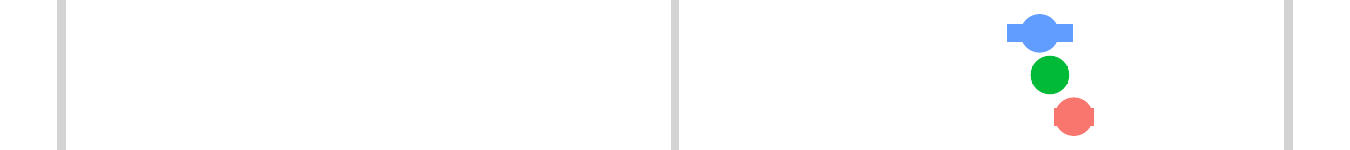} \\ 
\method{RaFeD} & {\cellcolor[HTML]{B0B0FF}{\textcolor[HTML]{000000}{Fit Prop.}}} & RandFeat & D & {\cellcolor[HTML]{61BA61}{\textcolor[HTML]{000000}{5s}}} & {\cellcolor[HTML]{FFF7B2}{\textcolor[HTML]{000000}{27m}}} & 13 & 10 & 6 & \inlinegraphics{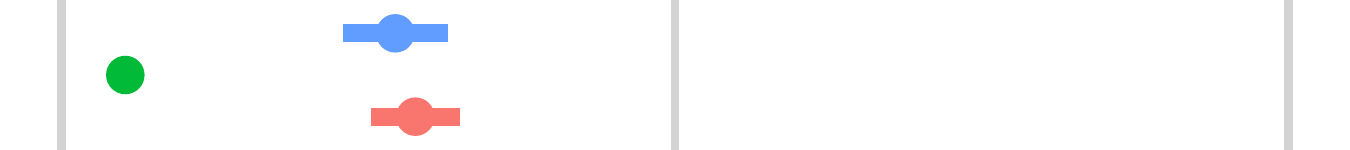} & 6 & 11 & 9 & \inlinegraphics{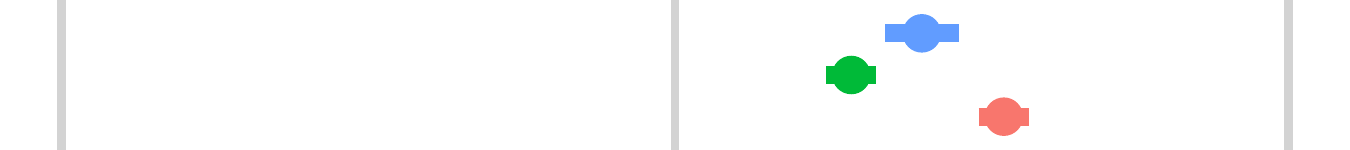} \\ 
\method{EsnS} & {\cellcolor[HTML]{B0B0FF}{\textcolor[HTML]{000000}{Fit Prop.}}} & ESN & S & {\cellcolor[HTML]{6ABE63}{\textcolor[HTML]{000000}{6s}}} & {\cellcolor[HTML]{FFEFA5}{\textcolor[HTML]{000000}{32m}}} & 14 & 14 & 10 & \inlinegraphics{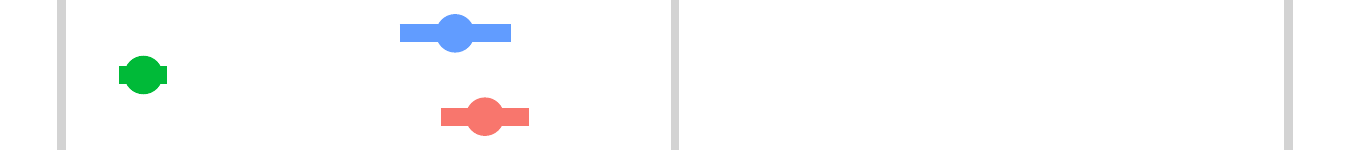} & 7 & 7 & \underline{3} & \inlinegraphics{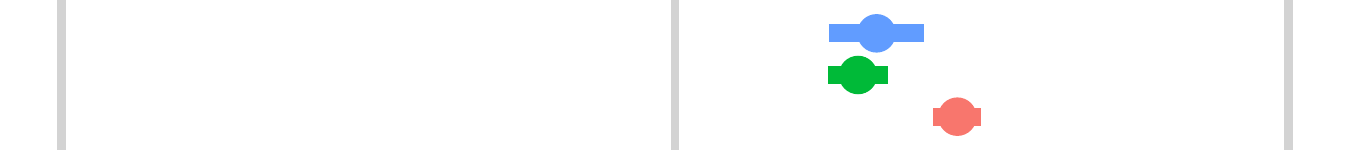} \\ 
\method{RaFeS} & {\cellcolor[HTML]{B0B0FF}{\textcolor[HTML]{000000}{Fit Prop.}}} & RandFeat & S & {\cellcolor[HTML]{5EB860}{\textcolor[HTML]{000000}{5s}}} & {\cellcolor[HTML]{FFF7B1}{\textcolor[HTML]{000000}{28m}}} & 15 & 12 & 5 & \inlinegraphics{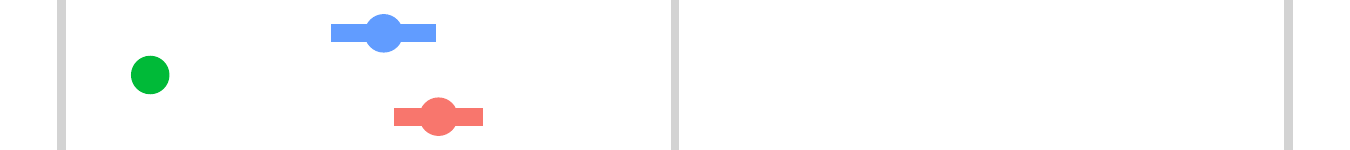} & \underline{3} & 8 & 11 & \inlinegraphics{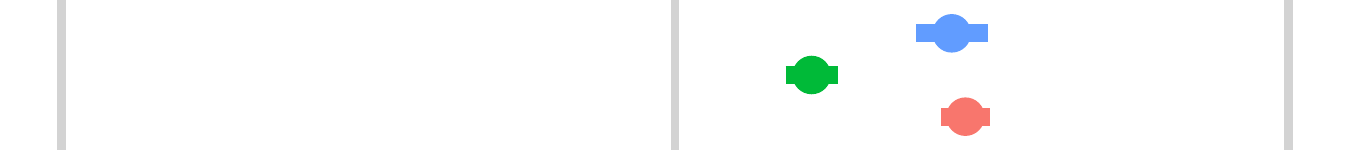} \\ 
\method{GpGp} & {\cellcolor[HTML]{00D8D8}{\textcolor[HTML]{000000}{Fit Solu.}}} & GP & GP & {\cellcolor[HTML]{B5E074}{\textcolor[HTML]{000000}{19s}}} & {\cellcolor[HTML]{F2FAAD}{\textcolor[HTML]{000000}{18m}}} & 16 & 17 & 13 & \inlinegraphics{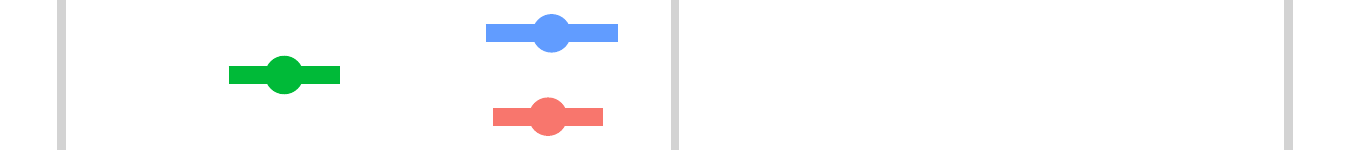} & 12 & 12 & 10 & \inlinegraphics{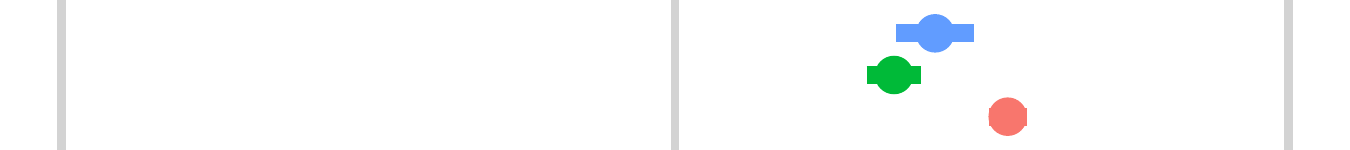} \\ 
\method{SpGp} & {\cellcolor[HTML]{00D8D8}{\textcolor[HTML]{000000}{Fit Solu.}}} & Spline & GP & {\cellcolor[HTML]{8DCE67}{\textcolor[HTML]{000000}{10s}}} & {\cellcolor[HTML]{78C465}{\textcolor[HTML]{000000}{3m}}} & 17 & 16 & 12 & \inlinegraphics{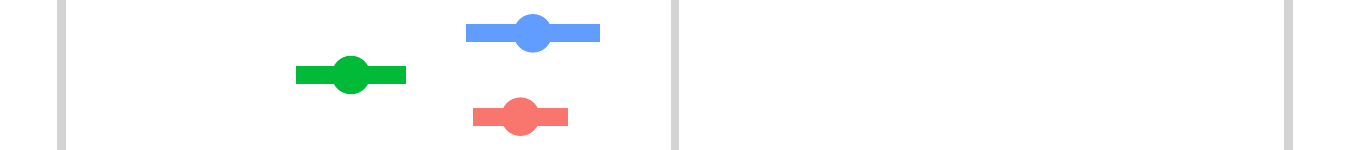} & 19 & 17 & 18 & \inlinegraphics{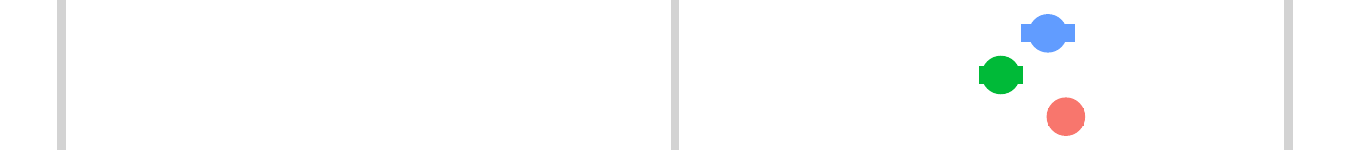} \\ 
\method{PgNetD} & {\cellcolor[HTML]{F0C0C0}{\textcolor[HTML]{000000}{Gr. Desc.}}} & NeuralNet & D & {\cellcolor[HTML]{FDAE61}{\textcolor[HTML]{000000}{7m}}} & {\cellcolor[HTML]{FFFFBF}{\textcolor[HTML]{000000}{23m}}} & 18 & 20 & 17 & \inlinegraphics{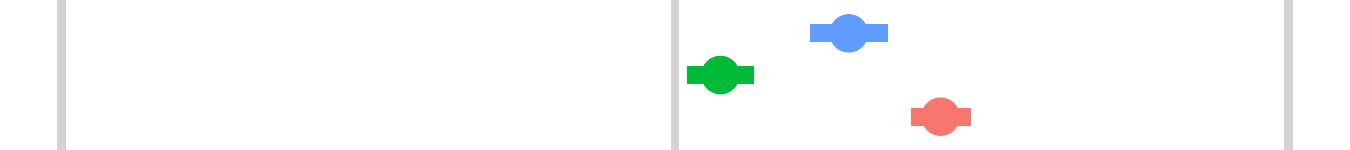} & 22 & 26 & 22 & \inlinegraphics{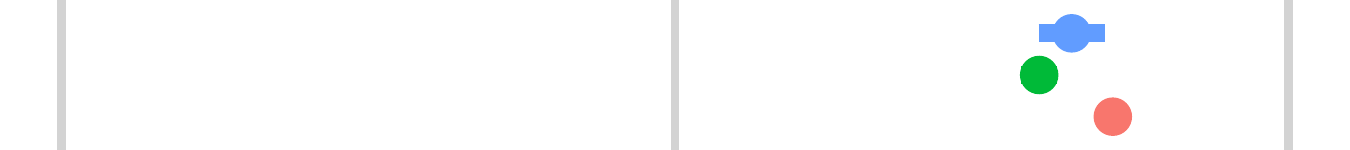} \\ 
\method{PgLlD} & {\cellcolor[HTML]{B0B0FF}{\textcolor[HTML]{000000}{Fit Prop.}}} & LocalLin & D & {\cellcolor[HTML]{84C966}{\textcolor[HTML]{000000}{8s}}} & {\cellcolor[HTML]{97D268}{\textcolor[HTML]{000000}{5m}}} & 19 & 19 & 19 & \inlinegraphics{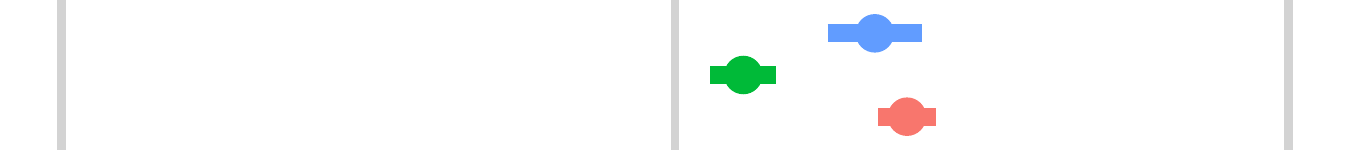} & 17 & 15 & 13 & \inlinegraphics{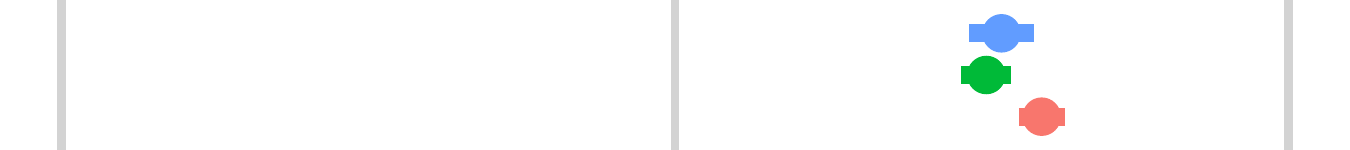} \\ 
\method{PgLlS} & {\cellcolor[HTML]{B0B0FF}{\textcolor[HTML]{000000}{Fit Prop.}}} & LocalLin & S & {\cellcolor[HTML]{7EC766}{\textcolor[HTML]{000000}{8s}}} & {\cellcolor[HTML]{93D068}{\textcolor[HTML]{000000}{5m}}} & 20 & 18 & 18 & \inlinegraphics{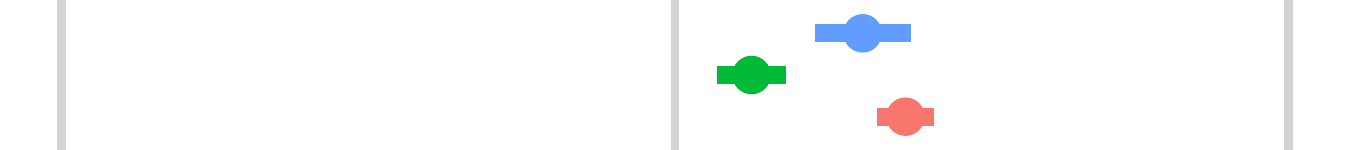} & 16 & 16 & 14 & \inlinegraphics{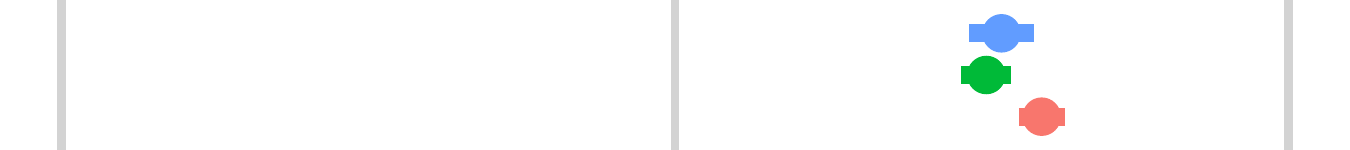} \\ 
\method{PgNetS} & {\cellcolor[HTML]{F0C0C0}{\textcolor[HTML]{000000}{Gr. Desc.}}} & NeuralNet & S & {\cellcolor[HTML]{FEBD6E}{\textcolor[HTML]{000000}{6m}}} & {\cellcolor[HTML]{D6EE89}{\textcolor[HTML]{000000}{11m}}} & 21 & 21 & 22 & \inlinegraphics{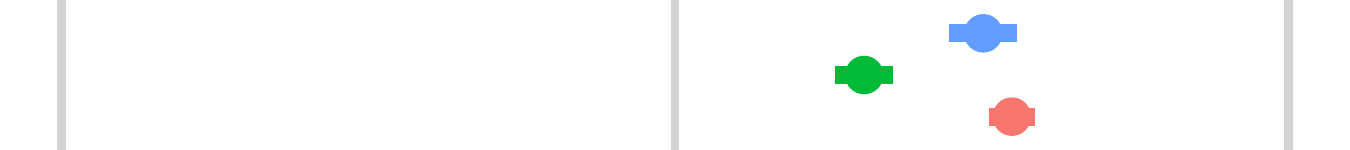} & 29 & 28 & 27 & \inlinegraphics{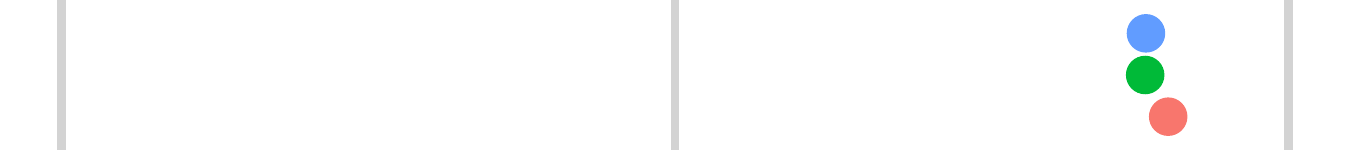} \\ 
\method{LlNn} & {\cellcolor[HTML]{00D8D8}{\textcolor[HTML]{000000}{Fit Solu.}}} & LocalLin & NN & {\cellcolor[HTML]{FFF4AC}{\textcolor[HTML]{000000}{1m}}} & {\cellcolor[HTML]{C9E880}{\textcolor[HTML]{000000}{9m}}} & 22 & 24 & 21 & \inlinegraphics{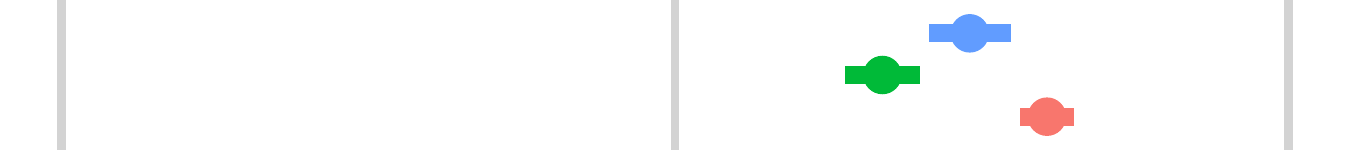} & 20 & 22 & 21 & \inlinegraphics{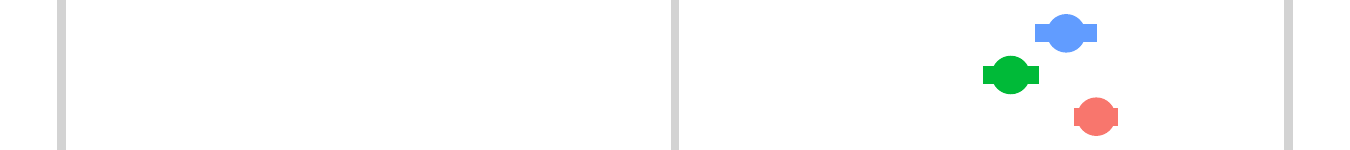} \\ 
\method{SpNn} & {\cellcolor[HTML]{00D8D8}{\textcolor[HTML]{000000}{Fit Solu.}}} & Spline &  & {\cellcolor[HTML]{49AC5A}{\textcolor[HTML]{FFFFFF}{3s}}} & {\cellcolor[HTML]{808080}{\textcolor[HTML]{FFFFFF}{}}} & 23 & 22 & 20 & \inlinegraphics{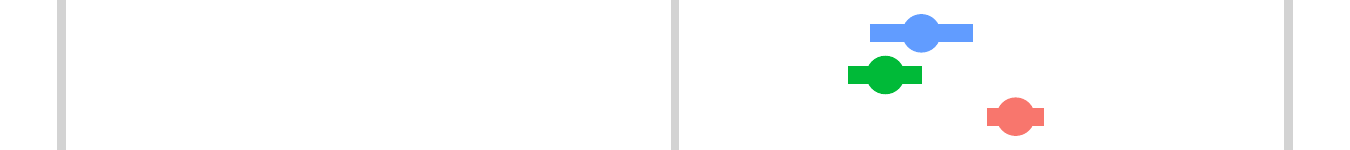} & 25 & 21 & 20 & \inlinegraphics{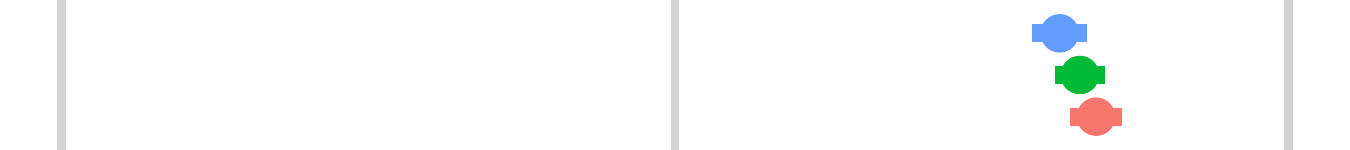} \\ 
\method{Node32} & {\cellcolor[HTML]{F0C0C0}{\textcolor[HTML]{000000}{Gr. Desc.}}} & NODE & bs32 & {\cellcolor[HTML]{F98F52}{\textcolor[HTML]{000000}{11m}}} & {\cellcolor[HTML]{FEB466}{\textcolor[HTML]{000000}{1h}}} & 24 & 23 & 26 & \inlinegraphics{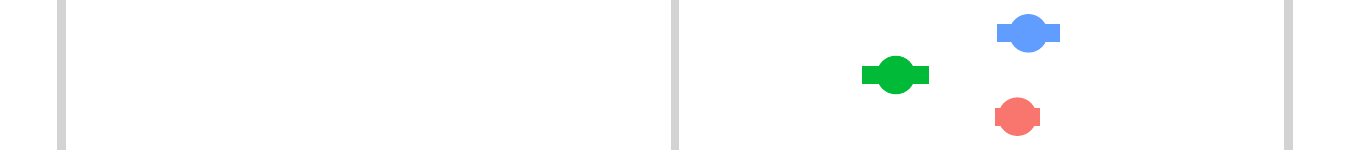} & 21 & 23 & 24 & \inlinegraphics{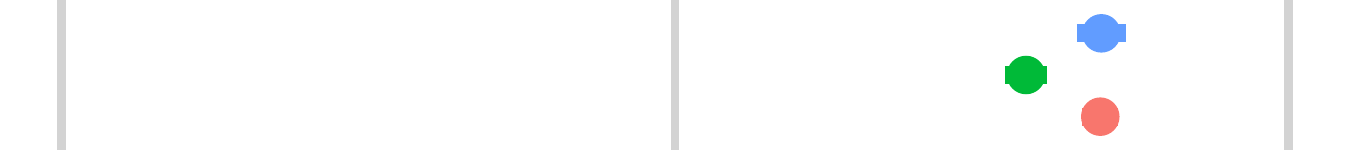} \\ 
\method{PwlNn} & {\cellcolor[HTML]{00D8D8}{\textcolor[HTML]{000000}{Fit Solu.}}} & PwLin & NN & {\cellcolor[HTML]{8DCD67}{\textcolor[HTML]{000000}{10s}}} & {\cellcolor[HTML]{808080}{\textcolor[HTML]{FFFFFF}{}}} & 25 & 26 & 23 & \inlinegraphics{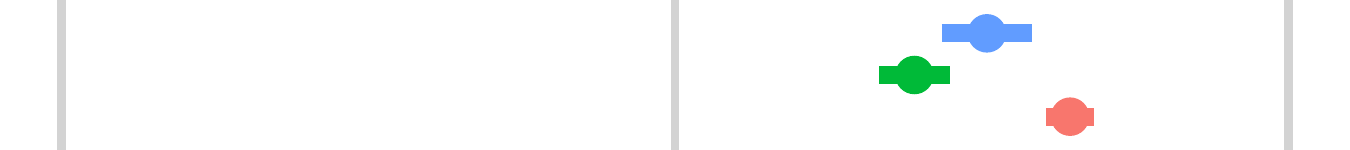} & 23 & 24 & 23 & \inlinegraphics{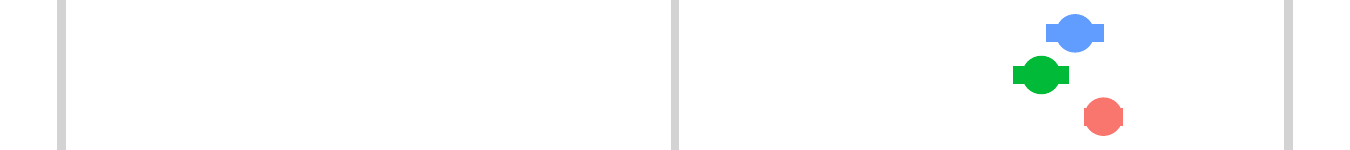} \\ 
\method{Node1} & {\cellcolor[HTML]{F0C0C0}{\textcolor[HTML]{000000}{Gr. Desc.}}} & NODE & bs1 & {\cellcolor[HTML]{A50026}{\textcolor[HTML]{FFFFFF}{1h}}} & {\cellcolor[HTML]{A50026}{\textcolor[HTML]{FFFFFF}{10h}}} & 26 & 25 & 25 & \inlinegraphics{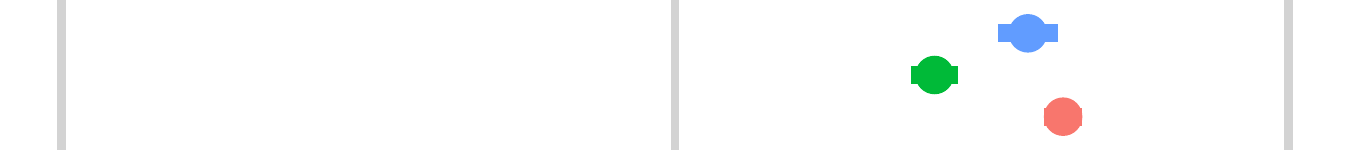} & 27 & 25 & 25 & \inlinegraphics{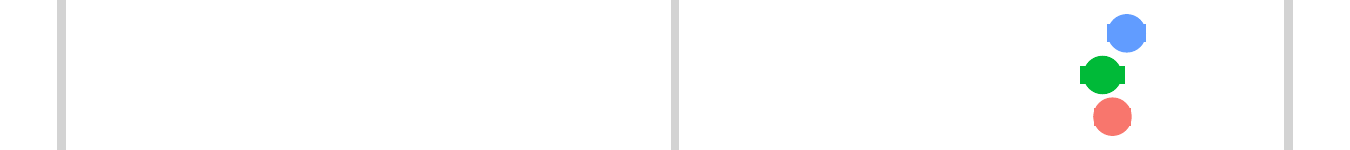} \\ 
\method{Analog} & {\cellcolor[HTML]{FFA0FF}{\textcolor[HTML]{000000}{Direct}}} & Analog &  & {\cellcolor[HTML]{006837}{\textcolor[HTML]{FFFFFF}{<1s}}} & {\cellcolor[HTML]{808080}{\textcolor[HTML]{FFFFFF}{}}} & 27 & 27 & 24 & \inlinegraphics{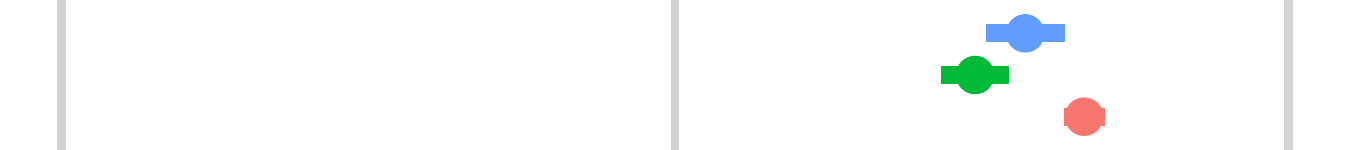} & 18 & 19 & 16 & \inlinegraphics{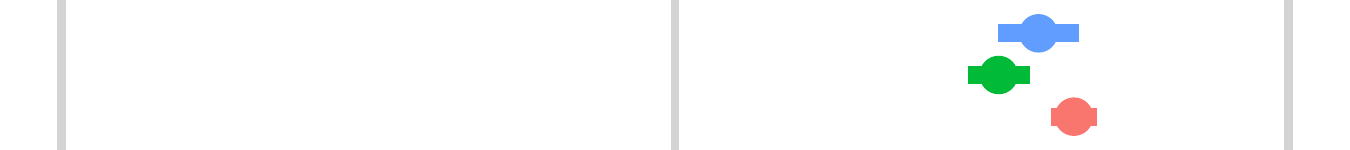} \\ 
\method{Gru} & {\cellcolor[HTML]{F0C0C0}{\textcolor[HTML]{000000}{Gr. Desc.}}} & GRU &  & {\cellcolor[HTML]{D52E27}{\textcolor[HTML]{FFFFFF}{44m}}} & {\cellcolor[HTML]{F57546}{\textcolor[HTML]{FFFFFF}{2h}}} & 28 & 28 & 27 & \inlinegraphics{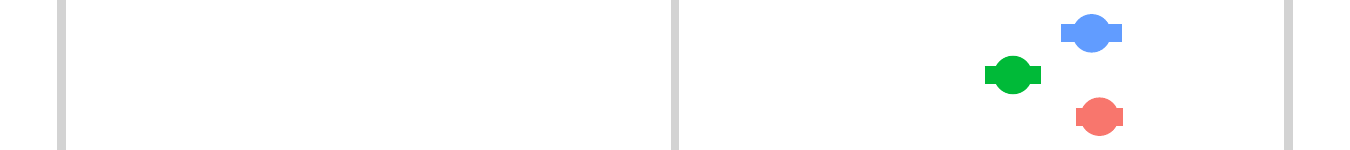} & 28 & 29 & 29 & \inlinegraphics{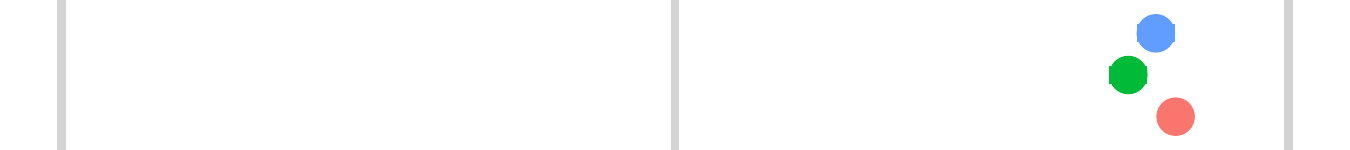} \\ 
\method{Trafo} & {\cellcolor[HTML]{F0C0C0}{\textcolor[HTML]{000000}{Gr. Desc.}}} & Transformer &  & {\cellcolor[HTML]{C52327}{\textcolor[HTML]{FFFFFF}{57m}}} & {\cellcolor[HTML]{E34C32}{\textcolor[HTML]{FFFFFF}{4h}}} & 29 & 29 & 29 & \inlinegraphics{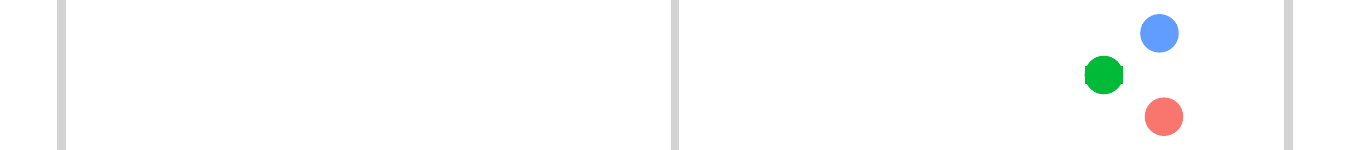} & 26 & 27 & 28 & \inlinegraphics{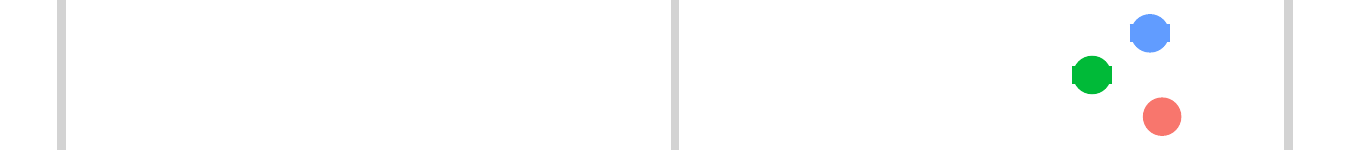} \\ 
\method{Rnn} & {\cellcolor[HTML]{F0C0C0}{\textcolor[HTML]{000000}{Gr. Desc.}}} & RNN &  & {\cellcolor[HTML]{F7824C}{\textcolor[HTML]{000000}{14m}}} & {\cellcolor[HTML]{FECD7B}{\textcolor[HTML]{000000}{59m}}} & 30 & 30 & 31 & \inlinegraphics{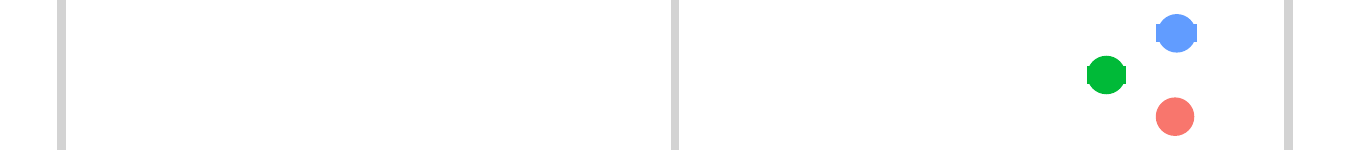} & 31 & 30 & 31 & \inlinegraphics{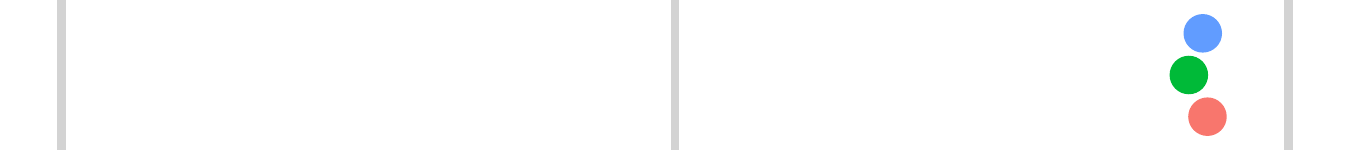} \\ 
\method{Lstm} & {\cellcolor[HTML]{F0C0C0}{\textcolor[HTML]{000000}{Gr. Desc.}}} & LSTM &  & {\cellcolor[HTML]{D12C27}{\textcolor[HTML]{FFFFFF}{46m}}} & {\cellcolor[HTML]{F77F4A}{\textcolor[HTML]{FFFFFF}{2h}}} & 31 & 31 & 30 & \inlinegraphics{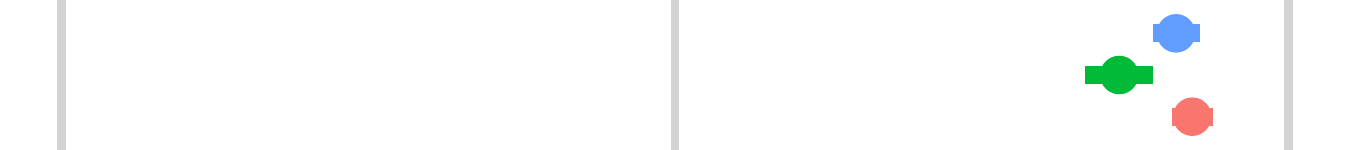} & 30 & 31 & 30 & \inlinegraphics{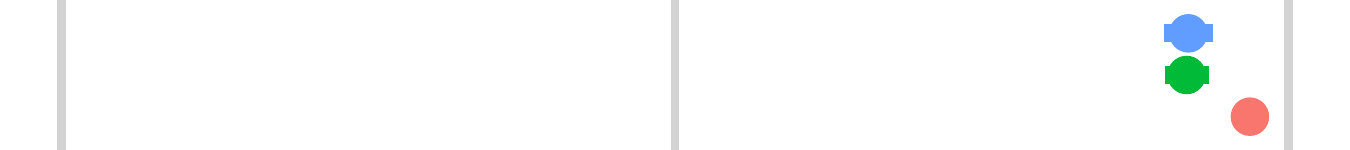} \\ 
\method{ConstL} & {\cellcolor[HTML]{FFA0FF}{\textcolor[HTML]{000000}{Direct}}} & Const & Last & {\cellcolor[HTML]{006837}{\textcolor[HTML]{FFFFFF}{<1s}}} & {\cellcolor[HTML]{808080}{\textcolor[HTML]{FFFFFF}{}}} & 32 & 32 & 32 & \inlinegraphics{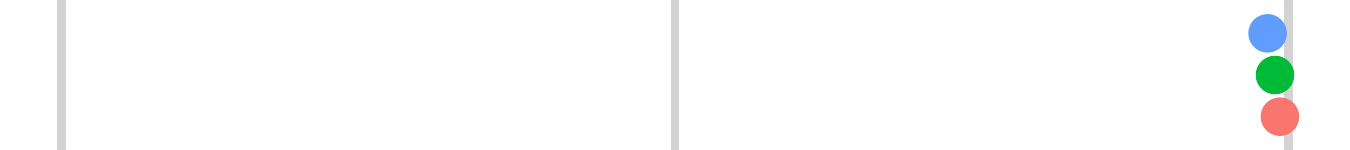} & 32 & 32 & 32 & \inlinegraphics{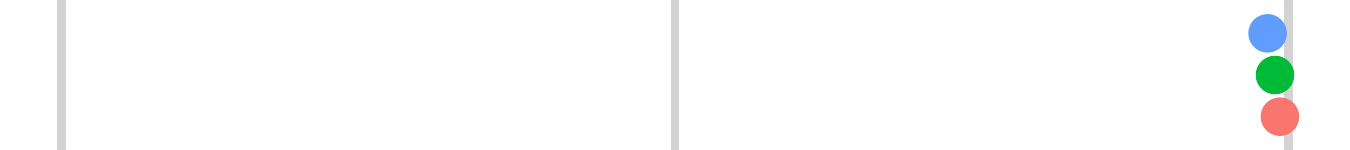} \\ 
\method{ConstM} & {\cellcolor[HTML]{FFA0FF}{\textcolor[HTML]{000000}{Direct}}} & Const & Mean & {\cellcolor[HTML]{006837}{\textcolor[HTML]{FFFFFF}{<1s}}} & {\cellcolor[HTML]{808080}{\textcolor[HTML]{FFFFFF}{}}} & 33 & 33 & 33 & \inlinegraphics{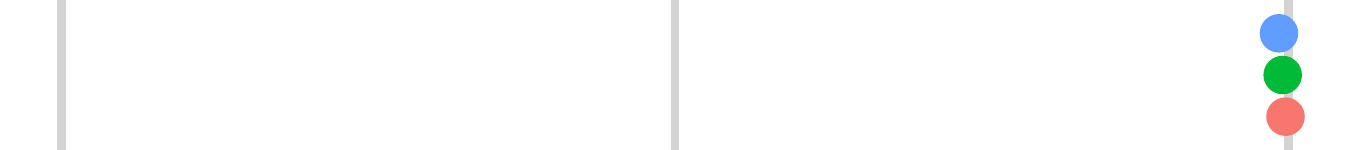} & 33 & 33 & 33 & \inlinegraphics{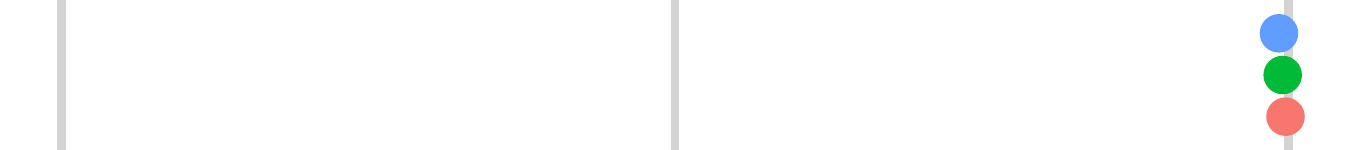} \\ 
\bottomrule
\end{tabular}
\end{center}

%% file: tbl/tableLorenzRndResults.tex
\begin{center}
\caption*{
{\large Cumulative Maximum Error for Test Data of \DeebLorenz{} with Random Timestep}
} 
\fontsize{8.0pt}{10pt}\selectfont
\fontfamily{phv}\selectfont
\renewcommand{\arraystretch}{1.05}
\setlength{\tabcolsep}{0.3em}
\rowcolors{2}{gray!20}{white}
\begin{tabular}{lcllrrrrrcrrrc}
\toprule
\multicolumn{4}{c}{Method} & \multicolumn{2}{c}{Compute} & \multicolumn{4}{c}{Noisefree} & \multicolumn{4}{c}{Noisy} \\ 
\cmidrule(lr){1-4} \cmidrule(lr){5-6} \cmidrule(lr){7-10} \cmidrule(lr){11-14}
Name & Class & Model & Variant & Test & Tune & \cellcolor[HTML]{00BA38}{S\#} & \cellcolor[HTML]{F8766D}{R\#} & \cellcolor[HTML]{619CFF}{N\#} & \CmeScale{} & \cellcolor[HTML]{00BA38}{S\#} & \cellcolor[HTML]{F8766D}{R\#} & \cellcolor[HTML]{619CFF}{N\#} & \CmeScale{} \\ 
\midrule\addlinespace[2.5pt]
\method{GpGp} & {\cellcolor[HTML]{00D8D8}{\textcolor[HTML]{000000}{Fit Solu.}}} & GP & GP & {\cellcolor[HTML]{B5E074}{\textcolor[HTML]{000000}{19s}}} & {\cellcolor[HTML]{F0F9AA}{\textcolor[HTML]{000000}{16m}}} & \textbf{\underline{1}} & \textbf{\underline{1}} & \textbf{\underline{1}} & \inlinegraphics{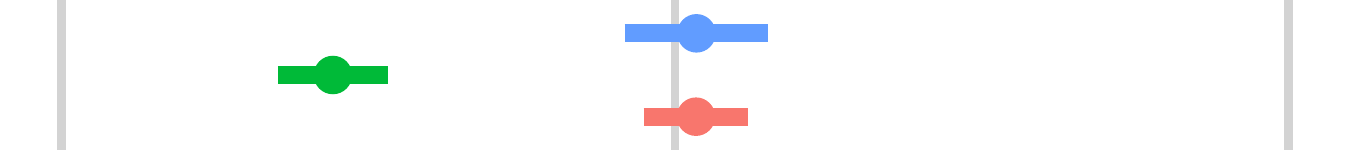} & \textbf{\underline{1}} & \textbf{\underline{1}} & \textbf{2} & \inlinegraphics{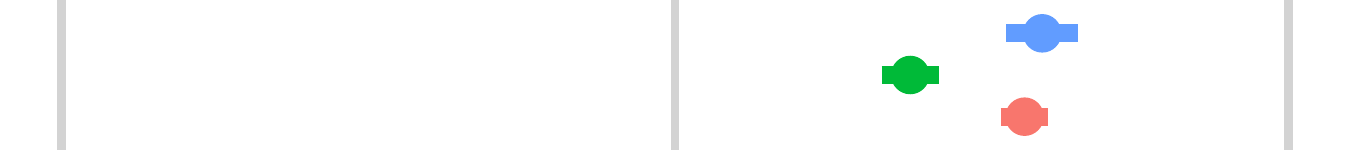} \\ 
\method{SpPo2} & {\cellcolor[HTML]{00D8D8}{\textcolor[HTML]{000000}{Fit Solu.}}} & Spline & Poly2 & {\cellcolor[HTML]{3AA556}{\textcolor[HTML]{FFFFFF}{2s}}} & {\cellcolor[HTML]{808080}{\textcolor[HTML]{FFFFFF}{}}} & \textbf{2} & \textbf{2} & 19 & \inlinegraphics{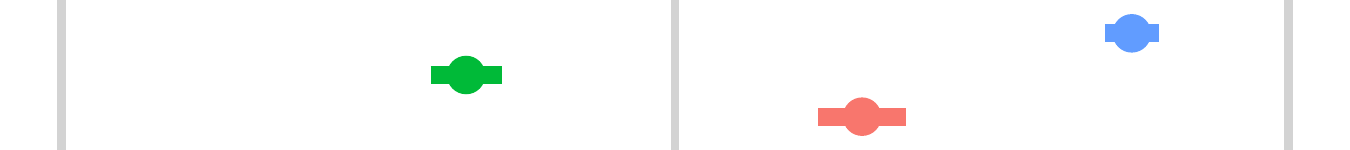} & 21 & 21 & 25 & \inlinegraphics{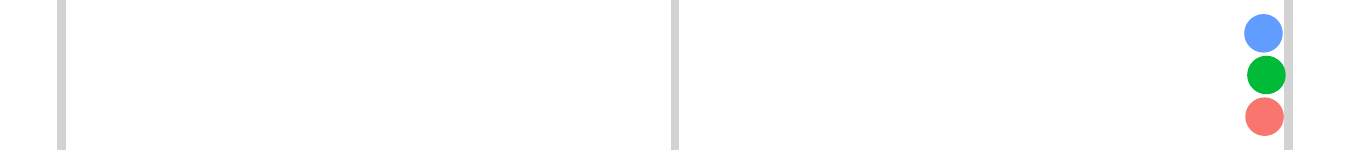} \\ 
\method{SpPo4} & {\cellcolor[HTML]{00D8D8}{\textcolor[HTML]{000000}{Fit Solu.}}} & Spline & Poly4 & {\cellcolor[HTML]{59B55F}{\textcolor[HTML]{FFFFFF}{4s}}} & {\cellcolor[HTML]{808080}{\textcolor[HTML]{FFFFFF}{}}} & \underline{3} & 4 & \underline{3} & \inlinegraphics{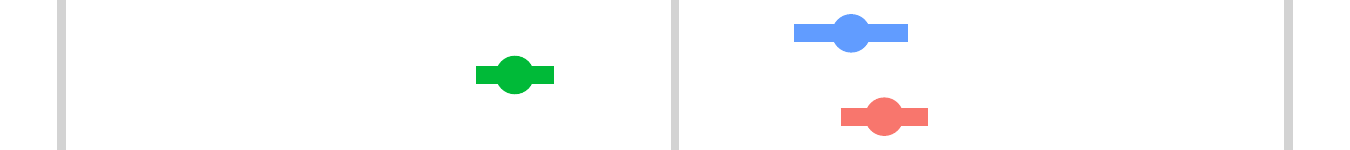} & 20 & 19 & 24 & \inlinegraphics{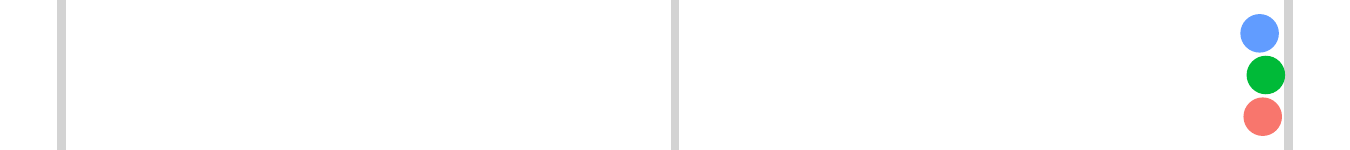} \\ 
\method{SpPo} & {\cellcolor[HTML]{00D8D8}{\textcolor[HTML]{000000}{Fit Solu.}}} & Spline & Poly & {\cellcolor[HTML]{5DB760}{\textcolor[HTML]{000000}{5s}}} & {\cellcolor[HTML]{77C465}{\textcolor[HTML]{000000}{3m}}} & 4 & \underline{3} & 6 & \inlinegraphics{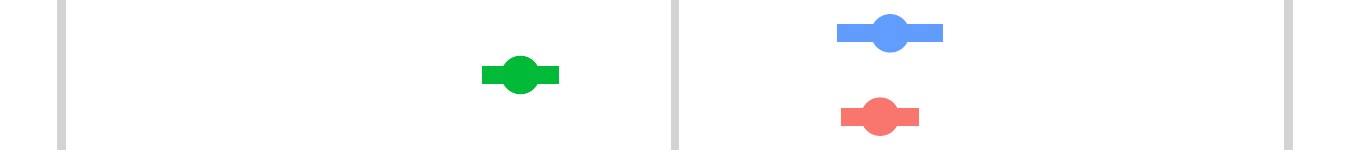} & 22 & 20 & 23 & \inlinegraphics{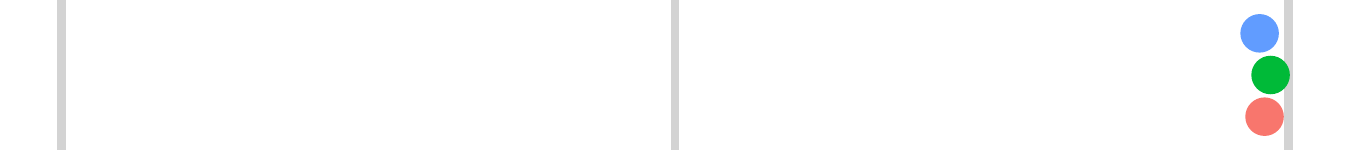} \\ 
\method{SINDy} & {\cellcolor[HTML]{00D8D8}{\textcolor[HTML]{000000}{Fit Solu.}}} & SINDy &  & {\cellcolor[HTML]{69BE63}{\textcolor[HTML]{000000}{6s}}} & {\cellcolor[HTML]{036F3B}{\textcolor[HTML]{FFFFFF}{53s}}} & 5 & 6 & 4 & \inlinegraphics{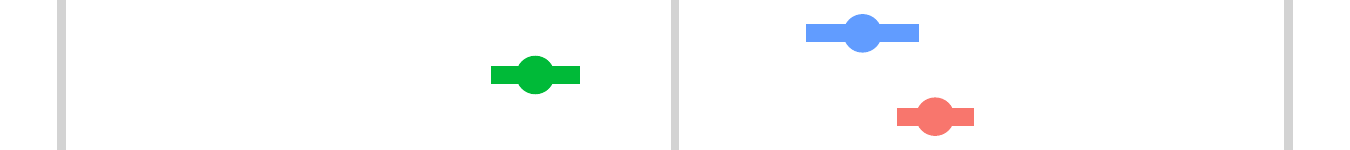} & 23 & 22 & 22 & \inlinegraphics{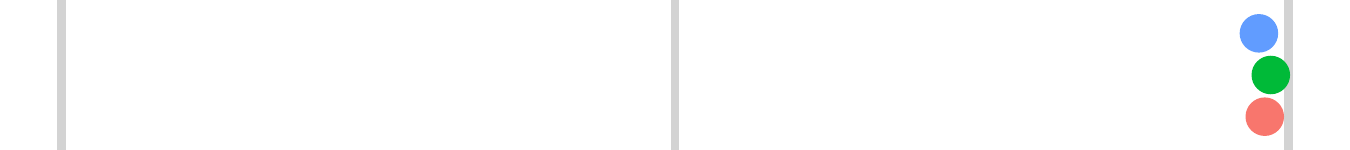} \\ 
\method{SINDyN} & {\cellcolor[HTML]{00D8D8}{\textcolor[HTML]{000000}{Fit Solu.}}} & SINDy & norm & {\cellcolor[HTML]{64BC62}{\textcolor[HTML]{000000}{5s}}} & {\cellcolor[HTML]{006837}{\textcolor[HTML]{FFFFFF}{47s}}} & 6 & 8 & 5 & \inlinegraphics{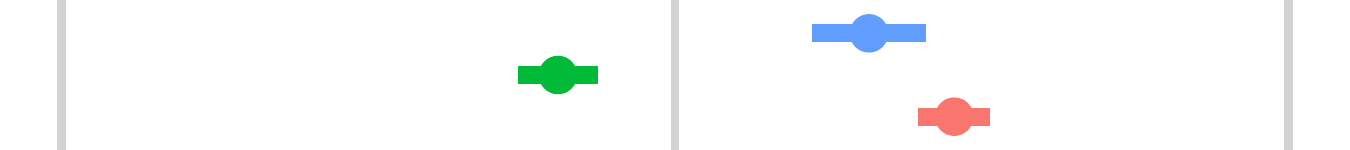} & 24 & 24 & 21 & \inlinegraphics{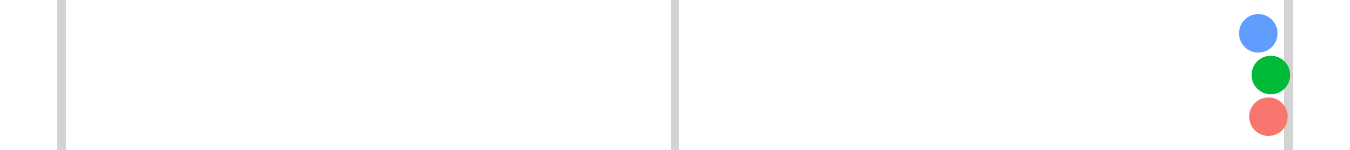} \\ 
\method{SpGp} & {\cellcolor[HTML]{00D8D8}{\textcolor[HTML]{000000}{Fit Solu.}}} & Spline & GP & {\cellcolor[HTML]{8BCD67}{\textcolor[HTML]{000000}{10s}}} & {\cellcolor[HTML]{7DC765}{\textcolor[HTML]{000000}{3m}}} & 7 & 5 & \textbf{2} & \inlinegraphics{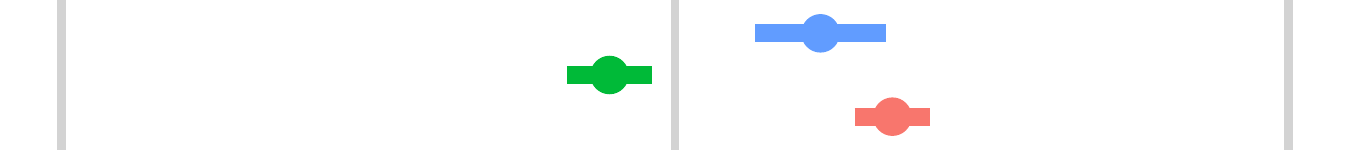} & 18 & 12 & 17 & \inlinegraphics{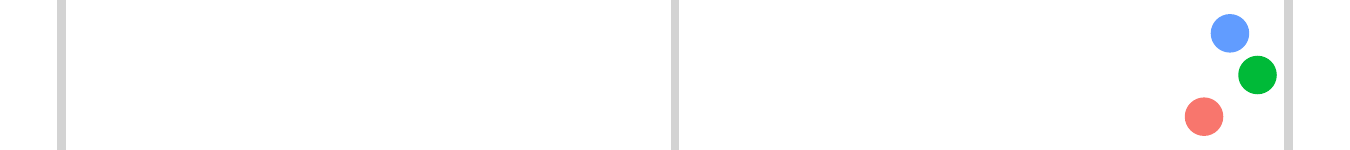} \\ 
\method{PgGpDT} & {\cellcolor[HTML]{B0B0FF}{\textcolor[HTML]{000000}{Fit Prop.}}} & GP & DT & {\cellcolor[HTML]{7DC765}{\textcolor[HTML]{000000}{8s}}} & {\cellcolor[HTML]{FFEB9D}{\textcolor[HTML]{000000}{32m}}} & 8 & 7 & 8 & \inlinegraphics{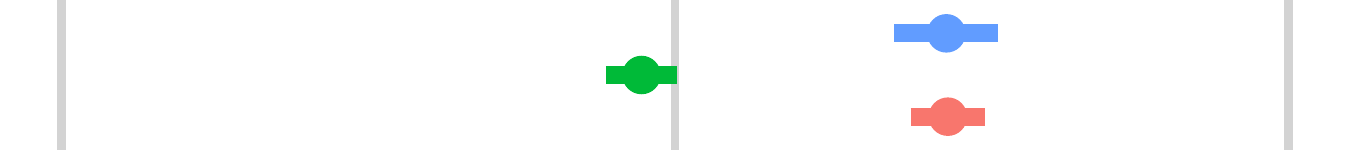} & 28 & 28 & 28 & \inlinegraphics{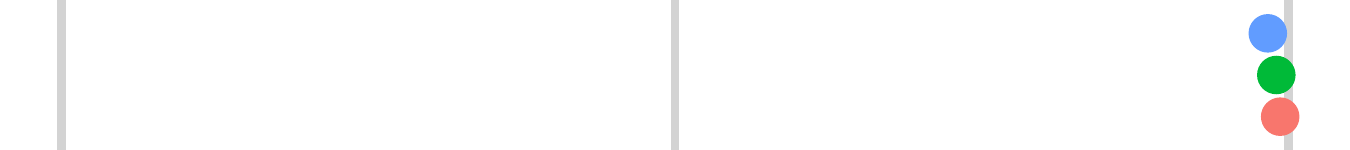} \\ 
\method{LinDT} & {\cellcolor[HTML]{B0B0FF}{\textcolor[HTML]{000000}{Fit Prop.}}} & Lin & DT & {\cellcolor[HTML]{5BB660}{\textcolor[HTML]{000000}{4s}}} & {\cellcolor[HTML]{B0DD71}{\textcolor[HTML]{000000}{6m}}} & 9 & 14 & 12 & \inlinegraphics{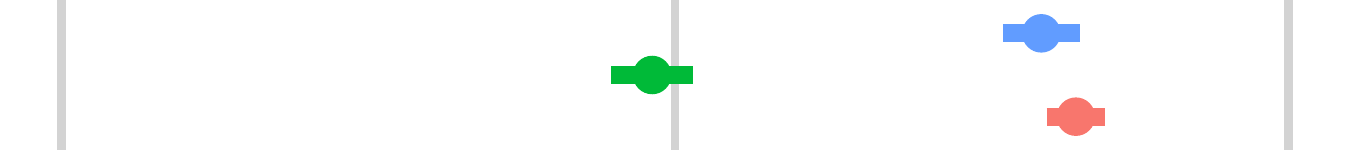} & 25 & 30 & 20 & \inlinegraphics{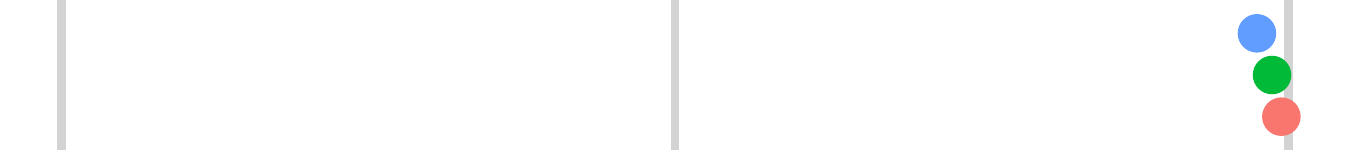} \\ 
\method{LinPo4T} & {\cellcolor[HTML]{B0B0FF}{\textcolor[HTML]{000000}{Fit Prop.}}} & Lin & Poly4T & {\cellcolor[HTML]{108546}{\textcolor[HTML]{FFFFFF}{1s}}} & {\cellcolor[HTML]{808080}{\textcolor[HTML]{FFFFFF}{}}} & 10 & 10 & 10 & \inlinegraphics{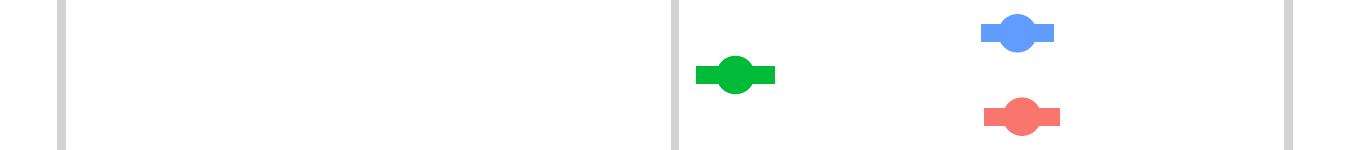} & 29 & 26 & 30 & \inlinegraphics{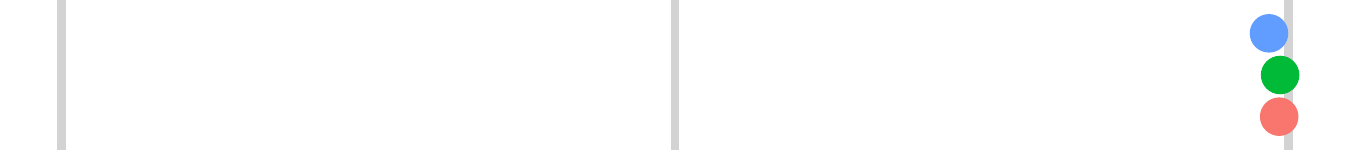} \\ 
\method{LinPo6T} & {\cellcolor[HTML]{B0B0FF}{\textcolor[HTML]{000000}{Fit Prop.}}} & Lin & Poly6T & {\cellcolor[HTML]{35A255}{\textcolor[HTML]{FFFFFF}{2s}}} & {\cellcolor[HTML]{808080}{\textcolor[HTML]{FFFFFF}{}}} & 11 & 12 & 18 & \inlinegraphics{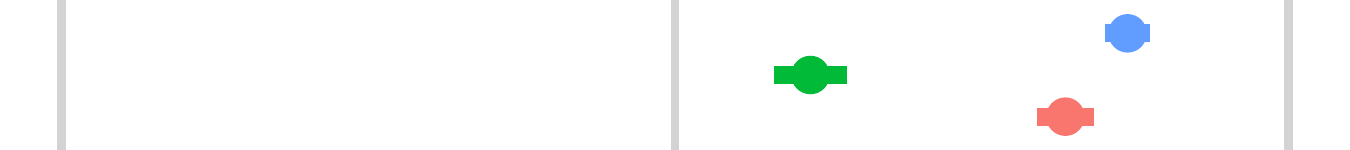} & 32 & 31 & 26 & \inlinegraphics{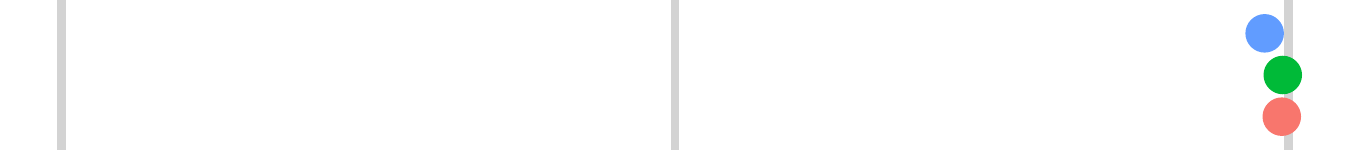} \\ 
\method{EsnDT} & {\cellcolor[HTML]{B0B0FF}{\textcolor[HTML]{000000}{Fit Prop.}}} & ESN & DT & {\cellcolor[HTML]{65BC63}{\textcolor[HTML]{000000}{5s}}} & {\cellcolor[HTML]{FFF2A9}{\textcolor[HTML]{000000}{28m}}} & 12 & 17 & 16 & \inlinegraphics{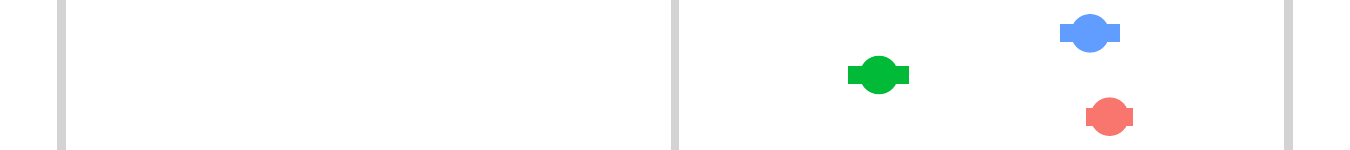} & 4 & 5 & 5 & \inlinegraphics{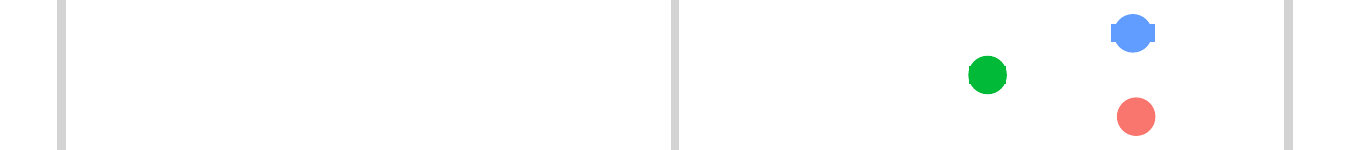} \\ 
\method{PgNetDT} & {\cellcolor[HTML]{F0C0C0}{\textcolor[HTML]{000000}{Gr. Desc.}}} & NeuralNet & DT & {\cellcolor[HTML]{FEBE6E}{\textcolor[HTML]{000000}{6m}}} & {\cellcolor[HTML]{F4FAAF}{\textcolor[HTML]{000000}{17m}}} & 13 & 13 & 11 & \inlinegraphics{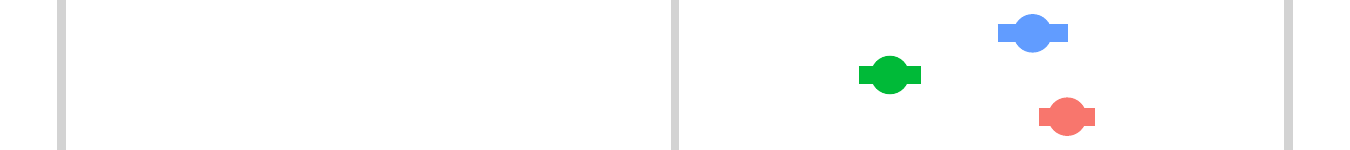} & 26 & 25 & 29 & \inlinegraphics{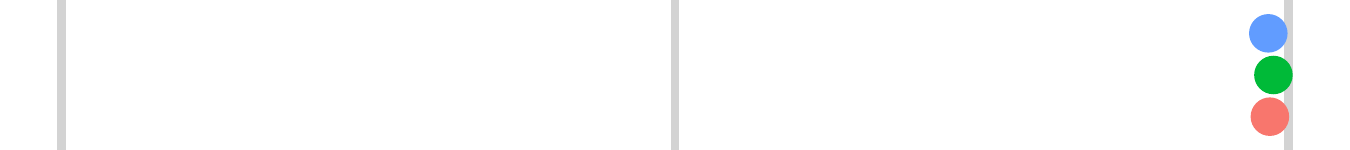} \\ 
\method{SpNn} & {\cellcolor[HTML]{00D8D8}{\textcolor[HTML]{000000}{Fit Solu.}}} & Spline &  & {\cellcolor[HTML]{44A959}{\textcolor[HTML]{FFFFFF}{3s}}} & {\cellcolor[HTML]{808080}{\textcolor[HTML]{FFFFFF}{}}} & 14 & 9 & 7 & \inlinegraphics{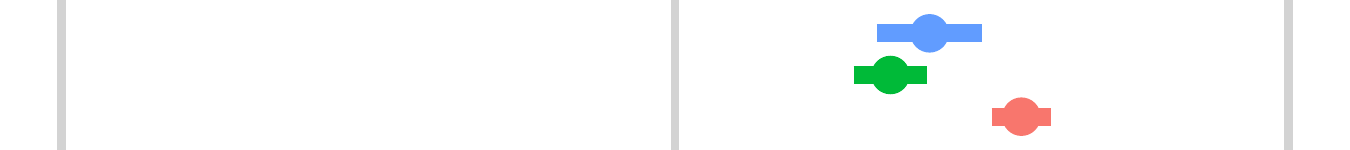} & 16 & 14 & 15 & \inlinegraphics{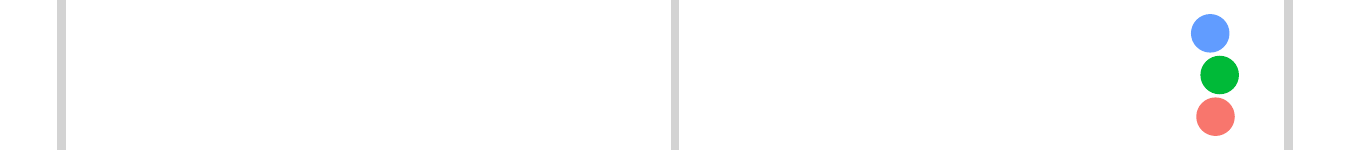} \\ 
\method{PgLlDT} & {\cellcolor[HTML]{B0B0FF}{\textcolor[HTML]{000000}{Fit Prop.}}} & LocalLin & DT & {\cellcolor[HTML]{79C565}{\textcolor[HTML]{000000}{7s}}} & {\cellcolor[HTML]{B7E075}{\textcolor[HTML]{000000}{7m}}} & 15 & 18 & 15 & \inlinegraphics{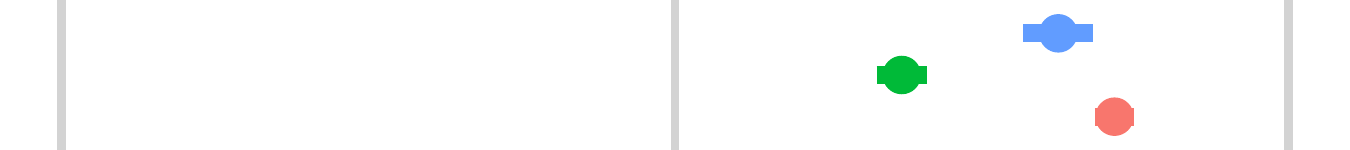} & 30 & 29 & 31 & \inlinegraphics{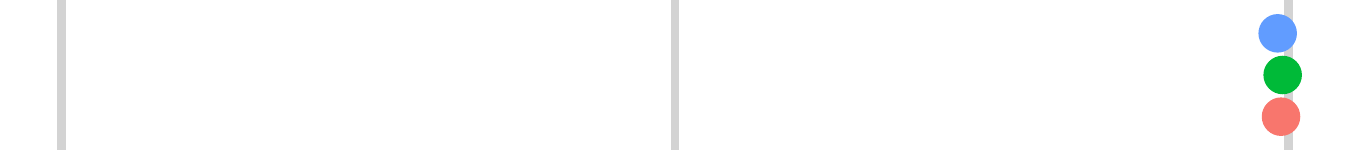} \\ 
\method{RaFeDT} & {\cellcolor[HTML]{B0B0FF}{\textcolor[HTML]{000000}{Fit Prop.}}} & RandFeat & DT & {\cellcolor[HTML]{5EB861}{\textcolor[HTML]{000000}{5s}}} & {\cellcolor[HTML]{FFF4AC}{\textcolor[HTML]{000000}{27m}}} & 16 & 15 & 14 & \inlinegraphics{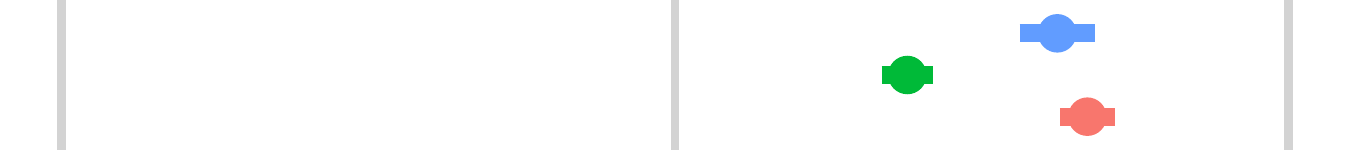} & \underline{3} & 6 & \underline{3} & \inlinegraphics{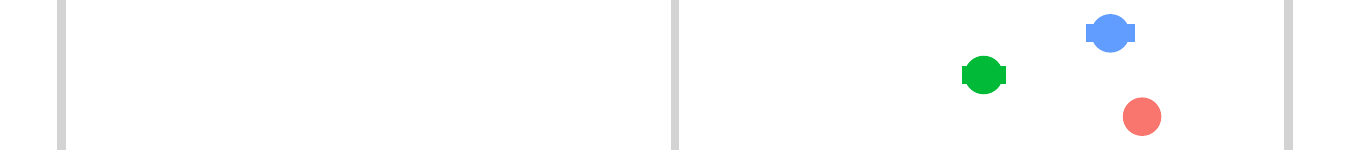} \\ 
\method{Node1} & {\cellcolor[HTML]{F0C0C0}{\textcolor[HTML]{000000}{Gr. Desc.}}} & NODE & bs1 & {\cellcolor[HTML]{A50026}{\textcolor[HTML]{FFFFFF}{1h}}} & {\cellcolor[HTML]{A50026}{\textcolor[HTML]{FFFFFF}{9h}}} & 17 & 11 & 13 & \inlinegraphics{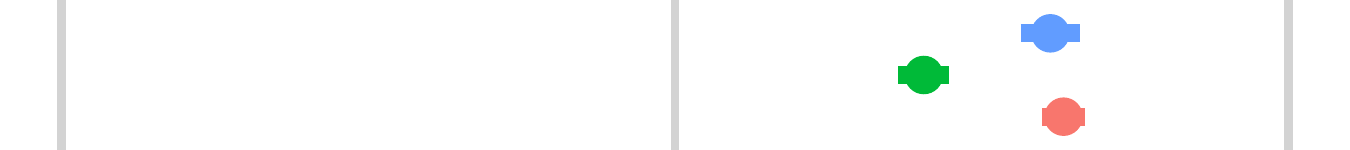} & 7 & 4 & 4 & \inlinegraphics{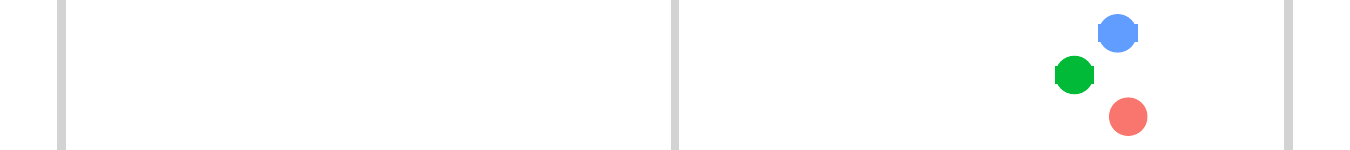} \\ 
\method{LinST} & {\cellcolor[HTML]{B0B0FF}{\textcolor[HTML]{000000}{Fit Prop.}}} & Lin & ST & {\cellcolor[HTML]{9AD369}{\textcolor[HTML]{000000}{12s}}} & {\cellcolor[HTML]{DFF193}{\textcolor[HTML]{000000}{12m}}} & 18 & 19 & 23 & \inlinegraphics{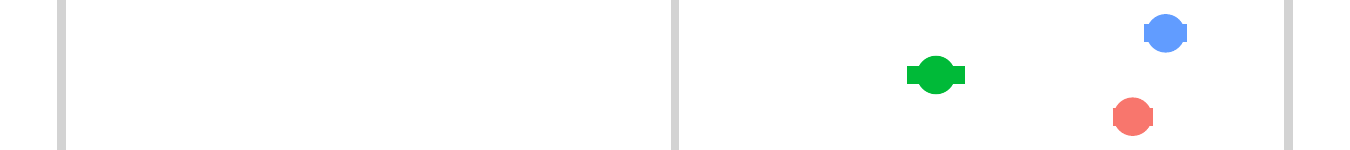} & 5 & \textbf{2} & 8 & \inlinegraphics{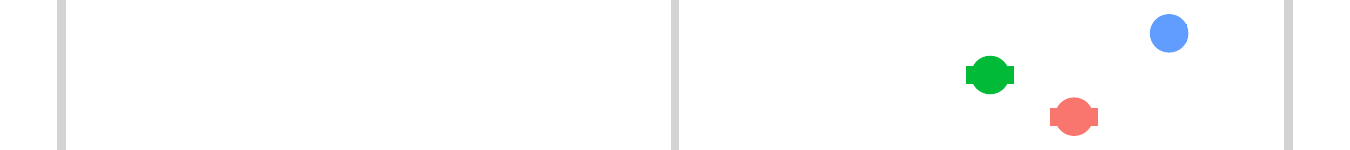} \\ 
\method{PwlNn} & {\cellcolor[HTML]{00D8D8}{\textcolor[HTML]{000000}{Fit Solu.}}} & PwLin & NN & {\cellcolor[HTML]{8FCE67}{\textcolor[HTML]{000000}{10s}}} & {\cellcolor[HTML]{808080}{\textcolor[HTML]{FFFFFF}{}}} & 19 & 20 & 17 & \inlinegraphics{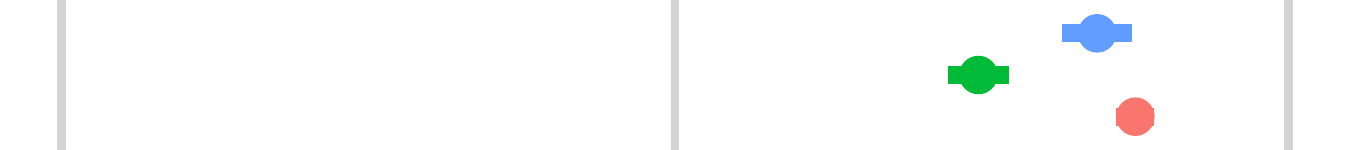} & 12 & 11 & 11 & \inlinegraphics{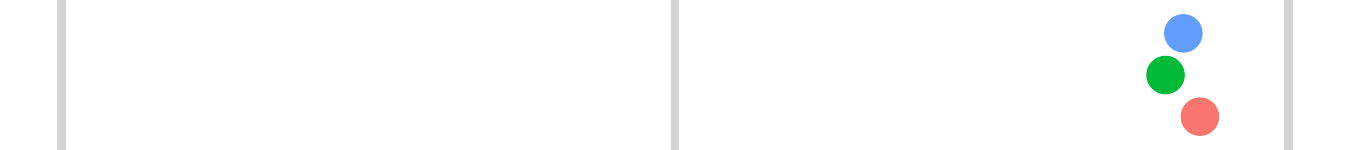} \\ 
\method{PgGpST} & {\cellcolor[HTML]{B0B0FF}{\textcolor[HTML]{000000}{Fit Prop.}}} & GP & ST & {\cellcolor[HTML]{A6D96A}{\textcolor[HTML]{000000}{15s}}} & {\cellcolor[HTML]{FFEDA1}{\textcolor[HTML]{000000}{31m}}} & 20 & 21 & 22 & \inlinegraphics{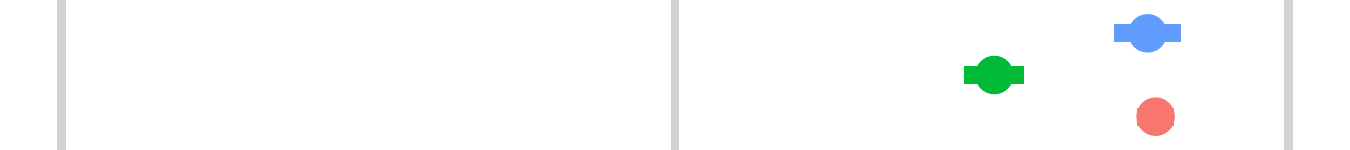} & 10 & 8 & 7 & \inlinegraphics{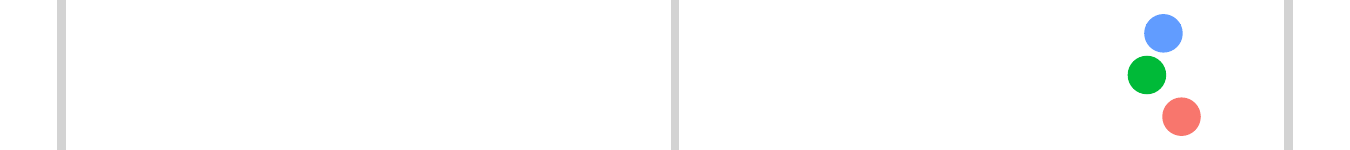} \\ 
\method{Analog} & {\cellcolor[HTML]{FFA0FF}{\textcolor[HTML]{000000}{Direct}}} & Analog &  & {\cellcolor[HTML]{006937}{\textcolor[HTML]{FFFFFF}{<1s}}} & {\cellcolor[HTML]{808080}{\textcolor[HTML]{FFFFFF}{}}} & 21 & 16 & 9 & \inlinegraphics{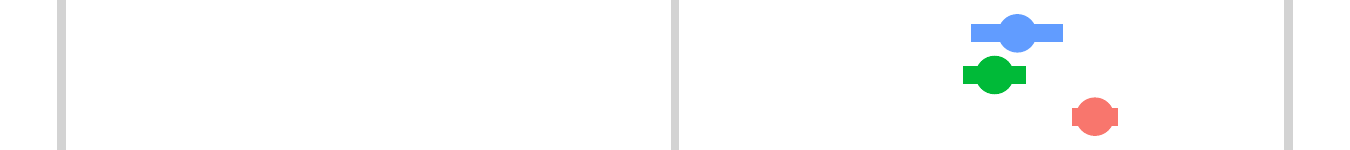} & 6 & \underline{3} & \textbf{\underline{1}} & \inlinegraphics{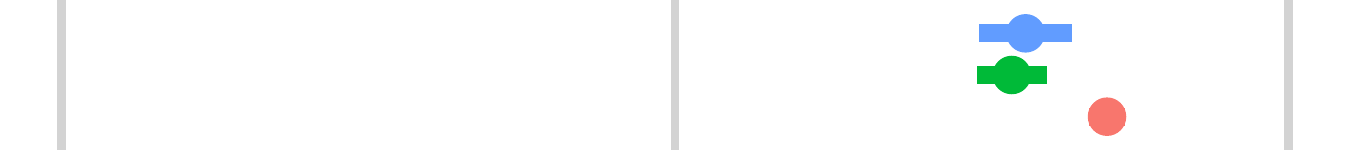} \\ 
\method{LlNn} & {\cellcolor[HTML]{00D8D8}{\textcolor[HTML]{000000}{Fit Solu.}}} & LocalLin & NN & {\cellcolor[HTML]{FFF5AE}{\textcolor[HTML]{000000}{1m}}} & {\cellcolor[HTML]{C9E881}{\textcolor[HTML]{000000}{9m}}} & 22 & 23 & 21 & \inlinegraphics{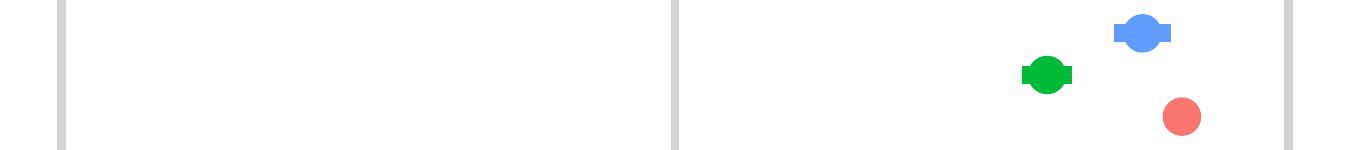} & 8 & 10 & 10 & \inlinegraphics{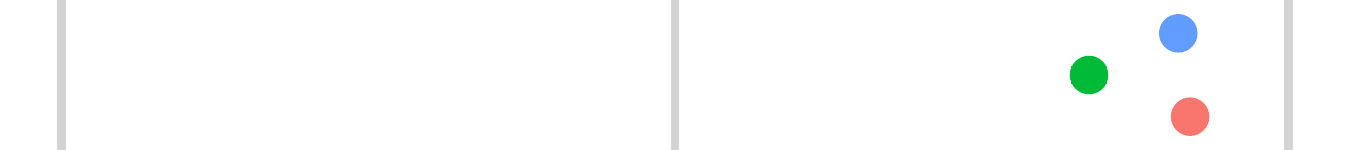} \\ 
\method{PgNetST} & {\cellcolor[HTML]{F0C0C0}{\textcolor[HTML]{000000}{Gr. Desc.}}} & NeuralNet & ST & {\cellcolor[HTML]{FCA55C}{\textcolor[HTML]{000000}{9m}}} & {\cellcolor[HTML]{FFFBB8}{\textcolor[HTML]{000000}{23m}}} & 23 & 22 & 20 & \inlinegraphics{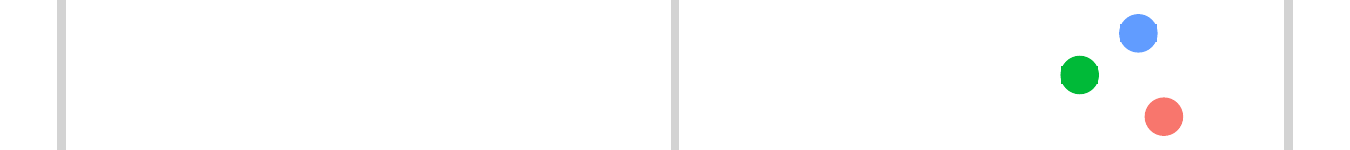} & 11 & 9 & 9 & \inlinegraphics{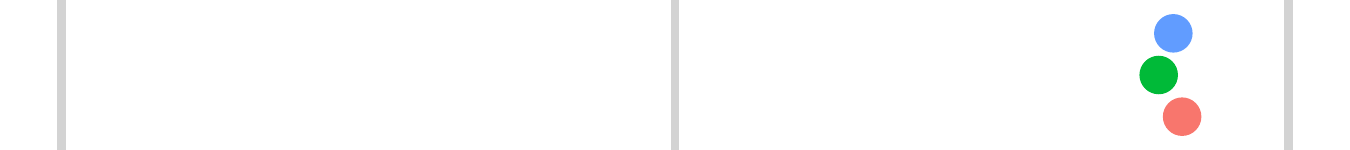} \\ 
\method{RaFeST} & {\cellcolor[HTML]{B0B0FF}{\textcolor[HTML]{000000}{Fit Prop.}}} & RandFeat & ST & {\cellcolor[HTML]{5EB860}{\textcolor[HTML]{000000}{5s}}} & {\cellcolor[HTML]{FFEFA3}{\textcolor[HTML]{000000}{30m}}} & 24 & 25 & 24 & \inlinegraphics{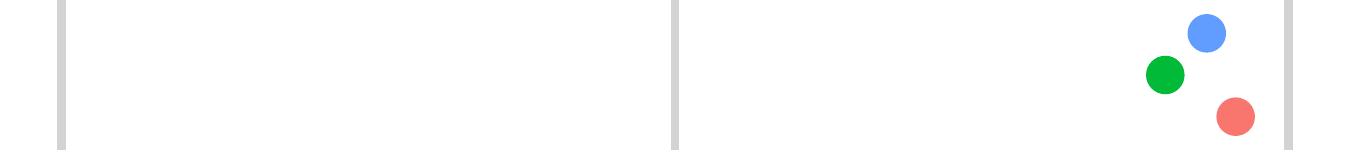} & \textbf{2} & 7 & 6 & \inlinegraphics{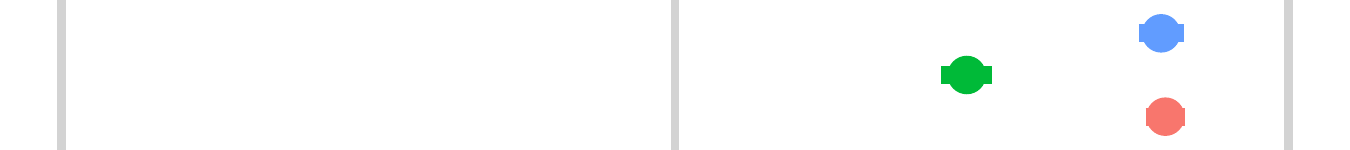} \\ 
\method{PgLlST} & {\cellcolor[HTML]{B0B0FF}{\textcolor[HTML]{000000}{Fit Prop.}}} & LocalLin & ST & {\cellcolor[HTML]{87CB67}{\textcolor[HTML]{000000}{9s}}} & {\cellcolor[HTML]{A8DA6C}{\textcolor[HTML]{000000}{5m}}} & 25 & 24 & 26 & \inlinegraphics{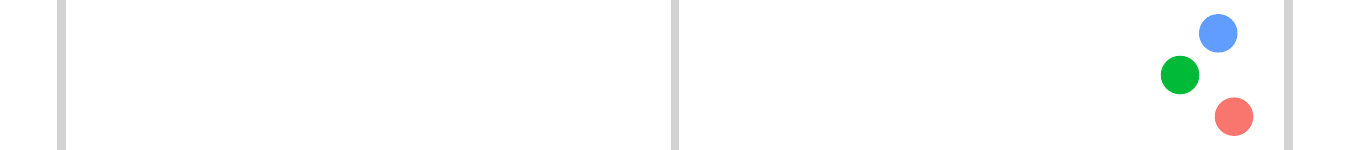} & 13 & 16 & 14 & \inlinegraphics{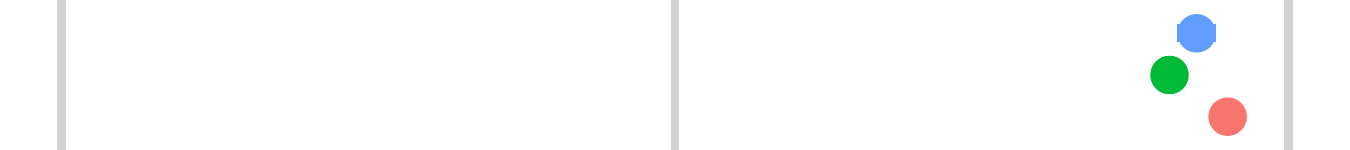} \\ 
\method{EsnST} & {\cellcolor[HTML]{B0B0FF}{\textcolor[HTML]{000000}{Fit Prop.}}} & ESN & ST & {\cellcolor[HTML]{68BE63}{\textcolor[HTML]{000000}{6s}}} & {\cellcolor[HTML]{FFF0A6}{\textcolor[HTML]{000000}{29m}}} & 26 & 26 & 25 & \inlinegraphics{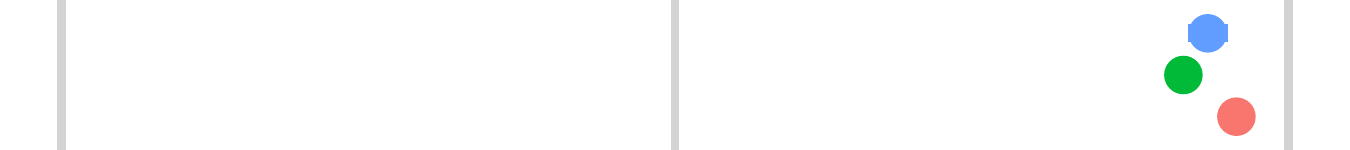} & 9 & 13 & 12 & \inlinegraphics{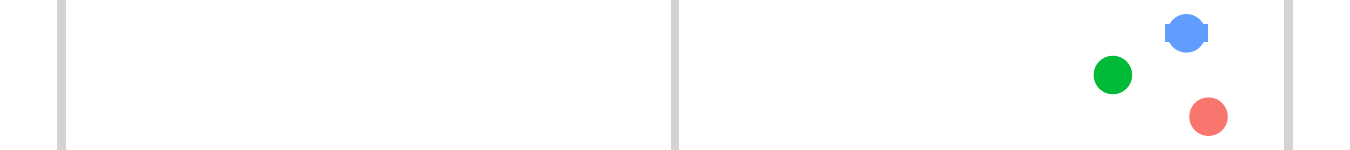} \\ 
\method{TrafoT} & {\cellcolor[HTML]{F0C0C0}{\textcolor[HTML]{000000}{Gr. Desc.}}} & Transformer & T & {\cellcolor[HTML]{A50026}{\textcolor[HTML]{FFFFFF}{1h}}} & {\cellcolor[HTML]{BF1E27}{\textcolor[HTML]{FFFFFF}{6h}}} & 27 & 28 & 28 & \inlinegraphics{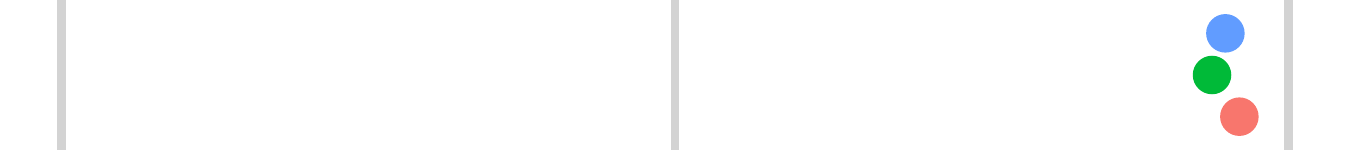} & 15 & 17 & 16 & \inlinegraphics{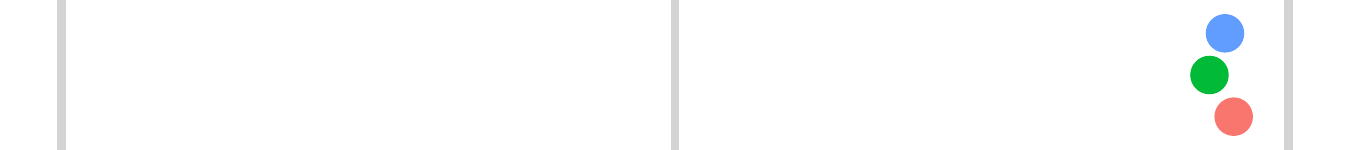} \\ 
\method{GruT} & {\cellcolor[HTML]{F0C0C0}{\textcolor[HTML]{000000}{Gr. Desc.}}} & GRU & T & {\cellcolor[HTML]{D73127}{\textcolor[HTML]{FFFFFF}{44m}}} & {\cellcolor[HTML]{F67B49}{\textcolor[HTML]{FFFFFF}{2h}}} & 28 & 27 & 27 & \inlinegraphics{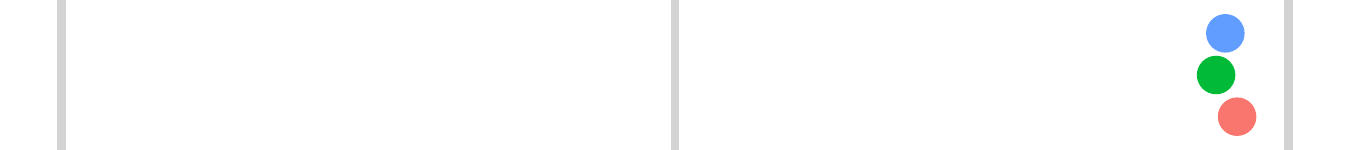} & 14 & 15 & 13 & \inlinegraphics{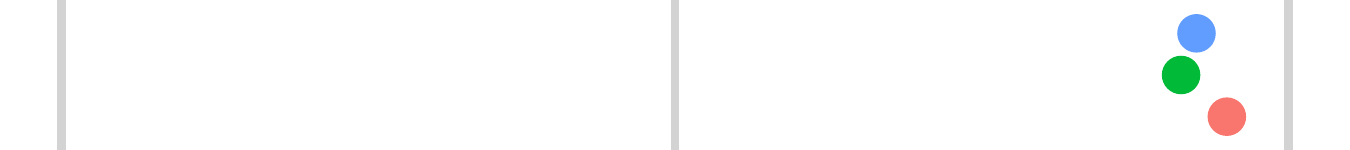} \\ 
\method{RnnT} & {\cellcolor[HTML]{F0C0C0}{\textcolor[HTML]{000000}{Gr. Desc.}}} & RNN & T & {\cellcolor[HTML]{F77C49}{\textcolor[HTML]{FFFFFF}{15m}}} & {\cellcolor[HTML]{FEC675}{\textcolor[HTML]{000000}{58m}}} & 29 & 29 & 29 & \inlinegraphics{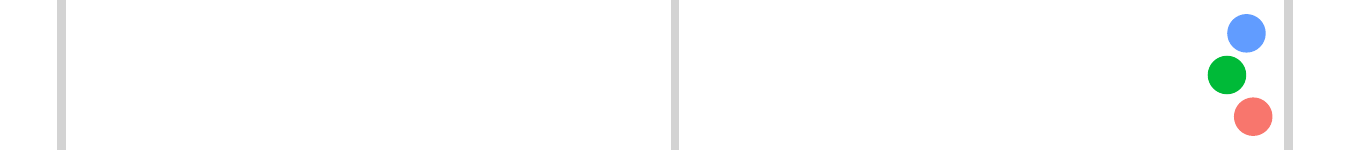} & 17 & 18 & 18 & \inlinegraphics{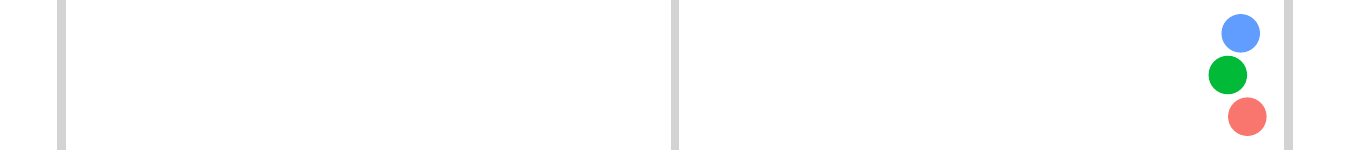} \\ 
\method{LstmT} & {\cellcolor[HTML]{F0C0C0}{\textcolor[HTML]{000000}{Gr. Desc.}}} & LSTM & T & {\cellcolor[HTML]{CE2927}{\textcolor[HTML]{FFFFFF}{52m}}} & {\cellcolor[HTML]{F67B49}{\textcolor[HTML]{FFFFFF}{2h}}} & 30 & 30 & 30 & \inlinegraphics{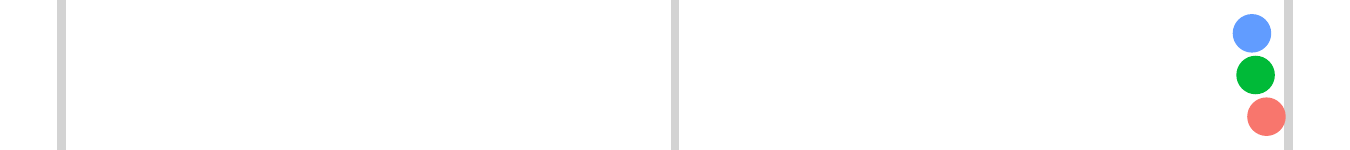} & 19 & 23 & 19 & \inlinegraphics{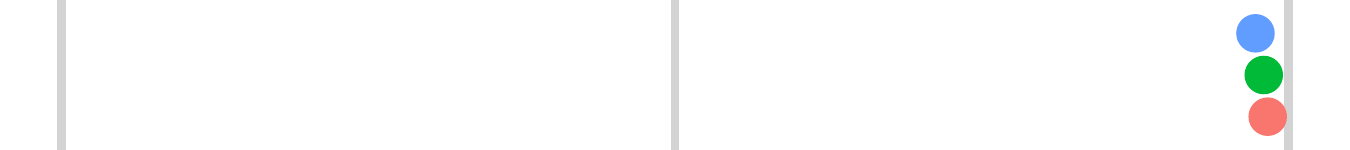} \\ 
\method{ConstL} & {\cellcolor[HTML]{FFA0FF}{\textcolor[HTML]{000000}{Direct}}} & Const & Last & {\cellcolor[HTML]{006837}{\textcolor[HTML]{FFFFFF}{<1s}}} & {\cellcolor[HTML]{808080}{\textcolor[HTML]{FFFFFF}{}}} & 31 & 31 & 31 & \inlinegraphics{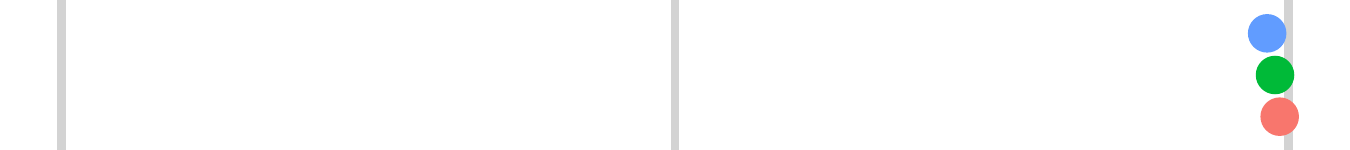} & 27 & 27 & 27 & \inlinegraphics{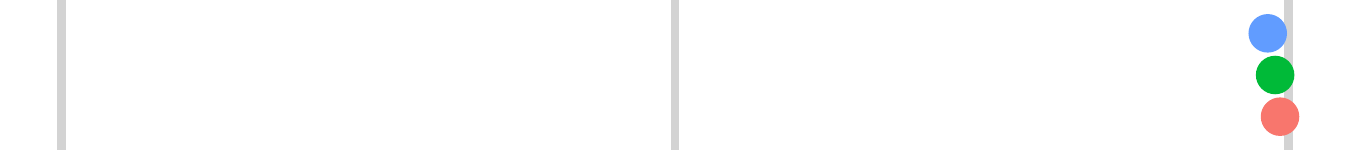} \\ 
\method{ConstM} & {\cellcolor[HTML]{FFA0FF}{\textcolor[HTML]{000000}{Direct}}} & Const & Mean & {\cellcolor[HTML]{006837}{\textcolor[HTML]{FFFFFF}{<1s}}} & {\cellcolor[HTML]{808080}{\textcolor[HTML]{FFFFFF}{}}} & 32 & 32 & 32 & \inlinegraphics{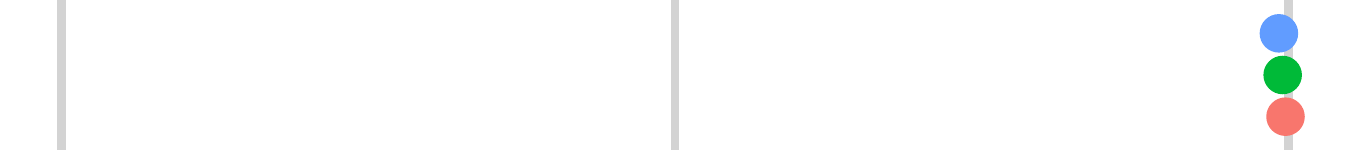} & 31 & 32 & 32 & \inlinegraphics{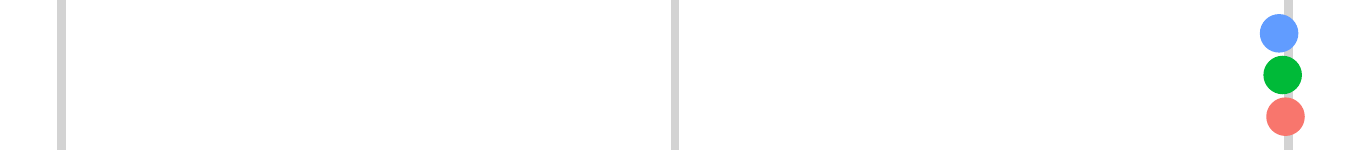} \\ 
\bottomrule
\end{tabular}
\end{center}

%% file: tbl/tableDystsResults.tex
\begin{center}
\caption*{
{\large Cumulative Maximum Error for Test Data of \Dysts{}}
} 
\fontsize{8.0pt}{10pt}\selectfont
\fontfamily{phv}\selectfont
\renewcommand{\arraystretch}{1.05}
\setlength{\tabcolsep}{0.3em}
\rowcolors{2}{gray!20}{white}
\begin{tabular}{lcllrrrrcrrc}
\toprule
\multicolumn{4}{c}{Method} & \multicolumn{2}{c}{Compute} & \multicolumn{3}{c}{Noisefree} & \multicolumn{3}{c}{Noisy} \\ 
\cmidrule(lr){1-4} \cmidrule(lr){5-6} \cmidrule(lr){7-9} \cmidrule(lr){10-12}
Name & Class & Model & Variant & Test & Tune & \# & medi & \CmeScale{} & \# & medi & \CmeScale{} \\ 
\midrule\addlinespace[2.5pt]
\method{SpPo} & {\cellcolor[HTML]{00D8D8}{\textcolor[HTML]{000000}{Fit Solu.}}} & Spline & Poly & {\cellcolor[HTML]{5CB760}{\textcolor[HTML]{000000}{3s}}} & {\cellcolor[HTML]{DCF090}{\textcolor[HTML]{000000}{2m}}} & \textbf{\underline{1}} & 0.00 & \inlinegraphics{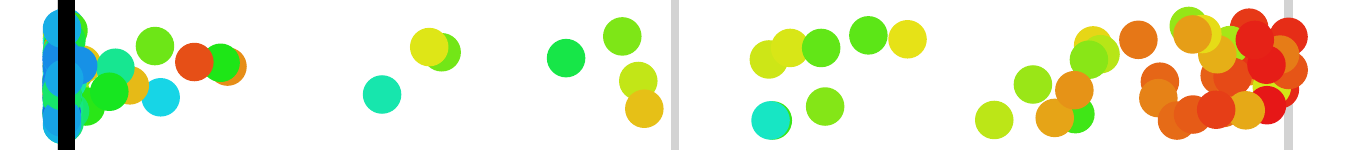} & 9 & 0.78 & \inlinegraphics{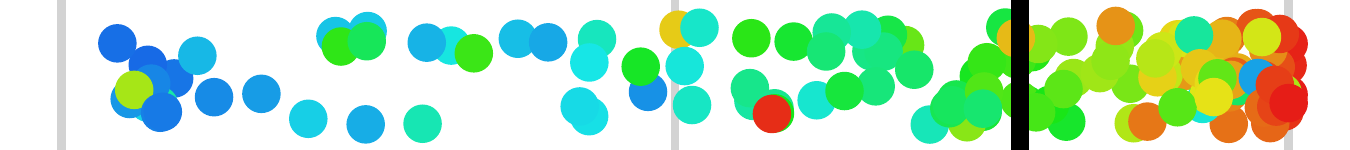} \\ 
\method{LinS} & {\cellcolor[HTML]{B0B0FF}{\textcolor[HTML]{000000}{Fit Prop.}}} & Lin & S & {\cellcolor[HTML]{0F8345}{\textcolor[HTML]{FFFFFF}{<1s}}} & {\cellcolor[HTML]{7FC766}{\textcolor[HTML]{000000}{49s}}} & \textbf{2} & 0.01 & \inlinegraphics{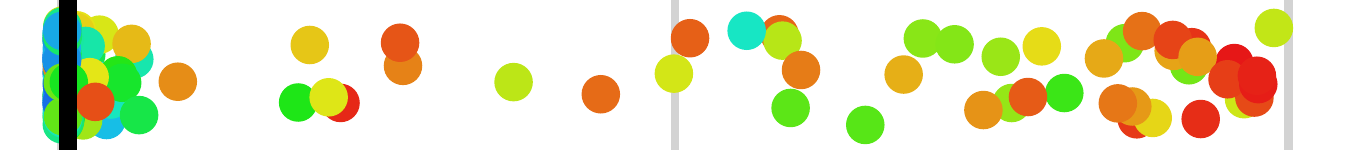} & 6 & 0.77 & \inlinegraphics{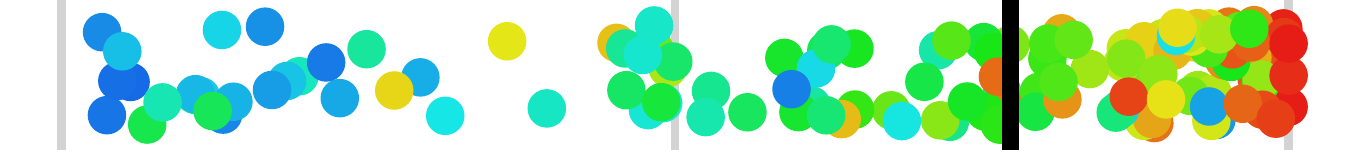} \\ 
\method{LinD} & {\cellcolor[HTML]{B0B0FF}{\textcolor[HTML]{000000}{Fit Prop.}}} & Lin & D & {\cellcolor[HTML]{0F8445}{\textcolor[HTML]{FFFFFF}{<1s}}} & {\cellcolor[HTML]{7CC665}{\textcolor[HTML]{000000}{48s}}} & \underline{3} & 0.01 & \inlinegraphics{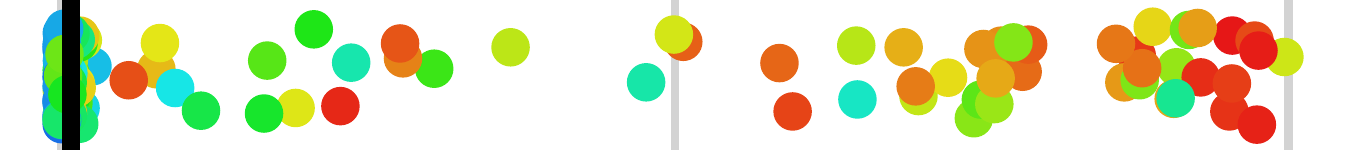} & 8 & 0.77 & \inlinegraphics{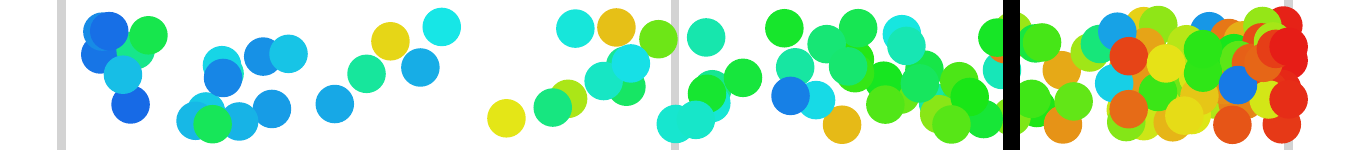} \\ 
\method{SpPo4} & {\cellcolor[HTML]{00D8D8}{\textcolor[HTML]{000000}{Fit Solu.}}} & Spline & Poly4 & {\cellcolor[HTML]{8CCD67}{\textcolor[HTML]{000000}{5s}}} & {\cellcolor[HTML]{808080}{\textcolor[HTML]{FFFFFF}{}}} & 4 & 0.01 & \inlinegraphics{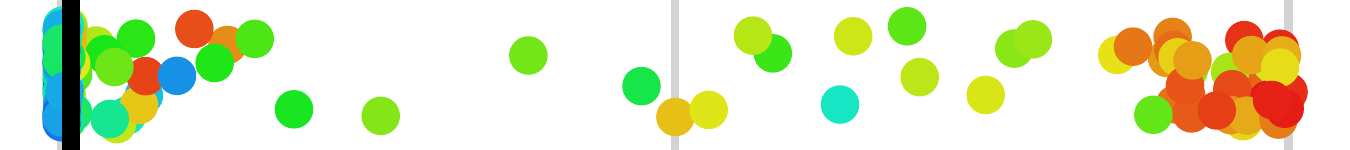} & 21 & 0.85 & \inlinegraphics{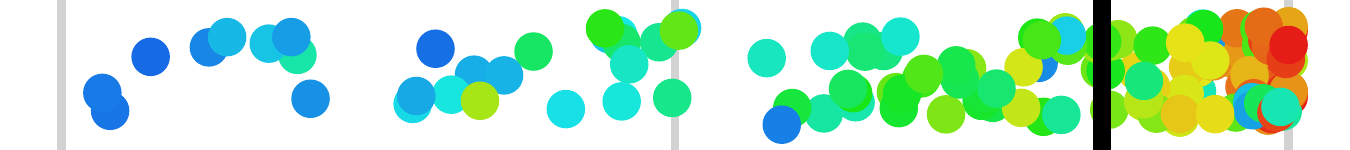} \\ 
\method{RaFeD} & {\cellcolor[HTML]{B0B0FF}{\textcolor[HTML]{000000}{Fit Prop.}}} & RandFeat & D & {\cellcolor[HTML]{118747}{\textcolor[HTML]{FFFFFF}{<1s}}} & {\cellcolor[HTML]{EAF6A2}{\textcolor[HTML]{000000}{3m}}} & 5 & 0.02 & \inlinegraphics{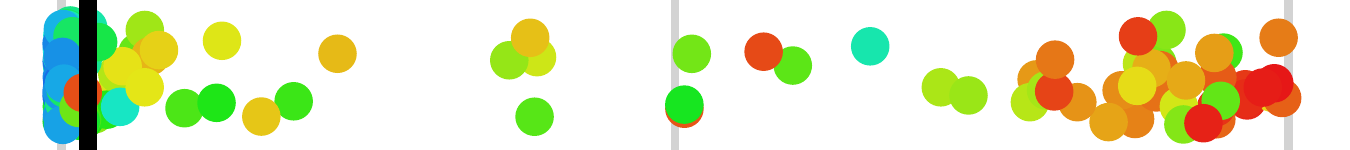} & 19 & 0.84 & \inlinegraphics{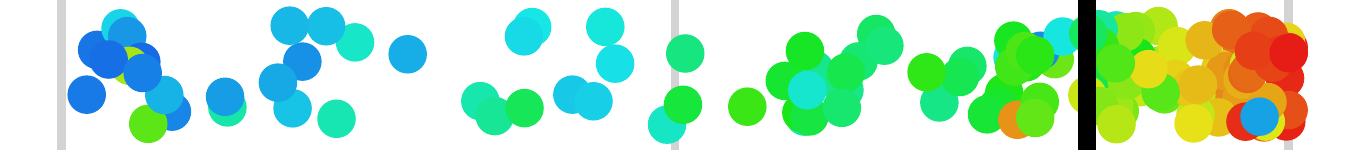} \\ 
\method{EsnD} & {\cellcolor[HTML]{B0B0FF}{\textcolor[HTML]{000000}{Fit Prop.}}} & ESN & D & {\cellcolor[HTML]{2B9E53}{\textcolor[HTML]{FFFFFF}{1s}}} & {\cellcolor[HTML]{F8FCB5}{\textcolor[HTML]{000000}{3m}}} & 6 & 0.02 & \inlinegraphics{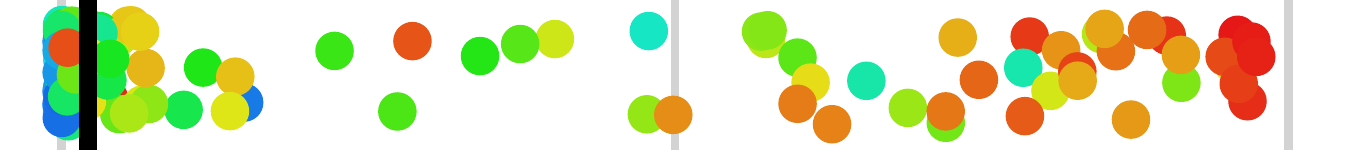} & \textbf{\underline{1}} & 0.73 & \inlinegraphics{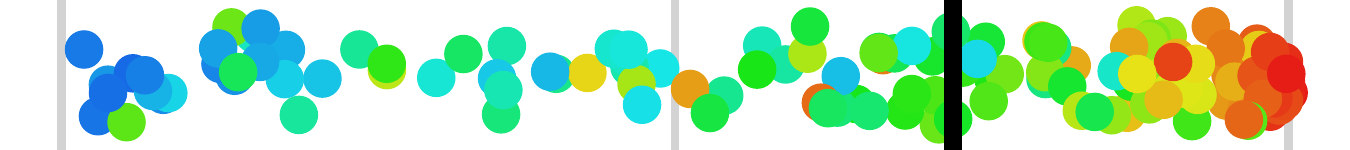} \\ 
\method{RaFeS} & {\cellcolor[HTML]{B0B0FF}{\textcolor[HTML]{000000}{Fit Prop.}}} & RandFeat & S & {\cellcolor[HTML]{118747}{\textcolor[HTML]{FFFFFF}{<1s}}} & {\cellcolor[HTML]{E9F6A1}{\textcolor[HTML]{000000}{3m}}} & 7 & 0.03 & \inlinegraphics{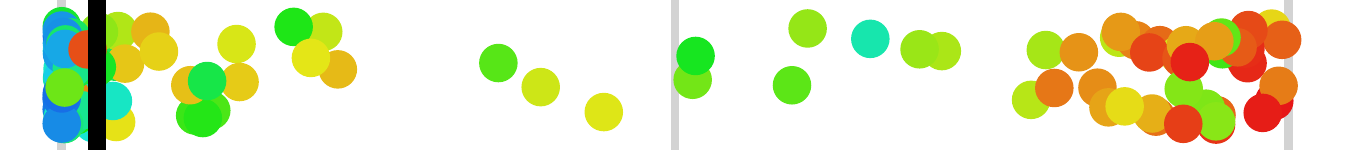} & 15 & 0.80 & \inlinegraphics{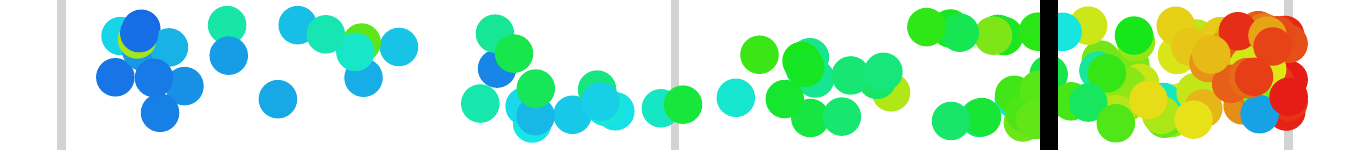} \\ 
\method{EsnS} & {\cellcolor[HTML]{B0B0FF}{\textcolor[HTML]{000000}{Fit Prop.}}} & ESN & S & {\cellcolor[HTML]{2E9F54}{\textcolor[HTML]{FFFFFF}{1s}}} & {\cellcolor[HTML]{FBFDB9}{\textcolor[HTML]{000000}{4m}}} & 8 & 0.03 & \inlinegraphics{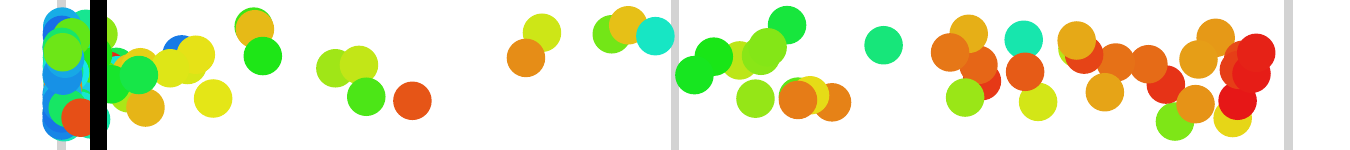} & \underline{3} & 0.76 & \inlinegraphics{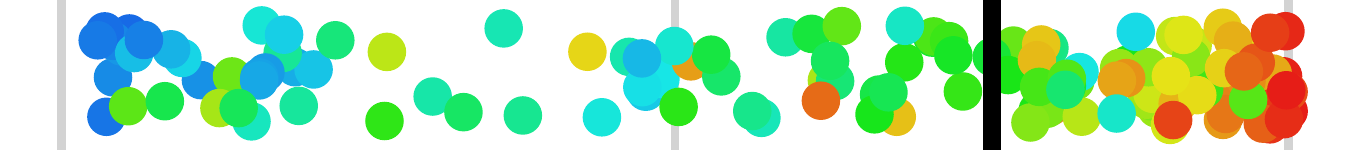} \\ 
\method{SINDy} & {\cellcolor[HTML]{00D8D8}{\textcolor[HTML]{000000}{Fit Solu.}}} & SINDy &  & {\cellcolor[HTML]{57B45E}{\textcolor[HTML]{FFFFFF}{2s}}} & {\cellcolor[HTML]{EFF8A9}{\textcolor[HTML]{000000}{3m}}} & 9 & 0.03 & \inlinegraphics{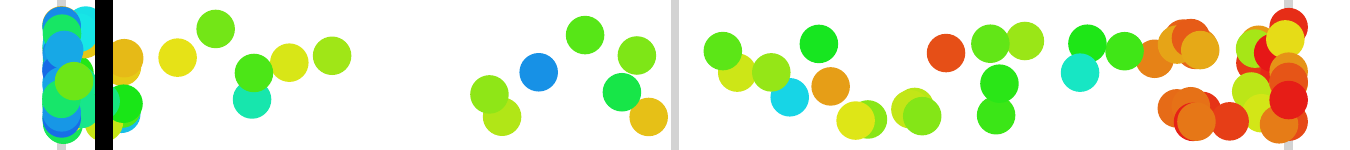} & 24 & 0.87 & \inlinegraphics{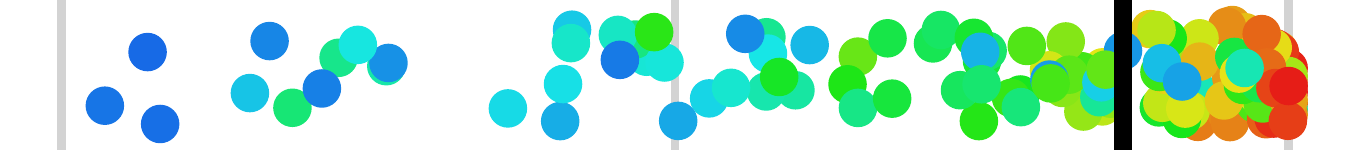} \\ 
\method{LinPo4} & {\cellcolor[HTML]{B0B0FF}{\textcolor[HTML]{000000}{Fit Prop.}}} & Lin & Poly4 & {\cellcolor[HTML]{08773F}{\textcolor[HTML]{FFFFFF}{<1s}}} & {\cellcolor[HTML]{808080}{\textcolor[HTML]{FFFFFF}{}}} & 10 & 0.04 & \inlinegraphics{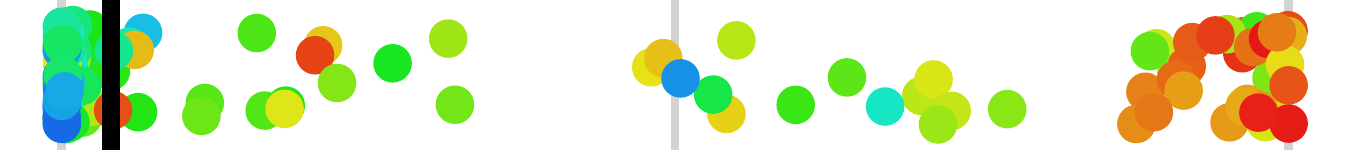} & 20 & 0.84 & \inlinegraphics{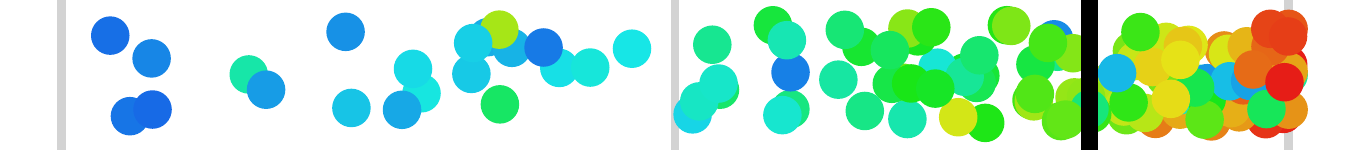} \\ 
\method{SINDyN} & {\cellcolor[HTML]{00D8D8}{\textcolor[HTML]{000000}{Fit Solu.}}} & SINDy & norm & {\cellcolor[HTML]{59B55F}{\textcolor[HTML]{FFFFFF}{2s}}} & {\cellcolor[HTML]{FCFEBA}{\textcolor[HTML]{000000}{4m}}} & 11 & 0.12 & \inlinegraphics{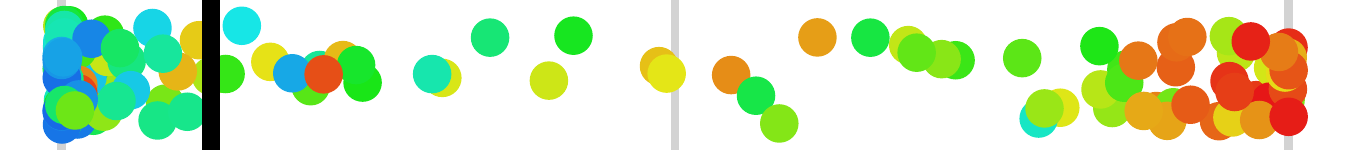} & 25 & 0.87 & \inlinegraphics{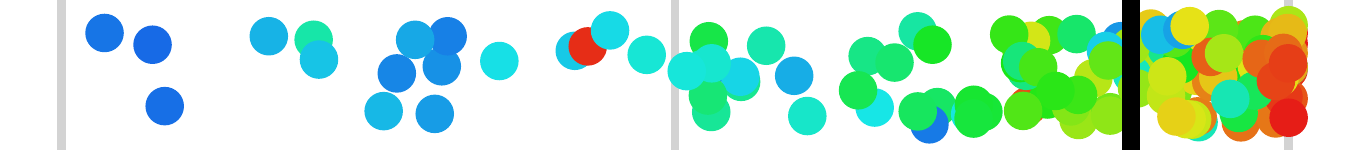} \\ 
\method{LinPo6} & {\cellcolor[HTML]{B0B0FF}{\textcolor[HTML]{000000}{Fit Prop.}}} & Lin & Poly6 & {\cellcolor[HTML]{47AB5A}{\textcolor[HTML]{FFFFFF}{2s}}} & {\cellcolor[HTML]{808080}{\textcolor[HTML]{FFFFFF}{}}} & 12 & 0.19 & \inlinegraphics{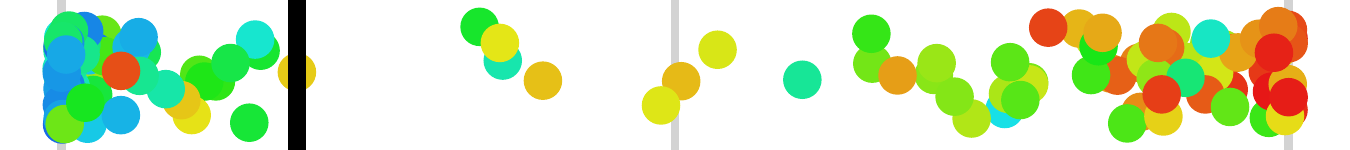} & 29 & 0.91 & \inlinegraphics{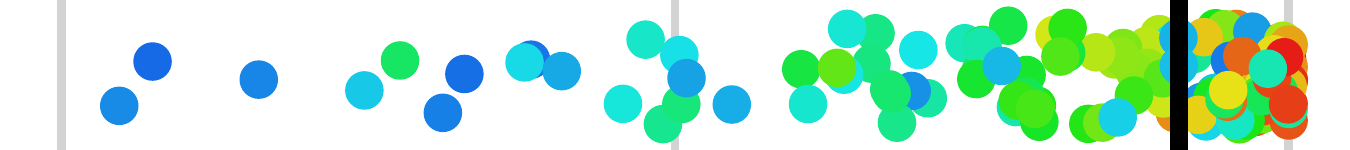} \\ 
\method{SpPo2} & {\cellcolor[HTML]{00D8D8}{\textcolor[HTML]{000000}{Fit Solu.}}} & Spline & Poly2 & {\cellcolor[HTML]{2B9E53}{\textcolor[HTML]{FFFFFF}{1s}}} & {\cellcolor[HTML]{808080}{\textcolor[HTML]{FFFFFF}{}}} & 13 & 0.21 & \inlinegraphics{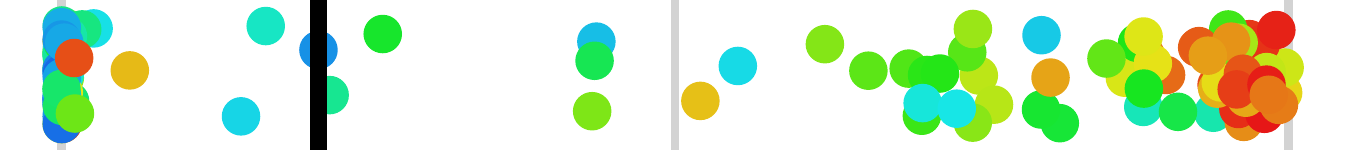} & 10 & 0.79 & \inlinegraphics{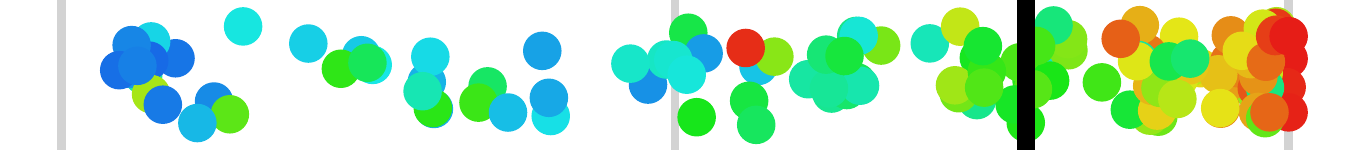} \\ 
\method{PgGpD} & {\cellcolor[HTML]{B0B0FF}{\textcolor[HTML]{000000}{Fit Prop.}}} & GP & D & {\cellcolor[HTML]{63BB62}{\textcolor[HTML]{000000}{3s}}} & {\cellcolor[HTML]{7DC665}{\textcolor[HTML]{000000}{48s}}} & 14 & 0.22 & \inlinegraphics{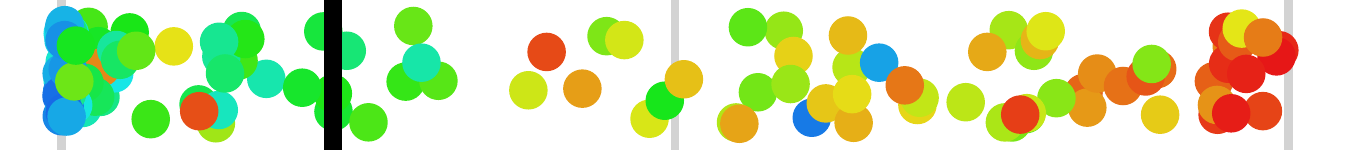} & 26 & 0.88 & \inlinegraphics{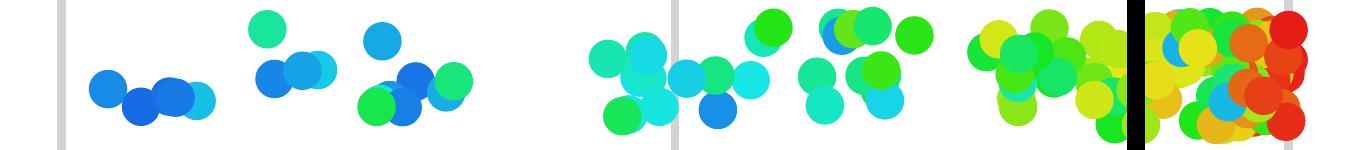} \\ 
\method{PgNetD} & {\cellcolor[HTML]{F0C0C0}{\textcolor[HTML]{000000}{Gr. Desc.}}} & NeuralNet & D & {\cellcolor[HTML]{FED682}{\textcolor[HTML]{000000}{1m}}} & {\cellcolor[HTML]{FAFDB9}{\textcolor[HTML]{000000}{4m}}} & 15 & 0.26 & \inlinegraphics{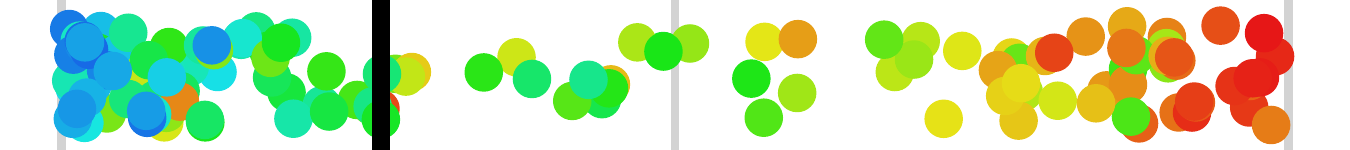} & 12 & 0.80 & \inlinegraphics{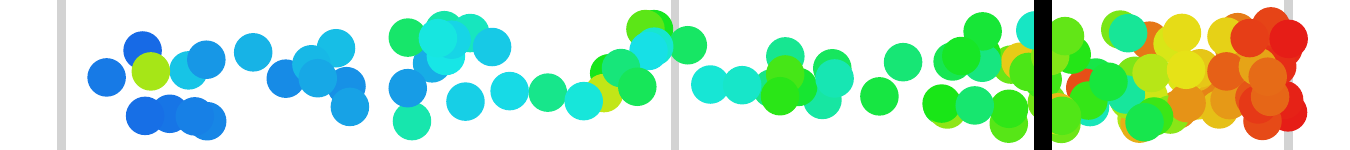} \\ 
\method{Gru} & {\cellcolor[HTML]{F0C0C0}{\textcolor[HTML]{000000}{Gr. Desc.}}} & GRU &  & {\cellcolor[HTML]{F46D43}{\textcolor[HTML]{FFFFFF}{4m}}} & {\cellcolor[HTML]{FB9856}{\textcolor[HTML]{000000}{19m}}} & 16 & 0.40 & \inlinegraphics{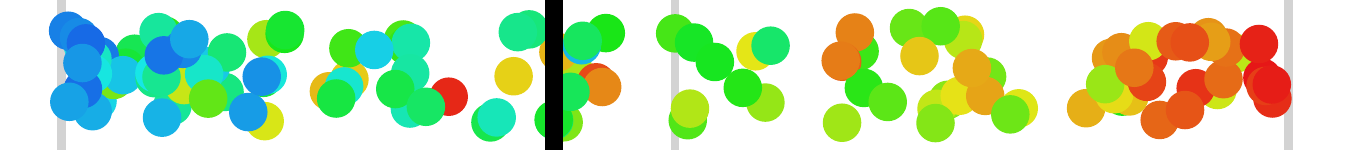} & \textbf{2} & 0.75 & \inlinegraphics{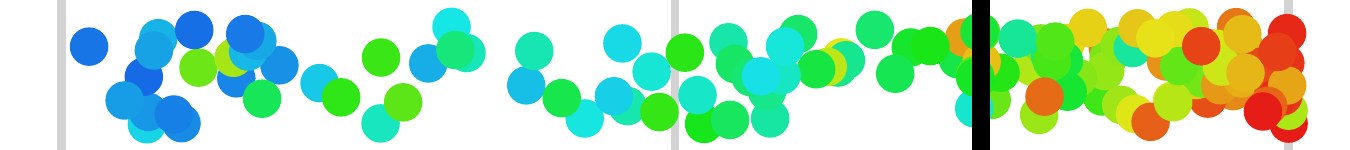} \\ 
\method{Node1} & {\cellcolor[HTML]{F0C0C0}{\textcolor[HTML]{000000}{Gr. Desc.}}} & NODE & bs1 & {\cellcolor[HTML]{A50026}{\textcolor[HTML]{FFFFFF}{19m}}} & {\cellcolor[HTML]{AE0C26}{\textcolor[HTML]{FFFFFF}{1h}}} & 17 & 0.42 & \inlinegraphics{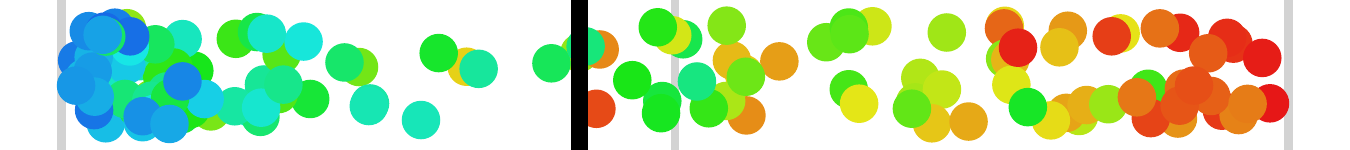} & 7 & 0.77 & \inlinegraphics{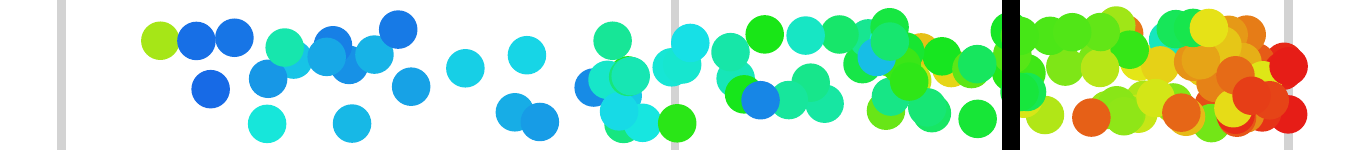} \\ 
\method{Node32} & {\cellcolor[HTML]{F0C0C0}{\textcolor[HTML]{000000}{Gr. Desc.}}} & NODE & bs32 & {\cellcolor[HTML]{F67A49}{\textcolor[HTML]{FFFFFF}{4m}}} & {\cellcolor[HTML]{F46D43}{\textcolor[HTML]{FFFFFF}{29m}}} & 18 & 0.44 & \inlinegraphics{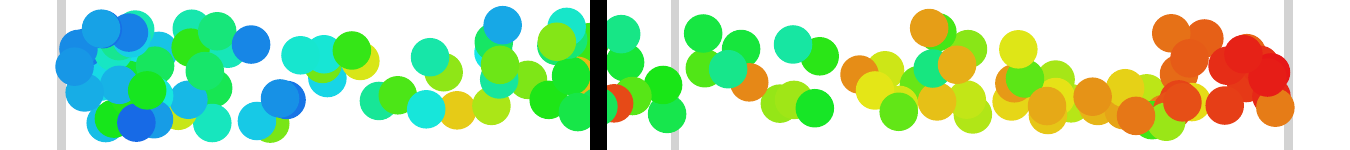} & 13 & 0.80 & \inlinegraphics{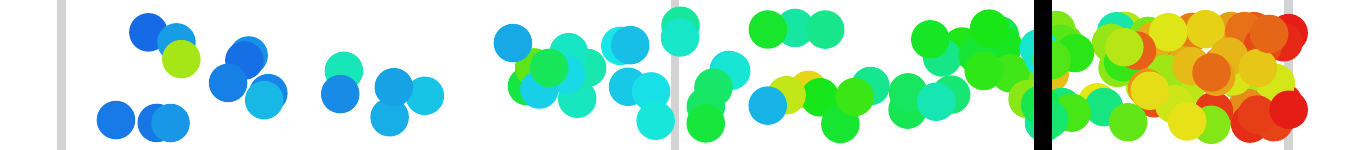} \\ 
\method{PgGpS} & {\cellcolor[HTML]{B0B0FF}{\textcolor[HTML]{000000}{Fit Prop.}}} & GP & S & {\cellcolor[HTML]{59B55F}{\textcolor[HTML]{FFFFFF}{2s}}} & {\cellcolor[HTML]{7BC665}{\textcolor[HTML]{000000}{47s}}} & 19 & 0.44 & \inlinegraphics{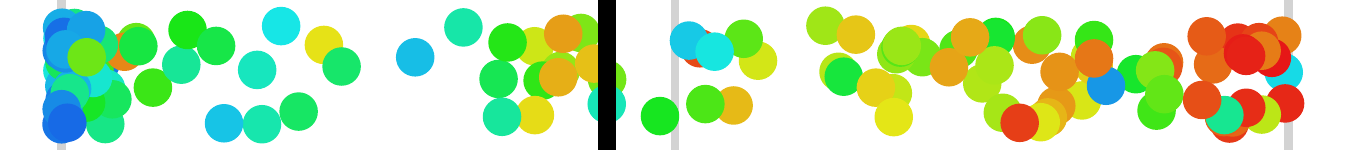} & 23 & 0.86 & \inlinegraphics{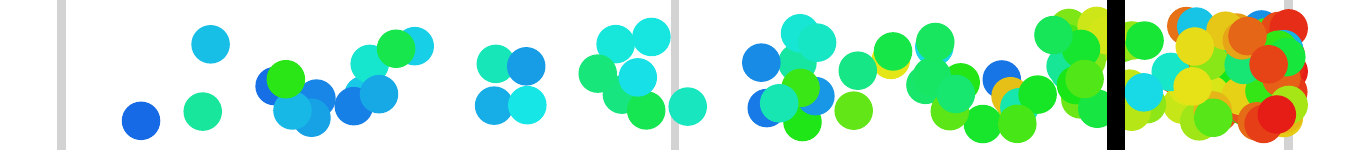} \\ 
\method{PgNetS} & {\cellcolor[HTML]{F0C0C0}{\textcolor[HTML]{000000}{Gr. Desc.}}} & NeuralNet & S & {\cellcolor[HTML]{FEDD88}{\textcolor[HTML]{000000}{1m}}} & {\cellcolor[HTML]{CFEB84}{\textcolor[HTML]{000000}{2m}}} & 20 & 0.46 & \inlinegraphics{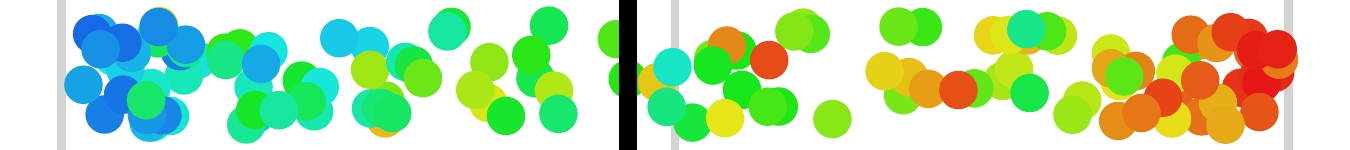} & 11 & 0.80 & \inlinegraphics{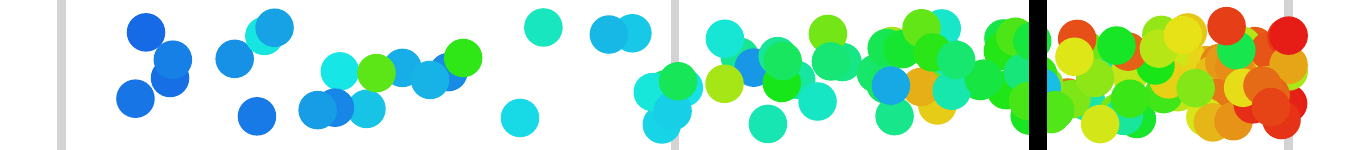} \\ 
\method{SpGp} & {\cellcolor[HTML]{00D8D8}{\textcolor[HTML]{000000}{Fit Solu.}}} & Spline & GP & {\cellcolor[HTML]{60B961}{\textcolor[HTML]{000000}{3s}}} & {\cellcolor[HTML]{5DB760}{\textcolor[HTML]{000000}{35s}}} & 21 & 0.54 & \inlinegraphics{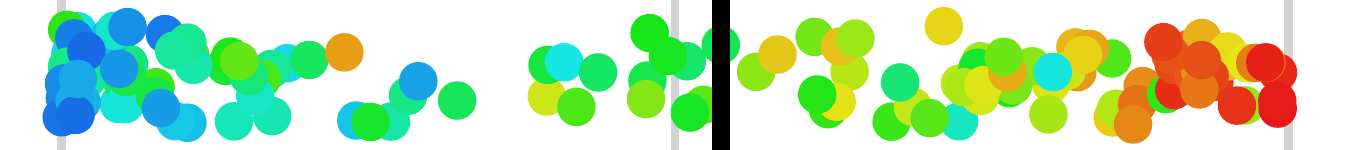} & 30 & 0.92 & \inlinegraphics{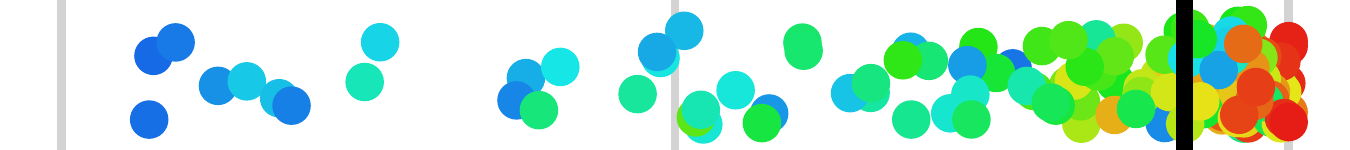} \\ 
\method{\_NBEAT} & {\cellcolor[HTML]{C0C0C0}{\textcolor[HTML]{000000}{Dysts}}} & NBEATS &  & {\cellcolor[HTML]{808080}{\textcolor[HTML]{FFFFFF}{}}} & {\cellcolor[HTML]{808080}{\textcolor[HTML]{FFFFFF}{}}} & 22 & 0.56 & \inlinegraphics{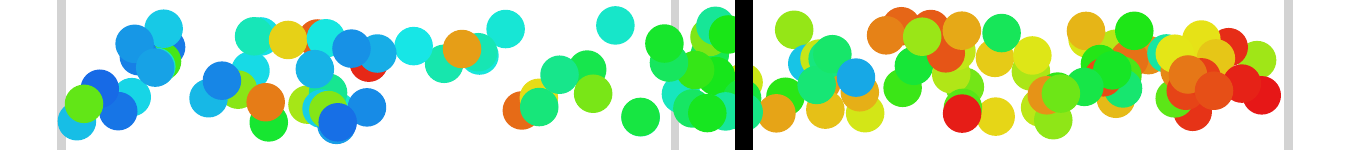} &  &  &  \\ 
\method{PgLlS} & {\cellcolor[HTML]{B0B0FF}{\textcolor[HTML]{000000}{Fit Prop.}}} & LocalLin & S & {\cellcolor[HTML]{67BD63}{\textcolor[HTML]{000000}{3s}}} & {\cellcolor[HTML]{006837}{\textcolor[HTML]{FFFFFF}{10s}}} & 23 & 0.57 & \inlinegraphics{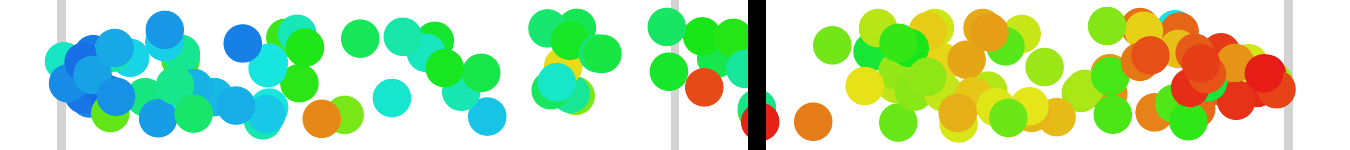} & 16 & 0.81 & \inlinegraphics{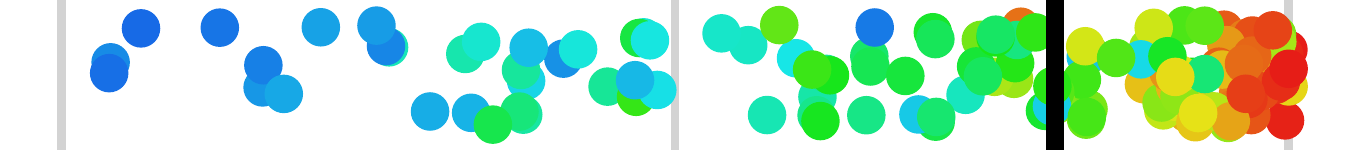} \\ 
\method{Analog} & {\cellcolor[HTML]{FFA0FF}{\textcolor[HTML]{000000}{Direct}}} & Analog &  & {\cellcolor[HTML]{006837}{\textcolor[HTML]{FFFFFF}{<1s}}} & {\cellcolor[HTML]{808080}{\textcolor[HTML]{FFFFFF}{}}} & 24 & 0.60 & \inlinegraphics{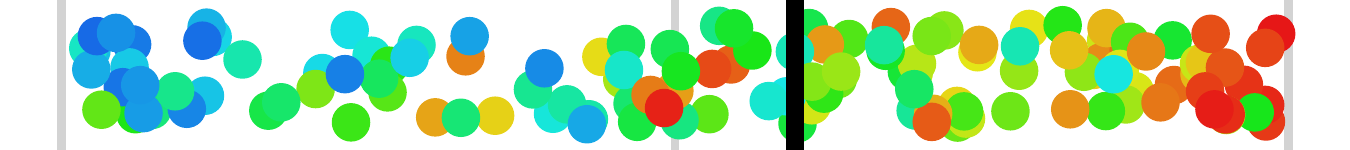} & 17 & 0.81 & \inlinegraphics{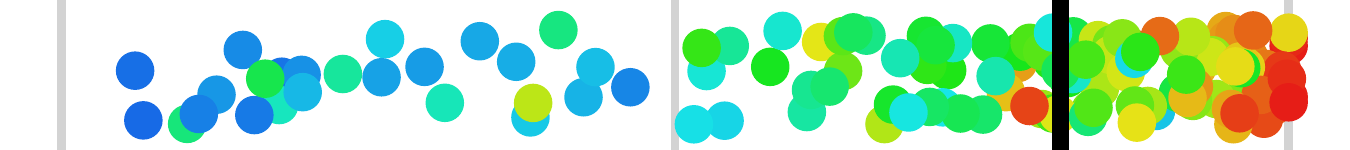} \\ 
\method{PwlNn} & {\cellcolor[HTML]{00D8D8}{\textcolor[HTML]{000000}{Fit Solu.}}} & PwLin & NN & {\cellcolor[HTML]{17924D}{\textcolor[HTML]{FFFFFF}{1s}}} & {\cellcolor[HTML]{808080}{\textcolor[HTML]{FFFFFF}{}}} & 25 & 0.60 & \inlinegraphics{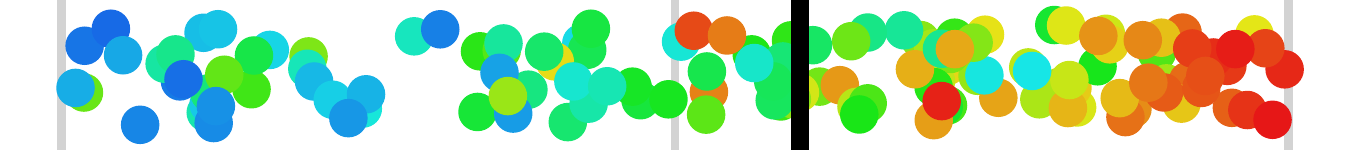} & 18 & 0.83 & \inlinegraphics{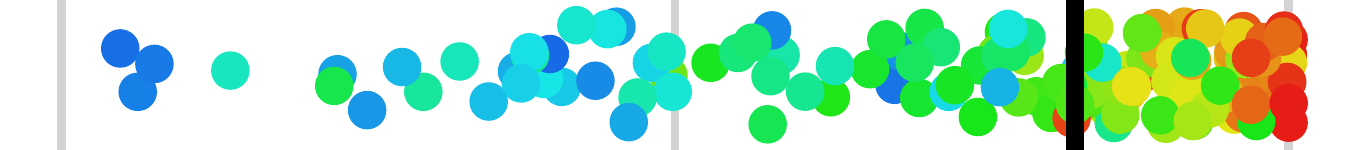} \\ 
\method{PgLlD} & {\cellcolor[HTML]{B0B0FF}{\textcolor[HTML]{000000}{Fit Prop.}}} & LocalLin & D & {\cellcolor[HTML]{65BC63}{\textcolor[HTML]{000000}{3s}}} & {\cellcolor[HTML]{006837}{\textcolor[HTML]{FFFFFF}{10s}}} & 26 & 0.61 & \inlinegraphics{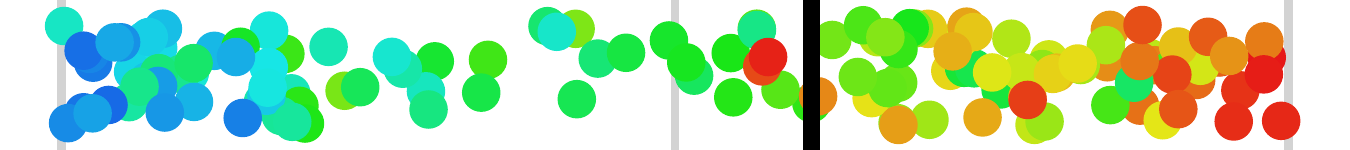} & 14 & 0.80 & \inlinegraphics{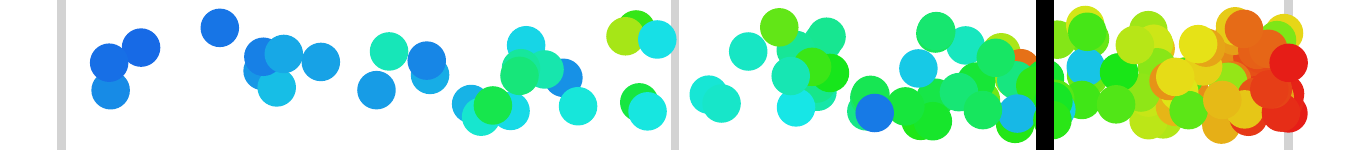} \\ 
\method{SpNn} & {\cellcolor[HTML]{00D8D8}{\textcolor[HTML]{000000}{Fit Solu.}}} & Spline &  & {\cellcolor[HTML]{19964F}{\textcolor[HTML]{FFFFFF}{1s}}} & {\cellcolor[HTML]{808080}{\textcolor[HTML]{FFFFFF}{}}} & 27 & 0.62 & \inlinegraphics{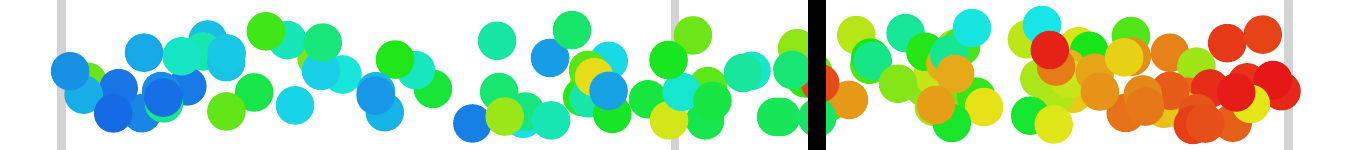} & 22 & 0.85 & \inlinegraphics{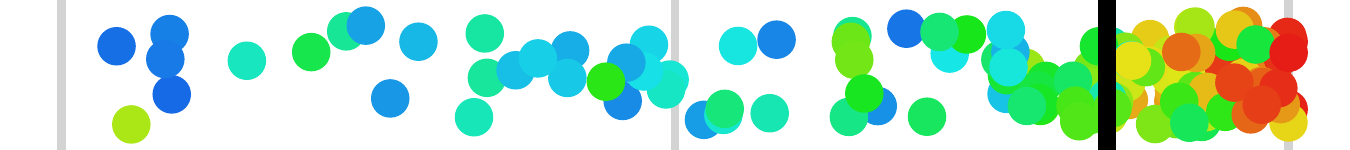} \\ 
\method{Rnn} & {\cellcolor[HTML]{F0C0C0}{\textcolor[HTML]{000000}{Gr. Desc.}}} & RNN &  & {\cellcolor[HTML]{FCA15B}{\textcolor[HTML]{000000}{2m}}} & {\cellcolor[HTML]{FFE594}{\textcolor[HTML]{000000}{7m}}} & 28 & 0.62 & \inlinegraphics{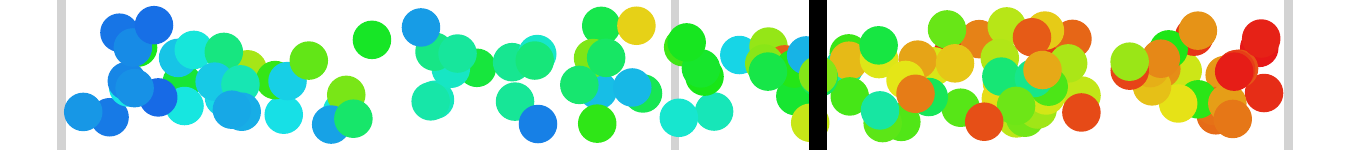} & 5 & 0.77 & \inlinegraphics{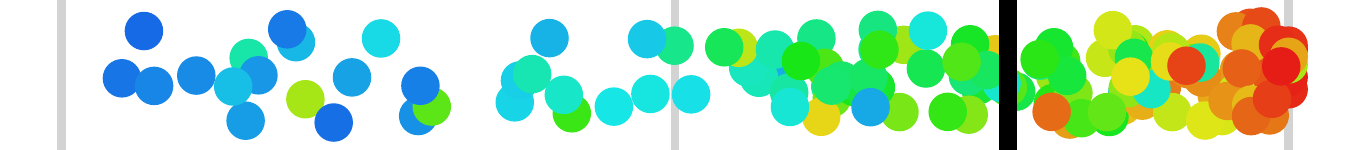} \\ 
\method{Trafo} & {\cellcolor[HTML]{F0C0C0}{\textcolor[HTML]{000000}{Gr. Desc.}}} & Transformer &  & {\cellcolor[HTML]{BF1E27}{\textcolor[HTML]{FFFFFF}{13m}}} & {\cellcolor[HTML]{A50026}{\textcolor[HTML]{FFFFFF}{1h}}} & 29 & 0.63 & \inlinegraphics{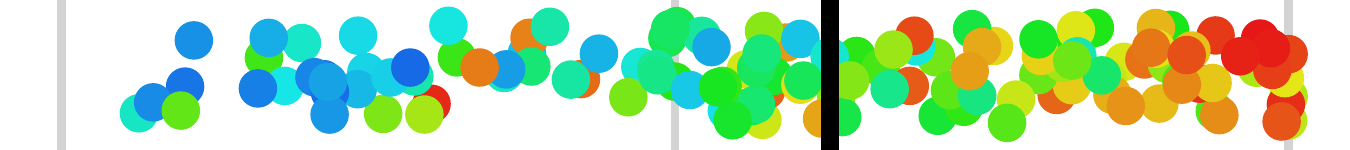} & 4 & 0.77 & \inlinegraphics{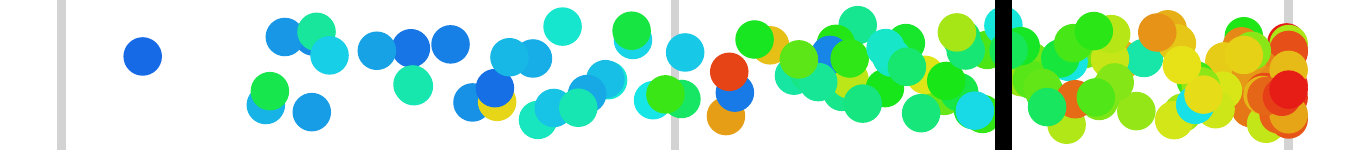} \\ 
\method{GpGp} & {\cellcolor[HTML]{00D8D8}{\textcolor[HTML]{000000}{Fit Solu.}}} & GP & GP & {\cellcolor[HTML]{63BB62}{\textcolor[HTML]{000000}{3s}}} & {\cellcolor[HTML]{CEEA84}{\textcolor[HTML]{000000}{2m}}} & 30 & 0.64 & \inlinegraphics{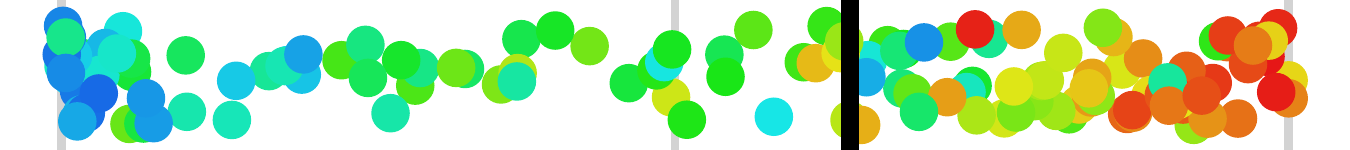} & 28 & 0.90 & \inlinegraphics{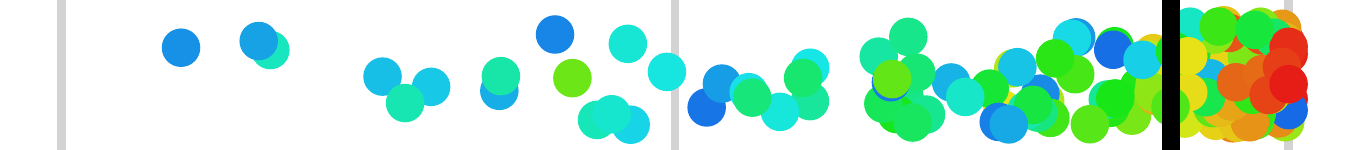} \\ 
\method{\_RaFo} & {\cellcolor[HTML]{C0C0C0}{\textcolor[HTML]{000000}{Dysts}}} & RandForest &  & {\cellcolor[HTML]{808080}{\textcolor[HTML]{FFFFFF}{}}} & {\cellcolor[HTML]{808080}{\textcolor[HTML]{FFFFFF}{}}} & 31 & 0.69 & \inlinegraphics{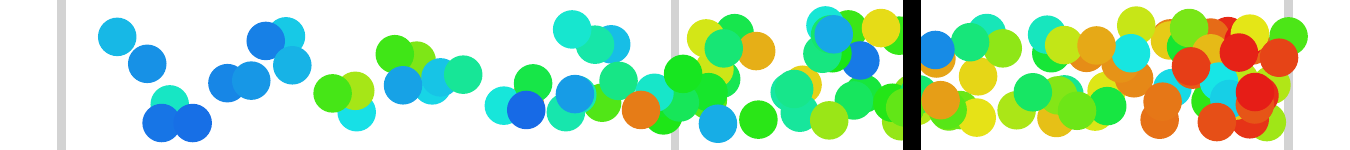} &  &  &  \\ 
\method{\_NHiTS} & {\cellcolor[HTML]{C0C0C0}{\textcolor[HTML]{000000}{Dysts}}} & NHiTS &  & {\cellcolor[HTML]{808080}{\textcolor[HTML]{FFFFFF}{}}} & {\cellcolor[HTML]{808080}{\textcolor[HTML]{FFFFFF}{}}} & 32 & 0.70 & \inlinegraphics{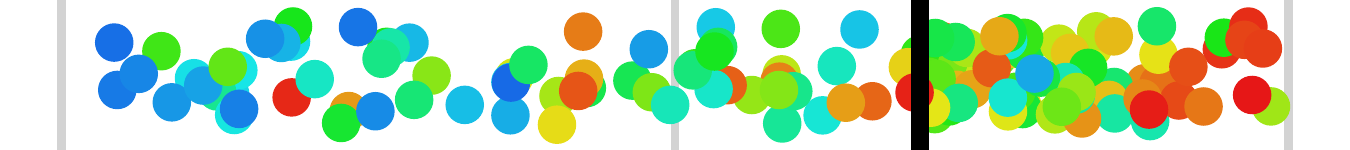} &  &  &  \\ 
\method{LlNn} & {\cellcolor[HTML]{00D8D8}{\textcolor[HTML]{000000}{Fit Solu.}}} & LocalLin & NN & {\cellcolor[HTML]{90CF67}{\textcolor[HTML]{000000}{6s}}} & {\cellcolor[HTML]{19964F}{\textcolor[HTML]{FFFFFF}{19s}}} & 33 & 0.73 & \inlinegraphics{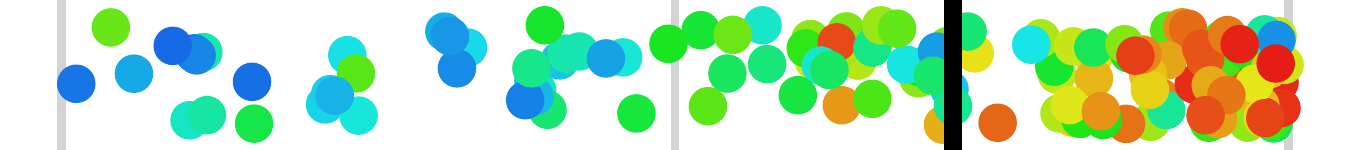} & 27 & 0.88 & \inlinegraphics{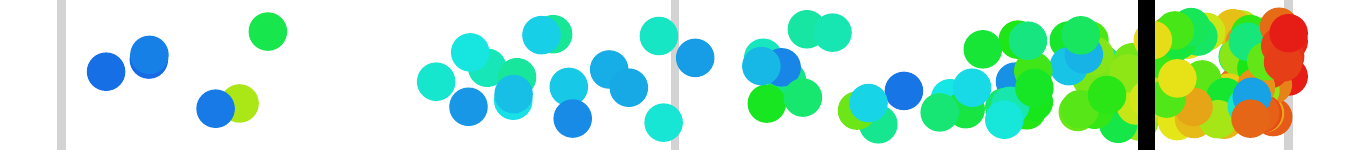} \\ 
\method{\_RNN} & {\cellcolor[HTML]{C0C0C0}{\textcolor[HTML]{000000}{Dysts}}} & RNN &  & {\cellcolor[HTML]{808080}{\textcolor[HTML]{FFFFFF}{}}} & {\cellcolor[HTML]{808080}{\textcolor[HTML]{FFFFFF}{}}} & 34 & 0.76 & \inlinegraphics{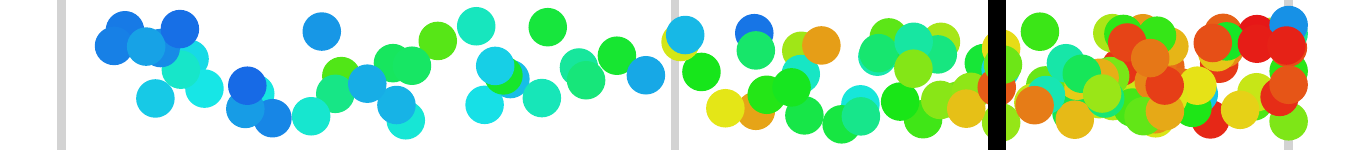} &  &  &  \\ 
\method{\_Node} & {\cellcolor[HTML]{C0C0C0}{\textcolor[HTML]{000000}{Dysts}}} & NeuralOde &  & {\cellcolor[HTML]{808080}{\textcolor[HTML]{FFFFFF}{}}} & {\cellcolor[HTML]{808080}{\textcolor[HTML]{FFFFFF}{}}} & 35 & 0.79 & \inlinegraphics{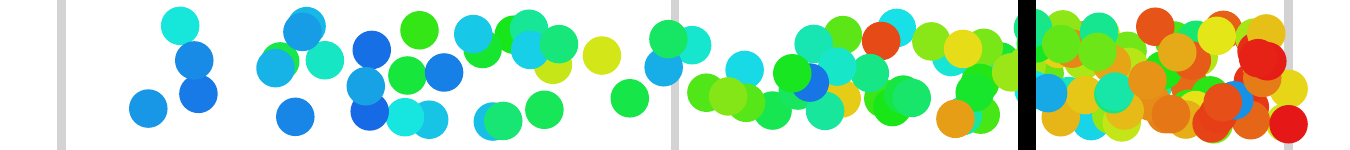} &  &  &  \\ 
\method{\_Esn} & {\cellcolor[HTML]{C0C0C0}{\textcolor[HTML]{000000}{Dysts}}} & Esn &  & {\cellcolor[HTML]{808080}{\textcolor[HTML]{FFFFFF}{}}} & {\cellcolor[HTML]{808080}{\textcolor[HTML]{FFFFFF}{}}} & 36 & 0.79 & \inlinegraphics{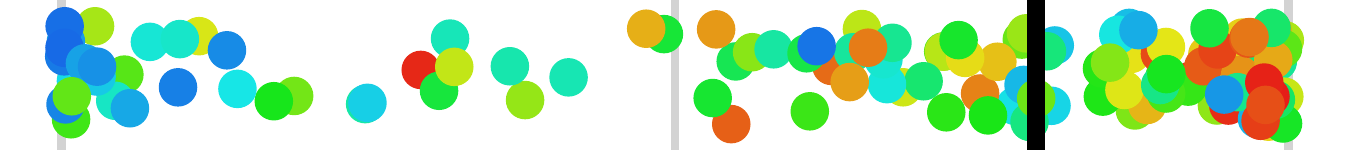} &  &  &  \\ 
\method{\_Nvar} & {\cellcolor[HTML]{C0C0C0}{\textcolor[HTML]{000000}{Dysts}}} & Nvar &  & {\cellcolor[HTML]{808080}{\textcolor[HTML]{FFFFFF}{}}} & {\cellcolor[HTML]{808080}{\textcolor[HTML]{FFFFFF}{}}} & 37 & 0.80 & \inlinegraphics{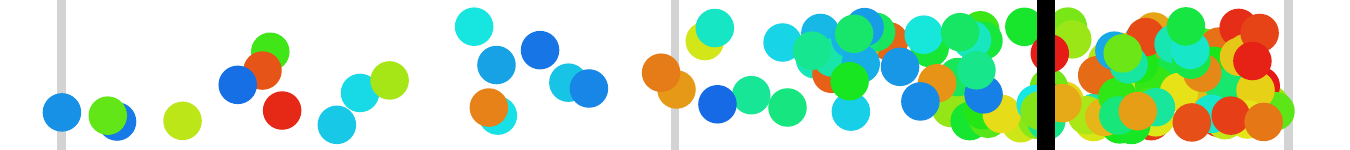} &  &  &  \\ 
\method{\_XGB} & {\cellcolor[HTML]{C0C0C0}{\textcolor[HTML]{000000}{Dysts}}} & XGB &  & {\cellcolor[HTML]{808080}{\textcolor[HTML]{FFFFFF}{}}} & {\cellcolor[HTML]{808080}{\textcolor[HTML]{FFFFFF}{}}} & 38 & 0.81 & \inlinegraphics{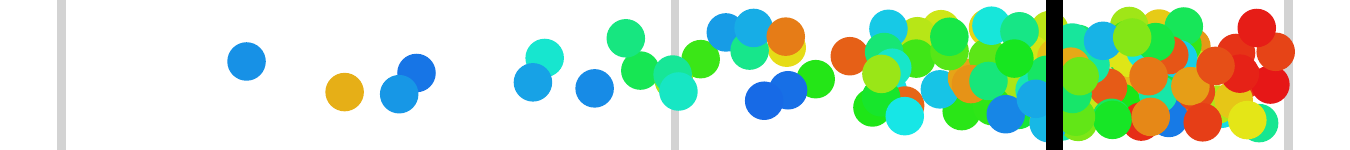} &  &  &  \\ 
\method{\_Trafo} & {\cellcolor[HTML]{C0C0C0}{\textcolor[HTML]{000000}{Dysts}}} & Transformer &  & {\cellcolor[HTML]{808080}{\textcolor[HTML]{FFFFFF}{}}} & {\cellcolor[HTML]{808080}{\textcolor[HTML]{FFFFFF}{}}} & 39 & 0.81 & \inlinegraphics{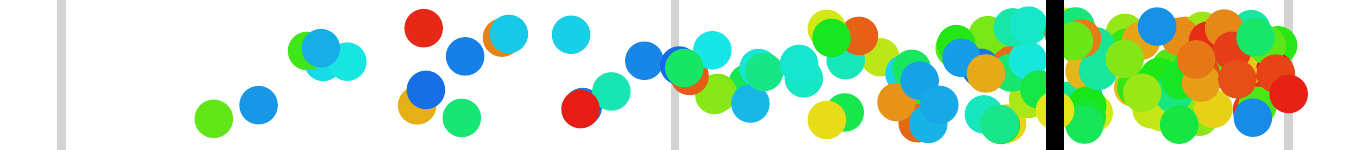} &  &  &  \\ 
\method{\_LinRe} & {\cellcolor[HTML]{C0C0C0}{\textcolor[HTML]{000000}{Dysts}}} & LinearReg &  & {\cellcolor[HTML]{808080}{\textcolor[HTML]{FFFFFF}{}}} & {\cellcolor[HTML]{808080}{\textcolor[HTML]{FFFFFF}{}}} & 40 & 0.84 & \inlinegraphics{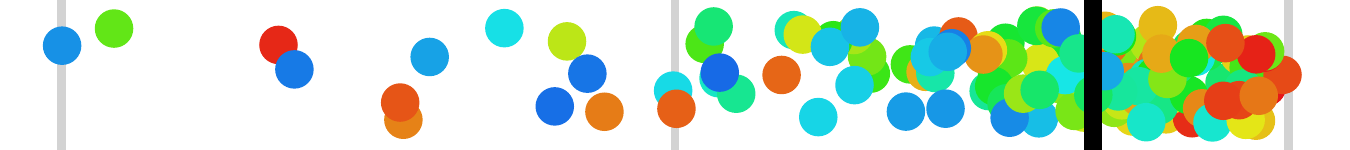} &  &  &  \\ 
\method{\_NLin} & {\cellcolor[HTML]{C0C0C0}{\textcolor[HTML]{000000}{Dysts}}} & NLinear &  & {\cellcolor[HTML]{808080}{\textcolor[HTML]{FFFFFF}{}}} & {\cellcolor[HTML]{808080}{\textcolor[HTML]{FFFFFF}{}}} & 41 & 0.88 & \inlinegraphics{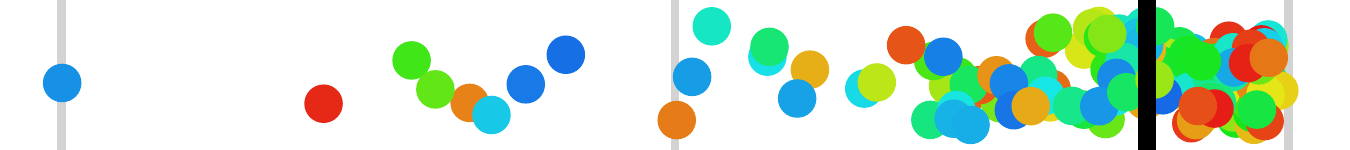} &  &  &  \\ 
\method{\_BlRNN} & {\cellcolor[HTML]{C0C0C0}{\textcolor[HTML]{000000}{Dysts}}} & BlockRNN &  & {\cellcolor[HTML]{808080}{\textcolor[HTML]{FFFFFF}{}}} & {\cellcolor[HTML]{808080}{\textcolor[HTML]{FFFFFF}{}}} & 42 & 0.90 & \inlinegraphics{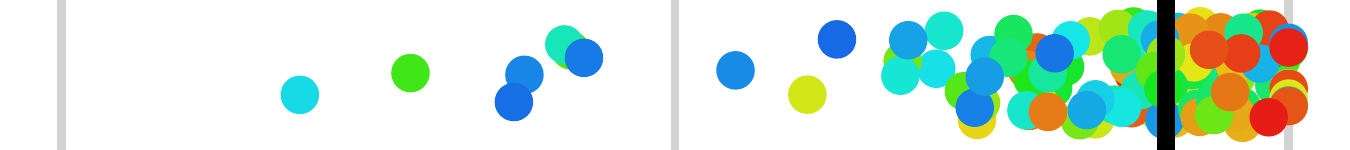} &  &  &  \\ 
\method{\_DLin} & {\cellcolor[HTML]{C0C0C0}{\textcolor[HTML]{000000}{Dysts}}} & DLinear &  & {\cellcolor[HTML]{808080}{\textcolor[HTML]{FFFFFF}{}}} & {\cellcolor[HTML]{808080}{\textcolor[HTML]{FFFFFF}{}}} & 43 & 0.90 & \inlinegraphics{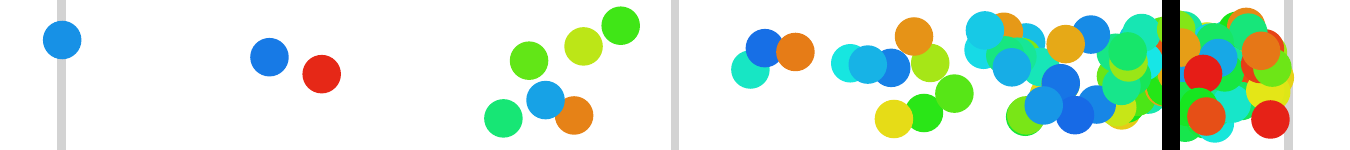} &  &  &  \\ 
\method{Lstm} & {\cellcolor[HTML]{F0C0C0}{\textcolor[HTML]{000000}{Gr. Desc.}}} & LSTM &  & {\cellcolor[HTML]{F0663F}{\textcolor[HTML]{FFFFFF}{5m}}} & {\cellcolor[HTML]{FCA85E}{\textcolor[HTML]{000000}{16m}}} & 44 & 0.91 & \inlinegraphics{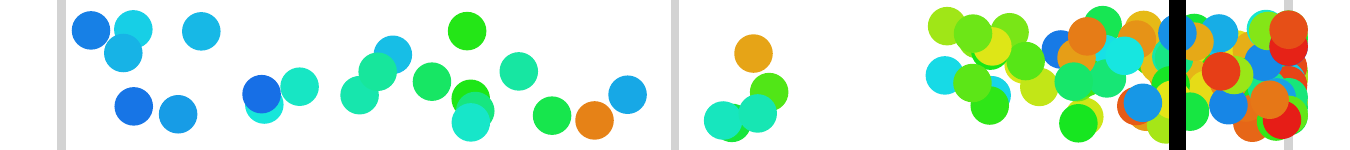} & 32 & 0.96 & \inlinegraphics{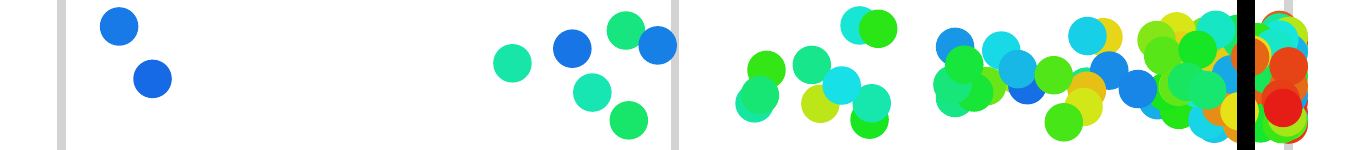} \\ 
\method{\_TCN} & {\cellcolor[HTML]{C0C0C0}{\textcolor[HTML]{000000}{Dysts}}} & TCN &  & {\cellcolor[HTML]{808080}{\textcolor[HTML]{FFFFFF}{}}} & {\cellcolor[HTML]{808080}{\textcolor[HTML]{FFFFFF}{}}} & 45 & 0.94 & \inlinegraphics{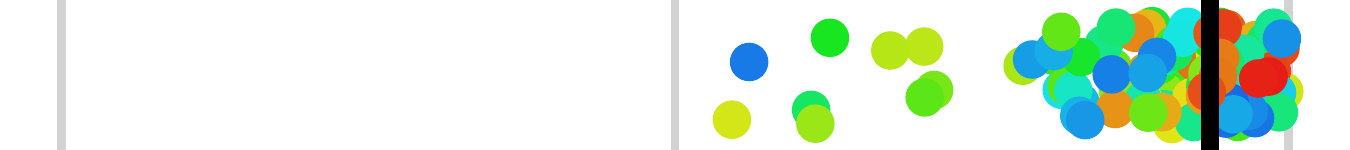} &  &  &  \\ 
\method{ConstL} & {\cellcolor[HTML]{FFA0FF}{\textcolor[HTML]{000000}{Direct}}} & Const & Last & {\cellcolor[HTML]{07763E}{\textcolor[HTML]{FFFFFF}{<1s}}} & {\cellcolor[HTML]{808080}{\textcolor[HTML]{FFFFFF}{}}} & 46 & 0.96 & \inlinegraphics{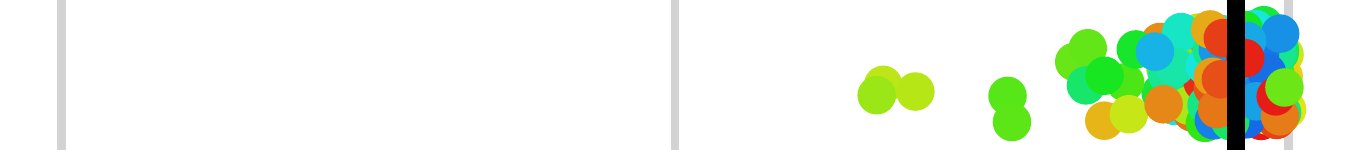} & 31 & 0.96 & \inlinegraphics{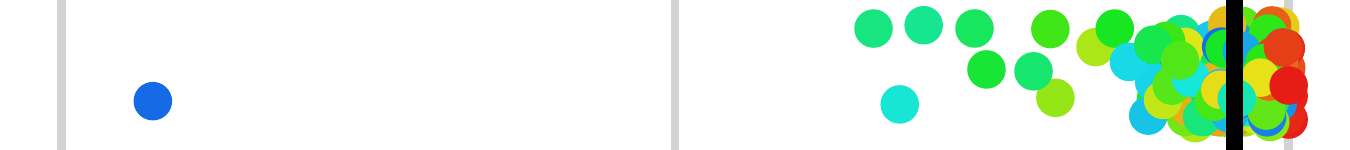} \\ 
\method{\_Kalma} & {\cellcolor[HTML]{C0C0C0}{\textcolor[HTML]{000000}{Dysts}}} &  &  & {\cellcolor[HTML]{808080}{\textcolor[HTML]{FFFFFF}{}}} & {\cellcolor[HTML]{808080}{\textcolor[HTML]{FFFFFF}{}}} & 47 & 0.98 & \inlinegraphics{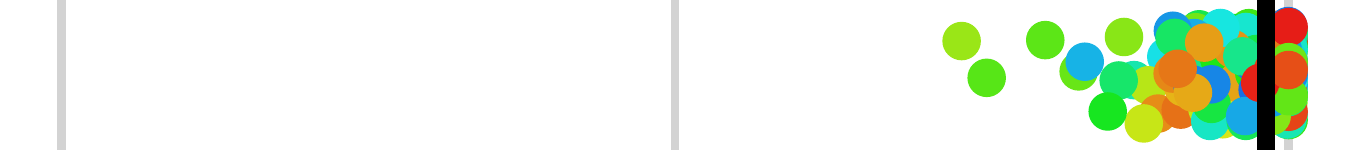} &  &  &  \\ 
\method{ConstM} & {\cellcolor[HTML]{FFA0FF}{\textcolor[HTML]{000000}{Direct}}} & Const & Mean & {\cellcolor[HTML]{148D4A}{\textcolor[HTML]{FFFFFF}{<1s}}} & {\cellcolor[HTML]{808080}{\textcolor[HTML]{FFFFFF}{}}} & 48 & 1.00 & \inlinegraphics{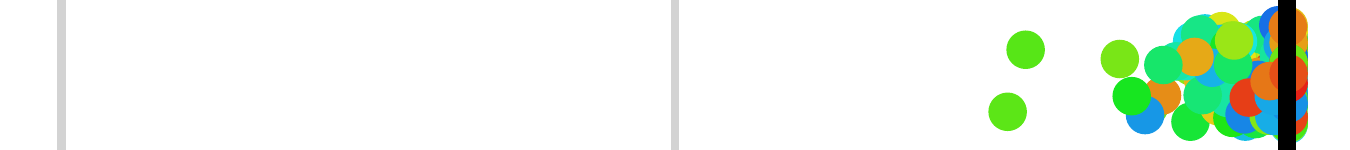} & 33 & 1.00 & \inlinegraphics{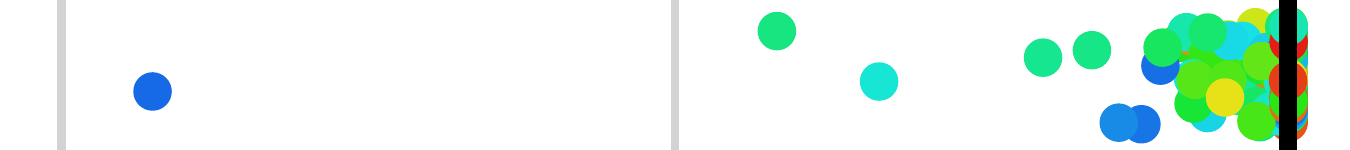} \\ 
\bottomrule
\end{tabular}
\end{center}

%% file: sec_conclusion.tex
\section{Conclusion}\label{sec:conclusion}
\begin{figure}
	\begin{center}
		{Median of Ranks vs Compute Time for \DeebLorenz{}}
		
		\vspace{0.5cm}
		
		\includegraphics[width=\textwidth]{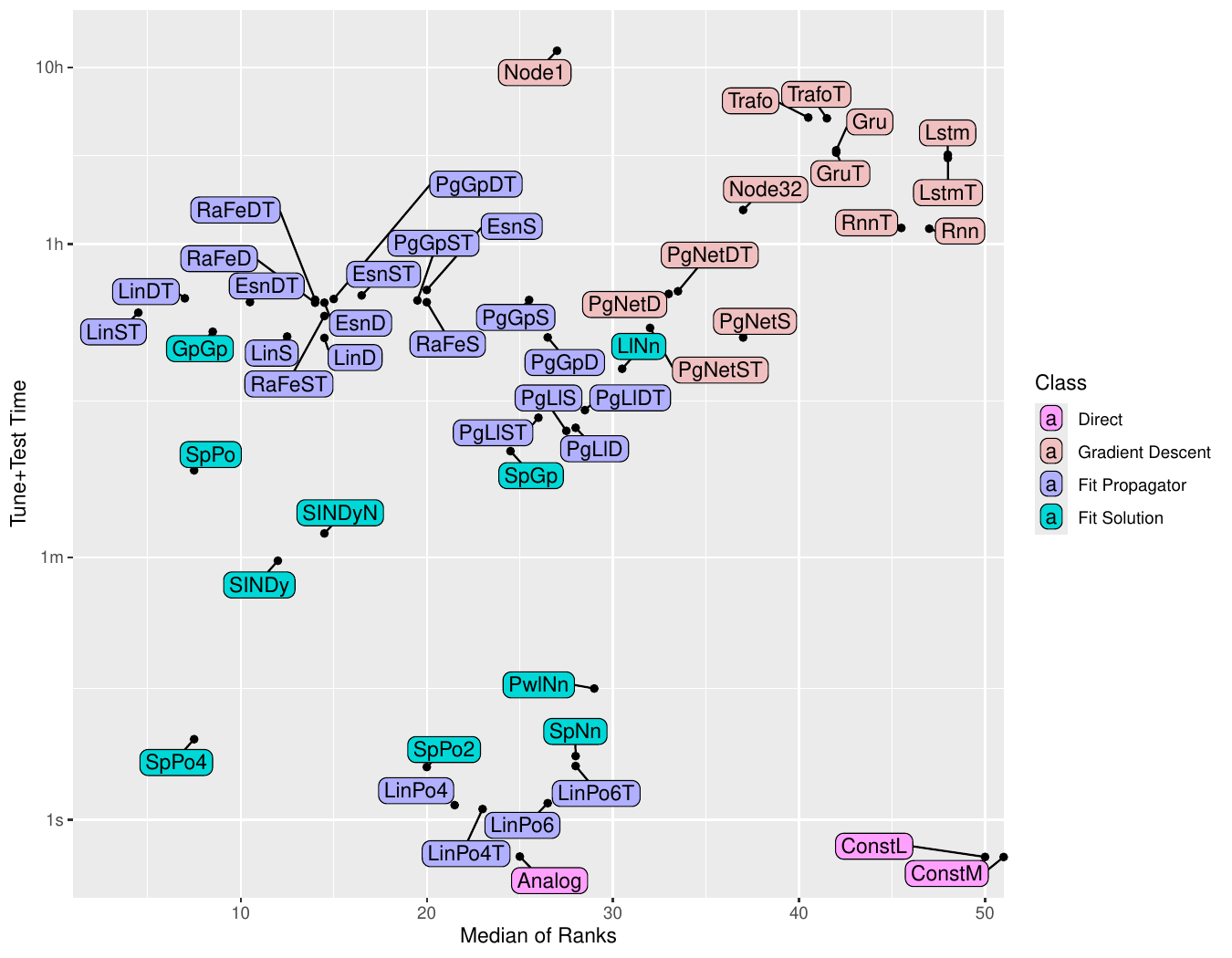}
	\end{center}
	\caption{Overall comparison of the considered methods in terms of their ranks and computational efficiency. For the horizontal axis, for each method, we take the median over their ranks according to the Cumulative Maximum Error ($\cme$) on the test data in the 12 different settings of \DeebLorenz{} (3 systems with 4 observation schemes each), see \cref{tbl:ranks:lorenz:cme}. For the vertical axis, we take the sum over the tune and test time per time series and average it over 12 settings of \DeebLorenz{}. Note that the median rank depends strongly on our specific selection of the 12 settings.}\label{fig:medianRankVsTime}
\end{figure}
This study provides a comprehensive evaluation of various methods for time series prediction across several low-dimensional chaotic dynamical systems, demonstrating that lightweight, simple methods consistently outperform more complex, computationally intensive algorithms (see \cref{fig:medianRankVsTime}). Methods based on the solution smoother approach or simple estimators of propagator maps showed both strong predictive performance and low computational costs. In contrast, more complex machine learning models that are trained using gradient descent generally underperformed. 

Hyperparameter optimization is shown to be a key factor for improving accuracy in some methods, with optimized methods achieving errors several orders of magnitude lower than those reported in a previous study with limited tuning. Additionally, noise and random timestep designs were found to significantly impact performance, with different methods excelling under different conditions.

Furthermore, this work highlights the importance of robust evaluation practices. Repeating experiments is essential to ensure statistical significance. The use of appropriate error metrics, such as the newly introduced cumulative maximum error ($\cme$), is necessary to capture relevant performance differences.

In general, meaningful evaluation of a method’s performance requires a carefully constructed study design that incorporates appropriate hyperparameter optimization, relevant baseline methods, informative error metrics, uncertainty quantification, and fair comparison practices. However, many existing studies fail to address one or more of these aspects, whereas we strive to adhere to these principles. Our findings challenge the increasing focus on complex, computationally intensive machine learning models and show that these models are often ill-suited for predicting low-dimensional chaotic systems. Instead, simpler and more efficient approaches tailored to the specific problem domain, frequently deliver superior performance.

%% file: sec_app_data.tex
\section{The Databases}\label{sec:app:data}
\subsection{\Dysts{}}\label{sec:app:data:dysts}
The \Dysts{} database consists of 133 chaotic systems with state space dimension between $3$ and $10$, see \cref{tbl:dysts:dim}. For a detailed description of the \Dysts{} database, see \cite{Gilpin21}. Here we comment on some peculiarities of the database, in particular the noisy version.
\begin{table}
    \begin{center}
        \begin{tabular}{l|ccccc}
        dimension & 3 & 4 & 5 & 6 & 10\\
        \hline
        \# systems & 102 & 19 & 2 & 3 & 7
    \end{tabular}
    \end{center}
    \caption{Number of systems in \Dysts{} with a given dimension.}\label{tbl:dysts:dim}
\end{table}

\begin{enumerate}[label=(\roman*)]
    \item
    In the noisy version of the data, the system  \model{MacArthur} is divergent (oscillating state values in the order of magnitude $\pm10^{100}$ and higher). Moreover, for \model{CircadianRhythm} and \model{DoublePendulum}, the variance of the solutions increases by multiple orders of magnitude between the noisefree and noisy version.
    \item 
    Curiously, the oscillating behavior of \model{MacArthur} makes this system rather predictable in the noisy case with a best $\cme$ of $0.055$ achieved by \method{EsnS}, which is lower than the best noisefree $\cme$ of $0.26$ achieved by \method{PgNetD}.
    \item
    For the noisefree testing dataset, the most difficult to learn systems seem to be \model{LidDrivenCavityFlow}, \model{IkedaDelay}, \model{DoublePendulum}, and \model{ForcedVanDerPol}. These all have values $0.85 < \cme < 0.96$ for the best methods in this study, while $100$ out of $133$ systems have a $\cme$ below $0.1$ for the best method.
    \item
    For the following systems from the noisy testing dataset, learning seems impossible (best method with $\cme > 0.99$):
    \model{ArnoldBeltramiChildress}, \model{ArnoldWeb}, \model{BickleyJet}. Out of $133$ noisy systems, $21$ have a $\cme$ below $0.1$ and $14$ have a $\cme$ above $0.9$.
\end{enumerate}
\subsection{\DeebLorenz{}}\label{sec:app:data:deenlorenz}
The \DeebLorenz{} database consists of three systems and four observation schemes and is separated into tuning and testing data with 10 and 100 replications, respectively.

Each system is described by an autonomous, first-order, three-dimensional ODE of the form
\begin{equation*}
    \dot u(t) = f(u(t))\eqcm\quad\text{for } t\in\R\eqcm
\end{equation*}
where $f \colon \R^3 \to \R^3$ is the vector field $f$ and the solution $u\colon\R\to\R^3$ describes the state of the system over time. For the three systems, the vector field $f$ are created as follows:
\begin{itemize}
    \item \textsc{Lorenz63std}:
        The Lorenz63 system in its standard definition \cite{Lorenz1963}, i.e.,
        \begin{equation}\label{eq:lorenz}
            f(u) =
            \begin{pmatrix}
                \sigma(u_2-u_1)\\
                u_1(\rho - u_3) - u_2\\
                u_1 u_2 - \beta u_3
            \end{pmatrix}
        \end{equation}
        with $\sigma = 10$, $\rho = 28$, $\beta = 8/3$.
    \item \textsc{Lorenz63random}:
        As \textsc{Lorenz63std}, but for each replication, the three parameters are drawn uniformly at random from an interval,
        \begin{equation}\label{eq:lorenzrandom}
            \sigma \sim \ms{Unif}([5, 15])
            \eqcm\quad
            \rho  \sim \ms{Unif}([20, 80])
            \eqcm\quad
            \beta \sim \ms{Unif}([2, 6])
            \eqfs
        \end{equation}
        These intervals are chosen so that the resulting systems exhibit chaotic behavior.
     \item \textsc{Lorenz63nonpar}:
        As \textsc{Lorenz63std}, but the parameters depend on the state, $\sigma = \sigma(u)$, $\rho = \rho(u)$, $\beta = \beta(u)$.
        For each replication, the functions $\sigma, \rho, \beta\colon \R^3 \to \R$ are sampled from a Gaussian process so that the (non-constant) parameter values are in (or at least close to) the intervals sampled from in \eqref{eq:lorenzrandom}.
        To  make sure that the sampled vector fields lead to interesting systems, we reject results where the trajectory of given initial states seem to approach a fixed point. The resulting trajectories all seem to exhibit chaotic behavior.
\end{itemize}
Note that the vector fields of \textsc{Lorenz63std} and \textsc{Lorenz63random} are polynomials of degree two with 23 out of 30 coefficients equal to zero, whereas instances of the vector fields of \textsc{Lorenz63nonpar} cannot be described by a polynomial of finite degree.

To obtain one replication of one of the systems, $f$ is set or sampled as described above. Then initial conditions $u_0$ are sampled uniformly at random from the Lorenz63 attractor (as a proxy for the attractors of the other variations of the Lorenz63 system). The initial value problem $\dot u = f(u)$, $u(0) = u_0$ is solved using a fourth order Runge--Kutta (RK4) numerical ODE solver\footnote{The R function \texttt{deSolve::ode(..., method = "rk4")}.} with constant timestep of $10^{-3}$ (one order of magnitude lower than the observation timesteps). If timesteps are random, they are sampled from an exponential distribution $\stepsize_i \sim \ms{Exp}(\lambda)$ with rate parameter $\lambda = \stepsize_0^{-1}$ and $\stepsize_0 = 10^{-2}$. If the stepsize is constant, we set $\stepsize_i = \stepsize_0$. We set $t_0 = 0$ and $t_i = t_{i-1} + \stepsize_i$. The last observation $t_n$ is chosen such that $t_{n+1}\geq T$ where $T = 100$. The number of observations is $n = 10^4$ (exactly, for constant timesteps, and, in expectation, for random time steps). For noisy observations, we draw independent noise from a multivariate normal distribution, $\epsilon_i \sim \mc N(0, \sigma^2 I_3)$ with standard deviation $\sigma = 0.1$, where $I_3$ is the $3\times 3$ identity matrix. For the noisefree observation scheme, we set $\epsilon_i = 0$. Then the observations are $Y_i = u(t_i) + \epsilon_i$. For testing, we also record $u(t_j)$, where $t_j = T + (j-n) \stepsize_0$, $j = n+1, \dots, n+m$, and $m = 10^3$.

%% file: sec_app_methods.tex
\section{Methods}\label{sec:app:methods}
\subsection{Implementation}
All methods are implemented in R \cite{R}, except for \method{Node}, which is implemented in Julia \cite{bezanson2017julia}. Furthermore,  \method{PgNet*}, \method{Trafo*}, \method{Rnn*}, \method{Lstm*}, \method{Gru*} use Keras \cite{chollet2015keras} (via its R-interface).
All methods were trained on CPUs (Intel Xeon Processor E5-2667 v3) with the exception of the indicated methods in \cref{fig:results:lorenz:length}, which utilize GPUs (NVIDIA H100).
All code for running the methods is available in the public git repositories on GitHub as listed in \cref{tbl:github}. The predictions for the test and validation data are available at \url{https://doi.org/10.5281/zenodo.12999941}.
\begin{table}
    \begin{center}
        \begin{tabular}{r|l}
            chroetz/DEEBdata & generation of DeebLorenz \\
            chroetz/DEEBcmd & command line interface for the other packages \\
            chroetz/DEEBesti & estimation methods\\
            chroetz/DEEBeval & evaluate predictions \\
            chroetz/DEEBtrajs & handling the csv-files containing observations and predictions\\
            chroetz/DEEB.jl & the Julia implementation of the neural ODE\\
            chroetz/ConfigOpts & handling the json-files containing settings for methods
        \end{tabular}
    \end{center}
    \caption{GitHub repositories containing the code used in this study.}\label{tbl:github}
\end{table}
\subsection{Normalization}\label{sec:normalize}
For all methods except \method{SINDy}, we calculate an affine linear normalization from the training data:
Let $\hat \mu = \frac1n \sum_{i=1}^{n} Y_i$ be the empirical mean and $\hat \Sigma = \frac{1}{n-1}\sum_{i=1}^{n} (Y_i - \hat\mu)(Y_i - \hat\mu)\tr$ the empirical covariance matrix. Then the normalized data is
\begin{equation}
    \bar Y_i := \hat\Sigma^{-\frac12}\br{Y_i - \hat\mu}
    \eqfs
\end{equation}
It has mean zero and an identity covariance matrix.
A prediction $\bar u(t_j)$ from a given method trained on the normalized data is transformed back by
\begin{equation}
    \hat u(t_j) := \hat\Sigma^{\frac12} \bar u(t_j) + \hat\mu
    \eqfs
\end{equation}
To not destroy potential sparsity, we only scale the inputs to \method{SINDy} by the inverse of the square root of $\hat\sigma^2 = \frac{1}{n-1}\sum_{i=1}^{n} Y_i\tr Y_i$.
\subsection{Different Variants of Methods}\label{sec:propgator}
For propagator based methods, a mapping from the current state to the next one is learned. The target is either directly the next state (suffix \method{S}) or the increment scaled by the inverse of the time step (suffix \method{D}). Additionally to the current (and potentially past) state(s), the time step can also serves as a predictor (suffix \method{T}). See also \cref{ssec:methods}.
\subsection{Hand-Tuning and Hyperparameter Grids}\label{sec:hyper}
The hyperparameter grids presented below are selected to perform well across all systems while maintaining bounded computational costs. Many advanced machine learning models, such as transformers, recurrent networks, and neural ODEs, have numerous variants. In this study, we employ standard, out-of-the-box architectures for all methods, applying minimal manual adjustments and relying on automatic hyperparameter tuning, as detailed below. While extensive fine-tuning of specific model variants could potentially improve performance for individual systems, our approach prioritizes fairness in comparison across models. This design also caters to practitioners with limited time and computational resources when applying machine learning to data from dynamical systems.
\subsection{\method{Analog}}
\begin{itemize}
	\item Parameters: margin $\omega\in \N$
	\item tuning: categorical, persistent: $\omega\in\{1, 10, 100\}$.
	\item Description: Initialize $x \leftarrow u(T)$ and $j_0 \leftarrow 0$. Let $k = \argmin_{i\in\nnset{n-\omega}} \euclof{Y_i - x}$. Set $\hat u(t_{j_0+j}) = Y_{k+j}$ for $j\in\nnset{j_{\ms{max}}}$, $j_{\ms{max}} := \min(n-k, m)$. If the prediction time series is not yet complete, repeat with $x \leftarrow Y_{k + j_{\ms{max}}}$ and $j_0 \leftarrow j_0 + j_{\ms{max}}$.
\end{itemize}
\subsection{\method{Node1}, \method{Node32}}
\begin{itemize}
	\item Parameters:
	\begin{itemize}
		\item number of hidden layers: $L$
		\item width of hidden layers: $W$
		\item activation function: swish
		\item weight decay $10^{-6}$
		\item ODE solver steps for the loss: $S$
		\item batch size: $1$ or $32$ as indicated by the suffix of the method name (and one experiment marked by \textsuperscript{\textit{*}} in \cref{fig:results:lorenz:length} with $1024$)
   		\item epochs: $400$
		\item Optimizer: AdamW
		\item learning rate: $10^{-3}$
		\item internal training--validation to use weights with lowest validation error: 85\% -- 15\%
	\end{itemize}
	\item Tuning:
	\begin{itemize}
		\item categorical, yielding: $(L,W)\in\{(2, 32), (4, 32), (2, 128)\}$
		\item scalar, exponential (factor $2$): $S = 2, 4$, $S\in[2, 64]$
	\end{itemize}
	\item Description: See \cite{chen18}. We use a Julia implementation.
    \item
    Note on Batch Size for Variable Timesteps: A batch size greater than $1$ requires equal computational steps across all elements within the batch. When timesteps vary between different segments of the training time series, this requirement is not satisfied. Forcing training in such cases by setting the timestep to a constant value leads to significantly larger error values compared to using a batch size of one. Consequently, we do not report results for \method{Node32} in the case of random timesteps.
    \item Note on Larger Networks: We evaluated wider and deeper networks using the validation dataset of \model{Lorenz63std}. However, these larger architectures either resulted in increased errors or yielded error reductions that were not statistically significant ($p > 0.1$).
\end{itemize}
\subsection{\method{PgGp*}}
\begin{itemize}
	\item Parameters:
	\begin{itemize}
		\item bandwidth: $h$
		\item kernel function: $x\mapsto\exp(x^2/(2h^2))$
		\item regularization: $\lambda$
	\end{itemize}
	\item Tuning:
	\begin{itemize}
		\item scalar, exponential (factor $2$): $h = 0.05, 0.2, 0.8$, $10^{-4} \leq h \leq 10$
		\item scalar, exponential (factor $10$): $\lambda = 10^{-12}, 10^{-8}, 10^{-4}$, $10^{-15} \leq \lambda\leq  10^2$
	\end{itemize}
	\item Description: A propagator estimator (section \ref{sec:propgator}) using a Gaussian process \cite{Rasmussen2006Gaussian}.
\end{itemize}
\subsection{\method{PgNet*}}
\begin{itemize}
	\item Parameters:
	\begin{itemize}
		\item batch size: $32$ (and one experiment marked by \textsuperscript{\textit{*}} in \cref{fig:results:lorenz:length} with $1024$)
		\item epochs: $1000$,
		\item training -- validation split: 90\% -- 10\%
		\item architecture (tuple of layer width): $w$
		\item activation function: swish
		\item learning rate $2\cdot10^{-4}$
	\end{itemize}
	\item Tuning: categorical, yielding: $w = (32, 64, 64, 32), (128, 128), (64, 128, 64)$
	\item Description: A propagator estimator (section \ref{sec:propgator}) using a vanilla feed-forward neural network implemented using Keras.
\end{itemize}
\subsection{\method{PgLl*}}
\begin{itemize}
	\item Parameters:
	\begin{itemize}
		\item kernel function: $x\mapsto\exp(x^2/h^2)$
		\item bandwidth: $h$
		\item neighbors: $k=50$
	\end{itemize}
	\item Tuning:
	\begin{itemize}
		\item scalar, exponential (factor $2$): $h = 0.05, 0.2, 0.8$, $h \in [10^{-4}, 10]$
	\end{itemize}
	\item Description: A propagator estimator (section \ref{sec:propgator}) using a local linear estimator. Locality is introduced by the kernel weights and a $k$-nearest neighbors restriction. The latter is used to keep the computational demand low.
\end{itemize}
\subsection{\method{Lin*}}\label{sec:meth:lin}
\begin{itemize}
	\item Parameters:
	\begin{itemize}
		\item past steps $K$
		\item skip steps $s$
		\item polynomial degree $\ell$
		\item penalty $\lambda$
	\end{itemize}
	\item Tuning:
	\begin{itemize}
		\item scalar, linear (step $1$): $K = 0, 1, 4$, $0\leq K \leq 32$
		\item scalar, linear (step $1$): $s = 1, 2$, $1\leq s\leq 9$
		\item scalar, linear (step $1$): $\ell = 1, 4$, $1 \leq \ell \leq 8$
		\item scalar, exponential (factor $10$): $\lambda = 10^{-12}, 10^{-8}, 10^{-4}$, $10^{-15} \leq \lambda \leq 10^2$
	\end{itemize}
	\item Description: A propagator estimator (section \ref{sec:propgator}) using Ridge regression (linear regression with $L_2$-penalty with weight $\lambda$) on polynomial features of degree at most $\ell$ of the current at time $t_i$ and past states $t_{i - sk}$, $k \in \nnset K$.
\end{itemize}
\subsection{\method{LinPo4}, \method{LinPo6}, \method{LinPo4T}, \method{LinPo6T}}
\begin{itemize}
	\item Parameters: as in section \ref{sec:meth:lin} with $K = 0$, $\ell = 4$ and $\ell = 6$, $\lambda = 0$
	\item Tuning: none
	\item Description: As \method{LinD} and \method{LinDT} (section \ref{sec:meth:lin}), respectively, but with fixed hyperparameters.
\end{itemize}
\subsection{\method{RaFe*}}\label{sec:meth:rafe}
\begin{itemize}
	\item Parameters:
	\begin{itemize}
		\item number of neurons: $400$
		\item input weight scale $c$
		\item penalty $\lambda$
		\item forward skip $\psi$
		\item random seed $r$
	\end{itemize}
	\item Tuning:
	\begin{itemize}
		\item scalar, exponential (factor $2$): $c = 0.025, 0.1, 0.4$, $10^{-7} \leq c \leq 10^2$
		\item scalar, exponential (factor $10$): $\lambda = 10^{-12}, 10^{-8}, 10^{-4}$, $10^{-15} \leq \lambda \leq 10^2$
		\item scalar, exponential (factor $2$): $\psi = 0, 1$, $1 \leq \psi \leq 64$
		\item categorical, persistent: $r = 1, 2, 3, 4$
	\end{itemize}
	\item Description: A propagator estimator (section \ref{sec:propgator}) using random feature regression, i.e., Ridge regression on features created from an untrained vanilla feed forward neural network with 1 layer of $\ell$ neurons. The forward skip $\psi$ modifies the target to $u(t) \mapsto u(t + (1+\psi)\stepsize)$, which can help in cases of high measurement noise and small stepsizes.
\end{itemize}
\subsection{\method{Esn*}}
\begin{itemize}
	\item Parameters:
	\begin{itemize}
		\item number of neurons: $400$
		\item node degree: $6$
		\item spectral radius: $0.1$
		\item input weight scale $c$
		\item penalty $\lambda$
		\item forward skip $\psi$
		\item random seed $r$
	\end{itemize}
	\item Tuning: same as \method{RaFe*} (section \ref{sec:meth:rafe})
	\item Description: See \cite{jaeger2001echo}.
\end{itemize}
\subsection{\method{Trafo*}}
\begin{itemize}
	\item Parameters:
	\begin{itemize}
		\item context length: $\ell$,
		\item position dimension: $d_{\ms{pos}}$
		\item head size: $s_{\ms{head}}$
		\item number of heads: $n_{\ms{head}}$
		\item number of blocks: $n_{\ms{block}} = 4$
		\item number of neurons in the MLP-layers: $n_{\ms{mlp}}$
		\item dropout parameter: $0.1$
		\item learning rate: $\lambda$
		\item training--validation split: 90\% -- 10\%
		\item epochs: $400$
		\item batch size: $32$ (and one experiment marked by \textsuperscript{\textit{*}} in \cref{fig:results:lorenz:length} with $1024$)
		\item positional encoding: sinusoidal
        \item loss: MSE
        \item optimizer: Adam
	\end{itemize}
	\item Tuning: We test the hyperparameter combinations $(\ell, d_{\ms{pos}}, s_{\ms{head}},  n_{\ms{head}}, n_{\ms{mlp}}, \lambda)$ given in the  rows of \cref{tbl:trafo}.
	\item Description: We use a Keras-implementation of a transformer network. As an input, we take the states at times $t_{k-\ell+1}, \dots, t_{k}$ and add a sinusoidal positional encoding of dimension $d_{\ms{pos}}$ by appending it to each of the state vectors. The network consist of $n_{\ms{block}}$ consecutive blocks consisting of an attention layer (followed by layer normalization) and a dense MLP-layer with ReLu activation (followed by layer normalization).  The attention layers have $n_{\ms{head}}$ heads of size $s_{\ms{head}}$.
\end{itemize}
\begin{table}
    \begin{center}
        \begin{tabular}{c|c|c|c|c|c}
            $\ell$ & $d_{\ms{pos}}$ & $s_{\ms{head}}$ & $n_{\ms{head}}$ & $n_{\ms{mlp}}$ & $\lambda$\\
            \hline
            $64$ & $16$ & $64$ & $ 4$ & $32$ & $10^{-3}$ \\
            $32$ & $ 8$ & $32$ & $16$ & $32$ & $10^{-4}$ \\
            $32$ & $32$ & $16$ & $16$ & $32$ & $10^{-4}$ \\
            $32$ & $ 8$ & $16$ & $16$ & $32$ & $10^{-3}$ \\
            $32$ & $ 8$ & $32$ & $ 8$ & $64$ & $10^{-4}$ \\
        \end{tabular}
    \end{center}
    \caption{Hyperparameters for the Transformer \method{Trafo}.}\label{tbl:trafo}
\end{table}
\subsection{\method{PwNn}}
\begin{itemize}
	\item Parameters: none.
	\item Tuning: none.
	\item Description: A solution smoother using piece-wise linear interpolation for estimating the solution $u$ and a nearest neighbor interpolation for estimating the vector field $f$.
\end{itemize}
\subsection{\method{SpNn}}
\begin{itemize}
	\item Parameters: none.
	\item Tuning: none.
	\item Description: A solution smoother using a cubic spline interpolation for estimating the solution $u$ and a nearest neighbor interpolation for estimating the vector field $f$.
\end{itemize}
\subsection{\method{LlNn}}
\begin{itemize}
	\item Parameters:
	\begin{itemize}
		\item kernel function: $x\mapsto\exp(x^2/h^2)$
		\item bandwidth: $h$
	\end{itemize}
	\item Tuning:
	\begin{itemize}
		\item scalar, exponential (factor $2$): $h = 0.05, 0.2, 0.8$, $10^{-4} \leq h \leq 10$
	\end{itemize}
	\item Description: A solution smoother using local linear regression for estimating the solution $u$ and a nearest neighbor interpolation for estimating the vector field $f$.
\end{itemize}
\subsection{\method{SpPo}}\label{sec:meth:sppo}
\begin{itemize}
	\item Parameters:
	\begin{itemize}
		\item polynomial degree $\ell$
		\item penalty $\lambda$
	\end{itemize}
	\item Tuning:
	\begin{itemize}
		\item scalar, linear (step $1$): $\ell = 2, 3, 4$, $1 \leq \ell \leq 8$
		\item scalar, exponential (factor $10$): $\lambda = 10^{-12}, 10^{-8}, 10^{-4}$, $10^{-15} \leq \lambda \leq 10^2$
	\end{itemize}
	\item Description: A solution smoother using a cubic spline interpolation for estimating the solution $u$ and Ridge regression with polynomial features for estimating the vector field $f$.
\end{itemize}
\subsection{\method{SpPo2}, \method{SpPo4}}
\begin{itemize}
	\item Parameters: as in section \ref{sec:meth:sppo} with $\ell = 2$ and $\ell = 4$, $\lambda = 0$
	\item Tuning: none
	\item Description: As \method{SpPo*} (section \ref{sec:meth:sppo}), but with fixed hyperparameters.
\end{itemize}
\subsection{\method{SpGp}}
\begin{itemize}
	\item Parameters:
	\begin{itemize}
		\item bandwidth: $h$
		\item neighbors:$k= 50$
		\item kernel function: $x\mapsto\exp(x^2/(2h^2))$
		\item regularization: $\lambda$
	\end{itemize}
	\item Tuning:
	\begin{itemize}
		\item scalar, exponential (factor $2$): $h = 0.05, 0.2, 0.8$, $10^{-4} \leq h \leq 10$
		\item scalar, exponential (factor $10$): $\lambda = 10^{-12}, 10^{-8}, 10^{-4}$, $10^{-15} \leq \lambda\leq  10^2$
	\end{itemize}
	\item Description: A solution smoother using a cubic spline interpolation for estimating the solution $u$ and Gaussian process regression for estimating the vector field $f$. To make the Gaussian process computationally efficient, we localize it by only considering the $k$ nearest neighbors for an evaluation.
\end{itemize}
\subsection{\method{GpGp}}
\begin{itemize}
	\item Parameters:
	\begin{itemize}
		\item bandwidth: $h$
		\item neighbors:$k= 50$
		\item kernel function: $x\mapsto\exp(x^2/(2h^2))$
		\item regularization: $\lambda$
		\item solution bandwidth: $0.1$
		\item solution regularization $\mu$
	\end{itemize}
	\item Tuning:
	\begin{itemize}
		\item scalar, exponential (factor $2$): $h = 0.05, 0.2, 0.8$, $10^{-4} \leq h \leq 10$
		\item scalar, exponential (factor $10$): $\lambda = 10^{-12}, 10^{-8}, 10^{-4}$, $10^{-15} \leq \lambda\leq  10^2$
		\item scalar, exponential (factor $10$): $\mu = 10^{-12}, 10^{-8}, 10^{-4}$, $10^{-15} \leq \mu\leq  10^2$
	\end{itemize}
	\item Description:  A solution smoother using Gaussian process regression for estimating the solution $u$ and Gaussian process regression for estimating the vector field $f$. To make the second Gaussian process computationally efficient, we localize it by only considering the $k$ nearest neighbors for an evaluation.
\end{itemize}
\subsection{\method{SINDy}, \method{SINDyN}}
\begin{itemize}
	\item Parameters:
	\begin{itemize}
		\item polynomial degree: $5$
		\item thresholding iterations: $100$
		\item threshold $\tau$
	\end{itemize}
	\item Tuning:
	\begin{itemize}
		\item scalar, exponential (factor $2$): $\tau = 0.04, 0.16, 0.64$, $10^{-7} \leq \tau \leq 10^2$
	\end{itemize}
	\item Description: A solution smoother using a cubic spline interpolation for estimating the solution $u$ and sparse linear regression with polynomial features for estimating the vector field $f$. See \cite{Brunton2016}.
\end{itemize}
\subsection{\method{Rnn*}, \method{Lstm*}, \method{Gru*}}
\begin{itemize}
    \item Parameters:
    \begin{itemize}
        \item batch size: 32 (and one experiment marked by \textsuperscript{\textit{*}} in \cref{fig:results:lorenz:length} with $1024$)
        \item epochs: 100,
        \item training -- validation split: 90\% -- 10\%
        \item architecture (tuple of layer width): $w$
        \item activation function: swish
        \item learning rate $10^{-3}$
    \end{itemize}
    \item Tuning: categorical, yielding: $w = (128), (64, 64), (32, 64, 32)$
    \item Description: Recurrent neural network architectures implemented using Keras.
\end{itemize}

%% file: sec_app_smape.tex
\section{On The Symmetric Mean Absolute Percent Error}\label{sec:app:smape}
In this section, we show some mathematical properties of the symmetric mean absolute percent error, $\smape$.

Let $S>0$ and $d\in\N$. Let $u, v\colon[0, S] \to \R^d$ be measurable functions such that there is no $t\in[0, S]$ such that $u(t) = 0 = v(t)$. Then $\smape(v,u)$ is defined as
\begin{equation*}
    \smape(v, u) := \frac{2\cdot 100}{S} \int_{0}^{S}\frac{\euclOf{v(t) - u(t)}}{\euclOf{v(t)} + \euclOf{u(t)}} \dl t
    \eqcm
\end{equation*}
where $\euclof{\cdot}$ denotes the Euclidean norm. Note that the case $u(t) = 0 = v(t)$ has to be excluded so that there is no division by zero.

The $\smape$ has following properties:
\begin{enumerate}[label=(\roman*)]
    \item $\smape(v,u) \in [0, 200]$
    \item Symmetry: $\smape(v,u) = \smape(u,v)$
    \item $\smape(v,u) = 0$ if and only if $u(t) = v(t)$ for almost all $t\in[0,S]$
    \item $\smape(v,u) = 200$ if and only if there is $\alpha\colon[0,S]\to(-\infty,0]$ such that $v(t) = \alpha(t)u(t)$ for almost all $t\in[0,S]$
    \item If $v_n \colon [0,S]\to\R^d$ for $n\in\N$ forms a sequence of functions with $\lim_{n\to\infty} \inf_{t\in[0,S]} \euclof{v_n(t)} = \infty$ and $\sup_{t\in[0,S]} \euclof{u(t)} < \infty$, then $\lim_{n\to\infty} \smape(v_n, u) = 200$
\end{enumerate}
%
\begin{proof}\mbox{}
\begin{enumerate}[label=(\roman*)]
    \item Non-negativity and triangle inequality of the Euclidean norm.
    \item Trivial.
    \item Clearly, $\smape(v,u) = 0$ if and only if $\euclof{v(t) - u(t)} = 0$ for almost all $t\in[0, S]$.
    \item On one hand, if such a function $\alpha$ exists, we have $\euclof{v(t) - u(t)} = (1+\alpha(t))\euclof{u(t)} = \euclof{v(t)} + \euclof{u(t)}$. Thus, $\smape(v,u) = 200$. On the other hand, if $\smape(v,u) = 200$, we must have $\euclof{v(t) - u(t)} = \euclof{v(t)} + \euclof{u(t)}$ for almost all $t\in[0,S]$. As the triangle inequality can only be an equality for linearly dependent vectors, there must be $\alpha(t)\in\R$ such that $v(t) = \alpha(t)u(t)$. Furthermore, $\alpha(t)$ must be non-positive, as we require $\abs{\alpha(t)-1} = \abs{\alpha(t)} + 1$.
    \item 
    Let $a_n := \inf_{t\in[0,S]} \euclof{v_n(t)}$ and $B := \sup_{t\in[0,S]} \euclof{u(t)}$. Then 
    \begin{equation*}
        \lim_{n\to\infty} \smape(v_n, u) \geq 200 \lim_{n\to\infty} \frac{a_n-B}{a_n+B} = 200
        \eqfs
    \end{equation*} 
\end{enumerate}
\end{proof}

%% file: sec_app_results.tex
\section{Further Details of the Results}\label{sec:app:tabresults}
In this section we list additional tables and plots the show the results of the simulation study in more detail.
Mean error values ($\cme$, $\smape$, $\tvalid$) for \DeebLorenz{} are shown in \cref{tbl:values:lorenz:cme,tbl:values:lorenz:smape,tbl:values:lorenz:tvalid}. The respective rankings are displayed in \cref{tbl:ranks:lorenz:cme,tbl:ranks:lorenz:smape,tbl:ranks:lorenz:tvalid}.
Furthermore, \cref{tbl:ranks:allscores} shows the ranks for all error metrics side by side and highlights the strongest disagreement.
For \Dysts{}, the median error values over all systems are shown in \cref{tbl:Dysts:medi}. These tables also include the values of the error metrics on the tuning data.
\cref{tbl:ovsT} and \cref{tbl:SvsD} detail the performances of the different variants of propagator based methods.
Results of statistical tests for distinguishing the performances of methods on \DeebLorenz{} are shown in \cref{fig:pValues} and \cref{fig:pValues10}.
\cref{fig:plane:const:lorenzStd,fig:plane:const:lorenzNonparam,fig:plane:rand:lorenzStd,fig:plane:rand:lorenzRandom,fig:plane:rand:lorenzNonparam} show variants of \cref{fig:plane:const:lorenzRandom} for the other systems and observation schemes of \DeebLorenz{}.
Finally, \cref{fig:results:lorenz:length} displays results for the system \model{Lorenz63std} of \DeebLorenz{} with reduced and increased data size.

%% file: app_tables_figures.tex
\newpage

\begin{table}
    \begin{center}
        \begin{minipage}{0.99\textwidth}
            \input{tbl/CumMaxErr_values_lorenz12.tex}

        \end{minipage}
    \end{center}
    \caption{Mean Cumulative Maximum Error ($\cme$) for the three systems of \DeebLorenz{}: \model{Lorenz63std} (S, green), \model{Lorenz63random} (R, red), \model{Lorenz63nonpar} (N, blue).}\label{tbl:values:lorenz:cme}
\end{table}

\begin{table}
    \begin{center}
        \begin{minipage}{0.99\textwidth}
            \input{tbl/sMAPE_values_lorenz12.tex}

        \end{minipage}
    \end{center}
    \caption{Mean $\smape$ for the three systems of \DeebLorenz{}: \model{Lorenz63std} (S, green), \model{Lorenz63random} (R, red), \model{Lorenz63nonpar} (N, blue).}\label{tbl:values:lorenz:smape}
\end{table}

\begin{table}
    \begin{center}
        \begin{minipage}{0.99\textwidth}
            \input{tbl/ValidTime_values_lorenz12.tex}

        \end{minipage}
    \end{center}
    \caption{Mean $\tvalid$ for the three systems of \DeebLorenz{}: \model{Lorenz63std} (S, green), \model{Lorenz63random} (R, red), \model{Lorenz63nonpar} (N, blue).}\label{tbl:values:lorenz:tvalid}
\end{table}

\begin{table}
    \begin{center}
        \begin{minipage}{0.99\textwidth}
            \input{tbl/CumMaxErr_ranks_lorenz12.tex}

        \end{minipage}
    \end{center}
    \caption{Ranking according to mean Cumulative Maximum Error ($\cme$) for the three systems of \DeebLorenz{}: \model{Lorenz63std} (S, green), \model{Lorenz63random} (R, red), \model{Lorenz63nonpar} (N, blue).}\label{tbl:ranks:lorenz:cme}
\end{table}

\begin{table}
    \begin{center}
        \begin{minipage}{0.99\textwidth}
            \input{tbl/sMAPE_ranks_lorenz12.tex}

        \end{minipage}
    \end{center}
    \caption{Ranking according to mean $\smape$ for the three systems of \DeebLorenz{}: \model{Lorenz63std} (S, green), \model{Lorenz63random} (R, red), \model{Lorenz63nonpar} (N, blue).}\label{tbl:ranks:lorenz:smape}
\end{table}

\begin{table}
    \begin{center}
        \begin{minipage}{0.99\textwidth}
            \input{tbl/ValidTime_ranks_lorenz12.tex}

        \end{minipage}
    \end{center}
    \caption{Ranking according to mean $\tvalid$ for the three systems of \DeebLorenz{}: \model{Lorenz63std} (S, green), \model{Lorenz63random} (R, red), \model{Lorenz63nonpar} (N, blue).}\label{tbl:ranks:lorenz:tvalid}
\end{table}

\begin{table}
    \begin{center}
        \begin{minipage}{0.99\textwidth}
            \input{tbl/lorenz12_ranks_allScores.tex}

        \end{minipage}
    \end{center}
    \caption{Ranks of all error metrics for the three systems of \DeebLorenz{}: \model{Lorenz63std} (S, green), \model{Lorenz63random} (R, red), \model{Lorenz63nonpar} (N, blue). The cells with the largest differences in ranks are shown in lighter color.}\label{tbl:ranks:allscores}
\end{table}

\begin{table}
    \begin{center}
        \begin{minipage}[t]{0.49\textwidth}
            \input{tbl/DeebDbDystsNoisefree_medianScores.tex}

        \end{minipage}
        \begin{minipage}[t]{0.49\textwidth}
            \input{tbl/DeebDbDystsNoisy_medianScores.tex}

        \end{minipage}
    \end{center}
    \caption{Median error metrics of tuned methods on validation and testing datasets for the \Dysts{} database. Here, $\tvalid$ is normalized so that the best value is $1$ to make different systems comparable.}\label{tbl:Dysts:medi}
\end{table}

\begin{table}
    \begin{center}
        \begin{minipage}{0.99\textwidth}
            \input{tbl/cme_ovsT_lorenz.tex}

        \end{minipage}
    \end{center}
    \caption{Relative differences, $2(\emptyset - T)/(\emptyset + T)$, in $\cme$ between default ($\emptyset$) and timestep input ($T$) variants of propagator estimators on the three systems of \DeebLorenz{} --- \model{Lorenz63std} (S, green), \model{Lorenz63random} (R, red), \model{Lorenz63nonpar} (N, blue).}\label{tbl:ovsT}
\end{table}

\begin{table}
    \begin{center}
        \begin{minipage}{0.99\textwidth}
            \input{tbl/cme_SvsD_lorenz.tex}

        \end{minipage}
    \end{center}
    \caption{Relative differences, $2(S - D)/(S + D)$, in $\cme$  between state ($S$) and difference quotient ($D$) variants of propagator estimators on the three systems of \DeebLorenz{} --- \model{Lorenz63std} (S, green), \model{Lorenz63random} (R, red), \model{Lorenz63nonpar} (N, blue) --- and median of the \Dysts{} testing datasets.}\label{tbl:SvsD}
\end{table}

\begin{figure}
    \begin{center}
        \begin{minipage}{0.99\textwidth}
            \begingroup
            \fontfamily{phv}\selectfont
            \renewcommand{\arraystretch}{1.05}
            \setlength{\tabcolsep}{0.3em}
            \fontsize{8.0pt}{10pt}\selectfont
            \begin{longtable*}{lcccc}
                \caption*{
                    {\large \DeebLorenz{} (100 repetitions), $\cme$, $t$-test, $p$-values}
                } \\
                \toprule
                &  \multicolumn{2}{c}{Constant $\stepsize$} & \multicolumn{2}{c}{Random $\stepsize$} \\
                \cmidrule(lr){2-3} \cmidrule(lr){4-5}
                & Noisefree & Noisy & Noisefree & Noisy \\
                \midrule\addlinespace[2.5pt]
                \raisebox{1.5cm}{\SymbolLorenzStandard{}}&
                \includegraphics[width=0.23\textwidth]{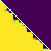}&
                \includegraphics[width=0.23\textwidth]{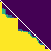}&
                \includegraphics[width=0.23\textwidth]{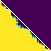}&
                \includegraphics[width=0.23\textwidth]{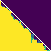}\\
                \raisebox{1.5cm}{\SymbolLorenzRandom{}}&
                \includegraphics[width=0.23\textwidth]{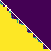}&
                \includegraphics[width=0.23\textwidth]{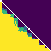}&
                \includegraphics[width=0.23\textwidth]{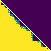}&
                \includegraphics[width=0.23\textwidth]{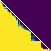}\\
                \raisebox{1.5cm}{\SymbolLorenzNonparam{}}&
                \includegraphics[width=0.23\textwidth]{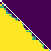}&
                \includegraphics[width=0.23\textwidth]{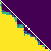}&
                \includegraphics[width=0.23\textwidth]{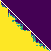}&
                \includegraphics[width=0.23\textwidth]{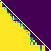}\\
                \bottomrule
            \end{longtable*}
            \endgroup
        \end{minipage}
    \end{center}
    \caption{For \DeebLorenz{}, the $p$-values for paired, one-sided $t$-tests of the null hypothesis $\cme($\method{ColumnMethod}$) \geq \cme($\method{RowMethod}$)$. The colors from yellow to dark blue indicate $p$-values in the intervals $[0, 0.001]$, $(0.001, 0.01]$, $(0.01, 0.05]$, $(0.05, 1]$. That is, a pixel in dark blue color indicates that the method corresponding to its column index is not significantly better than the method corresponding to its row index. If the color is yellow, then the difference is highly significant. The methods for the rows and columns are sorted by rank (see \cref{tbl:ranks:lorenz:cme}), with the best method at the top and on the left, respectively. The plots show that rankings are mostly statistically significant up to perturbations of a few positions.}\label{fig:pValues}
\end{figure}

\begin{figure}
    \begin{center}
        \begin{minipage}{0.99\textwidth}
            \begingroup
            \fontfamily{phv}\selectfont
            \renewcommand{\arraystretch}{1.05}
            \setlength{\tabcolsep}{0.3em}
            \fontsize{8.0pt}{10pt}\selectfont
            \begin{longtable*}{lcccc}
                \caption*{
                    {\large \DeebLorenz{} (10 repetitions), $\cme$, $t$-test, $p$-values}
                } \\
                \toprule
                &  \multicolumn{2}{c}{Constant $\stepsize$} & \multicolumn{2}{c}{Random $\stepsize$} \\
                \cmidrule(lr){2-3} \cmidrule(lr){4-5}
                & Noisefree & Noisy & Noisefree & Noisy \\
                \midrule\addlinespace[2.5pt]
                \raisebox{1.5cm}{\SymbolLorenzStandard{}}&
                \includegraphics[width=0.23\textwidth]{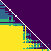}&
                \includegraphics[width=0.23\textwidth]{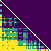}&
                \includegraphics[width=0.23\textwidth]{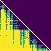}&
                \includegraphics[width=0.23\textwidth]{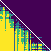}\\
                \raisebox{1.5cm}{\SymbolLorenzRandom{}}&
                \includegraphics[width=0.23\textwidth]{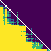}&
                \includegraphics[width=0.23\textwidth]{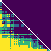}&
                \includegraphics[width=0.23\textwidth]{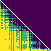}&
                \includegraphics[width=0.23\textwidth]{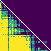}\\
                \raisebox{1.5cm}{\SymbolLorenzNonparam{}}&
                \includegraphics[width=0.23\textwidth]{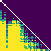}&
                \includegraphics[width=0.23\textwidth]{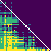}&
                \includegraphics[width=0.23\textwidth]{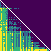}&
                \includegraphics[width=0.23\textwidth]{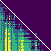}\\
                \bottomrule
            \end{longtable*}
            \endgroup
        \end{minipage}
    \end{center}
    \caption{Same as \cref{fig:pValues} but with only 10 instead of 100 repetitions. The methods are ordered as in \cref{fig:pValues}, i.e., according to \cref{tbl:ranks:lorenz:cme}. The $p$-values are larger than in \cref{fig:pValues}. This plot shows that doing only 10 repetitions leads to largely indistinguishable method performances.}\label{fig:pValues10}
\end{figure}

\begin{figure}
	\begin{center}
		{Cumulative Maximum Error for Test Data of \model{Lorenz63std} with Constant Timestep}

		\vspace{0.5cm}

		\includegraphics[width=\textwidth]{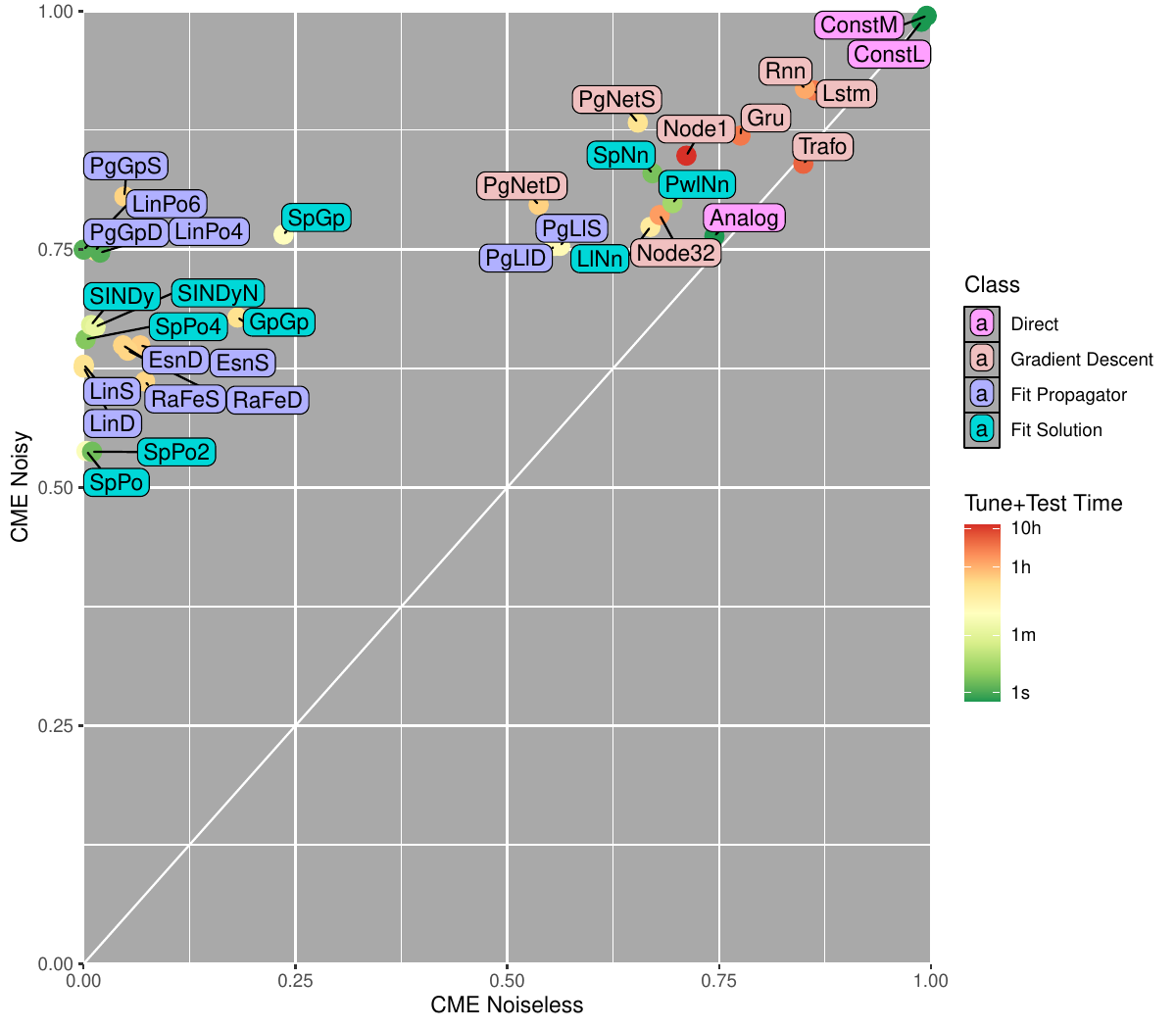}
	\end{center}
	\caption{Same as \cref{fig:plane:const:lorenzRandom} but for \model{Lorenz63std} with Constant Timestep.}\label{fig:plane:const:lorenzStd}
\end{figure}

\begin{figure}
	\begin{center}
		{Cumulative Maximum Error for Test Data of \model{Lorenz63nonpar} with Constant Timestep}

		\vspace{0.5cm}

		\includegraphics[width=\textwidth]{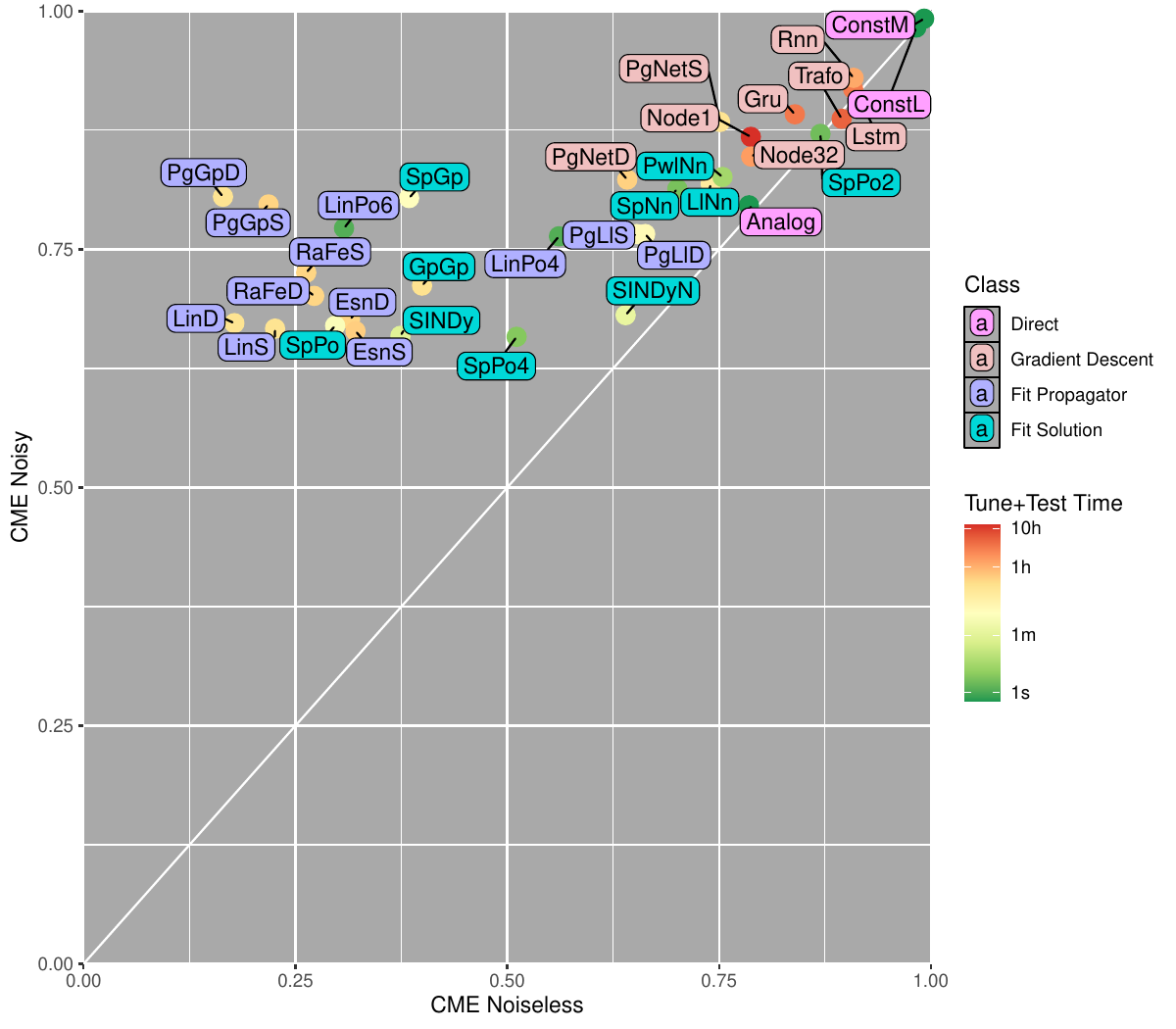}
	\end{center}
	\caption{Same as \cref{fig:plane:const:lorenzRandom} but for \model{Lorenz63nonpar} with Constant Timestep.}\label{fig:plane:const:lorenzNonparam}
\end{figure}

\begin{figure}
	\begin{center}
		{Cumulative Maximum Error for Test Data of \model{Lorenz63std} with Random Timestep}

		\vspace{0.5cm}

		\includegraphics[width=\textwidth]{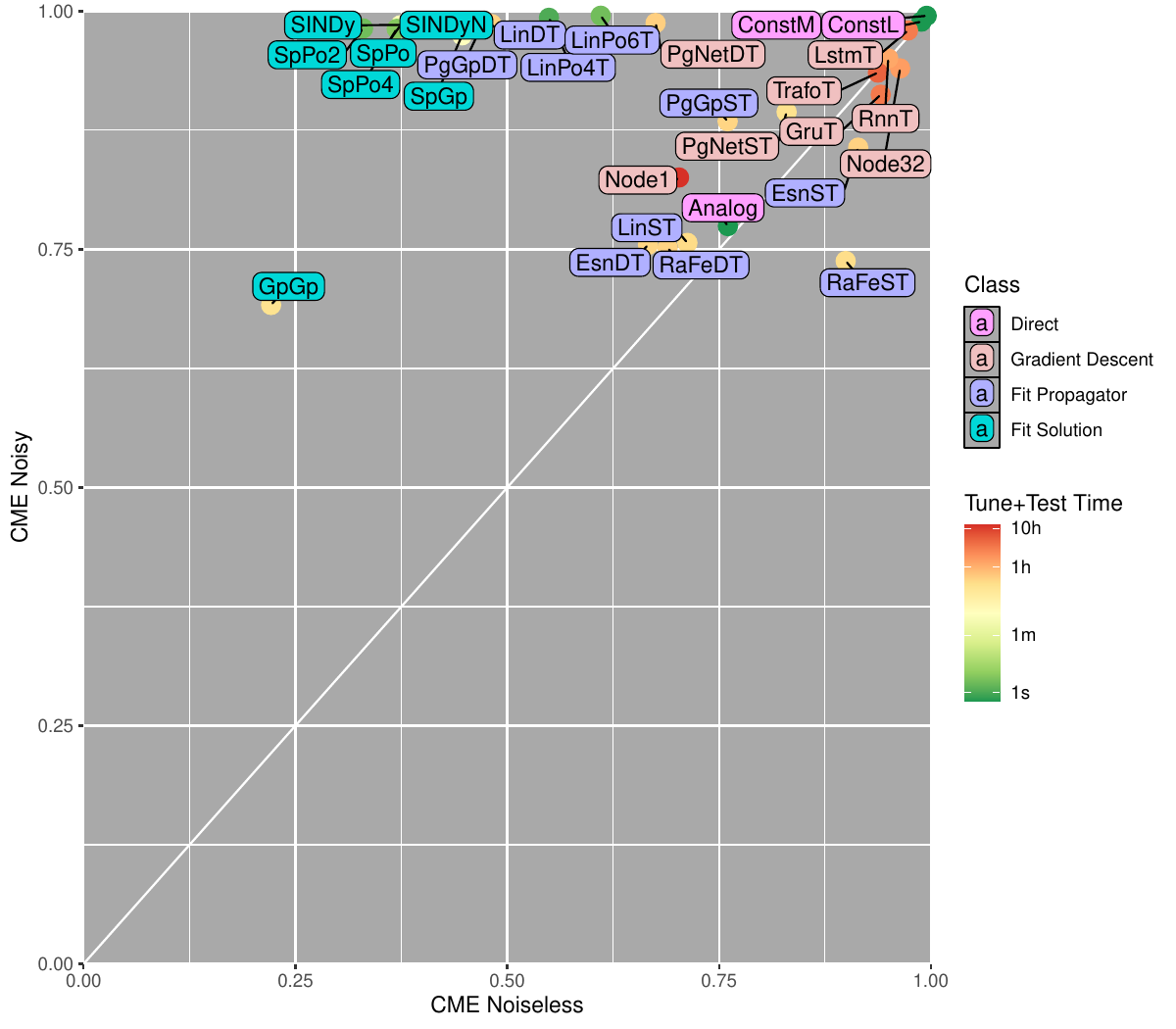}
	\end{center}
	\caption{Same as \cref{fig:plane:const:lorenzRandom} but for \model{Lorenz63std} with Random Timestep.}\label{fig:plane:rand:lorenzStd}
\end{figure}

\begin{figure}
	\begin{center}
		{Cumulative Maximum Error for Test Data of \model{Lorenz63random} with Random Timestep}

		\vspace{0.5cm}

		\includegraphics[width=\textwidth]{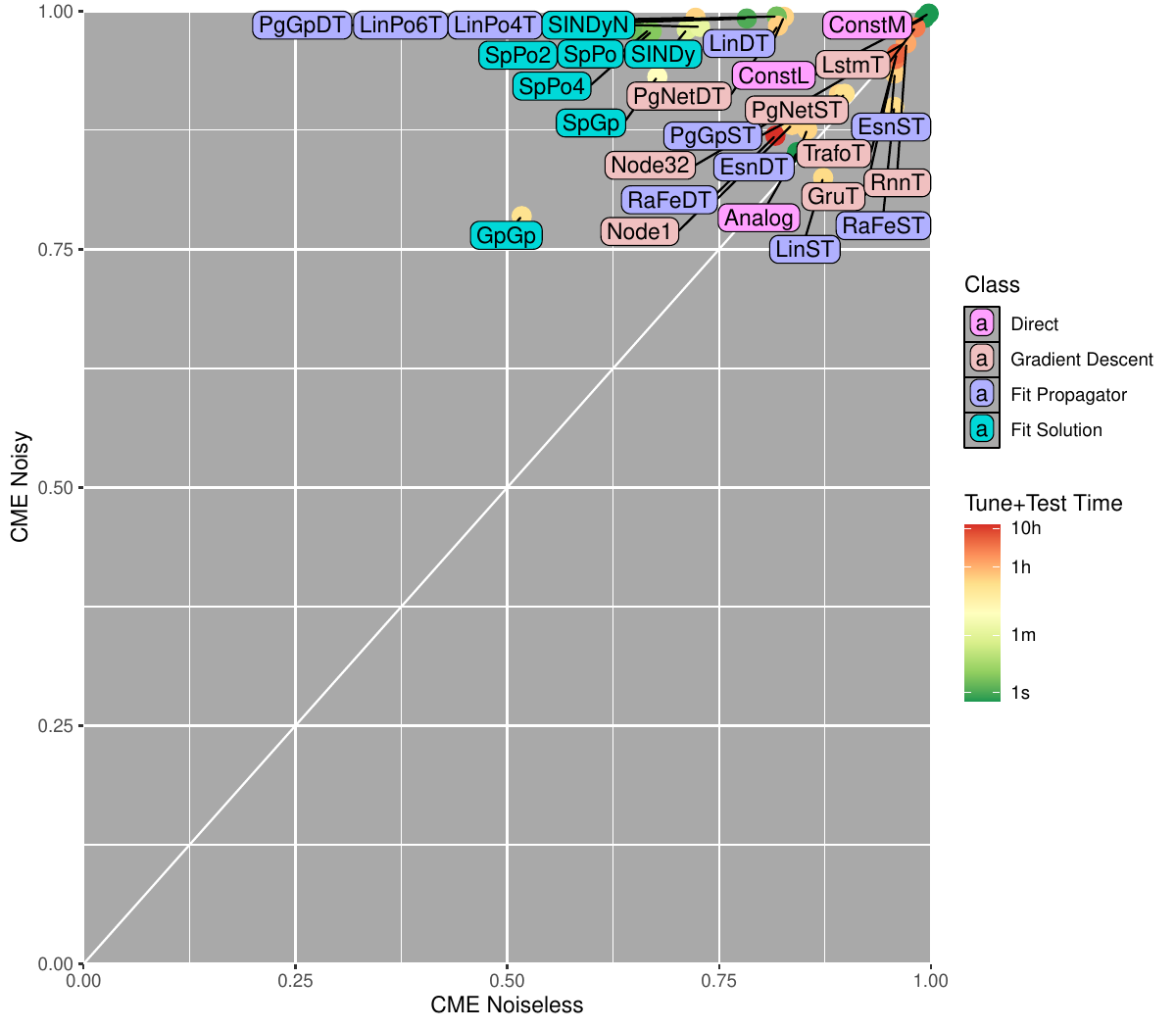}
	\end{center}
	\caption{Same as \cref{fig:plane:const:lorenzRandom} but for \model{Lorenz63random} with Random Timestep.}\label{fig:plane:rand:lorenzRandom}
\end{figure}

\begin{figure}
	\begin{center}
		{Cumulative Maximum Error for Test Data of \model{Lorenz63nonpar} with Random Timestep}

		\vspace{0.5cm}

		\includegraphics[width=\textwidth]{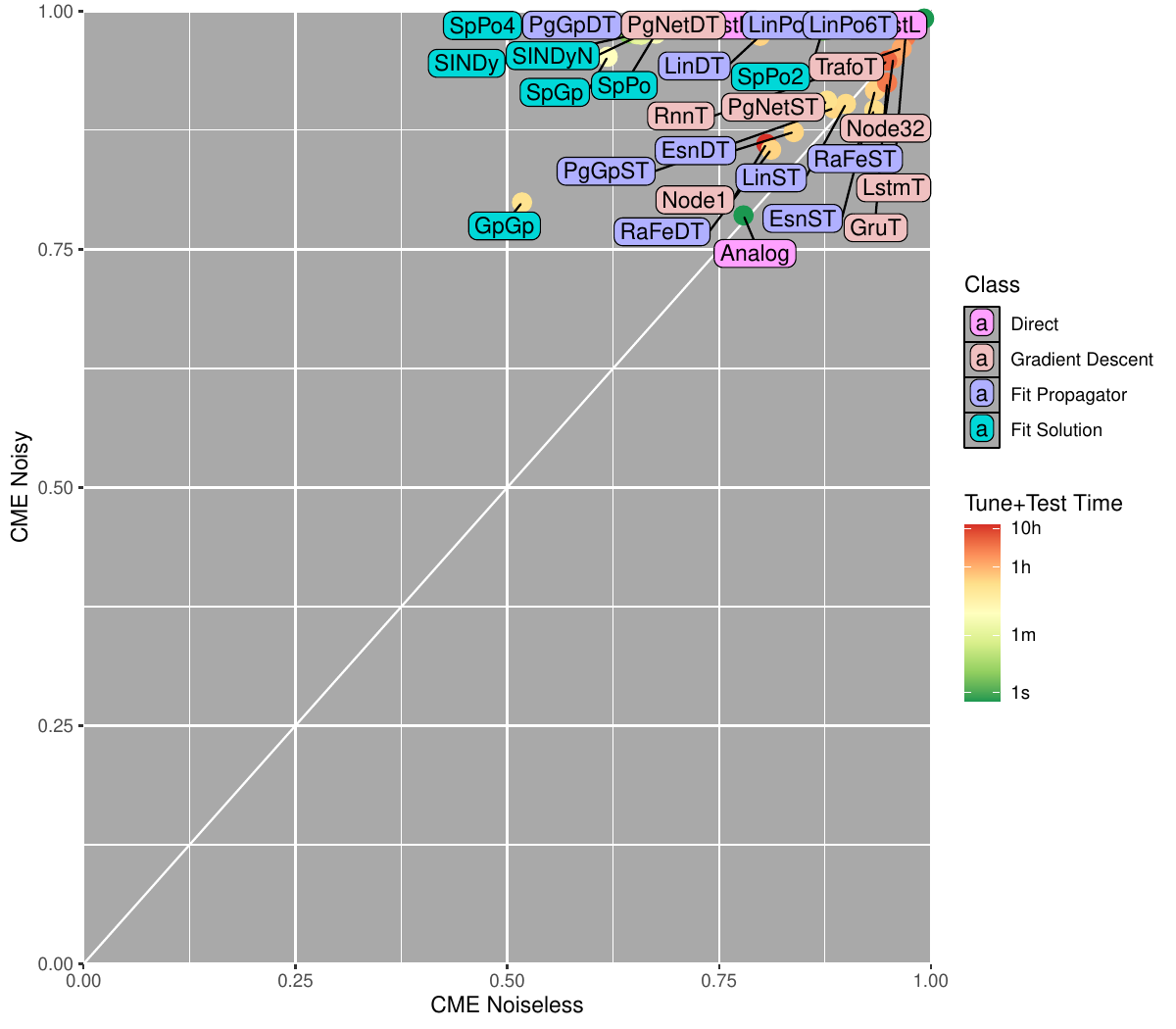}
	\end{center}
	\caption{Same as \cref{fig:plane:const:lorenzRandom} but for \model{Lorenz63nonpar} with Random Timestep.}\label{fig:plane:rand:lorenzNonparam}
\end{figure}

\begin{table}
	\begin{center}
		\input{tbl/tableLorenzLength.tex}

	\end{center}
	\caption{Cumulative Maximum Errors ($\cme$) for the test data of \model{Lorenz63std} with data increased and reduced by a factor of $10$. The setup $n = 10^4$ is the default used for all other experiments in the database \DeebLorenz{}. The precise error values shown here for $n = 10^4$ may diverge slightly from those shown in \cref{tbl:values:lorenz:cme} as new time series were sampled for this experiment. To make all results within this table comparable, shorter time series ($n \in \{10^3, 10^4\}$) are created from the longest ones ($n=10^5$) by removing data points from the beginning of the time series, so that the forecasting task is evaluated on the exact same part of the time series for all $n$.\\
    Some models, such as \method{EsnD}, do not improve when increasing the number of observations from $n=10^4$ to $n=10^5$, indicating a saturation for the give model complexity. The reservoir size of \method{EsnD} was fixed to $400$, which yields good results across all systems considered in the main part of the article. Increasing the reservoir size to $1000$ indeed decreases the error for $n=10^5$ observations ($\cme$ of $0.031$ for noisefree and $0.563$ for noisy data).}
	\label{fig:results:lorenz:length}
\end{table}

%% file: tbl/CumMaxErr_values_lorenz12.tex
\begin{center}
\caption*{
{\large \DeebLorenz{}, $\cme$, Values}
} 
\fontsize{8.0pt}{10pt}\selectfont
\fontfamily{phv}\selectfont
\renewcommand{\arraystretch}{1.05}
\setlength{\tabcolsep}{0.3em}
\begin{tabular}{lrrrrrrrrrrrrr}
\toprule
 &  & \multicolumn{6}{c}{Constant $\stepsize$} & \multicolumn{6}{c}{Random $\stepsize$} \\ 
\cmidrule(lr){3-8} \cmidrule(lr){9-14}
 &  & \multicolumn{3}{c}{Noisefree} & \multicolumn{3}{c}{Noisy} & \multicolumn{3}{c}{Noisefree} & \multicolumn{3}{c}{Noisy} \\ 
\cmidrule(lr){3-5} \cmidrule(lr){6-8} \cmidrule(lr){9-11} \cmidrule(lr){12-14}
Method & Median & \cellcolor[HTML]{00BA38}{S} & \cellcolor[HTML]{F8766D}{R} & \cellcolor[HTML]{619CFF}{N} & \cellcolor[HTML]{00BA38}{S} & \cellcolor[HTML]{F8766D}{R} & \cellcolor[HTML]{619CFF}{N} & \cellcolor[HTML]{00BA38}{S} & \cellcolor[HTML]{F8766D}{R} & \cellcolor[HTML]{619CFF}{N} & \cellcolor[HTML]{00BA38}{S} & \cellcolor[HTML]{F8766D}{R} & \cellcolor[HTML]{619CFF}{N} \\ 
\midrule\addlinespace[2.5pt]
\method{GpGp} & {\cellcolor[HTML]{29798E}{\textcolor[HTML]{FFFFFF}{0.60}}} & {\cellcolor[HTML]{85D54A}{\textcolor[HTML]{000000}{0.18}}} & {\cellcolor[HTML]{23A983}{\textcolor[HTML]{FFFFFF}{0.40}}} & {\cellcolor[HTML]{22A884}{\textcolor[HTML]{FFFFFF}{0.40}}} & {\cellcolor[HTML]{32658E}{\textcolor[HTML]{FFFFFF}{0.68}}} & {\cellcolor[HTML]{3E4C8A}{\textcolor[HTML]{FFFFFF}{0.77}}} & {\cellcolor[HTML]{365C8D}{\textcolor[HTML]{FFFFFF}{0.71}}} & {\cellcolor[HTML]{6DCE59}{\textcolor[HTML]{000000}{0.22}}} & {\cellcolor[HTML]{228C8D}{\textcolor[HTML]{FFFFFF}{0.52}}} & {\cellcolor[HTML]{228C8D}{\textcolor[HTML]{FFFFFF}{0.52}}} & {\cellcolor[HTML]{33628D}{\textcolor[HTML]{FFFFFF}{0.69}}} & {\cellcolor[HTML]{3F4889}{\textcolor[HTML]{FFFFFF}{0.78}}} & {\cellcolor[HTML]{414487}{\textcolor[HTML]{FFFFFF}{0.80}}} \\
\bottomrule
\end{tabular}
\end{center}

%% file: tbl/sMAPE_values_lorenz12.tex
\begin{center}
\caption*{
{\large \DeebLorenz{}, $\smape$, Values}
} 
\fontsize{8.0pt}{10pt}\selectfont
\fontfamily{phv}\selectfont
\renewcommand{\arraystretch}{1.05}
\setlength{\tabcolsep}{0.3em}
\begin{tabular}{lrrrrrrrrrrrrr}
\toprule
 &  & \multicolumn{6}{c}{Constant $\stepsize$} & \multicolumn{6}{c}{Random $\stepsize$} \\ 
\cmidrule(lr){3-8} \cmidrule(lr){9-14}
 &  & \multicolumn{3}{c}{Noisefree} & \multicolumn{3}{c}{Noisy} & \multicolumn{3}{c}{Noisefree} & \multicolumn{3}{c}{Noisy} \\ 
\cmidrule(lr){3-5} \cmidrule(lr){6-8} \cmidrule(lr){9-11} \cmidrule(lr){12-14}
Method & Median & \cellcolor[HTML]{00BA38}{S} & \cellcolor[HTML]{F8766D}{R} & \cellcolor[HTML]{619CFF}{N} & \cellcolor[HTML]{00BA38}{S} & \cellcolor[HTML]{F8766D}{R} & \cellcolor[HTML]{619CFF}{N} & \cellcolor[HTML]{00BA38}{S} & \cellcolor[HTML]{F8766D}{R} & \cellcolor[HTML]{619CFF}{N} & \cellcolor[HTML]{00BA38}{S} & \cellcolor[HTML]{F8766D}{R} & \cellcolor[HTML]{619CFF}{N} \\ 
\midrule\addlinespace[2.5pt]
\method{GpGp} & {\cellcolor[HTML]{7FD34E}{\textcolor[HTML]{000000}{38}}} & {\cellcolor[HTML]{DFE318}{\textcolor[HTML]{000000}{9.4}}} & {\cellcolor[HTML]{B1DD2E}{\textcolor[HTML]{000000}{23}}} & {\cellcolor[HTML]{ADDC30}{\textcolor[HTML]{000000}{24}}} & {\cellcolor[HTML]{74D055}{\textcolor[HTML]{000000}{42}}} & {\cellcolor[HTML]{64CB5E}{\textcolor[HTML]{000000}{47}}} & {\cellcolor[HTML]{75D054}{\textcolor[HTML]{000000}{42}}} & {\cellcolor[HTML]{D2E21B}{\textcolor[HTML]{000000}{13}}} & {\cellcolor[HTML]{96D83F}{\textcolor[HTML]{000000}{31}}} & {\cellcolor[HTML]{89D548}{\textcolor[HTML]{000000}{35}}} & {\cellcolor[HTML]{68CC5C}{\textcolor[HTML]{000000}{46}}} & {\cellcolor[HTML]{5DC863}{\textcolor[HTML]{000000}{50}}} & {\cellcolor[HTML]{46C06F}{\textcolor[HTML]{000000}{59}}} \\
\bottomrule
\end{tabular}
\end{center}

%% file: tbl/ValidTime_values_lorenz12.tex
\begin{center}
\caption*{
{\large \DeebLorenz{}, $\tvalid$, Values}
} 
\fontsize{8.0pt}{10pt}\selectfont
\fontfamily{phv}\selectfont
\renewcommand{\arraystretch}{1.05}
\setlength{\tabcolsep}{0.3em}
\begin{tabular}{lrrrrrrrrrrrrr}
\toprule
 &  & \multicolumn{6}{c}{Constant $\stepsize$} & \multicolumn{6}{c}{Random $\stepsize$} \\ 
\cmidrule(lr){3-8} \cmidrule(lr){9-14}
 &  & \multicolumn{3}{c}{Noisefree} & \multicolumn{3}{c}{Noisy} & \multicolumn{3}{c}{Noisefree} & \multicolumn{3}{c}{Noisy} \\ 
\cmidrule(lr){3-5} \cmidrule(lr){6-8} \cmidrule(lr){9-11} \cmidrule(lr){12-14}
Method & Median & \cellcolor[HTML]{00BA38}{S} & \cellcolor[HTML]{F8766D}{R} & \cellcolor[HTML]{619CFF}{N} & \cellcolor[HTML]{00BA38}{S} & \cellcolor[HTML]{F8766D}{R} & \cellcolor[HTML]{619CFF}{N} & \cellcolor[HTML]{00BA38}{S} & \cellcolor[HTML]{F8766D}{R} & \cellcolor[HTML]{619CFF}{N} & \cellcolor[HTML]{00BA38}{S} & \cellcolor[HTML]{F8766D}{R} & \cellcolor[HTML]{619CFF}{N} \\ 
\midrule\addlinespace[2.5pt]
\method{ConstM} & {\cellcolor[HTML]{440154}{\textcolor[HTML]{FFFFFF}{0.0}}} & {\cellcolor[HTML]{440154}{\textcolor[HTML]{FFFFFF}{0.0}}} & {\cellcolor[HTML]{440154}{\textcolor[HTML]{FFFFFF}{0.0}}} & {\cellcolor[HTML]{440154}{\textcolor[HTML]{FFFFFF}{0.0}}} & {\cellcolor[HTML]{440154}{\textcolor[HTML]{FFFFFF}{0.0}}} & {\cellcolor[HTML]{440154}{\textcolor[HTML]{FFFFFF}{0.0}}} & {\cellcolor[HTML]{440154}{\textcolor[HTML]{FFFFFF}{0.0}}} & {\cellcolor[HTML]{440154}{\textcolor[HTML]{FFFFFF}{0.0}}} & {\cellcolor[HTML]{440154}{\textcolor[HTML]{FFFFFF}{0.0}}} & {\cellcolor[HTML]{440154}{\textcolor[HTML]{FFFFFF}{0.0}}} & {\cellcolor[HTML]{440154}{\textcolor[HTML]{FFFFFF}{0.0}}} & {\cellcolor[HTML]{440154}{\textcolor[HTML]{FFFFFF}{0.0}}} & {\cellcolor[HTML]{440154}{\textcolor[HTML]{FFFFFF}{0.0}}} \\
\bottomrule
\end{tabular}
\end{center}

%% file: tbl/CumMaxErr_ranks_lorenz12.tex
\begin{center}
\caption*{
{\large \DeebLorenz{}, $\cme$, Ranks}
} 
\fontsize{8.0pt}{10pt}\selectfont
\fontfamily{phv}\selectfont
\renewcommand{\arraystretch}{1.05}
\setlength{\tabcolsep}{0.3em}
\begin{tabular}{lrrrrrrrrrrrrr}
\toprule
 &  & \multicolumn{6}{c}{Constant $\stepsize$} & \multicolumn{6}{c}{Random $\stepsize$} \\ 
\cmidrule(lr){3-8} \cmidrule(lr){9-14}
 &  & \multicolumn{3}{c}{Noisefree} & \multicolumn{3}{c}{Noisy} & \multicolumn{3}{c}{Noisefree} & \multicolumn{3}{c}{Noisy} \\ 
\cmidrule(lr){3-5} \cmidrule(lr){6-8} \cmidrule(lr){9-11} \cmidrule(lr){12-14}
Method & Median & \cellcolor[HTML]{00BA38}{S} & \cellcolor[HTML]{F8766D}{R} & \cellcolor[HTML]{619CFF}{N} & \cellcolor[HTML]{00BA38}{S} & \cellcolor[HTML]{F8766D}{R} & \cellcolor[HTML]{619CFF}{N} & \cellcolor[HTML]{00BA38}{S} & \cellcolor[HTML]{F8766D}{R} & \cellcolor[HTML]{619CFF}{N} & \cellcolor[HTML]{00BA38}{S} & \cellcolor[HTML]{F8766D}{R} & \cellcolor[HTML]{619CFF}{N} \\ 
\midrule\addlinespace[2.5pt]
\method{LinST} & {\cellcolor[HTML]{D0E11C}{\textcolor[HTML]{000000}{4.5}}} & {\cellcolor[HTML]{D7E219}{\textcolor[HTML]{000000}{4}}} & {\cellcolor[HTML]{F1E51D}{\textcolor[HTML]{000000}{2}}} & {\cellcolor[HTML]{AEDC30}{\textcolor[HTML]{000000}{7}}} & {\cellcolor[HTML]{D7E219}{\textcolor[HTML]{000000}{4}}} & {\cellcolor[HTML]{E4E419}{\textcolor[HTML]{000000}{3}}} & {\cellcolor[HTML]{F1E51D}{\textcolor[HTML]{000000}{2}}} & {\cellcolor[HTML]{33B67A}{\textcolor[HTML]{FFFFFF}{18}}} & {\cellcolor[HTML]{20A386}{\textcolor[HTML]{FFFFFF}{22}}} & {\cellcolor[HTML]{2A788E}{\textcolor[HTML]{FFFFFF}{31}}} & {\cellcolor[HTML]{C9E11F}{\textcolor[HTML]{000000}{5}}} & {\cellcolor[HTML]{F1E51D}{\textcolor[HTML]{000000}{2}}} & {\cellcolor[HTML]{86D549}{\textcolor[HTML]{000000}{10}}} \\
\bottomrule
\end{tabular}
\end{center}

%% file: tbl/sMAPE_ranks_lorenz12.tex
\begin{center}
\caption*{
{\large \DeebLorenz{}, $\smape$, Ranks}
} 
\fontsize{8.0pt}{10pt}\selectfont
\fontfamily{phv}\selectfont
\renewcommand{\arraystretch}{1.05}
\setlength{\tabcolsep}{0.3em}
\begin{tabular}{lrrrrrrrrrrrrr}
\toprule
 &  & \multicolumn{6}{c}{Constant $\stepsize$} & \multicolumn{6}{c}{Random $\stepsize$} \\ 
\cmidrule(lr){3-8} \cmidrule(lr){9-14}
 &  & \multicolumn{3}{c}{Noisefree} & \multicolumn{3}{c}{Noisy} & \multicolumn{3}{c}{Noisefree} & \multicolumn{3}{c}{Noisy} \\ 
\cmidrule(lr){3-5} \cmidrule(lr){6-8} \cmidrule(lr){9-11} \cmidrule(lr){12-14}
Method & Median & \cellcolor[HTML]{00BA38}{S} & \cellcolor[HTML]{F8766D}{R} & \cellcolor[HTML]{619CFF}{N} & \cellcolor[HTML]{00BA38}{S} & \cellcolor[HTML]{F8766D}{R} & \cellcolor[HTML]{619CFF}{N} & \cellcolor[HTML]{00BA38}{S} & \cellcolor[HTML]{F8766D}{R} & \cellcolor[HTML]{619CFF}{N} & \cellcolor[HTML]{00BA38}{S} & \cellcolor[HTML]{F8766D}{R} & \cellcolor[HTML]{619CFF}{N} \\ 
\midrule\addlinespace[2.5pt]
\method{LinST} & {\cellcolor[HTML]{C2DF23}{\textcolor[HTML]{000000}{5.5}}} & {\cellcolor[HTML]{D7E219}{\textcolor[HTML]{000000}{4}}} & {\cellcolor[HTML]{F1E51D}{\textcolor[HTML]{000000}{2}}} & {\cellcolor[HTML]{BCDF27}{\textcolor[HTML]{000000}{6}}} & {\cellcolor[HTML]{BCDF27}{\textcolor[HTML]{000000}{6}}} & {\cellcolor[HTML]{C9E11F}{\textcolor[HTML]{000000}{5}}} & {\cellcolor[HTML]{FDE725}{\textcolor[HTML]{000000}{1}}} & {\cellcolor[HTML]{26AD81}{\textcolor[HTML]{FFFFFF}{20}}} & {\cellcolor[HTML]{1F9A8A}{\textcolor[HTML]{FFFFFF}{24}}} & {\cellcolor[HTML]{2E6F8E}{\textcolor[HTML]{FFFFFF}{33}}} & {\cellcolor[HTML]{C9E11F}{\textcolor[HTML]{000000}{5}}} & {\cellcolor[HTML]{E4E419}{\textcolor[HTML]{000000}{3}}} & {\cellcolor[HTML]{33B67A}{\textcolor[HTML]{FFFFFF}{18}}} \\
\bottomrule
\end{tabular}
\end{center}

%% file: tbl/ValidTime_ranks_lorenz12.tex
\begin{center}
\caption*{
{\large \DeebLorenz{}, $\tvalid$, Ranks}
} 
\fontsize{8.0pt}{10pt}\selectfont
\fontfamily{phv}\selectfont
\renewcommand{\arraystretch}{1.05}
\setlength{\tabcolsep}{0.3em}
\begin{tabular}{lrrrrrrrrrrrrr}
\toprule
 &  & \multicolumn{6}{c}{Constant $\stepsize$} & \multicolumn{6}{c}{Random $\stepsize$} \\ 
\cmidrule(lr){3-8} \cmidrule(lr){9-14}
 &  & \multicolumn{3}{c}{Noisefree} & \multicolumn{3}{c}{Noisy} & \multicolumn{3}{c}{Noisefree} & \multicolumn{3}{c}{Noisy} \\ 
\cmidrule(lr){3-5} \cmidrule(lr){6-8} \cmidrule(lr){9-11} \cmidrule(lr){12-14}
Method & Median & \cellcolor[HTML]{00BA38}{S} & \cellcolor[HTML]{F8766D}{R} & \cellcolor[HTML]{619CFF}{N} & \cellcolor[HTML]{00BA38}{S} & \cellcolor[HTML]{F8766D}{R} & \cellcolor[HTML]{619CFF}{N} & \cellcolor[HTML]{00BA38}{S} & \cellcolor[HTML]{F8766D}{R} & \cellcolor[HTML]{619CFF}{N} & \cellcolor[HTML]{00BA38}{S} & \cellcolor[HTML]{F8766D}{R} & \cellcolor[HTML]{619CFF}{N} \\ 
\midrule\addlinespace[2.5pt]
\method{LinST} & {\cellcolor[HTML]{D0E11C}{\textcolor[HTML]{000000}{4.5}}} & {\cellcolor[HTML]{FDE725}{\textcolor[HTML]{000000}{1}}} & {\cellcolor[HTML]{F1E51D}{\textcolor[HTML]{000000}{2}}} & {\cellcolor[HTML]{C9E11F}{\textcolor[HTML]{000000}{5}}} & {\cellcolor[HTML]{D7E219}{\textcolor[HTML]{000000}{4}}} & {\cellcolor[HTML]{D7E219}{\textcolor[HTML]{000000}{4}}} & {\cellcolor[HTML]{F1E51D}{\textcolor[HTML]{000000}{2}}} & {\cellcolor[HTML]{2CB17E}{\textcolor[HTML]{FFFFFF}{19}}} & {\cellcolor[HTML]{1F9F88}{\textcolor[HTML]{FFFFFF}{23}}} & {\cellcolor[HTML]{2A788E}{\textcolor[HTML]{FFFFFF}{31}}} & {\cellcolor[HTML]{C9E11F}{\textcolor[HTML]{000000}{5}}} & {\cellcolor[HTML]{F1E51D}{\textcolor[HTML]{000000}{2}}} & {\cellcolor[HTML]{62CB5F}{\textcolor[HTML]{000000}{13}}} \\
\bottomrule
\end{tabular}
\end{center}

%% file: tbl/lorenz12_ranks_allScores.tex
\begin{center}
\caption*{
{\large Ranks for $\cme$, $\smape$, $\tvalid$}
} 
\fontsize{7.0pt}{9pt}\selectfont
\fontfamily{phv}\selectfont
\renewcommand{\arraystretch}{1.05}
\setlength{\tabcolsep}{0.3em}
\begin{tabular}{lrrrrrrrrrrrr}
\toprule
 & \multicolumn{6}{c}{Constant $\stepsize$} & \multicolumn{6}{c}{Random $\stepsize$} \\ 
\cmidrule(lr){2-7} \cmidrule(lr){8-13}
 & \multicolumn{3}{c}{Noisefree} & \multicolumn{3}{c}{Noisy} & \multicolumn{3}{c}{Noisefree} & \multicolumn{3}{c}{Noisy} \\ 
\cmidrule(lr){2-4} \cmidrule(lr){5-7} \cmidrule(lr){8-10} \cmidrule(lr){11-13}
Method & \cellcolor[HTML]{00BA38}{S} & \cellcolor[HTML]{F8766D}{R} & \cellcolor[HTML]{619CFF}{N} & \cellcolor[HTML]{00BA38}{S} & \cellcolor[HTML]{F8766D}{R} & \cellcolor[HTML]{619CFF}{N} & \cellcolor[HTML]{00BA38}{S} & \cellcolor[HTML]{F8766D}{R} & \cellcolor[HTML]{619CFF}{N} & \cellcolor[HTML]{00BA38}{S} & \cellcolor[HTML]{F8766D}{R} & \cellcolor[HTML]{619CFF}{N} \\ 
\midrule\addlinespace[2.5pt]
\method{Analog} & {\cellcolor[HTML]{440154}{\textcolor[HTML]{FFFFFF}{41,41,41}}} & {\cellcolor[HTML]{440154}{\textcolor[HTML]{FFFFFF}{41,41,41}}} & {\cellcolor[HTML]{440154}{\textcolor[HTML]{FFFFFF}{37,37,38}}} & {\cellcolor[HTML]{277F8E}{\textcolor[HTML]{FFFFFF}{29,18,29}}} & {\cellcolor[HTML]{46337E}{\textcolor[HTML]{FFFFFF}{31,29,32}}} & {\cellcolor[HTML]{365C8D}{\textcolor[HTML]{FFFFFF}{25,21,25}}} & {\cellcolor[HTML]{46337E}{\textcolor[HTML]{FFFFFF}{25,22,25}}} & {\cellcolor[HTML]{365C8D}{\textcolor[HTML]{FFFFFF}{16,13,18}}} & {\cellcolor[HTML]{46337E}{\textcolor[HTML]{FFFFFF}{9,8,10}}} & {\cellcolor[HTML]{46337E}{\textcolor[HTML]{FFFFFF}{6,6,4}}} & {\cellcolor[HTML]{440154}{\textcolor[HTML]{FFFFFF}{3,2,3}}} & {\cellcolor[HTML]{440154}{\textcolor[HTML]{FFFFFF}{1,1,2}}} \\
\bottomrule
\end{tabular}
\end{center}

%% file: tbl/DeebDbDystsNoisefree_medianScores.tex
\begin{center}
\caption*{
{\large \Dysts{}, Noisefree}
} 
\fontsize{8.0pt}{10pt}\selectfont
\fontfamily{phv}\selectfont
\renewcommand{\arraystretch}{1.05}
\setlength{\tabcolsep}{0.3em}
\begin{tabular}{lrrrrrr}
\toprule
 & \multicolumn{3}{c}{Validation} & \multicolumn{3}{c}{Test} \\ 
\cmidrule(lr){2-4} \cmidrule(lr){5-7}
Method & $\cme$ & $\smape$ & $\tvalid$ & $\cme$ & $\smape$ & $\tvalid$ \\ 
\midrule\addlinespace[2.5pt]
\method{Analog} & {\cellcolor[HTML]{2A768E}{\textcolor[HTML]{FFFFFF}{0.61}}} & {\cellcolor[HTML]{4DC26C}{\textcolor[HTML]{000000}{37}}} & {\cellcolor[HTML]{482576}{\textcolor[HTML]{FFFFFF}{0.10}}} & {\cellcolor[HTML]{29798E}{\textcolor[HTML]{FFFFFF}{0.60}}} & {\cellcolor[HTML]{57C766}{\textcolor[HTML]{000000}{34}}} & {\cellcolor[HTML]{472E7C}{\textcolor[HTML]{FFFFFF}{0.13}}} \\ 
\method{ConstL} & {\cellcolor[HTML]{471265}{\textcolor[HTML]{FFFFFF}{0.95}}} & {\cellcolor[HTML]{424186}{\textcolor[HTML]{FFFFFF}{110}}} & {\cellcolor[HTML]{450558}{\textcolor[HTML]{FFFFFF}{0.010}}} & {\cellcolor[HTML]{471164}{\textcolor[HTML]{FFFFFF}{0.96}}} & {\cellcolor[HTML]{453681}{\textcolor[HTML]{FFFFFF}{110}}} & {\cellcolor[HTML]{450558}{\textcolor[HTML]{FFFFFF}{0.010}}} \\ 
\method{ConstM} & {\cellcolor[HTML]{440155}{\textcolor[HTML]{FFFFFF}{1.0}}} & {\cellcolor[HTML]{453581}{\textcolor[HTML]{FFFFFF}{110}}} & {\cellcolor[HTML]{440154}{\textcolor[HTML]{FFFFFF}{0}}} & {\cellcolor[HTML]{440155}{\textcolor[HTML]{FFFFFF}{1.0}}} & {\cellcolor[HTML]{414487}{\textcolor[HTML]{FFFFFF}{100}}} & {\cellcolor[HTML]{440154}{\textcolor[HTML]{FFFFFF}{0}}} \\ 
\method{EsnD} & {\cellcolor[HTML]{FAE723}{\textcolor[HTML]{000000}{0.0047}}} & {\cellcolor[HTML]{FCE724}{\textcolor[HTML]{000000}{0.16}}} & {\cellcolor[HTML]{FDE725}{\textcolor[HTML]{000000}{1.0}}} & {\cellcolor[HTML]{F0E51C}{\textcolor[HTML]{000000}{0.022}}} & {\cellcolor[HTML]{FAE722}{\textcolor[HTML]{000000}{0.67}}} & {\cellcolor[HTML]{FDE725}{\textcolor[HTML]{000000}{1.0}}} \\ 
\method{EsnS} & {\cellcolor[HTML]{FBE723}{\textcolor[HTML]{000000}{0.0040}}} & {\cellcolor[HTML]{FDE725}{\textcolor[HTML]{000000}{0.12}}} & {\cellcolor[HTML]{FDE725}{\textcolor[HTML]{000000}{1.0}}} & {\cellcolor[HTML]{EBE51A}{\textcolor[HTML]{000000}{0.030}}} & {\cellcolor[HTML]{F8E621}{\textcolor[HTML]{000000}{0.96}}} & {\cellcolor[HTML]{FDE725}{\textcolor[HTML]{000000}{1.0}}} \\ 
\method{GpGp} & {\cellcolor[HTML]{55C568}{\textcolor[HTML]{000000}{0.26}}} & {\cellcolor[HTML]{BBDE27}{\textcolor[HTML]{000000}{13}}} & {\cellcolor[HTML]{31668E}{\textcolor[HTML]{FFFFFF}{0.33}}} & {\cellcolor[HTML]{2E6E8E}{\textcolor[HTML]{FFFFFF}{0.64}}} & {\cellcolor[HTML]{4FC46A}{\textcolor[HTML]{000000}{36}}} & {\cellcolor[HTML]{414287}{\textcolor[HTML]{FFFFFF}{0.20}}} \\ 
\method{Gru} & {\cellcolor[HTML]{4DC26C}{\textcolor[HTML]{000000}{0.28}}} & {\cellcolor[HTML]{BADE28}{\textcolor[HTML]{000000}{13}}} & {\cellcolor[HTML]{277E8E}{\textcolor[HTML]{FFFFFF}{0.42}}} & {\cellcolor[HTML]{22A884}{\textcolor[HTML]{FFFFFF}{0.40}}} & {\cellcolor[HTML]{A1DA38}{\textcolor[HTML]{000000}{18}}} & {\cellcolor[HTML]{2D718E}{\textcolor[HTML]{FFFFFF}{0.37}}} \\ 
\method{LinD} & {\cellcolor[HTML]{FCE724}{\textcolor[HTML]{000000}{0.0013}}} & {\cellcolor[HTML]{FDE725}{\textcolor[HTML]{000000}{0.066}}} & {\cellcolor[HTML]{FDE725}{\textcolor[HTML]{000000}{1.0}}} & {\cellcolor[HTML]{F8E621}{\textcolor[HTML]{000000}{0.0079}}} & {\cellcolor[HTML]{FBE723}{\textcolor[HTML]{000000}{0.41}}} & {\cellcolor[HTML]{FDE725}{\textcolor[HTML]{000000}{1.0}}} \\ 
\method{LinPo4} & {\cellcolor[HTML]{E9E51A}{\textcolor[HTML]{000000}{0.033}}} & {\cellcolor[HTML]{F6E620}{\textcolor[HTML]{000000}{1.5}}} & {\cellcolor[HTML]{FDE725}{\textcolor[HTML]{000000}{1.0}}} & {\cellcolor[HTML]{E4E419}{\textcolor[HTML]{000000}{0.040}}} & {\cellcolor[HTML]{F8E621}{\textcolor[HTML]{000000}{1.1}}} & {\cellcolor[HTML]{FDE725}{\textcolor[HTML]{000000}{1.0}}} \\ 
\method{LinPo6} & {\cellcolor[HTML]{1F988B}{\textcolor[HTML]{FFFFFF}{0.47}}} & {\cellcolor[HTML]{9BD93C}{\textcolor[HTML]{000000}{19}}} & {\cellcolor[HTML]{218F8D}{\textcolor[HTML]{FFFFFF}{0.49}}} & {\cellcolor[HTML]{7FD34E}{\textcolor[HTML]{000000}{0.19}}} & {\cellcolor[HTML]{C1DF24}{\textcolor[HTML]{000000}{12}}} & {\cellcolor[HTML]{1E9C89}{\textcolor[HTML]{FFFFFF}{0.55}}} \\ 
\method{LinS} & {\cellcolor[HTML]{FCE724}{\textcolor[HTML]{000000}{0.0011}}} & {\cellcolor[HTML]{FDE725}{\textcolor[HTML]{000000}{0.035}}} & {\cellcolor[HTML]{FDE725}{\textcolor[HTML]{000000}{1.0}}} & {\cellcolor[HTML]{FAE722}{\textcolor[HTML]{000000}{0.0054}}} & {\cellcolor[HTML]{FCE724}{\textcolor[HTML]{000000}{0.23}}} & {\cellcolor[HTML]{FDE725}{\textcolor[HTML]{000000}{1.0}}} \\ 
\method{LlNn} & {\cellcolor[HTML]{2E6E8E}{\textcolor[HTML]{FFFFFF}{0.64}}} & {\cellcolor[HTML]{48C16E}{\textcolor[HTML]{000000}{38}}} & {\cellcolor[HTML]{472C7A}{\textcolor[HTML]{FFFFFF}{0.12}}} & {\cellcolor[HTML]{38598C}{\textcolor[HTML]{FFFFFF}{0.73}}} & {\cellcolor[HTML]{29AF7F}{\textcolor[HTML]{FFFFFF}{48}}} & {\cellcolor[HTML]{482071}{\textcolor[HTML]{FFFFFF}{0.085}}} \\ 
\method{Lstm} & {\cellcolor[HTML]{3E4B8A}{\textcolor[HTML]{FFFFFF}{0.77}}} & {\cellcolor[HTML]{1FA187}{\textcolor[HTML]{FFFFFF}{56}}} & {\cellcolor[HTML]{481668}{\textcolor[HTML]{FFFFFF}{0.055}}} & {\cellcolor[HTML]{482173}{\textcolor[HTML]{FFFFFF}{0.91}}} & {\cellcolor[HTML]{2F6C8E}{\textcolor[HTML]{FFFFFF}{85}}} & {\cellcolor[HTML]{440154}{\textcolor[HTML]{FFFFFF}{0}}} \\ 
\method{Node1} & {\cellcolor[HTML]{87D549}{\textcolor[HTML]{000000}{0.18}}} & {\cellcolor[HTML]{D1E11C}{\textcolor[HTML]{000000}{8.9}}} & {\cellcolor[HTML]{21A685}{\textcolor[HTML]{FFFFFF}{0.59}}} & {\cellcolor[HTML]{1FA287}{\textcolor[HTML]{FFFFFF}{0.42}}} & {\cellcolor[HTML]{B7DE2A}{\textcolor[HTML]{000000}{14}}} & {\cellcolor[HTML]{2B758E}{\textcolor[HTML]{FFFFFF}{0.39}}} \\ 
\method{Node32} & {\cellcolor[HTML]{55C667}{\textcolor[HTML]{000000}{0.26}}} & {\cellcolor[HTML]{C3DF22}{\textcolor[HTML]{000000}{12}}} & {\cellcolor[HTML]{24868E}{\textcolor[HTML]{FFFFFF}{0.46}}} & {\cellcolor[HTML]{1F9F88}{\textcolor[HTML]{FFFFFF}{0.44}}} & {\cellcolor[HTML]{85D54A}{\textcolor[HTML]{000000}{24}}} & {\cellcolor[HTML]{375B8D}{\textcolor[HTML]{FFFFFF}{0.28}}} \\ 
\method{PgGpD} & {\cellcolor[HTML]{DCE318}{\textcolor[HTML]{000000}{0.052}}} & {\cellcolor[HTML]{F5E61F}{\textcolor[HTML]{000000}{1.9}}} & {\cellcolor[HTML]{F4E61E}{\textcolor[HTML]{000000}{0.98}}} & {\cellcolor[HTML]{6DCE59}{\textcolor[HTML]{000000}{0.22}}} & {\cellcolor[HTML]{D5E21A}{\textcolor[HTML]{000000}{8.1}}} & {\cellcolor[HTML]{1E9C89}{\textcolor[HTML]{FFFFFF}{0.55}}} \\ 
\method{PgGpS} & {\cellcolor[HTML]{A7DB35}{\textcolor[HTML]{000000}{0.13}}} & {\cellcolor[HTML]{E4E419}{\textcolor[HTML]{000000}{5.3}}} & {\cellcolor[HTML]{59C864}{\textcolor[HTML]{000000}{0.74}}} & {\cellcolor[HTML]{1F9E89}{\textcolor[HTML]{FFFFFF}{0.44}}} & {\cellcolor[HTML]{98D83E}{\textcolor[HTML]{000000}{20}}} & {\cellcolor[HTML]{2F6B8E}{\textcolor[HTML]{FFFFFF}{0.35}}} \\ 
\method{PgLlD} & {\cellcolor[HTML]{3ABA76}{\textcolor[HTML]{000000}{0.32}}} & {\cellcolor[HTML]{95D840}{\textcolor[HTML]{000000}{20}}} & {\cellcolor[HTML]{297A8E}{\textcolor[HTML]{FFFFFF}{0.41}}} & {\cellcolor[HTML]{2B758E}{\textcolor[HTML]{FFFFFF}{0.61}}} & {\cellcolor[HTML]{41BD72}{\textcolor[HTML]{000000}{40}}} & {\cellcolor[HTML]{365C8D}{\textcolor[HTML]{FFFFFF}{0.29}}} \\ 
\method{PgLlS} & {\cellcolor[HTML]{3FBC73}{\textcolor[HTML]{000000}{0.31}}} & {\cellcolor[HTML]{BFDF25}{\textcolor[HTML]{000000}{12}}} & {\cellcolor[HTML]{297B8E}{\textcolor[HTML]{FFFFFF}{0.41}}} & {\cellcolor[HTML]{26818E}{\textcolor[HTML]{FFFFFF}{0.57}}} & {\cellcolor[HTML]{51C46A}{\textcolor[HTML]{000000}{35}}} & {\cellcolor[HTML]{3E4B8A}{\textcolor[HTML]{FFFFFF}{0.23}}} \\ 
\method{PgNetD} & {\cellcolor[HTML]{9DD93B}{\textcolor[HTML]{000000}{0.15}}} & {\cellcolor[HTML]{D4E21A}{\textcolor[HTML]{000000}{8.3}}} & {\cellcolor[HTML]{2EB37C}{\textcolor[HTML]{FFFFFF}{0.65}}} & {\cellcolor[HTML]{57C766}{\textcolor[HTML]{000000}{0.26}}} & {\cellcolor[HTML]{C0DF25}{\textcolor[HTML]{000000}{12}}} & {\cellcolor[HTML]{24868E}{\textcolor[HTML]{FFFFFF}{0.46}}} \\ 
\method{PgNetS} & {\cellcolor[HTML]{53C569}{\textcolor[HTML]{000000}{0.27}}} & {\cellcolor[HTML]{B8DE29}{\textcolor[HTML]{000000}{14}}} & {\cellcolor[HTML]{25838E}{\textcolor[HTML]{FFFFFF}{0.45}}} & {\cellcolor[HTML]{1F998A}{\textcolor[HTML]{FFFFFF}{0.46}}} & {\cellcolor[HTML]{9ED93A}{\textcolor[HTML]{000000}{19}}} & {\cellcolor[HTML]{32658E}{\textcolor[HTML]{FFFFFF}{0.32}}} \\ 
\method{PwlNn} & {\cellcolor[HTML]{2B758E}{\textcolor[HTML]{FFFFFF}{0.61}}} & {\cellcolor[HTML]{5AC864}{\textcolor[HTML]{000000}{33}}} & {\cellcolor[HTML]{472E7C}{\textcolor[HTML]{FFFFFF}{0.13}}} & {\cellcolor[HTML]{2A788E}{\textcolor[HTML]{FFFFFF}{0.60}}} & {\cellcolor[HTML]{55C568}{\textcolor[HTML]{000000}{35}}} & {\cellcolor[HTML]{453882}{\textcolor[HTML]{FFFFFF}{0.16}}} \\ 
\method{RaFeD} & {\cellcolor[HTML]{FBE723}{\textcolor[HTML]{000000}{0.0036}}} & {\cellcolor[HTML]{FCE724}{\textcolor[HTML]{000000}{0.14}}} & {\cellcolor[HTML]{FDE725}{\textcolor[HTML]{000000}{1.0}}} & {\cellcolor[HTML]{F0E51C}{\textcolor[HTML]{000000}{0.022}}} & {\cellcolor[HTML]{F9E621}{\textcolor[HTML]{000000}{0.92}}} & {\cellcolor[HTML]{FDE725}{\textcolor[HTML]{000000}{1.0}}} \\ 
\method{RaFeS} & {\cellcolor[HTML]{FBE723}{\textcolor[HTML]{000000}{0.0039}}} & {\cellcolor[HTML]{FCE724}{\textcolor[HTML]{000000}{0.19}}} & {\cellcolor[HTML]{FDE725}{\textcolor[HTML]{000000}{1.0}}} & {\cellcolor[HTML]{EBE51B}{\textcolor[HTML]{000000}{0.029}}} & {\cellcolor[HTML]{F9E622}{\textcolor[HTML]{000000}{0.84}}} & {\cellcolor[HTML]{FDE725}{\textcolor[HTML]{000000}{1.0}}} \\ 
\method{Rnn} & {\cellcolor[HTML]{20928C}{\textcolor[HTML]{FFFFFF}{0.49}}} & {\cellcolor[HTML]{75D054}{\textcolor[HTML]{000000}{27}}} & {\cellcolor[HTML]{3E4B8A}{\textcolor[HTML]{FFFFFF}{0.23}}} & {\cellcolor[HTML]{2B748E}{\textcolor[HTML]{FFFFFF}{0.62}}} & {\cellcolor[HTML]{45BF70}{\textcolor[HTML]{000000}{39}}} & {\cellcolor[HTML]{424186}{\textcolor[HTML]{FFFFFF}{0.19}}} \\ 
\method{SINDy} & {\cellcolor[HTML]{F7E620}{\textcolor[HTML]{000000}{0.010}}} & {\cellcolor[HTML]{FCE724}{\textcolor[HTML]{000000}{0.37}}} & {\cellcolor[HTML]{FDE725}{\textcolor[HTML]{000000}{1.0}}} & {\cellcolor[HTML]{E8E419}{\textcolor[HTML]{000000}{0.035}}} & {\cellcolor[HTML]{F6E620}{\textcolor[HTML]{000000}{1.4}}} & {\cellcolor[HTML]{FDE725}{\textcolor[HTML]{000000}{1.0}}} \\ 
\method{SINDyN} & {\cellcolor[HTML]{D8E219}{\textcolor[HTML]{000000}{0.058}}} & {\cellcolor[HTML]{F3E61E}{\textcolor[HTML]{000000}{2.3}}} & {\cellcolor[HTML]{E4E419}{\textcolor[HTML]{000000}{0.96}}} & {\cellcolor[HTML]{ADDC30}{\textcolor[HTML]{000000}{0.12}}} & {\cellcolor[HTML]{E0E318}{\textcolor[HTML]{000000}{6.0}}} & {\cellcolor[HTML]{65CB5E}{\textcolor[HTML]{000000}{0.76}}} \\ 
\method{SpGp} & {\cellcolor[HTML]{58C765}{\textcolor[HTML]{000000}{0.26}}} & {\cellcolor[HTML]{B8DE29}{\textcolor[HTML]{000000}{14}}} & {\cellcolor[HTML]{2C718E}{\textcolor[HTML]{FFFFFF}{0.37}}} & {\cellcolor[HTML]{24878E}{\textcolor[HTML]{FFFFFF}{0.54}}} & {\cellcolor[HTML]{71CF57}{\textcolor[HTML]{000000}{28}}} & {\cellcolor[HTML]{375B8D}{\textcolor[HTML]{FFFFFF}{0.28}}} \\ 
\method{SpNn} & {\cellcolor[HTML]{2D718E}{\textcolor[HTML]{FFFFFF}{0.63}}} & {\cellcolor[HTML]{45C06F}{\textcolor[HTML]{000000}{38}}} & {\cellcolor[HTML]{472D7B}{\textcolor[HTML]{FFFFFF}{0.13}}} & {\cellcolor[HTML]{2B748E}{\textcolor[HTML]{FFFFFF}{0.62}}} & {\cellcolor[HTML]{50C46A}{\textcolor[HTML]{000000}{36}}} & {\cellcolor[HTML]{453882}{\textcolor[HTML]{FFFFFF}{0.16}}} \\ 
\method{SpPo} & {\cellcolor[HTML]{FCE724}{\textcolor[HTML]{000000}{0.0022}}} & {\cellcolor[HTML]{FDE725}{\textcolor[HTML]{000000}{0.061}}} & {\cellcolor[HTML]{FDE725}{\textcolor[HTML]{000000}{1.0}}} & {\cellcolor[HTML]{FBE723}{\textcolor[HTML]{000000}{0.0041}}} & {\cellcolor[HTML]{FCE724}{\textcolor[HTML]{000000}{0.18}}} & {\cellcolor[HTML]{FDE725}{\textcolor[HTML]{000000}{1.0}}} \\ 
\method{SpPo2} & {\cellcolor[HTML]{76D054}{\textcolor[HTML]{000000}{0.21}}} & {\cellcolor[HTML]{EAE51A}{\textcolor[HTML]{000000}{4.0}}} & {\cellcolor[HTML]{25848E}{\textcolor[HTML]{FFFFFF}{0.45}}} & {\cellcolor[HTML]{74D055}{\textcolor[HTML]{000000}{0.21}}} & {\cellcolor[HTML]{EBE51B}{\textcolor[HTML]{000000}{3.8}}} & {\cellcolor[HTML]{4AC16D}{\textcolor[HTML]{000000}{0.71}}} \\ 
\method{SpPo4} & {\cellcolor[HTML]{F6E620}{\textcolor[HTML]{000000}{0.012}}} & {\cellcolor[HTML]{FCE724}{\textcolor[HTML]{000000}{0.35}}} & {\cellcolor[HTML]{FDE725}{\textcolor[HTML]{000000}{1.0}}} & {\cellcolor[HTML]{F8E621}{\textcolor[HTML]{000000}{0.0079}}} & {\cellcolor[HTML]{FBE723}{\textcolor[HTML]{000000}{0.40}}} & {\cellcolor[HTML]{FDE725}{\textcolor[HTML]{000000}{1.0}}} \\ 
\method{Trafo} & {\cellcolor[HTML]{1E9D89}{\textcolor[HTML]{FFFFFF}{0.45}}} & {\cellcolor[HTML]{78D152}{\textcolor[HTML]{000000}{26}}} & {\cellcolor[HTML]{472E7C}{\textcolor[HTML]{FFFFFF}{0.13}}} & {\cellcolor[HTML]{2C718E}{\textcolor[HTML]{FFFFFF}{0.63}}} & {\cellcolor[HTML]{41BD72}{\textcolor[HTML]{000000}{40}}} & {\cellcolor[HTML]{463480}{\textcolor[HTML]{FFFFFF}{0.15}}} \\ 
\method{\_BlRNN} & {\cellcolor[HTML]{808080}{\textcolor[HTML]{FFFFFF}{}}} & {\cellcolor[HTML]{808080}{\textcolor[HTML]{FFFFFF}{}}} & {\cellcolor[HTML]{808080}{\textcolor[HTML]{FFFFFF}{}}} & {\cellcolor[HTML]{482475}{\textcolor[HTML]{FFFFFF}{0.90}}} & {\cellcolor[HTML]{34618D}{\textcolor[HTML]{FFFFFF}{90}}} & {\cellcolor[HTML]{460A5D}{\textcolor[HTML]{FFFFFF}{0.025}}} \\ 
\method{\_DLin} & {\cellcolor[HTML]{808080}{\textcolor[HTML]{FFFFFF}{}}} & {\cellcolor[HTML]{808080}{\textcolor[HTML]{FFFFFF}{}}} & {\cellcolor[HTML]{808080}{\textcolor[HTML]{FFFFFF}{}}} & {\cellcolor[HTML]{482374}{\textcolor[HTML]{FFFFFF}{0.90}}} & {\cellcolor[HTML]{297B8E}{\textcolor[HTML]{FFFFFF}{76}}} & {\cellcolor[HTML]{460A5D}{\textcolor[HTML]{FFFFFF}{0.025}}} \\ 
\method{\_Esn} & {\cellcolor[HTML]{808080}{\textcolor[HTML]{FFFFFF}{}}} & {\cellcolor[HTML]{808080}{\textcolor[HTML]{FFFFFF}{}}} & {\cellcolor[HTML]{808080}{\textcolor[HTML]{FFFFFF}{}}} & {\cellcolor[HTML]{404588}{\textcolor[HTML]{FFFFFF}{0.79}}} & {\cellcolor[HTML]{365C8D}{\textcolor[HTML]{FFFFFF}{93}}} & {\cellcolor[HTML]{482979}{\textcolor[HTML]{FFFFFF}{0.12}}} \\ 
\method{\_Kalma} & {\cellcolor[HTML]{808080}{\textcolor[HTML]{FFFFFF}{}}} & {\cellcolor[HTML]{808080}{\textcolor[HTML]{FFFFFF}{}}} & {\cellcolor[HTML]{808080}{\textcolor[HTML]{FFFFFF}{}}} & {\cellcolor[HTML]{46085C}{\textcolor[HTML]{FFFFFF}{0.98}}} & {\cellcolor[HTML]{481D6F}{\textcolor[HTML]{FFFFFF}{120}}} & {\cellcolor[HTML]{440154}{\textcolor[HTML]{FFFFFF}{0}}} \\ 
\method{\_LinRe} & {\cellcolor[HTML]{808080}{\textcolor[HTML]{FFFFFF}{}}} & {\cellcolor[HTML]{808080}{\textcolor[HTML]{FFFFFF}{}}} & {\cellcolor[HTML]{808080}{\textcolor[HTML]{FFFFFF}{}}} & {\cellcolor[HTML]{453882}{\textcolor[HTML]{FFFFFF}{0.84}}} & {\cellcolor[HTML]{20928C}{\textcolor[HTML]{FFFFFF}{64}}} & {\cellcolor[HTML]{481769}{\textcolor[HTML]{FFFFFF}{0.060}}} \\ 
\method{\_NBEAT} & {\cellcolor[HTML]{808080}{\textcolor[HTML]{FFFFFF}{}}} & {\cellcolor[HTML]{808080}{\textcolor[HTML]{FFFFFF}{}}} & {\cellcolor[HTML]{808080}{\textcolor[HTML]{FFFFFF}{}}} & {\cellcolor[HTML]{26828E}{\textcolor[HTML]{FFFFFF}{0.56}}} & {\cellcolor[HTML]{74D055}{\textcolor[HTML]{000000}{27}}} & {\cellcolor[HTML]{472E7C}{\textcolor[HTML]{FFFFFF}{0.13}}} \\ 
\method{\_NHiTS} & {\cellcolor[HTML]{808080}{\textcolor[HTML]{FFFFFF}{}}} & {\cellcolor[HTML]{808080}{\textcolor[HTML]{FFFFFF}{}}} & {\cellcolor[HTML]{808080}{\textcolor[HTML]{FFFFFF}{}}} & {\cellcolor[HTML]{34608D}{\textcolor[HTML]{FFFFFF}{0.70}}} & {\cellcolor[HTML]{45C06F}{\textcolor[HTML]{000000}{38}}} & {\cellcolor[HTML]{482576}{\textcolor[HTML]{FFFFFF}{0.10}}} \\ 
\method{\_NLin} & {\cellcolor[HTML]{808080}{\textcolor[HTML]{FFFFFF}{}}} & {\cellcolor[HTML]{808080}{\textcolor[HTML]{FFFFFF}{}}} & {\cellcolor[HTML]{808080}{\textcolor[HTML]{FFFFFF}{}}} & {\cellcolor[HTML]{482979}{\textcolor[HTML]{FFFFFF}{0.88}}} & {\cellcolor[HTML]{27808E}{\textcolor[HTML]{FFFFFF}{74}}} & {\cellcolor[HTML]{46085C}{\textcolor[HTML]{FFFFFF}{0.020}}} \\ 
\method{\_Node} & {\cellcolor[HTML]{808080}{\textcolor[HTML]{FFFFFF}{}}} & {\cellcolor[HTML]{808080}{\textcolor[HTML]{FFFFFF}{}}} & {\cellcolor[HTML]{808080}{\textcolor[HTML]{FFFFFF}{}}} & {\cellcolor[HTML]{3E4B8A}{\textcolor[HTML]{FFFFFF}{0.77}}} & {\cellcolor[HTML]{32B67A}{\textcolor[HTML]{FFFFFF}{44}}} & {\cellcolor[HTML]{471265}{\textcolor[HTML]{FFFFFF}{0.045}}} \\ 
\method{\_Nvar} & {\cellcolor[HTML]{808080}{\textcolor[HTML]{FFFFFF}{}}} & {\cellcolor[HTML]{808080}{\textcolor[HTML]{FFFFFF}{}}} & {\cellcolor[HTML]{808080}{\textcolor[HTML]{FFFFFF}{}}} & {\cellcolor[HTML]{414387}{\textcolor[HTML]{FFFFFF}{0.80}}} & {\cellcolor[HTML]{23A983}{\textcolor[HTML]{FFFFFF}{51}}} & {\cellcolor[HTML]{48196B}{\textcolor[HTML]{FFFFFF}{0.065}}} \\ 
\method{\_RaFo} & {\cellcolor[HTML]{808080}{\textcolor[HTML]{FFFFFF}{}}} & {\cellcolor[HTML]{808080}{\textcolor[HTML]{FFFFFF}{}}} & {\cellcolor[HTML]{808080}{\textcolor[HTML]{FFFFFF}{}}} & {\cellcolor[HTML]{34618D}{\textcolor[HTML]{FFFFFF}{0.69}}} & {\cellcolor[HTML]{27AD81}{\textcolor[HTML]{FFFFFF}{49}}} & {\cellcolor[HTML]{481769}{\textcolor[HTML]{FFFFFF}{0.060}}} \\ 
\method{\_RNN} & {\cellcolor[HTML]{808080}{\textcolor[HTML]{FFFFFF}{}}} & {\cellcolor[HTML]{808080}{\textcolor[HTML]{FFFFFF}{}}} & {\cellcolor[HTML]{808080}{\textcolor[HTML]{FFFFFF}{}}} & {\cellcolor[HTML]{3C4F8A}{\textcolor[HTML]{FFFFFF}{0.76}}} & {\cellcolor[HTML]{22A884}{\textcolor[HTML]{FFFFFF}{52}}} & {\cellcolor[HTML]{482173}{\textcolor[HTML]{FFFFFF}{0.090}}} \\ 
\method{\_TCN} & {\cellcolor[HTML]{808080}{\textcolor[HTML]{FFFFFF}{}}} & {\cellcolor[HTML]{808080}{\textcolor[HTML]{FFFFFF}{}}} & {\cellcolor[HTML]{808080}{\textcolor[HTML]{FFFFFF}{}}} & {\cellcolor[HTML]{48196B}{\textcolor[HTML]{FFFFFF}{0.94}}} & {\cellcolor[HTML]{443A83}{\textcolor[HTML]{FFFFFF}{110}}} & {\cellcolor[HTML]{46075A}{\textcolor[HTML]{FFFFFF}{0.015}}} \\ 
\method{\_Trafo} & {\cellcolor[HTML]{808080}{\textcolor[HTML]{FFFFFF}{}}} & {\cellcolor[HTML]{808080}{\textcolor[HTML]{FFFFFF}{}}} & {\cellcolor[HTML]{808080}{\textcolor[HTML]{FFFFFF}{}}} & {\cellcolor[HTML]{424186}{\textcolor[HTML]{FFFFFF}{0.81}}} & {\cellcolor[HTML]{1F9E89}{\textcolor[HTML]{FFFFFF}{58}}} & {\cellcolor[HTML]{46075A}{\textcolor[HTML]{FFFFFF}{0.015}}} \\ 
\method{\_XGB} & {\cellcolor[HTML]{808080}{\textcolor[HTML]{FFFFFF}{}}} & {\cellcolor[HTML]{808080}{\textcolor[HTML]{FFFFFF}{}}} & {\cellcolor[HTML]{808080}{\textcolor[HTML]{FFFFFF}{}}} & {\cellcolor[HTML]{424186}{\textcolor[HTML]{FFFFFF}{0.81}}} & {\cellcolor[HTML]{25848E}{\textcolor[HTML]{FFFFFF}{71}}} & {\cellcolor[HTML]{481467}{\textcolor[HTML]{FFFFFF}{0.050}}} \\ 
\bottomrule
\end{tabular}
\end{center}

%% file: tbl/DeebDbDystsNoisy_medianScores.tex
\begin{center}
\caption*{
{\large \Dysts{}, Noisy}
} 
\fontsize{8.0pt}{10pt}\selectfont
\fontfamily{phv}\selectfont
\renewcommand{\arraystretch}{1.05}
\setlength{\tabcolsep}{0.3em}
\begin{tabular}{lrrrrrr}
\toprule
 & \multicolumn{3}{c}{Validation} & \multicolumn{3}{c}{Test} \\ 
\cmidrule(lr){2-4} \cmidrule(lr){5-7}
Method & $\cme$ & $\smape$ & $\tvalid$ & $\cme$ & $\smape$ & $\tvalid$ \\ 
\midrule\addlinespace[2.5pt]
\method{Analog} & {\cellcolor[HTML]{3C508B}{\textcolor[HTML]{FFFFFF}{0.76}}} & {\cellcolor[HTML]{22A884}{\textcolor[HTML]{FFFFFF}{52}}} & {\cellcolor[HTML]{450558}{\textcolor[HTML]{FFFFFF}{0.010}}} & {\cellcolor[HTML]{423F85}{\textcolor[HTML]{FFFFFF}{0.81}}} & {\cellcolor[HTML]{1F9F88}{\textcolor[HTML]{FFFFFF}{57}}} & {\cellcolor[HTML]{440356}{\textcolor[HTML]{FFFFFF}{0.0050}}} \\ 
\method{ConstL} & {\cellcolor[HTML]{471063}{\textcolor[HTML]{FFFFFF}{0.96}}} & {\cellcolor[HTML]{38598C}{\textcolor[HTML]{FFFFFF}{94}}} & {\cellcolor[HTML]{450558}{\textcolor[HTML]{FFFFFF}{0.010}}} & {\cellcolor[HTML]{471264}{\textcolor[HTML]{FFFFFF}{0.96}}} & {\cellcolor[HTML]{3E4989}{\textcolor[HTML]{FFFFFF}{100}}} & {\cellcolor[HTML]{450558}{\textcolor[HTML]{FFFFFF}{0.010}}} \\ 
\method{ConstM} & {\cellcolor[HTML]{440154}{\textcolor[HTML]{FFFFFF}{1.0}}} & {\cellcolor[HTML]{3F4889}{\textcolor[HTML]{FFFFFF}{100}}} & {\cellcolor[HTML]{440154}{\textcolor[HTML]{FFFFFF}{0}}} & {\cellcolor[HTML]{440154}{\textcolor[HTML]{FFFFFF}{1.0}}} & {\cellcolor[HTML]{472C7A}{\textcolor[HTML]{FFFFFF}{110}}} & {\cellcolor[HTML]{440154}{\textcolor[HTML]{FFFFFF}{0}}} \\ 
\method{EsnD} & {\cellcolor[HTML]{2CB17E}{\textcolor[HTML]{FFFFFF}{0.36}}} & {\cellcolor[HTML]{A8DB34}{\textcolor[HTML]{000000}{17}}} & {\cellcolor[HTML]{433E85}{\textcolor[HTML]{FFFFFF}{0.18}}} & {\cellcolor[HTML]{38598C}{\textcolor[HTML]{FFFFFF}{0.73}}} & {\cellcolor[HTML]{21A585}{\textcolor[HTML]{FFFFFF}{54}}} & {\cellcolor[HTML]{482374}{\textcolor[HTML]{FFFFFF}{0.095}}} \\ 
\method{EsnS} & {\cellcolor[HTML]{26AD81}{\textcolor[HTML]{FFFFFF}{0.38}}} & {\cellcolor[HTML]{A9DB33}{\textcolor[HTML]{000000}{17}}} & {\cellcolor[HTML]{424186}{\textcolor[HTML]{FFFFFF}{0.19}}} & {\cellcolor[HTML]{3C508B}{\textcolor[HTML]{FFFFFF}{0.76}}} & {\cellcolor[HTML]{22A884}{\textcolor[HTML]{FFFFFF}{52}}} & {\cellcolor[HTML]{482173}{\textcolor[HTML]{FFFFFF}{0.090}}} \\ 
\method{GpGp} & {\cellcolor[HTML]{2C728E}{\textcolor[HTML]{FFFFFF}{0.63}}} & {\cellcolor[HTML]{3CBC74}{\textcolor[HTML]{000000}{41}}} & {\cellcolor[HTML]{481467}{\textcolor[HTML]{FFFFFF}{0.050}}} & {\cellcolor[HTML]{482374}{\textcolor[HTML]{FFFFFF}{0.90}}} & {\cellcolor[HTML]{2F6B8E}{\textcolor[HTML]{FFFFFF}{85}}} & {\cellcolor[HTML]{46085C}{\textcolor[HTML]{FFFFFF}{0.020}}} \\ 
\method{Gru} & {\cellcolor[HTML]{2C718E}{\textcolor[HTML]{FFFFFF}{0.63}}} & {\cellcolor[HTML]{32B67A}{\textcolor[HTML]{FFFFFF}{44}}} & {\cellcolor[HTML]{482576}{\textcolor[HTML]{FFFFFF}{0.10}}} & {\cellcolor[HTML]{3B528B}{\textcolor[HTML]{FFFFFF}{0.75}}} & {\cellcolor[HTML]{20A386}{\textcolor[HTML]{FFFFFF}{54}}} & {\cellcolor[HTML]{482374}{\textcolor[HTML]{FFFFFF}{0.095}}} \\ 
\method{LinD} & {\cellcolor[HTML]{20928C}{\textcolor[HTML]{FFFFFF}{0.49}}} & {\cellcolor[HTML]{80D34D}{\textcolor[HTML]{000000}{25}}} & {\cellcolor[HTML]{433D84}{\textcolor[HTML]{FFFFFF}{0.18}}} & {\cellcolor[HTML]{3E4B8A}{\textcolor[HTML]{FFFFFF}{0.77}}} & {\cellcolor[HTML]{1E9D89}{\textcolor[HTML]{FFFFFF}{58}}} & {\cellcolor[HTML]{48196B}{\textcolor[HTML]{FFFFFF}{0.065}}} \\ 
\method{LinPo4} & {\cellcolor[HTML]{453882}{\textcolor[HTML]{FFFFFF}{0.84}}} & {\cellcolor[HTML]{228B8D}{\textcolor[HTML]{FFFFFF}{68}}} & {\cellcolor[HTML]{471265}{\textcolor[HTML]{FFFFFF}{0.045}}} & {\cellcolor[HTML]{453882}{\textcolor[HTML]{FFFFFF}{0.84}}} & {\cellcolor[HTML]{21918C}{\textcolor[HTML]{FFFFFF}{65}}} & {\cellcolor[HTML]{471265}{\textcolor[HTML]{FFFFFF}{0.045}}} \\ 
\method{LinPo6} & {\cellcolor[HTML]{482677}{\textcolor[HTML]{FFFFFF}{0.89}}} & {\cellcolor[HTML]{297A8E}{\textcolor[HTML]{FFFFFF}{77}}} & {\cellcolor[HTML]{470C5F}{\textcolor[HTML]{FFFFFF}{0.030}}} & {\cellcolor[HTML]{482173}{\textcolor[HTML]{FFFFFF}{0.91}}} & {\cellcolor[HTML]{2A788E}{\textcolor[HTML]{FFFFFF}{78}}} & {\cellcolor[HTML]{460A5D}{\textcolor[HTML]{FFFFFF}{0.025}}} \\ 
\method{LinS} & {\cellcolor[HTML]{1E9B8A}{\textcolor[HTML]{FFFFFF}{0.46}}} & {\cellcolor[HTML]{7ED34F}{\textcolor[HTML]{000000}{25}}} & {\cellcolor[HTML]{453882}{\textcolor[HTML]{FFFFFF}{0.16}}} & {\cellcolor[HTML]{3E4C8A}{\textcolor[HTML]{FFFFFF}{0.77}}} & {\cellcolor[HTML]{1F9E89}{\textcolor[HTML]{FFFFFF}{57}}} & {\cellcolor[HTML]{481668}{\textcolor[HTML]{FFFFFF}{0.055}}} \\ 
\method{LlNn} & {\cellcolor[HTML]{3F4889}{\textcolor[HTML]{FFFFFF}{0.79}}} & {\cellcolor[HTML]{20A386}{\textcolor[HTML]{FFFFFF}{54}}} & {\cellcolor[HTML]{471265}{\textcolor[HTML]{FFFFFF}{0.045}}} & {\cellcolor[HTML]{472A7A}{\textcolor[HTML]{FFFFFF}{0.88}}} & {\cellcolor[HTML]{277E8E}{\textcolor[HTML]{FFFFFF}{75}}} & {\cellcolor[HTML]{460A5D}{\textcolor[HTML]{FFFFFF}{0.025}}} \\ 
\method{Lstm} & {\cellcolor[HTML]{424086}{\textcolor[HTML]{FFFFFF}{0.81}}} & {\cellcolor[HTML]{26828E}{\textcolor[HTML]{FFFFFF}{73}}} & {\cellcolor[HTML]{470E61}{\textcolor[HTML]{FFFFFF}{0.035}}} & {\cellcolor[HTML]{470E61}{\textcolor[HTML]{FFFFFF}{0.96}}} & {\cellcolor[HTML]{472D7B}{\textcolor[HTML]{FFFFFF}{110}}} & {\cellcolor[HTML]{440154}{\textcolor[HTML]{FFFFFF}{0}}} \\ 
\method{Node1} & {\cellcolor[HTML]{2A768E}{\textcolor[HTML]{FFFFFF}{0.61}}} & {\cellcolor[HTML]{35B779}{\textcolor[HTML]{FFFFFF}{43}}} & {\cellcolor[HTML]{482374}{\textcolor[HTML]{FFFFFF}{0.095}}} & {\cellcolor[HTML]{3E4B8A}{\textcolor[HTML]{FFFFFF}{0.77}}} & {\cellcolor[HTML]{1FA187}{\textcolor[HTML]{FFFFFF}{56}}} & {\cellcolor[HTML]{482071}{\textcolor[HTML]{FFFFFF}{0.085}}} \\ 
\method{Node32} & {\cellcolor[HTML]{34618D}{\textcolor[HTML]{FFFFFF}{0.69}}} & {\cellcolor[HTML]{34B679}{\textcolor[HTML]{FFFFFF}{44}}} & {\cellcolor[HTML]{481C6E}{\textcolor[HTML]{FFFFFF}{0.075}}} & {\cellcolor[HTML]{414487}{\textcolor[HTML]{FFFFFF}{0.80}}} & {\cellcolor[HTML]{1F9A8A}{\textcolor[HTML]{FFFFFF}{60}}} & {\cellcolor[HTML]{48196B}{\textcolor[HTML]{FFFFFF}{0.065}}} \\ 
\method{PgGpD} & {\cellcolor[HTML]{31688E}{\textcolor[HTML]{FFFFFF}{0.67}}} & {\cellcolor[HTML]{31B57B}{\textcolor[HTML]{FFFFFF}{45}}} & {\cellcolor[HTML]{481769}{\textcolor[HTML]{FFFFFF}{0.060}}} & {\cellcolor[HTML]{472D7B}{\textcolor[HTML]{FFFFFF}{0.88}}} & {\cellcolor[HTML]{25838E}{\textcolor[HTML]{FFFFFF}{72}}} & {\cellcolor[HTML]{470C5F}{\textcolor[HTML]{FFFFFF}{0.030}}} \\ 
\method{PgGpS} & {\cellcolor[HTML]{31688E}{\textcolor[HTML]{FFFFFF}{0.67}}} & {\cellcolor[HTML]{37B878}{\textcolor[HTML]{000000}{43}}} & {\cellcolor[HTML]{481668}{\textcolor[HTML]{FFFFFF}{0.055}}} & {\cellcolor[HTML]{46327E}{\textcolor[HTML]{FFFFFF}{0.86}}} & {\cellcolor[HTML]{228B8D}{\textcolor[HTML]{FFFFFF}{68}}} & {\cellcolor[HTML]{470C5F}{\textcolor[HTML]{FFFFFF}{0.030}}} \\ 
\method{PgLlD} & {\cellcolor[HTML]{2D718E}{\textcolor[HTML]{FFFFFF}{0.63}}} & {\cellcolor[HTML]{28AE80}{\textcolor[HTML]{FFFFFF}{49}}} & {\cellcolor[HTML]{482071}{\textcolor[HTML]{FFFFFF}{0.085}}} & {\cellcolor[HTML]{414387}{\textcolor[HTML]{FFFFFF}{0.80}}} & {\cellcolor[HTML]{238A8D}{\textcolor[HTML]{FFFFFF}{68}}} & {\cellcolor[HTML]{48196B}{\textcolor[HTML]{FFFFFF}{0.065}}} \\ 
\method{PgLlS} & {\cellcolor[HTML]{2D718E}{\textcolor[HTML]{FFFFFF}{0.63}}} & {\cellcolor[HTML]{32B67A}{\textcolor[HTML]{FFFFFF}{44}}} & {\cellcolor[HTML]{481B6D}{\textcolor[HTML]{FFFFFF}{0.070}}} & {\cellcolor[HTML]{424086}{\textcolor[HTML]{FFFFFF}{0.81}}} & {\cellcolor[HTML]{218E8D}{\textcolor[HTML]{FFFFFF}{66}}} & {\cellcolor[HTML]{481769}{\textcolor[HTML]{FFFFFF}{0.060}}} \\ 
\method{PgNetD} & {\cellcolor[HTML]{39568C}{\textcolor[HTML]{FFFFFF}{0.73}}} & {\cellcolor[HTML]{2BB17E}{\textcolor[HTML]{FFFFFF}{47}}} & {\cellcolor[HTML]{481B6D}{\textcolor[HTML]{FFFFFF}{0.070}}} & {\cellcolor[HTML]{414487}{\textcolor[HTML]{FFFFFF}{0.80}}} & {\cellcolor[HTML]{20A386}{\textcolor[HTML]{FFFFFF}{55}}} & {\cellcolor[HTML]{481E70}{\textcolor[HTML]{FFFFFF}{0.080}}} \\ 
\method{PgNetS} & {\cellcolor[HTML]{39558C}{\textcolor[HTML]{FFFFFF}{0.74}}} & {\cellcolor[HTML]{1E9B8A}{\textcolor[HTML]{FFFFFF}{59}}} & {\cellcolor[HTML]{482071}{\textcolor[HTML]{FFFFFF}{0.085}}} & {\cellcolor[HTML]{404588}{\textcolor[HTML]{FFFFFF}{0.80}}} & {\cellcolor[HTML]{20938C}{\textcolor[HTML]{FFFFFF}{63}}} & {\cellcolor[HTML]{481467}{\textcolor[HTML]{FFFFFF}{0.050}}} \\ 
\method{PwlNn} & {\cellcolor[HTML]{414387}{\textcolor[HTML]{FFFFFF}{0.80}}} & {\cellcolor[HTML]{1FA287}{\textcolor[HTML]{FFFFFF}{55}}} & {\cellcolor[HTML]{481467}{\textcolor[HTML]{FFFFFF}{0.050}}} & {\cellcolor[HTML]{443C84}{\textcolor[HTML]{FFFFFF}{0.83}}} & {\cellcolor[HTML]{21918C}{\textcolor[HTML]{FFFFFF}{65}}} & {\cellcolor[HTML]{481467}{\textcolor[HTML]{FFFFFF}{0.050}}} \\ 
\method{RaFeD} & {\cellcolor[HTML]{1E9C89}{\textcolor[HTML]{FFFFFF}{0.45}}} & {\cellcolor[HTML]{9ED93A}{\textcolor[HTML]{000000}{19}}} & {\cellcolor[HTML]{482878}{\textcolor[HTML]{FFFFFF}{0.11}}} & {\cellcolor[HTML]{443983}{\textcolor[HTML]{FFFFFF}{0.84}}} & {\cellcolor[HTML]{23898E}{\textcolor[HTML]{FFFFFF}{69}}} & {\cellcolor[HTML]{481467}{\textcolor[HTML]{FFFFFF}{0.050}}} \\ 
\method{RaFeS} & {\cellcolor[HTML]{1E9D89}{\textcolor[HTML]{FFFFFF}{0.45}}} & {\cellcolor[HTML]{9CD93B}{\textcolor[HTML]{000000}{19}}} & {\cellcolor[HTML]{482878}{\textcolor[HTML]{FFFFFF}{0.11}}} & {\cellcolor[HTML]{414287}{\textcolor[HTML]{FFFFFF}{0.80}}} & {\cellcolor[HTML]{25838E}{\textcolor[HTML]{FFFFFF}{72}}} & {\cellcolor[HTML]{481467}{\textcolor[HTML]{FFFFFF}{0.050}}} \\ 
\method{Rnn} & {\cellcolor[HTML]{355E8D}{\textcolor[HTML]{FFFFFF}{0.71}}} & {\cellcolor[HTML]{28AE80}{\textcolor[HTML]{FFFFFF}{49}}} & {\cellcolor[HTML]{482071}{\textcolor[HTML]{FFFFFF}{0.085}}} & {\cellcolor[HTML]{3E4C8A}{\textcolor[HTML]{FFFFFF}{0.77}}} & {\cellcolor[HTML]{1F998A}{\textcolor[HTML]{FFFFFF}{60}}} & {\cellcolor[HTML]{481B6D}{\textcolor[HTML]{FFFFFF}{0.070}}} \\ 
\method{SINDy} & {\cellcolor[HTML]{46337F}{\textcolor[HTML]{FFFFFF}{0.86}}} & {\cellcolor[HTML]{404688}{\textcolor[HTML]{FFFFFF}{100}}} & {\cellcolor[HTML]{471265}{\textcolor[HTML]{FFFFFF}{0.045}}} & {\cellcolor[HTML]{472F7D}{\textcolor[HTML]{FFFFFF}{0.87}}} & {\cellcolor[HTML]{3A548C}{\textcolor[HTML]{FFFFFF}{96}}} & {\cellcolor[HTML]{470E61}{\textcolor[HTML]{FFFFFF}{0.035}}} \\ 
\method{SINDyN} & {\cellcolor[HTML]{443B84}{\textcolor[HTML]{FFFFFF}{0.83}}} & {\cellcolor[HTML]{1F998A}{\textcolor[HTML]{FFFFFF}{60}}} & {\cellcolor[HTML]{481668}{\textcolor[HTML]{FFFFFF}{0.055}}} & {\cellcolor[HTML]{472E7C}{\textcolor[HTML]{FFFFFF}{0.87}}} & {\cellcolor[HTML]{24878E}{\textcolor[HTML]{FFFFFF}{70}}} & {\cellcolor[HTML]{470E61}{\textcolor[HTML]{FFFFFF}{0.035}}} \\ 
\method{SpGp} & {\cellcolor[HTML]{424086}{\textcolor[HTML]{FFFFFF}{0.81}}} & {\cellcolor[HTML]{24AA83}{\textcolor[HTML]{FFFFFF}{51}}} & {\cellcolor[HTML]{470E61}{\textcolor[HTML]{FFFFFF}{0.035}}} & {\cellcolor[HTML]{482071}{\textcolor[HTML]{FFFFFF}{0.92}}} & {\cellcolor[HTML]{2C718E}{\textcolor[HTML]{FFFFFF}{82}}} & {\cellcolor[HTML]{46085C}{\textcolor[HTML]{FFFFFF}{0.020}}} \\ 
\method{SpNn} & {\cellcolor[HTML]{424186}{\textcolor[HTML]{FFFFFF}{0.81}}} & {\cellcolor[HTML]{1F998A}{\textcolor[HTML]{FFFFFF}{60}}} & {\cellcolor[HTML]{471063}{\textcolor[HTML]{FFFFFF}{0.040}}} & {\cellcolor[HTML]{463480}{\textcolor[HTML]{FFFFFF}{0.85}}} & {\cellcolor[HTML]{20938C}{\textcolor[HTML]{FFFFFF}{63}}} & {\cellcolor[HTML]{470E61}{\textcolor[HTML]{FFFFFF}{0.035}}} \\ 
\method{SpPo} & {\cellcolor[HTML]{2F6C8E}{\textcolor[HTML]{FFFFFF}{0.65}}} & {\cellcolor[HTML]{3FBC73}{\textcolor[HTML]{000000}{40}}} & {\cellcolor[HTML]{482374}{\textcolor[HTML]{FFFFFF}{0.095}}} & {\cellcolor[HTML]{3E4989}{\textcolor[HTML]{FFFFFF}{0.78}}} & {\cellcolor[HTML]{22A884}{\textcolor[HTML]{FFFFFF}{52}}} & {\cellcolor[HTML]{48196B}{\textcolor[HTML]{FFFFFF}{0.065}}} \\ 
\method{SpPo2} & {\cellcolor[HTML]{404688}{\textcolor[HTML]{FFFFFF}{0.79}}} & {\cellcolor[HTML]{21A585}{\textcolor[HTML]{FFFFFF}{54}}} & {\cellcolor[HTML]{481668}{\textcolor[HTML]{FFFFFF}{0.055}}} & {\cellcolor[HTML]{3F4889}{\textcolor[HTML]{FFFFFF}{0.79}}} & {\cellcolor[HTML]{24AA83}{\textcolor[HTML]{FFFFFF}{51}}} & {\cellcolor[HTML]{481467}{\textcolor[HTML]{FFFFFF}{0.050}}} \\ 
\method{SpPo4} & {\cellcolor[HTML]{424086}{\textcolor[HTML]{FFFFFF}{0.81}}} & {\cellcolor[HTML]{228D8D}{\textcolor[HTML]{FFFFFF}{67}}} & {\cellcolor[HTML]{481769}{\textcolor[HTML]{FFFFFF}{0.060}}} & {\cellcolor[HTML]{453581}{\textcolor[HTML]{FFFFFF}{0.85}}} & {\cellcolor[HTML]{1F958B}{\textcolor[HTML]{FFFFFF}{62}}} & {\cellcolor[HTML]{471265}{\textcolor[HTML]{FFFFFF}{0.045}}} \\ 
\method{Trafo} & {\cellcolor[HTML]{2C718E}{\textcolor[HTML]{FFFFFF}{0.63}}} & {\cellcolor[HTML]{36B779}{\textcolor[HTML]{FFFFFF}{43}}} & {\cellcolor[HTML]{470C5F}{\textcolor[HTML]{FFFFFF}{0.030}}} & {\cellcolor[HTML]{3D4D8A}{\textcolor[HTML]{FFFFFF}{0.77}}} & {\cellcolor[HTML]{1FA088}{\textcolor[HTML]{FFFFFF}{57}}} & {\cellcolor[HTML]{471063}{\textcolor[HTML]{FFFFFF}{0.040}}} \\ 
\bottomrule
\end{tabular}
\end{center}

%% file: tbl/cme_ovsT_lorenz.tex
\begin{center}
\caption*{
{\large Relative Differences in $\cme$ When Adding the Timestep to the Input ($\emptyset \to T$)}
} 
\fontsize{8.0pt}{10pt}\selectfont
\fontfamily{phv}\selectfont
\renewcommand{\arraystretch}{1.05}
\setlength{\tabcolsep}{0.3em}
\begin{tabular}{lrrrrrrrrrrrrr}
\toprule
 &  & \multicolumn{6}{c}{Constant $\stepsize$} & \multicolumn{6}{c}{Random $\stepsize$} \\ 
\cmidrule(lr){3-8} \cmidrule(lr){9-14}
 &  & \multicolumn{3}{c}{Noisefree} & \multicolumn{3}{c}{Noisy} & \multicolumn{3}{c}{Noisefree} & \multicolumn{3}{c}{Noisy} \\ 
\cmidrule(lr){3-5} \cmidrule(lr){6-8} \cmidrule(lr){9-11} \cmidrule(lr){12-14}
Method &  & \cellcolor[HTML]{00BA38}{S} & \cellcolor[HTML]{F8766D}{R} & \cellcolor[HTML]{619CFF}{N} & \cellcolor[HTML]{00BA38}{S} & \cellcolor[HTML]{F8766D}{R} & \cellcolor[HTML]{619CFF}{N} & \cellcolor[HTML]{00BA38}{S} & \cellcolor[HTML]{F8766D}{R} & \cellcolor[HTML]{619CFF}{N} & \cellcolor[HTML]{00BA38}{S} & \cellcolor[HTML]{F8766D}{R} & \cellcolor[HTML]{619CFF}{N} \\ 
\midrule\addlinespace[2.5pt]
\multicolumn{2}{r}{$T<\emptyset$}  & 22\% & 50\% & 39\% & 28\% & 61\% & 44\% & 94\% & 83\% & 89\% & 83\% & 78\% & 72\% \\ 
\method{EsnD} & 67\% & {\cellcolor[HTML]{DDEFEC}{\textcolor[HTML]{000000}{0.05}}} & {\cellcolor[HTML]{F6F2E8}{\textcolor[HTML]{000000}{-0.03}}} & {\cellcolor[HTML]{F6ECD4}{\textcolor[HTML]{000000}{-0.07}}} & {\cellcolor[HTML]{EAF2F1}{\textcolor[HTML]{000000}{0.03}}} & {\cellcolor[HTML]{F4F5F5}{\textcolor[HTML]{000000}{0.00}}} & {\cellcolor[HTML]{E8F2F0}{\textcolor[HTML]{000000}{0.03}}} & {\cellcolor[HTML]{C7EAE5}{\textcolor[HTML]{000000}{0.10}}} & {\cellcolor[HTML]{F5F4F2}{\textcolor[HTML]{000000}{-0.01}}} & {\cellcolor[HTML]{F2F4F4}{\textcolor[HTML]{000000}{0.01}}} & {\cellcolor[HTML]{E8F2F0}{\textcolor[HTML]{000000}{0.03}}} & {\cellcolor[HTML]{F0F4F3}{\textcolor[HTML]{000000}{0.01}}} & {\cellcolor[HTML]{F5F4F0}{\textcolor[HTML]{000000}{-0.01}}} \\ 
\method{EsnS} & 75\% & {\cellcolor[HTML]{543005}{\textcolor[HTML]{FFFFFF}{-0.57}}} & {\cellcolor[HTML]{BBE5DF}{\textcolor[HTML]{000000}{0.12}}} & {\cellcolor[HTML]{F6EFDD}{\textcolor[HTML]{000000}{-0.05}}} & {\cellcolor[HTML]{DCEFEC}{\textcolor[HTML]{000000}{0.05}}} & {\cellcolor[HTML]{F6F3EC}{\textcolor[HTML]{000000}{-0.02}}} & {\cellcolor[HTML]{F0F4F3}{\textcolor[HTML]{000000}{0.01}}} & {\cellcolor[HTML]{F3F5F4}{\textcolor[HTML]{000000}{0.00}}} & {\cellcolor[HTML]{F1F4F4}{\textcolor[HTML]{000000}{0.01}}} & {\cellcolor[HTML]{F3F4F4}{\textcolor[HTML]{000000}{0.01}}} & {\cellcolor[HTML]{DAEEEB}{\textcolor[HTML]{000000}{0.06}}} & {\cellcolor[HTML]{E9F2F1}{\textcolor[HTML]{000000}{0.03}}} & {\cellcolor[HTML]{ECF3F2}{\textcolor[HTML]{000000}{0.02}}} \\ 
\method{Gru} & 50\% & {\cellcolor[HTML]{F6F2E9}{\textcolor[HTML]{000000}{-0.02}}} & {\cellcolor[HTML]{F6F2EA}{\textcolor[HTML]{000000}{-0.02}}} & {\cellcolor[HTML]{F6F3EC}{\textcolor[HTML]{000000}{-0.02}}} & {\cellcolor[HTML]{F5F5F5}{\textcolor[HTML]{000000}{0.00}}} & {\cellcolor[HTML]{F5F5F5}{\textcolor[HTML]{000000}{0.00}}} & {\cellcolor[HTML]{F5F5F5}{\textcolor[HTML]{000000}{0.00}}} & {\cellcolor[HTML]{F3F5F4}{\textcolor[HTML]{000000}{0.00}}} & {\cellcolor[HTML]{F1F4F3}{\textcolor[HTML]{000000}{0.01}}} & {\cellcolor[HTML]{F4F5F5}{\textcolor[HTML]{000000}{0.00}}} & {\cellcolor[HTML]{E5F1EF}{\textcolor[HTML]{000000}{0.04}}} & {\cellcolor[HTML]{EEF3F3}{\textcolor[HTML]{000000}{0.01}}} & {\cellcolor[HTML]{E6F2F0}{\textcolor[HTML]{000000}{0.03}}} \\ 
\method{LinD} & 75\% & {\cellcolor[HTML]{543005}{\textcolor[HTML]{FFFFFF}{-0.57}}} & {\cellcolor[HTML]{F0DDAF}{\textcolor[HTML]{000000}{-0.13}}} & {\cellcolor[HTML]{F5F5F5}{\textcolor[HTML]{000000}{0.00}}} & {\cellcolor[HTML]{E4F1EF}{\textcolor[HTML]{000000}{0.04}}} & {\cellcolor[HTML]{F4F5F5}{\textcolor[HTML]{000000}{0.00}}} & {\cellcolor[HTML]{CDEBE7}{\textcolor[HTML]{000000}{0.09}}} & {\cellcolor[HTML]{015147}{\textcolor[HTML]{FFFFFF}{0.45}}} & {\cellcolor[HTML]{DCEFEC}{\textcolor[HTML]{000000}{0.06}}} & {\cellcolor[HTML]{DDEFEC}{\textcolor[HTML]{000000}{0.05}}} & {\cellcolor[HTML]{F2F4F4}{\textcolor[HTML]{000000}{0.01}}} & {\cellcolor[HTML]{F5F4F1}{\textcolor[HTML]{000000}{-0.01}}} & {\cellcolor[HTML]{EDF3F2}{\textcolor[HTML]{000000}{0.02}}} \\ 
\method{LinPo4} & 42\% & {\cellcolor[HTML]{543005}{\textcolor[HTML]{FFFFFF}{-1.70}}} & {\cellcolor[HTML]{C38937}{\textcolor[HTML]{FFFFFF}{-0.29}}} & {\cellcolor[HTML]{F6F0E4}{\textcolor[HTML]{000000}{-0.03}}} & {\cellcolor[HTML]{F5F5F5}{\textcolor[HTML]{000000}{0.00}}} & {\cellcolor[HTML]{F5F5F5}{\textcolor[HTML]{000000}{0.00}}} & {\cellcolor[HTML]{F5F5F4}{\textcolor[HTML]{000000}{0.00}}} & {\cellcolor[HTML]{52AAA1}{\textcolor[HTML]{FFFFFF}{0.26}}} & {\cellcolor[HTML]{CAEBE6}{\textcolor[HTML]{000000}{0.09}}} & {\cellcolor[HTML]{DBEFEC}{\textcolor[HTML]{000000}{0.06}}} & {\cellcolor[HTML]{F5F5F5}{\textcolor[HTML]{000000}{0.00}}} & {\cellcolor[HTML]{F5F5F5}{\textcolor[HTML]{000000}{0.00}}} & {\cellcolor[HTML]{F5F5F5}{\textcolor[HTML]{000000}{0.00}}} \\ 
\method{LinPo6} & 67\% & {\cellcolor[HTML]{543005}{\textcolor[HTML]{FFFFFF}{-2.00}}} & {\cellcolor[HTML]{543005}{\textcolor[HTML]{FFFFFF}{-1.73}}} & {\cellcolor[HTML]{543005}{\textcolor[HTML]{FFFFFF}{-0.89}}} & {\cellcolor[HTML]{F2F4F4}{\textcolor[HTML]{000000}{0.01}}} & {\cellcolor[HTML]{F1F4F4}{\textcolor[HTML]{000000}{0.01}}} & {\cellcolor[HTML]{F6E9C8}{\textcolor[HTML]{000000}{-0.09}}} & {\cellcolor[HTML]{016057}{\textcolor[HTML]{FFFFFF}{0.41}}} & {\cellcolor[HTML]{C0E7E1}{\textcolor[HTML]{000000}{0.11}}} & {\cellcolor[HTML]{EFF3F3}{\textcolor[HTML]{000000}{0.01}}} & {\cellcolor[HTML]{F4F5F5}{\textcolor[HTML]{000000}{0.00}}} & {\cellcolor[HTML]{F4F5F5}{\textcolor[HTML]{000000}{0.00}}} & {\cellcolor[HTML]{F1F4F4}{\textcolor[HTML]{000000}{0.01}}} \\ 
\method{LinS} & 100\% & {\cellcolor[HTML]{F5F5F5}{\textcolor[HTML]{000000}{0.00}}} & {\cellcolor[HTML]{F5F5F5}{\textcolor[HTML]{000000}{0.00}}} & {\cellcolor[HTML]{F5F5F5}{\textcolor[HTML]{000000}{0.00}}} & {\cellcolor[HTML]{E2F0EE}{\textcolor[HTML]{000000}{0.04}}} & {\cellcolor[HTML]{F4F5F5}{\textcolor[HTML]{000000}{0.00}}} & {\cellcolor[HTML]{DEEFED}{\textcolor[HTML]{000000}{0.05}}} & {\cellcolor[HTML]{62B6AC}{\textcolor[HTML]{000000}{0.24}}} & {\cellcolor[HTML]{CEECE7}{\textcolor[HTML]{000000}{0.09}}} & {\cellcolor[HTML]{E0F0EE}{\textcolor[HTML]{000000}{0.05}}} & {\cellcolor[HTML]{8ED2C8}{\textcolor[HTML]{000000}{0.18}}} & {\cellcolor[HTML]{A7DDD4}{\textcolor[HTML]{000000}{0.15}}} & {\cellcolor[HTML]{E2F0EE}{\textcolor[HTML]{000000}{0.04}}} \\ 
\method{Lstm} & 50\% & {\cellcolor[HTML]{F6F2E9}{\textcolor[HTML]{000000}{-0.02}}} & {\cellcolor[HTML]{F3F5F4}{\textcolor[HTML]{000000}{0.00}}} & {\cellcolor[HTML]{F4F5F5}{\textcolor[HTML]{000000}{0.00}}} & {\cellcolor[HTML]{F5F4F3}{\textcolor[HTML]{000000}{0.00}}} & {\cellcolor[HTML]{E9F2F1}{\textcolor[HTML]{000000}{0.03}}} & {\cellcolor[HTML]{F5F5F4}{\textcolor[HTML]{000000}{0.00}}} & {\cellcolor[HTML]{F5F4F3}{\textcolor[HTML]{000000}{0.00}}} & {\cellcolor[HTML]{F5F5F5}{\textcolor[HTML]{000000}{0.00}}} & {\cellcolor[HTML]{F4F5F5}{\textcolor[HTML]{000000}{0.00}}} & {\cellcolor[HTML]{F5F3EE}{\textcolor[HTML]{000000}{-0.01}}} & {\cellcolor[HTML]{F5F5F5}{\textcolor[HTML]{000000}{0.00}}} & {\cellcolor[HTML]{F5F4F3}{\textcolor[HTML]{000000}{0.00}}} \\ 
\method{PgGpD} & 42\% & {\cellcolor[HTML]{F5F5F5}{\textcolor[HTML]{000000}{0.00}}} & {\cellcolor[HTML]{F5F5F5}{\textcolor[HTML]{000000}{0.00}}} & {\cellcolor[HTML]{F5F5F5}{\textcolor[HTML]{000000}{0.00}}} & {\cellcolor[HTML]{F5F3EE}{\textcolor[HTML]{000000}{-0.01}}} & {\cellcolor[HTML]{E3F1EF}{\textcolor[HTML]{000000}{0.04}}} & {\cellcolor[HTML]{F5F5F5}{\textcolor[HTML]{000000}{0.00}}} & {\cellcolor[HTML]{004135}{\textcolor[HTML]{FFFFFF}{0.49}}} & {\cellcolor[HTML]{7FCCC0}{\textcolor[HTML]{000000}{0.20}}} & {\cellcolor[HTML]{96D6CC}{\textcolor[HTML]{000000}{0.17}}} & {\cellcolor[HTML]{F5F5F4}{\textcolor[HTML]{000000}{0.00}}} & {\cellcolor[HTML]{F6F2EB}{\textcolor[HTML]{000000}{-0.02}}} & {\cellcolor[HTML]{F1F4F4}{\textcolor[HTML]{000000}{0.01}}} \\ 
\method{PgGpS} & 50\% & {\cellcolor[HTML]{F5F5F5}{\textcolor[HTML]{000000}{0.00}}} & {\cellcolor[HTML]{F5F5F5}{\textcolor[HTML]{000000}{0.00}}} & {\cellcolor[HTML]{F5F5F5}{\textcolor[HTML]{000000}{0.00}}} & {\cellcolor[HTML]{F5F5F5}{\textcolor[HTML]{000000}{0.00}}} & {\cellcolor[HTML]{F5F5F5}{\textcolor[HTML]{000000}{0.00}}} & {\cellcolor[HTML]{F5F5F5}{\textcolor[HTML]{000000}{0.00}}} & {\cellcolor[HTML]{9DD8CF}{\textcolor[HTML]{000000}{0.16}}} & {\cellcolor[HTML]{D8EEEB}{\textcolor[HTML]{000000}{0.06}}} & {\cellcolor[HTML]{DFF0ED}{\textcolor[HTML]{000000}{0.05}}} & {\cellcolor[HTML]{F1F4F4}{\textcolor[HTML]{000000}{0.01}}} & {\cellcolor[HTML]{DDEFED}{\textcolor[HTML]{000000}{0.05}}} & {\cellcolor[HTML]{E5F1EF}{\textcolor[HTML]{000000}{0.04}}} \\ 
\method{PgLlD} & 58\% & {\cellcolor[HTML]{F6E8C4}{\textcolor[HTML]{000000}{-0.10}}} & {\cellcolor[HTML]{F5F5F5}{\textcolor[HTML]{000000}{0.00}}} & {\cellcolor[HTML]{F6EFE0}{\textcolor[HTML]{000000}{-0.04}}} & {\cellcolor[HTML]{F5F5F4}{\textcolor[HTML]{000000}{0.00}}} & {\cellcolor[HTML]{F5F5F4}{\textcolor[HTML]{000000}{0.00}}} & {\cellcolor[HTML]{F5F5F5}{\textcolor[HTML]{000000}{0.00}}} & {\cellcolor[HTML]{B7E3DD}{\textcolor[HTML]{000000}{0.12}}} & {\cellcolor[HTML]{EAF2F1}{\textcolor[HTML]{000000}{0.03}}} & {\cellcolor[HTML]{E1F0EE}{\textcolor[HTML]{000000}{0.04}}} & {\cellcolor[HTML]{F5F5F5}{\textcolor[HTML]{000000}{0.00}}} & {\cellcolor[HTML]{F5F5F5}{\textcolor[HTML]{000000}{0.00}}} & {\cellcolor[HTML]{F5F5F5}{\textcolor[HTML]{000000}{0.00}}} \\ 
\method{PgLlS} & 83\% & {\cellcolor[HTML]{BCE5DF}{\textcolor[HTML]{000000}{0.12}}} & {\cellcolor[HTML]{E8F2F0}{\textcolor[HTML]{000000}{0.03}}} & {\cellcolor[HTML]{DCEFEC}{\textcolor[HTML]{000000}{0.05}}} & {\cellcolor[HTML]{F5F4F2}{\textcolor[HTML]{000000}{-0.01}}} & {\cellcolor[HTML]{F3F5F4}{\textcolor[HTML]{000000}{0.00}}} & {\cellcolor[HTML]{F5F5F4}{\textcolor[HTML]{000000}{0.00}}} & {\cellcolor[HTML]{E8F2F0}{\textcolor[HTML]{000000}{0.03}}} & {\cellcolor[HTML]{EFF4F3}{\textcolor[HTML]{000000}{0.01}}} & {\cellcolor[HTML]{EFF4F3}{\textcolor[HTML]{000000}{0.01}}} & {\cellcolor[HTML]{EDF3F2}{\textcolor[HTML]{000000}{0.02}}} & {\cellcolor[HTML]{F1F4F4}{\textcolor[HTML]{000000}{0.01}}} & {\cellcolor[HTML]{F2F4F4}{\textcolor[HTML]{000000}{0.01}}} \\ 
\method{PgNetD} & 50\% & {\cellcolor[HTML]{F6EFDE}{\textcolor[HTML]{000000}{-0.05}}} & {\cellcolor[HTML]{E4F1EF}{\textcolor[HTML]{000000}{0.04}}} & {\cellcolor[HTML]{E8F2F0}{\textcolor[HTML]{000000}{0.03}}} & {\cellcolor[HTML]{F6F3EC}{\textcolor[HTML]{000000}{-0.02}}} & {\cellcolor[HTML]{F5F4F1}{\textcolor[HTML]{000000}{-0.01}}} & {\cellcolor[HTML]{F5F3ED}{\textcolor[HTML]{000000}{-0.02}}} & {\cellcolor[HTML]{A0DAD0}{\textcolor[HTML]{000000}{0.16}}} & {\cellcolor[HTML]{CBEBE6}{\textcolor[HTML]{000000}{0.09}}} & {\cellcolor[HTML]{C9EBE6}{\textcolor[HTML]{000000}{0.10}}} & {\cellcolor[HTML]{F5F5F5}{\textcolor[HTML]{000000}{0.00}}} & {\cellcolor[HTML]{F5F5F5}{\textcolor[HTML]{000000}{0.00}}} & {\cellcolor[HTML]{F5F5F4}{\textcolor[HTML]{000000}{0.00}}} \\ 
\method{PgNetS} & 75\% & {\cellcolor[HTML]{F6F1E6}{\textcolor[HTML]{000000}{-0.03}}} & {\cellcolor[HTML]{F2F4F4}{\textcolor[HTML]{000000}{0.01}}} & {\cellcolor[HTML]{F5F4EF}{\textcolor[HTML]{000000}{-0.01}}} & {\cellcolor[HTML]{F5F5F4}{\textcolor[HTML]{000000}{0.00}}} & {\cellcolor[HTML]{F1F4F3}{\textcolor[HTML]{000000}{0.01}}} & {\cellcolor[HTML]{F0F4F3}{\textcolor[HTML]{000000}{0.01}}} & {\cellcolor[HTML]{BCE6DF}{\textcolor[HTML]{000000}{0.12}}} & {\cellcolor[HTML]{D8EEEB}{\textcolor[HTML]{000000}{0.06}}} & {\cellcolor[HTML]{D4EDE9}{\textcolor[HTML]{000000}{0.07}}} & {\cellcolor[HTML]{E2F1EE}{\textcolor[HTML]{000000}{0.04}}} & {\cellcolor[HTML]{DEF0ED}{\textcolor[HTML]{000000}{0.05}}} & {\cellcolor[HTML]{E3F1EF}{\textcolor[HTML]{000000}{0.04}}} \\ 
\method{RaFeD} & 58\% & {\cellcolor[HTML]{F5F4F2}{\textcolor[HTML]{000000}{-0.01}}} & {\cellcolor[HTML]{F5F3EE}{\textcolor[HTML]{000000}{-0.01}}} & {\cellcolor[HTML]{F6F0E4}{\textcolor[HTML]{000000}{-0.03}}} & {\cellcolor[HTML]{F5F5F3}{\textcolor[HTML]{000000}{0.00}}} & {\cellcolor[HTML]{F5F5F5}{\textcolor[HTML]{000000}{0.00}}} & {\cellcolor[HTML]{F5F5F5}{\textcolor[HTML]{000000}{0.00}}} & {\cellcolor[HTML]{D2EDE9}{\textcolor[HTML]{000000}{0.08}}} & {\cellcolor[HTML]{E2F1EE}{\textcolor[HTML]{000000}{0.04}}} & {\cellcolor[HTML]{EBF3F1}{\textcolor[HTML]{000000}{0.02}}} & {\cellcolor[HTML]{DBEFEC}{\textcolor[HTML]{000000}{0.06}}} & {\cellcolor[HTML]{ECF3F2}{\textcolor[HTML]{000000}{0.02}}} & {\cellcolor[HTML]{EDF3F2}{\textcolor[HTML]{000000}{0.02}}} \\ 
\method{RaFeS} & 58\% & {\cellcolor[HTML]{F5F5F4}{\textcolor[HTML]{000000}{0.00}}} & {\cellcolor[HTML]{CBEBE6}{\textcolor[HTML]{000000}{0.09}}} & {\cellcolor[HTML]{F5F4F0}{\textcolor[HTML]{000000}{-0.01}}} & {\cellcolor[HTML]{F6EEDA}{\textcolor[HTML]{000000}{-0.05}}} & {\cellcolor[HTML]{F6F0E0}{\textcolor[HTML]{000000}{-0.04}}} & {\cellcolor[HTML]{F5F5F5}{\textcolor[HTML]{000000}{0.00}}} & {\cellcolor[HTML]{F1F4F3}{\textcolor[HTML]{000000}{0.01}}} & {\cellcolor[HTML]{F5F5F5}{\textcolor[HTML]{000000}{0.00}}} & {\cellcolor[HTML]{F4F5F5}{\textcolor[HTML]{000000}{0.00}}} & {\cellcolor[HTML]{72C2B7}{\textcolor[HTML]{000000}{0.22}}} & {\cellcolor[HTML]{D8EEEB}{\textcolor[HTML]{000000}{0.06}}} & {\cellcolor[HTML]{E2F0EE}{\textcolor[HTML]{000000}{0.04}}} \\ 
\method{Rnn} & 75\% & {\cellcolor[HTML]{F4F5F5}{\textcolor[HTML]{000000}{0.00}}} & {\cellcolor[HTML]{F4F5F5}{\textcolor[HTML]{000000}{0.00}}} & {\cellcolor[HTML]{F0F4F3}{\textcolor[HTML]{000000}{0.01}}} & {\cellcolor[HTML]{F5F5F5}{\textcolor[HTML]{000000}{0.00}}} & {\cellcolor[HTML]{F5F5F5}{\textcolor[HTML]{000000}{0.00}}} & {\cellcolor[HTML]{F3F4F4}{\textcolor[HTML]{000000}{0.01}}} & {\cellcolor[HTML]{EEF3F2}{\textcolor[HTML]{000000}{0.02}}} & {\cellcolor[HTML]{F5F5F5}{\textcolor[HTML]{000000}{0.00}}} & {\cellcolor[HTML]{F5F4F1}{\textcolor[HTML]{000000}{-0.01}}} & {\cellcolor[HTML]{F1F4F4}{\textcolor[HTML]{000000}{0.01}}} & {\cellcolor[HTML]{F4F5F5}{\textcolor[HTML]{000000}{0.00}}} & {\cellcolor[HTML]{F5F5F3}{\textcolor[HTML]{000000}{0.00}}} \\ 
\method{Trafo} & 42\% & {\cellcolor[HTML]{F5F5F3}{\textcolor[HTML]{000000}{0.00}}} & {\cellcolor[HTML]{F5F4F0}{\textcolor[HTML]{000000}{-0.01}}} & {\cellcolor[HTML]{EEF3F3}{\textcolor[HTML]{000000}{0.01}}} & {\cellcolor[HTML]{F6F1E7}{\textcolor[HTML]{000000}{-0.03}}} & {\cellcolor[HTML]{F5F4F1}{\textcolor[HTML]{000000}{-0.01}}} & {\cellcolor[HTML]{F5F4F3}{\textcolor[HTML]{000000}{0.00}}} & {\cellcolor[HTML]{EEF3F2}{\textcolor[HTML]{000000}{0.02}}} & {\cellcolor[HTML]{F5F5F4}{\textcolor[HTML]{000000}{0.00}}} & {\cellcolor[HTML]{F5F5F4}{\textcolor[HTML]{000000}{0.00}}} & {\cellcolor[HTML]{EEF3F3}{\textcolor[HTML]{000000}{0.01}}} & {\cellcolor[HTML]{F1F4F4}{\textcolor[HTML]{000000}{0.01}}} & {\cellcolor[HTML]{EEF3F3}{\textcolor[HTML]{000000}{0.02}}} \\ 
\bottomrule
\end{tabular}
\end{center}

%% file: tbl/cme_SvsD_lorenz.tex
\begin{center}
\caption*{
{\large Relative Differences in $\cme$ Between State ($S$) and Diff. Quotient Target ($D$)}
} 
\fontsize{8.0pt}{10pt}\selectfont
\fontfamily{phv}\selectfont
\renewcommand{\arraystretch}{1.05}
\setlength{\tabcolsep}{0.3em}
\begin{tabular}{lrrrrrrrrrrrrrrr}
\toprule
 &  & \multicolumn{2}{c}{\Dysts{}} & \multicolumn{12}{c}{\DeebLorenz{}} \\ 
\cmidrule(lr){3-4} \cmidrule(lr){5-16}
 &  & \multicolumn{2}{c}{Constant $\stepsize$} & \multicolumn{6}{c}{Constant $\stepsize$} & \multicolumn{6}{c}{Random $\stepsize$} \\ 
\cmidrule(lr){3-4} \cmidrule(lr){5-10} \cmidrule(lr){11-16}
 &  & Noisefree & Noisy & \multicolumn{3}{c}{Noisefree} & \multicolumn{3}{c}{Noisy} & \multicolumn{3}{c}{Noisefree} & \multicolumn{3}{c}{Noisy} \\ 
\cmidrule(lr){3-3} \cmidrule(lr){4-4} \cmidrule(lr){5-7} \cmidrule(lr){8-10} \cmidrule(lr){11-13} \cmidrule(lr){14-16}
Method &  & Median & Median & \cellcolor[HTML]{00BA38}{S} & \cellcolor[HTML]{F8766D}{R} & \cellcolor[HTML]{619CFF}{N} & \cellcolor[HTML]{00BA38}{S} & \cellcolor[HTML]{F8766D}{R} & \cellcolor[HTML]{619CFF}{N} & \cellcolor[HTML]{00BA38}{S} & \cellcolor[HTML]{F8766D}{R} & \cellcolor[HTML]{619CFF}{N} & \cellcolor[HTML]{00BA38}{S} & \cellcolor[HTML]{F8766D}{R} & \cellcolor[HTML]{619CFF}{N} \\ 
\midrule\addlinespace[2.5pt]
\multicolumn{2}{r}{$D<S$}  & 67\% & 33\% & 92\% & 50\% & 67\% & 58\% & 58\% & 58\% & 100\% & 100\% & 100\% & 25\% & 33\% & 33\% \\ 
\method{Esn} & 86\% & {\cellcolor[HTML]{2F9088}{\textcolor[HTML]{FFFFFF}{0.31}}} & {\cellcolor[HTML]{E2F0EE}{\textcolor[HTML]{000000}{0.04}}} & {\cellcolor[HTML]{1A7971}{\textcolor[HTML]{FFFFFF}{0.36}}} & {\cellcolor[HTML]{EEF3F2}{\textcolor[HTML]{000000}{0.02}}} & {\cellcolor[HTML]{ECF3F2}{\textcolor[HTML]{000000}{0.02}}} & {\cellcolor[HTML]{F5F5F5}{\textcolor[HTML]{000000}{0.00}}} & {\cellcolor[HTML]{F2F4F4}{\textcolor[HTML]{000000}{0.01}}} & {\cellcolor[HTML]{F6F3EC}{\textcolor[HTML]{000000}{-0.02}}} & {\cellcolor[HTML]{72C2B7}{\textcolor[HTML]{000000}{0.22}}} & {\cellcolor[HTML]{B3E2DB}{\textcolor[HTML]{000000}{0.13}}} & {\cellcolor[HTML]{C2E8E2}{\textcolor[HTML]{000000}{0.11}}} & {\cellcolor[HTML]{9FDAD0}{\textcolor[HTML]{000000}{0.16}}} & {\cellcolor[HTML]{D1ECE8}{\textcolor[HTML]{000000}{0.08}}} & {\cellcolor[HTML]{D0ECE8}{\textcolor[HTML]{000000}{0.08}}} \\ 
\method{EsnT} & 75\% & {\cellcolor[HTML]{808080}{\textcolor[HTML]{FFFFFF}{}}} & {\cellcolor[HTML]{808080}{\textcolor[HTML]{FFFFFF}{}}} & {\cellcolor[HTML]{003C30}{\textcolor[HTML]{FFFFFF}{0.93}}} & {\cellcolor[HTML]{F1DEB1}{\textcolor[HTML]{000000}{-0.13}}} & {\cellcolor[HTML]{F4F5F5}{\textcolor[HTML]{000000}{0.00}}} & {\cellcolor[HTML]{F6F1E6}{\textcolor[HTML]{000000}{-0.03}}} & {\cellcolor[HTML]{EAF2F1}{\textcolor[HTML]{000000}{0.02}}} & {\cellcolor[HTML]{F5F5F5}{\textcolor[HTML]{000000}{0.00}}} & {\cellcolor[HTML]{2F9088}{\textcolor[HTML]{FFFFFF}{0.31}}} & {\cellcolor[HTML]{BDE6E0}{\textcolor[HTML]{000000}{0.11}}} & {\cellcolor[HTML]{C1E8E2}{\textcolor[HTML]{000000}{0.11}}} & {\cellcolor[HTML]{B5E2DB}{\textcolor[HTML]{000000}{0.13}}} & {\cellcolor[HTML]{D7EEEB}{\textcolor[HTML]{000000}{0.07}}} & {\cellcolor[HTML]{DFF0ED}{\textcolor[HTML]{000000}{0.05}}} \\ 
\method{Lin} & 43\% & {\cellcolor[HTML]{975B12}{\textcolor[HTML]{FFFFFF}{-0.38}}} & {\cellcolor[HTML]{F5F5F4}{\textcolor[HTML]{000000}{0.00}}} & {\cellcolor[HTML]{003C30}{\textcolor[HTML]{FFFFFF}{1.24}}} & {\cellcolor[HTML]{5E3607}{\textcolor[HTML]{FFFFFF}{-0.48}}} & {\cellcolor[HTML]{65B8AE}{\textcolor[HTML]{000000}{0.24}}} & {\cellcolor[HTML]{F3F4F4}{\textcolor[HTML]{000000}{0.00}}} & {\cellcolor[HTML]{F5F5F5}{\textcolor[HTML]{000000}{0.00}}} & {\cellcolor[HTML]{F5F4F1}{\textcolor[HTML]{000000}{-0.01}}} & {\cellcolor[HTML]{90D3C9}{\textcolor[HTML]{000000}{0.18}}} & {\cellcolor[HTML]{CEECE7}{\textcolor[HTML]{000000}{0.08}}} & {\cellcolor[HTML]{BFE6E1}{\textcolor[HTML]{000000}{0.11}}} & {\cellcolor[HTML]{F6E9C8}{\textcolor[HTML]{000000}{-0.09}}} & {\cellcolor[HTML]{F6F1E5}{\textcolor[HTML]{000000}{-0.03}}} & {\cellcolor[HTML]{F6EEDB}{\textcolor[HTML]{000000}{-0.05}}} \\ 
\method{LinT} & 58\% & {\cellcolor[HTML]{808080}{\textcolor[HTML]{FFFFFF}{}}} & {\cellcolor[HTML]{808080}{\textcolor[HTML]{FFFFFF}{}}} & {\cellcolor[HTML]{003C30}{\textcolor[HTML]{FFFFFF}{0.82}}} & {\cellcolor[HTML]{543005}{\textcolor[HTML]{FFFFFF}{-0.60}}} & {\cellcolor[HTML]{65B8AE}{\textcolor[HTML]{000000}{0.24}}} & {\cellcolor[HTML]{F5F5F5}{\textcolor[HTML]{000000}{0.00}}} & {\cellcolor[HTML]{F5F5F5}{\textcolor[HTML]{000000}{0.00}}} & {\cellcolor[HTML]{E8F2F1}{\textcolor[HTML]{000000}{0.03}}} & {\cellcolor[HTML]{0A6C64}{\textcolor[HTML]{FFFFFF}{0.39}}} & {\cellcolor[HTML]{DCEFEC}{\textcolor[HTML]{000000}{0.05}}} & {\cellcolor[HTML]{BAE4DE}{\textcolor[HTML]{000000}{0.12}}} & {\cellcolor[HTML]{CC984B}{\textcolor[HTML]{FFFFFF}{-0.26}}} & {\cellcolor[HTML]{E3C787}{\textcolor[HTML]{000000}{-0.19}}} & {\cellcolor[HTML]{F6EBCF}{\textcolor[HTML]{000000}{-0.08}}} \\ 
\method{PgGp} & 57\% & {\cellcolor[HTML]{003C30}{\textcolor[HTML]{FFFFFF}{0.67}}} & {\cellcolor[HTML]{F6F3EC}{\textcolor[HTML]{000000}{-0.02}}} & {\cellcolor[HTML]{003C30}{\textcolor[HTML]{FFFFFF}{1.12}}} & {\cellcolor[HTML]{003C30}{\textcolor[HTML]{FFFFFF}{0.75}}} & {\cellcolor[HTML]{45A198}{\textcolor[HTML]{FFFFFF}{0.28}}} & {\cellcolor[HTML]{D4EDE9}{\textcolor[HTML]{000000}{0.07}}} & {\cellcolor[HTML]{F5F5F4}{\textcolor[HTML]{000000}{0.00}}} & {\cellcolor[HTML]{F5F4F0}{\textcolor[HTML]{000000}{-0.01}}} & {\cellcolor[HTML]{ADDFD7}{\textcolor[HTML]{000000}{0.14}}} & {\cellcolor[HTML]{D4EDEA}{\textcolor[HTML]{000000}{0.07}}} & {\cellcolor[HTML]{CFECE8}{\textcolor[HTML]{000000}{0.08}}} & {\cellcolor[HTML]{F5E7C1}{\textcolor[HTML]{000000}{-0.10}}} & {\cellcolor[HTML]{F5F3EF}{\textcolor[HTML]{000000}{-0.01}}} & {\cellcolor[HTML]{F6EDD6}{\textcolor[HTML]{000000}{-0.06}}} \\ 
\method{PgGpT} & 67\% & {\cellcolor[HTML]{808080}{\textcolor[HTML]{FFFFFF}{}}} & {\cellcolor[HTML]{808080}{\textcolor[HTML]{FFFFFF}{}}} & {\cellcolor[HTML]{003C30}{\textcolor[HTML]{FFFFFF}{1.12}}} & {\cellcolor[HTML]{003C30}{\textcolor[HTML]{FFFFFF}{0.75}}} & {\cellcolor[HTML]{45A198}{\textcolor[HTML]{FFFFFF}{0.28}}} & {\cellcolor[HTML]{DAEFEB}{\textcolor[HTML]{000000}{0.06}}} & {\cellcolor[HTML]{E4F1EF}{\textcolor[HTML]{000000}{0.04}}} & {\cellcolor[HTML]{F5F4F0}{\textcolor[HTML]{000000}{-0.01}}} & {\cellcolor[HTML]{004A3F}{\textcolor[HTML]{FFFFFF}{0.47}}} & {\cellcolor[HTML]{79C8BC}{\textcolor[HTML]{000000}{0.21}}} & {\cellcolor[HTML]{7DCBBF}{\textcolor[HTML]{000000}{0.20}}} & {\cellcolor[HTML]{F3E3BA}{\textcolor[HTML]{000000}{-0.11}}} & {\cellcolor[HTML]{F6EACB}{\textcolor[HTML]{000000}{-0.08}}} & {\cellcolor[HTML]{F6E9C8}{\textcolor[HTML]{000000}{-0.09}}} \\ 
\method{PgLl} & 50\% & {\cellcolor[HTML]{F6EBCF}{\textcolor[HTML]{000000}{-0.08}}} & {\cellcolor[HTML]{F0F4F3}{\textcolor[HTML]{000000}{0.01}}} & {\cellcolor[HTML]{F0F4F3}{\textcolor[HTML]{000000}{0.01}}} & {\cellcolor[HTML]{F5F5F4}{\textcolor[HTML]{000000}{0.00}}} & {\cellcolor[HTML]{F5F3ED}{\textcolor[HTML]{000000}{-0.02}}} & {\cellcolor[HTML]{F5F5F5}{\textcolor[HTML]{000000}{0.00}}} & {\cellcolor[HTML]{F5F5F5}{\textcolor[HTML]{000000}{0.00}}} & {\cellcolor[HTML]{F5F5F5}{\textcolor[HTML]{000000}{0.00}}} & {\cellcolor[HTML]{87D0C4}{\textcolor[HTML]{000000}{0.19}}} & {\cellcolor[HTML]{C9EAE6}{\textcolor[HTML]{000000}{0.10}}} & {\cellcolor[HTML]{BBE5DF}{\textcolor[HTML]{000000}{0.12}}} & {\cellcolor[HTML]{F6EBCD}{\textcolor[HTML]{000000}{-0.08}}} & {\cellcolor[HTML]{F6F0E3}{\textcolor[HTML]{000000}{-0.04}}} & {\cellcolor[HTML]{F6EDD6}{\textcolor[HTML]{000000}{-0.06}}} \\ 
\method{PgLlT} & 42\% & {\cellcolor[HTML]{808080}{\textcolor[HTML]{FFFFFF}{}}} & {\cellcolor[HTML]{808080}{\textcolor[HTML]{FFFFFF}{}}} & {\cellcolor[HTML]{DEC07B}{\textcolor[HTML]{000000}{-0.20}}} & {\cellcolor[HTML]{F6F1E6}{\textcolor[HTML]{000000}{-0.03}}} & {\cellcolor[HTML]{F4E3BB}{\textcolor[HTML]{000000}{-0.11}}} & {\cellcolor[HTML]{F3F4F4}{\textcolor[HTML]{000000}{0.00}}} & {\cellcolor[HTML]{F5F4F3}{\textcolor[HTML]{000000}{0.00}}} & {\cellcolor[HTML]{F4F5F5}{\textcolor[HTML]{000000}{0.00}}} & {\cellcolor[HTML]{429F97}{\textcolor[HTML]{FFFFFF}{0.28}}} & {\cellcolor[HTML]{C2E8E2}{\textcolor[HTML]{000000}{0.11}}} & {\cellcolor[HTML]{A5DCD3}{\textcolor[HTML]{000000}{0.15}}} & {\cellcolor[HTML]{F6E8C4}{\textcolor[HTML]{000000}{-0.10}}} & {\cellcolor[HTML]{F6EFDF}{\textcolor[HTML]{000000}{-0.04}}} & {\cellcolor[HTML]{F6ECD2}{\textcolor[HTML]{000000}{-0.07}}} \\ 
\method{PgNet} & 71\% & {\cellcolor[HTML]{003C30}{\textcolor[HTML]{FFFFFF}{0.56}}} & {\cellcolor[HTML]{F5F4F3}{\textcolor[HTML]{000000}{0.00}}} & {\cellcolor[HTML]{83CEC2}{\textcolor[HTML]{000000}{0.20}}} & {\cellcolor[HTML]{D1ECE9}{\textcolor[HTML]{000000}{0.08}}} & {\cellcolor[HTML]{9FD9D0}{\textcolor[HTML]{000000}{0.16}}} & {\cellcolor[HTML]{C5E9E4}{\textcolor[HTML]{000000}{0.10}}} & {\cellcolor[HTML]{DEEFED}{\textcolor[HTML]{000000}{0.05}}} & {\cellcolor[HTML]{D5EDEA}{\textcolor[HTML]{000000}{0.07}}} & {\cellcolor[HTML]{9AD7CE}{\textcolor[HTML]{000000}{0.16}}} & {\cellcolor[HTML]{D8EEEB}{\textcolor[HTML]{000000}{0.06}}} & {\cellcolor[HTML]{D0ECE8}{\textcolor[HTML]{000000}{0.08}}} & {\cellcolor[HTML]{F6EDD8}{\textcolor[HTML]{000000}{-0.06}}} & {\cellcolor[HTML]{F6F2E8}{\textcolor[HTML]{000000}{-0.03}}} & {\cellcolor[HTML]{F6F0E1}{\textcolor[HTML]{000000}{-0.04}}} \\ 
\method{PgNetT} & 75\% & {\cellcolor[HTML]{808080}{\textcolor[HTML]{FFFFFF}{}}} & {\cellcolor[HTML]{808080}{\textcolor[HTML]{FFFFFF}{}}} & {\cellcolor[HTML]{8ED2C8}{\textcolor[HTML]{000000}{0.18}}} & {\cellcolor[HTML]{C1E7E2}{\textcolor[HTML]{000000}{0.11}}} & {\cellcolor[HTML]{82CEC2}{\textcolor[HTML]{000000}{0.20}}} & {\cellcolor[HTML]{CDEBE7}{\textcolor[HTML]{000000}{0.09}}} & {\cellcolor[HTML]{E5F1EF}{\textcolor[HTML]{000000}{0.03}}} & {\cellcolor[HTML]{E1F0EE}{\textcolor[HTML]{000000}{0.04}}} & {\cellcolor[HTML]{7CCABE}{\textcolor[HTML]{000000}{0.21}}} & {\cellcolor[HTML]{CBEBE6}{\textcolor[HTML]{000000}{0.09}}} & {\cellcolor[HTML]{C5E9E4}{\textcolor[HTML]{000000}{0.10}}} & {\cellcolor[HTML]{F6E8C3}{\textcolor[HTML]{000000}{-0.10}}} & {\cellcolor[HTML]{F6EBCF}{\textcolor[HTML]{000000}{-0.08}}} & {\cellcolor[HTML]{F6EACC}{\textcolor[HTML]{000000}{-0.08}}} \\ 
\method{RaFe} & 71\% & {\cellcolor[HTML]{47A29A}{\textcolor[HTML]{FFFFFF}{0.28}}} & {\cellcolor[HTML]{F6F0E2}{\textcolor[HTML]{000000}{-0.04}}} & {\cellcolor[HTML]{2A8A82}{\textcolor[HTML]{FFFFFF}{0.33}}} & {\cellcolor[HTML]{D8EEEB}{\textcolor[HTML]{000000}{0.06}}} & {\cellcolor[HTML]{F6F0E3}{\textcolor[HTML]{000000}{-0.04}}} & {\cellcolor[HTML]{F6EEDB}{\textcolor[HTML]{000000}{-0.05}}} & {\cellcolor[HTML]{F6F0E0}{\textcolor[HTML]{000000}{-0.04}}} & {\cellcolor[HTML]{E6F1F0}{\textcolor[HTML]{000000}{0.03}}} & {\cellcolor[HTML]{82CEC2}{\textcolor[HTML]{000000}{0.20}}} & {\cellcolor[HTML]{CAEBE6}{\textcolor[HTML]{000000}{0.09}}} & {\cellcolor[HTML]{B9E4DE}{\textcolor[HTML]{000000}{0.12}}} & {\cellcolor[HTML]{A8DDD5}{\textcolor[HTML]{000000}{0.14}}} & {\cellcolor[HTML]{D8EEEB}{\textcolor[HTML]{000000}{0.06}}} & {\cellcolor[HTML]{D5EDEA}{\textcolor[HTML]{000000}{0.07}}} \\ 
\method{RaFeT} & 67\% & {\cellcolor[HTML]{808080}{\textcolor[HTML]{FFFFFF}{}}} & {\cellcolor[HTML]{808080}{\textcolor[HTML]{FFFFFF}{}}} & {\cellcolor[HTML]{2B8C84}{\textcolor[HTML]{FFFFFF}{0.32}}} & {\cellcolor[HTML]{F6EFE0}{\textcolor[HTML]{000000}{-0.04}}} & {\cellcolor[HTML]{F6EDD7}{\textcolor[HTML]{000000}{-0.06}}} & {\cellcolor[HTML]{F5F5F5}{\textcolor[HTML]{000000}{0.00}}} & {\cellcolor[HTML]{F5F5F5}{\textcolor[HTML]{000000}{0.00}}} & {\cellcolor[HTML]{E6F1F0}{\textcolor[HTML]{000000}{0.03}}} & {\cellcolor[HTML]{52AAA0}{\textcolor[HTML]{FFFFFF}{0.26}}} & {\cellcolor[HTML]{AFE0D8}{\textcolor[HTML]{000000}{0.13}}} & {\cellcolor[HTML]{ACDFD7}{\textcolor[HTML]{000000}{0.14}}} & {\cellcolor[HTML]{F6F3EC}{\textcolor[HTML]{000000}{-0.02}}} & {\cellcolor[HTML]{EBF3F2}{\textcolor[HTML]{000000}{0.02}}} & {\cellcolor[HTML]{E0F0ED}{\textcolor[HTML]{000000}{0.05}}} \\ 
\bottomrule
\end{tabular}
\end{center}

%% file: tbl/tableLorenzLength.tex
\begin{center}
\caption*{
{\large Cumulative Maximum Error for Test Data of \model{Lorenz63std}}
} 
\fontsize{8.0pt}{10pt}\selectfont
\fontfamily{phv}\selectfont
\renewcommand{\arraystretch}{1.05}
\setlength{\tabcolsep}{0.3em}
\rowcolors{2}{gray!20}{white}
\begin{tabular}{lcllrrrrrrrrrrrr}
\toprule
 &  &  &  & \multicolumn{6}{c}{Noisefree} & \multicolumn{6}{c}{Noisy} \\ 
\cmidrule(lr){5-10} \cmidrule(lr){11-16}
 &  &  &  & \multicolumn{2}{c}{$n=10^3$} & \multicolumn{2}{c}{$n=10^4$} & \multicolumn{2}{c}{$n=10^5$} & \multicolumn{2}{c}{$n=10^3$} & \multicolumn{2}{c}{$n=10^4$} & \multicolumn{2}{c}{$n=10^5$} \\ 
\cmidrule(lr){5-6} \cmidrule(lr){7-8} \cmidrule(lr){9-10} \cmidrule(lr){11-12} \cmidrule(lr){13-14} \cmidrule(lr){15-16}
Name & Class & Model & Variant & \# & $\cme$ & \# & $\cme$ & \# & $\cme$ & \# & $\cme$ & \# & $\cme$ & \# & $\cme$ \\ 
\midrule\addlinespace[2.5pt]
\method{LinPo6} & {\cellcolor[HTML]{B0B0FF}{\textcolor[HTML]{000000}{Fit Prop.}}} & Lin & Poly6 & {\cellcolor[HTML]{58C765}{\textcolor[HTML]{000000}{9}}} & {\cellcolor[HTML]{58C765}{\textcolor[HTML]{000000}{0.15}}} & {\cellcolor[HTML]{FDE725}{\textcolor[HTML]{000000}{1}}} & {\cellcolor[HTML]{FDE725}{\textcolor[HTML]{000000}{0.0000086}}} & {\cellcolor[HTML]{FDE725}{\textcolor[HTML]{000000}{1}}} & {\cellcolor[HTML]{FDE725}{\textcolor[HTML]{000000}{0.000024}}} & {\cellcolor[HTML]{287D8E}{\textcolor[HTML]{FFFFFF}{19}}} & {\cellcolor[HTML]{287D8E}{\textcolor[HTML]{FFFFFF}{0.85}}} & {\cellcolor[HTML]{20A386}{\textcolor[HTML]{FFFFFF}{14}}} & {\cellcolor[HTML]{20A386}{\textcolor[HTML]{FFFFFF}{0.76}}} & {\cellcolor[HTML]{20A386}{\textcolor[HTML]{FFFFFF}{14}}} & {\cellcolor[HTML]{20A386}{\textcolor[HTML]{FFFFFF}{0.74}}} \\
\bottomrule
\end{tabular}
\begin{minipage}{\linewidth}
\textsuperscript{\textit{*}}Batch size 1024. Trained on GPUs.\\
\end{minipage}
\end{center}